\documentclass[a4paper,11pt,oneside,titlepage]{book}
\usepackage[T1]{fontenc}		
\usepackage[utf8]{inputenc}		
\usepackage[english]{babel}		
\usepackage{geometry}			
\usepackage{setspace}			
\usepackage{fancyhdr}			
\usepackage{afterpage} 			
\usepackage{hyperref}			
\usepackage{url}
\usepackage{color}			
\usepackage{xcolor}			
\usepackage{enumerate}			
\usepackage{enumitem}			
\usepackage{graphicx}			
\usepackage{epstopdf}			
\usepackage{pstool}			
\usepackage{psfrag}
\usepackage{array}			
\usepackage{tabularx}			
\usepackage[bottom]{footmisc}		
\usepackage{booktabs}			
\usepackage{longtable}			
\usepackage{caption}			
\usepackage{float}			
\usepackage{subfloat}			
\usepackage{rotating}			
\usepackage{rotfloat}			
\usepackage{amsmath} 			
\usepackage{amssymb}			
\usepackage{cancel}                     
\usepackage{mleftright}
\usepackage{listings}                   
\usepackage[swapnames,norules,nouppercase]{frontespizio}
\usepackage[intoc]{nomencl}
\usepackage[acronym,toc,nomain,nopostdot,nonumberlist]{glossaries}
\usepackage[square, numbers, comma, sort&compress]{natbib}
\usepackage{verbatim}                   
\usepackage{wrapfig}                 
\usepackage{blindtext}
\usepackage{cleveref}
\usepackage{tikz}
\usepackage{subcaption}
\usepackage[normalem]{ulem}
\usepackage{multirow}
\usepackage{xcolor}
\usepackage{bm}
\usepackage{bbold}
\usepackage{cases}
\usepackage{fontawesome5}
\usepackage{colortbl}
\usepackage[nohyperlinks]{acronym}
\usepackage{pgfplots}
\usepgfplotslibrary{fillbetween}
\usetikzlibrary{arrows,matrix,chains,fit,positioning,shapes.multipart,backgrounds} 
\usetikzlibrary{decorations.pathreplacing}
\usetikzlibrary{fadings}
\usepackage{dblfloatfix}
\usepackage{empheq}
\preto\chapter{\glsresetall}
\newacronym{ML}{ML}{Machine Learning}
\newacronym{AI}{AI}{Artificial Intelligence}
\newacronym{DL}{DL}{Deep Learning}
\newacronym{DNN}{DNN}{Deep Neural Network}
\newacronym{XAI}{XAI}{eXplainable Artificial Intelligence}
\newacronym{MANN}{MANN}{Memory Augmented Neural Network}
\newacronym{NN}{NN}{Neural Network}
\newacronym{CNN}{CNN}{Convolutional Neural Network}
\newacronym{GNN}{GNN}{Graph Neural Network}
\newacronym{GCN}{GCN}{Graph Convolutional Network}
\newacronym{GAT}{GAT}{Graph Attention Network}
\newacronym{GIN}{GIN}{Graph Isomorphism Network}
\newacronym{VA}{VA}{Visual Analytics}
\newacronym{CoEx}{CoEx}{Composotional Explanations}
\newacronym{NetDissect}{NetDissect}{Network Dissection}
\newacronym{PIGNN}{PIGNN}{Prototype-based Interpretable Graph Neural Network}
\newacronym{SDNC}{SDNC}{Simplified Differentiable Neural Computer}
\newacronym{DNC}{DNC}{Differentiable Neural Computer}
\newacronym{MLP}{MLP}{Multi-Layer Perceptron}
\newacronym{CBM}{CBM}{Concept Bottleneck Models}
\newacronym{GCW}{GCW}{Graph Concept Whitening}

\usepackage{amsthm}

\theoremstyle{definition}
\newtheorem{definition}{Definition}[section]

\newcommand*{\wrong}[1]{%
	\tikz[baseline=(X.base)] \node[rectangle, fill=red!90!blue!60, rounded corners, inner sep=0.3mm] (X) {#1};%
}
\newcommand*{\prediction}[1]{%
	\tikz[baseline=(X.base)] \node[rectangle, fill=green, rounded corners, inner sep=0.3mm] (X) {#1};%
}
\newcommand*{\best}[1]{%
	\tikz[baseline=(X.base)] \node[rectangle, fill=cyan!90!black!100, rounded corners, inner sep=0.3mm] (X) {#1};%
}
\newcommand*{\worst}[1]{%
	\tikz[baseline=(X.base)] \node[rectangle, fill=orange, rounded corners, inner sep=0.3mm] (X) {#1};%
}
\tikzset{
  mymatrix/.style={matrix of nodes, nodes=typetag, row sep=1em},
  mycontainer/.style={draw=gray, inner sep=1ex},
  typetag/.style={draw=gray, inner sep=1ex, anchor=west},
  title/.style={draw=none, color=gray, inner sep=0pt},
  >=stealth',
  squarednode/.style={
    rectangle,
    rounded corners,
    draw=black, very thick,
    text centered},
  roundnode/.style={circle, draw=black!60, fill=white!5, very thick, minimum size=2mm},
  to/.style={
    ->,
    thick,
    shorten <=2pt,
    shorten >=2pt,},
  pre/.style={=stealth',semithick},
  arrowfancy/.style={->,shorten >=1pt,>=stealth',ultra thick,draw=CornflowerBlue,looseness=.8},
  post/.style={->,shorten >=1pt,>=stealth',semithick},
  vector/.style={
    rectangle split, rectangle split parts=#1, draw, anchor=center},
  from/.style={
    <-,
    thick,
    shorten <=2pt,
    shorten >=2pt,}
}

\usetikzlibrary{arrows,decorations.pathmorphing,backgrounds,positioning,fit,
petri, decorations.pathreplacing,shadows,calc}
\definecolor{echoreg}{HTML}{2cb1e1}
\tikzset{%
  cascadedd/.style = {%
    general shadow = {%
      shadow scale = 1,
      shadow xshift = -3ex,
      shadow yshift = 3ex,
      draw=black,
      thick,
      fill = white},
    general shadow = {%
      shadow scale = 1,
      shadow xshift = -2.5ex,
      shadow yshift = 2.5ex,
      draw=black,
      thick,
      fill = white},
    general shadow = {%
      shadow scale = 1,
      shadow xshift = -2ex,
      shadow yshift = 2ex,
      draw=black,
      thick,
      fill = white},
    general shadow = {%
      shadow scale = 1,
      shadow xshift = -1.5ex,
      shadow yshift = 1.5ex,
      draw=black,
      thick,
      fill = white},
    general shadow = {%
      shadow scale = 1,
      shadow xshift = -1ex,
      shadow yshift = 1ex,
      draw=black,
      thick,
      fill = white},
    general shadow = {%
      shadow scale = 1,
      shadow xshift = -.5ex,
      shadow yshift = .5ex,
      draw=black,
      thick,
      fill = white},
    fill = white, 
    draw,
    thick}}
    \tikzset{%
    cascadedd_images/.style = {%
      general shadow = {%
        shadow scale = 1,
        shadow xshift = -3ex,
        shadow yshift = 3ex,
        draw=black,
        thick,
        fill = white},
      general shadow = {%
        shadow scale = 1,
        shadow xshift = -2.5ex,
        shadow yshift = 2.5ex,
        draw=black,
        thick,
        fill = white},
      general shadow = {%
        shadow scale = 1,
        shadow xshift = -2ex,
        shadow yshift = 2ex,
        draw=black,
        thick,
        fill = white},
      general shadow = {%
        shadow scale = 1,
        shadow xshift = -1.5ex,
        shadow yshift = 1.5ex,
        draw=black,
        thick,
        fill = white},
      general shadow = {%
        shadow scale = 1,
        shadow xshift = -1ex,
        shadow yshift = 1ex,
        draw=black,
        thick,
        fill = white},
      general shadow = {%
        minimum width=1cm,
        minimum height=1cm,
        shadow scale = 1,
        shadow xshift = -.5ex,
        shadow yshift = .5ex,
        draw=black,
        thick,
        fill = white},
      fill = white, 
      draw,
      thick}}
    
    \tikzset{%
              base/.style = {rectangle, rounded corners, draw=black,thick,
                             text centered,fill = echoreg!30, font=\sffamily},
    input/.style = {base,minimum width=2cm,align=center, minimum height=1cm,
    draw=black, text width=3cm},
    supportset/.style={cascadedd, align=center,rounded corners, 
    minimum width=2cm, minimum height=1cm, text width=3cm },
    arrow/.style = {very thick,-stealth},
    architecture/.style={opacity=0.8,align=center,rectangle, draw, 
    rounded corners, thick, fill=red!30, text width=3cm,  minimum width=3cm, minimum height=0.8cm},}

    \tikzset{
  mymatrix/.style={matrix of nodes, nodes=typetag, row sep=1em},
  mycontainer/.style={draw=gray, inner sep=1ex},
  typetag/.style={draw=gray, inner sep=1ex, anchor=west},
  title/.style={draw=none, color=gray, inner sep=0pt},
  >=stealth',
  squarednode/.style={
    rectangle,
    rounded corners,
    draw=black, very thick,
    text centered},
  roundnode/.style={circle, draw=black!60, fill=white!5, very thick, minimum size=2mm},
  to/.style={
    ->,
    thick,
    shorten <=2pt,
    shorten >=2pt,},
  pre/.style={=stealth',semithick},
  arrowfancy/.style={->,shorten >=1pt,>=stealth',ultra thick,draw=CornflowerBlue,looseness=.8},
  post/.style={->,shorten >=1pt,>=stealth',semithick},
  vector/.style={
    rectangle split, rectangle split parts=#1, draw, anchor=center},
  from/.style={
    <-,
    thick,
    shorten <=2pt,
    shorten >=2pt,}
}

\DeclareMathOperator*{\argmax}{arg\,max}
\DeclareMathOperator*{\argmin}{arg\,min}
\fancypagestyle{plain}{
                        \fancyhf{}                              
                        \fancyfoot[R]{\thepage}                 
                        \renewcommand{\headrulewidth}{0pt}      
                        \renewcommand{\footrulewidth}{0.4pt}    
                        }                                       
\hypersetup{colorlinks=true, linkcolor=black, citecolor=black, filecolor=black, urlcolor=black}
\definecolor{sapienza}{RGB}{130,36,51} 
\definecolor{cust1}{RGB}{85,85,85}
\definecolor{cust2}{RGB}{212,212,212}
\makenomenclature 
\makeglossaries 
\usepackage{calligra}

\begin{document}
\frontmatter 	   
\pagestyle{empty}  

\begin{frontespizio}
\Preambolo{\renewcommand{\fronttitlefont}{\fontsize{24}{24}\bfseries}}

\Margini{4cm}{3cm}{3cm}{3cm}				
\Logo[4cm]{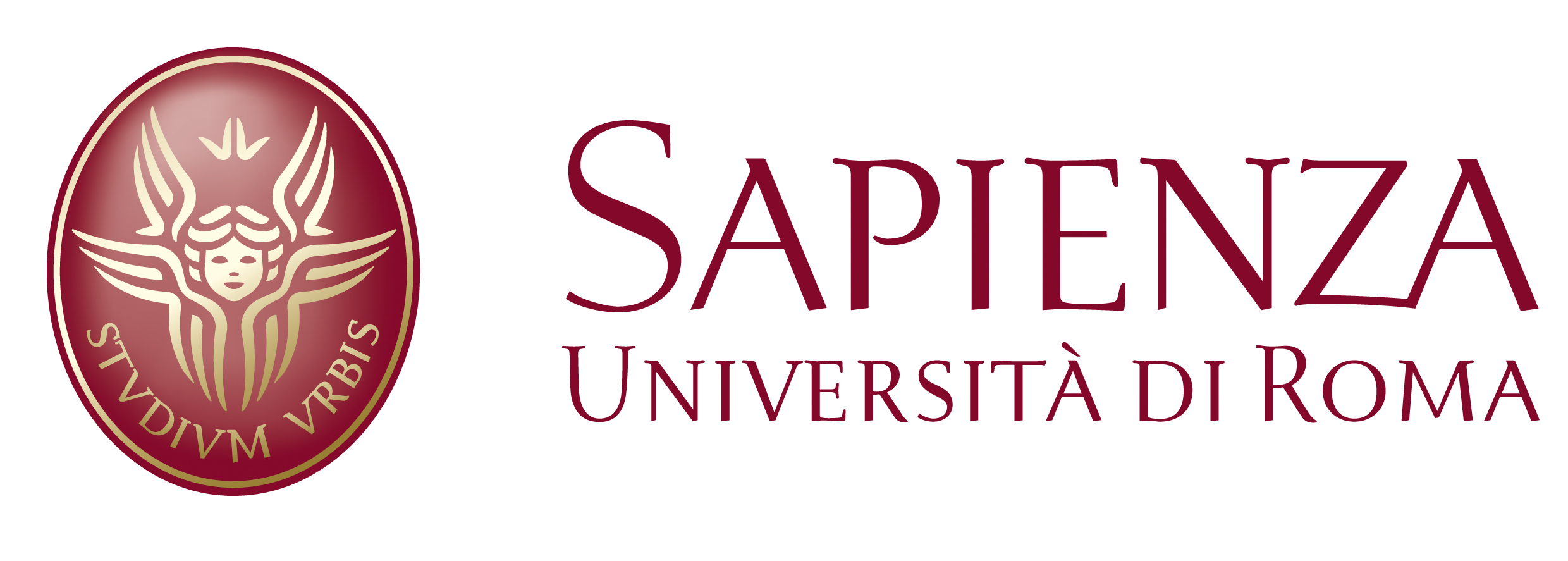}
\Istituzione{Sapienza University of Rome}
\Divisione{Department of Computer, Control and Management Engineering}		
\Scuola{PhD in Computer Engineering}
\Titoletto{Thesis For The Degree Of Doctor Of Philosophy}
\Titolo{Explaining Deep Neural Networks by Leveraging Intrinsic Methods}
\Punteggiatura{}					
\NCandidato{Candidate}					
\Preambolo{\renewcommand{\frontsmallfont}[1]{\small}}
\Candidato[]{Biagio La Rosa}
\Piede{Academic Year 2023-2024 (XXXVI cycle)}					
\end{frontespizio}
\IfFileExists{\jobname-frn.pdf}{}{%
\immediate\write18{pdflatex \jobname-frn}}
 
\clearpage




\newpage
\afterpage{\null\thispagestyle{empty}\clearpage} 
\thispagestyle{plain}			
\setlength{\parskip}{0pt plus 1.0pt}
\section*{Abstract}
Deep neural networks have been pivotal in driving AI advancements over the past decade, revolutionizing domains such as gaming, biology, autonomous systems, and voice and text assistants. 
Despite their impact, these networks are often regarded as black-box models due to their intricate structures and the absence of explanations for their decisions.  This opacity poses a significant challenge to AI systems' wider adoption and trustworthiness. This thesis addresses this issue by contributing to the field of eXplainable AI, focusing on enhancing the interpretability of deep neural networks.

The core contributions lie in introducing novel techniques aimed at making these networks more interpretable by leveraging an analysis of their inner workings. Specifically, the contributions are threefold. Firstly, the thesis introduces designs for self-explanatory deep neural networks, such as the integration of external memory for interpretability purposes and the usage of prototype and constraint-based layers across several domains. These proposed architectures are specifically designed to preserve most of the black-box networks, thereby maintaining or improving their performance. Secondly, this research delves into novel investigations on neurons within trained deep neural networks, shedding light on overlooked phenomena related to their activation values. Lastly, the thesis conducts an analysis of the application of explanatory techniques in the field of visual analytics, exploring the maturity of their adoption and the potential of these systems to convey explanations to users effectively.

In summary, this thesis contributes to the growing field of Explainable AI by proposing intrinsic techniques to enhance the interpretability of deep neural networks. By mitigating the opacity issue of deep neural networks and applying them to several different applications, the research aims to foster trust in AI systems and facilitate their wider adoption across several applications.


\thispagestyle{empty}
\mbox{} 
\cleardoublepage

\thispagestyle{plain}			
\section*{Acknowledgements}

I strongly believe that the goals achieved in life need to be shared with all the people you interact with. My life has been full of sliding doors, chaos, dreams, failures, and discoveries, and all of them involved other persons in the process. Therefore, thank you to all the people who met me in their lives, no matter if you like or hate me: each of you pushed me to become who I am.

Special thanks for their contribution to my research path must be given to:
\begin{itemize}
\item Prof. Roberto Capobianco, for supporting my freedom, my research, and guiding my studies. We embarked on this mad and risky journey of explainability together, and I really enjoyed the time under your supervision. Thank you for pushing me to go outside my comfort zone and for all your precious suggestions. The calls during the evening to discuss novel ideas and the long night for the IJCAI paper are examples of memories that I will remember forever! arigatō Roberto!
\item Prof. Leilani Gilpin for welcoming me as a visitor to your lab, for granting me freedom during the months there, and for being so supportive even in tasks where you were not supposed to be. You are a rare case in academia: your way of supervising and the environment that you built around me allowed me to exploit the best of my skills and reach the goals that I wanted to reach. Therefore, thank you for everything!
\item KRL group for our dinners and casual chats! My first year alone as a student and under the pandemic was hard, but thanks to your presence, I discovered what it means to be in a lab group of smart people. 
\item XAI-KRL group for becoming a house where discussing the field, ideas, and opinions and grow together.
\item the AWARE group (Marco, Graziano, and Prof. Santucci) for introducing me to a novel field and opening the doors to so many future directions and collaborations.
\item Prof. Daniele Nardi, for the chance to be in your lab, for your support whenever I needed it, and for being kind and nice to me! I really appreciate everything that you have done in these last three years!
\item Prof. Roberto Navigli, for introducing me to the field of AI and research when I was an undergraduate without any knowledge of AI. Your suggestions and your supervision definitely shaped my future, transforming me from a future software engineer to a future researcher. Your compliments and advice pushed me to surpass myself at each step of my career until today.
\end{itemize}

Thank you, Alessio Ragno, for becoming my French partner in crime in XAI. Your presence and our battles helped me to grow as a supervisor and as a reviewer 2. Because of you, I learned how to choose the best paper title and how to sell ice to a polar bear. Thank you, Michela Proietti, for being patient and kind in this last period. I'm sure you will become a great scientist; just relax and enjoy your journey!

Additionally, I would like to thank all the students that I supervised both for theses and projects: interacting with you allowed me to experiment and improve my skills as a supervisor and become a better researcher. Thank you to all the reviewers2 who rejected my papers and proposals

During this path, I had the chance to collaborate with great colleagues whose contribution and help to this thesis are valuable in multiple aspects. Other than my French partners Alessio and Michela, I thank Romain Giot, Graziano Basili, Sayo M. Makinwa, Enrico Bertini, Marco Angelini, Romain Bourqui, Rino Ragno, David Auber, and Giuseppe Santucci for contributing to the research presented in this thesis. I hope to collaborate again with all of you in the future.

Thank you to all my family for being supportive during my long academic travel. I hope that today you will be proud of me and you can finally relax. Thank you also to my friends for pushing me to prove to them that doing research is ``a job''. Finally, thank you to Roberta for being so patient during the writing process of this thesis and for postponing your drama. Thank you for encouraging me to make difficult decisions even if these decisions will impact your life and future. I really appreciate your help, and I hope we will walk together for a long time.
\cleardoublepage

\begin{singlespace}
 \tableofcontents 	
 \addcontentsline{toc}{chapter}{\listfigurename}
 \listoffigures
 \addcontentsline{toc}{chapter}{\listtablename}
 \listoftables

 \printnomenclature
 \printglossaries
\end{singlespace}

\mainmatter	  
\clearpage
\pagestyle{fancy} 
\renewcommand{\chaptermark}[1]{\markright{\chaptername\ \thechapter.\ #1}{}}
\renewcommand{\sectionmark}[1]{\markright{\thesection.\ #1}}
\lhead{} 
\chead{}                   
\rhead{\slshape \rightmark} 
\lfoot{}
\cfoot{} 
\rfoot{\thepage}          
\renewcommand{\headrulewidth}{0.4pt} 
\renewcommand{\footrulewidth}{0.4pt}

\chapter{Introduction}
\thispagestyle{plain}
\nomenclature{\textbf{a}}{$n$-Norm Distance}
\nomenclature{$\lVert \rVert ^n$}{$n$-Norm Distance}
\nomenclature{$\lVert \rVert$}{Square Distance}
\nomenclature{| |}{Cardinality}
\nomenclature{$abs()$}{Absolute Value}
\nomenclature{$\sigma_{sig}(x)$}{Sigmoid Function}
\nomenclature{$\sigma_{soft}(x)$}{Softmax Function}
\nomenclature{$\sigma_{soft^+}(x)$}{Softplus Function}
\nomenclature{$\sigma_{tanh}(x)$}{Tanh Function}
\nomenclature{$\circ$}{Element-wise Multiplication}
\nomenclature{$\oplus$}{Concatenation}
\nomenclature{$f_{FE}$}{Feature Extractor}
\nomenclature{$f_{CL}$}{Classifier}
\nomenclature{$\mathfrak{D}$}{Dataset}
\nomenclature{$\mathfrak{L}^l$}{Set of logical connections of arity $l$ }

\section{Introduction}
\label{chap:intro}
\gls{AI} is a field of computer science aimed at developing machines capable of solving tasks that typically require human intelligence. The first successful AI approaches were rooted in expert systems relying on rules and symbolic reasoning. However, despite early optimism, symbolic AI has revealed limited adaptation capabilities. These systems often assume perfect knowledge of tasks and disregard uncertainty or ambiguity in data. Consequently, pure symbolic AI struggles to manage complex tasks for which humans cannot describe the rules governing the phenomenon. For instance, tasks like translating text, recognizing images, and exploring unknown environments were deemed impossible to solve.

\gls{ML} emerged as a paradigm shift in AI research to mitigate these problems.
\gls{ML} provides algorithms capable of learning from data and improving performance over time without explicit knowledge of a particular phenomenon's rules. Classical \gls{ML} algorithms use statistical theory for pattern recognition on a set of data. Examples of this category include decision trees, logistic regression, and Support Vector Machines~\cite{Hearst1998}, which were the state-of-the-art tools to deal with complex tasks until recently.

In recent years, the internet's expansion and the availability of cheaper hardware, open-source platforms, and big data have enabled the collection of vast raw datasets. However, classical \gls{ML} struggles to fully exploit these datasets due to their size, complexity, and lack of explicit semantics associated with data. Conversely, \textbf{\glspl{DNN} are explicitly designed to process raw data and memorize large quantities of information} in the interconnections between network layers. Therefore, the field of \gls{DL}, which leverages \glspl{DNN}, has become the primary subfield of \gls{ML}.

A \gls{DNN} consists of thousands or millions of interconnected neurons. Since the work of \citet{Rosenblatt1958}, the design of neurons and \glspl{DNN} has become increasingly complex, involving functions such as non-linear operations, convolution, memorization, attention, and skip connections. The complexity of these designs enables \glspl{DNN} to achieve impressive performance in various tasks, often surpassing human performance. Games~\cite{Silver2017, Mnih2015}, vision ~\cite{Taigman2014, Redmon2016}, robotics ~\cite{Lillicrap2016, Levine2016}, and natural language processing~\cite{Vaswani2017, Brown2020} have all undergone a revolution in their respective field. Nowadays, mainstream applications for speech recognition, machine translation, and text generation are all powered by \gls{DL} systems.

However, the gain in performance comes at the cost of transparency. While symbolic systems are easy to understand in terms of encoded knowledge and decision process, classical \gls{ML} tends to be more opaque in both aspects. Indeed, keeping track of the learning process is challenging, and the explainability of these systems is often limited to extracting the learned decision process. But, even in these cases, there is a trade-off between complexity and transparency. For instance, when using large \gls{ML} models (e.g., a wide and deep decision tree), extracting explanations about the behavior that are easy to understand could be challenging since explanations can be extremely long. The challenge is exacerbated for \glspl{DNN}, where there is no semantic associated with data, and tracking how inputs are transformed into outputs is a nightmare due to the complexity of the interconnections. Therefore, \textbf{\glspl{DNN} are commonly referred to as black-boxes, where one feeds an input and receives an output without understanding the motivations behind the results}.

The field of \gls{XAI} aims to address the need for transparency and interpretability in machine learning and \gls{AI} systems. \gls{XAI} aims to provide insights into the inner workings of AI models, enabling users to understand and interpret their outputs effectively. \gls{XAI} encompasses a broad range of techniques, spanning from methods that highlight the most important parts of the input to those that extract the knowledge learned by machine learning models.

In classical \gls{ML}, \gls{XAI} methods mainly focus on providing concise explanations summarizing the already known rationale behind decisions. Examples of these methods are feature importance analysis~\cite{Strumbelj2013}, which identifies the most relevant features contributing to model predictions, and decision tree visualization~\cite{Teoh2003}, which provides a graphical representation of the decision-making process. 
In the context of \gls{DL}, \gls{XAI} methods deal with an unknown decision process and an unknown learned knowledge and aim at approximating, guessing, or probing the real model behavior. 

The first and most popular techniques in the area are the so-called \emph{extrinsic methods}, which approximate the behavior and generate explanations for \gls{DL} models by exploiting external means. For example, several techniques employ surrogate models~\cite{Lundberg2017, Ribeiro2016}, generative models ~\cite{BarredoArrieta2020}, or perturbation-based analysis~\cite{Fong2017} to approximate the decision process of the networks around a given point. Despite the flexibility and high compatibility with existing models, recent works~\cite{Adebayo2018, Rudin2019} argue against relying solely on extrinsic methods. Indeed, extrinsic methods often struggle to capture the complexity of model behavior of \glspl{DNN}, are biased by the selection of the external means, and require too much time to return a reliable approximation due to the complexity of the tasks managed by \glspl{DNN}.

To address this issue, researchers are starting to explore \emph{intrinsic methods}, which aim to enhance the interpretability of deep models by leveraging the inner workings of the models. This objective can be achieved by modifying the design of \glspl{DNN} to make it more explainable, adjusting their training process to produce explainable representations, or analyzing and connecting the working mechanisms of their components. These methods include techniques such as attention mechanisms~\cite{Bahdanau2014}, which highlight relevant input features, activations analysis~\cite{Bau2017,Mu2020}, and self-explainable DNNs~\cite{Chen2019, Chen2020}.
Intrinsic methods offer the advantage of directly linking interpretability to the model's design. They are generally faster than extrinsic methods and are more faithful to the model behavior. Nonetheless, they are usually tailored to specific settings (architectures, training processes, etc.) and, in the case of self-explainable DNNs, can incur performance trade-offs, thereby limiting the adoption from the general audience.

\textbf{This thesis contributes to the ongoing research effort on \gls{XAI} intrinsic methods by proposing methods that leverage \gls{DNN} inner workings for explaining \gls{DL}}. In this context, the thesis proposes multiple designs of self-explainable DNNs and a post-training method to investigate the neurons' recognition capabilities. \textbf{The goal of the proposed self-explainable DNNs is to reduce the performance trade-off and broaden the applicability of such methods}. To achieve the first goal, the proposed layers can be inserted into black-box models without disrupting their structure and preserving most of their representation power. The second goal is achieved either by extending approaches to novel domains~\cite{Chen2019, Wang2021tesnet}, treating the preceding layers as black-boxes (i.e., not exploiting specific shapes or structures), and evaluating the proposed techniques across several architectures. As a side product, the thesis also increases the diversity of approaches in self-explainable DNNs since it introduces a new family of architectures: memory-based self-explainable DNNs.

Similarly, \textbf{the proposed post-training method} shares the underlying principle of compatibility and reliance on \glspl{DNN} inner workings (i.e., activations) and \textbf{advances our knowledge of the semantics encoded in neurons}. Namely, it enables the investigation of broader settings than those explored in literature~\cite{Mu2020, Bau2017}, shedding light on novel phenomena related to the neurons' activation spectrum.

Finally, the thesis delves into the ongoing discourse on rendering the explanations usable and beneficial for users.
Recently, several interactive systems have emerged to connect users, \gls{DL} systems, and explanations by exploiting interactive interfaces~\cite{Wexler2019, Tenney2020} and dialogue systems powered by large language models~\cite{Slack2023}. This thesis contributes to this research by surveying the combination of XAI techniques and \gls{VA} systems. 
 \textbf{By surveying existing methodologies and advocating for the integration of XAI techniques into \gls{VA} systems, the thesis aims to increase the awareness of the \gls{XAI} and \gls{VA} communities of each other and foster a new alternative direction for interactive explanations.}

\section{Contributions}
The main contributions of this thesis to the field of \gls{XAI} are the following:
\begin{itemize}
    \item the introduction of a \textbf{novel prototype-based layer for \glspl{GNN}}, which generalizes prototype-based self-explainable \glspl{DNN}~\cite{Chen2019} in terms of domains, architectures, and assumptions. Differently from the current research, the layer represents prototypes as node embeddings, allowing its application to both graph and node classification tasks. Furthermore, our contribution includes a sparse explanation visualization for graphs that preserves the faithfulness of explanations while enhancing their understandability compared to commonly used visualization methods for prototype-based networks.
    
    
    \item the proposal of \textbf{novel memory modules designed to enhance the interpretability of existing neural networks while preserving their performance}.
    The utilization of memory modules for interpretability purposes represents a novelty in current literature. By operating within the same training settings as black-box models, these memory modules address some of the issues associated with the use of self-explainable DNNs. Additionally, these designs are flexible and can facilitate the retrieval of various types of explanations.

    \item the \textbf{application and adaptation of concept-whitening}~\cite{Chen2020} to the chemical domain and drug discovery tasks to ensure that latent representations encode semantics related to molecular properties.
   Our contribution includes the generalization of the method~\cite{Chen2020} to the graph and chemical domain, thereby expanding its applicability beyond the vision domain. The generalization includes the type of data, the covered architectural families, and the type of normalization layers, an area not explored in the literature. 
    
    \item the \textbf{design of a novel algorithm for computing compositional explanations of neurons behavior}. This algorithm enables explanations for a broader spectrum of neurons behavior than the one investigated in the literature and overcomes issues related to the computational complexity of previous approaches~\cite{Mu2020}. Our contributions include the discovery and discussion of novel phenomena related to neuron activations, as well as the design of novel metrics for measuring the properties of explanation methods for latent representation.

    \item a \textbf{review and analysis of \gls{VA} systems as a means to connect users and explanations} in an interactive environment without requiring XAI experts during the usage. By focusing on systems dealing with \glspl{DNN} and employing \gls{XAI} methods, the thesis contributes to identifying strengths and weaknesses of the proposed VA solutions, laying the foundation for future enhancements in the integration and collaborations between VA and XAI research.
\end{itemize}

\section{Thesis Organization}
The thesis is organized into five parts: \nameref{part:preliminaries}, \nameref{part:senn}, \nameref{part:neurons}, \nameref{part:va}, and \nameref{part:conclusion}. Each part is organized as follows:
\begin{itemize}
    \item \textbf{\Cref{part:preliminaries} \nameref{part:preliminaries}} provides the necessary background knowledge and terminology to comprehend the chapters of this thesis and reviews the related work on explainable AI applied to \glspl{DNN}. 
    \begin{itemize}
        \item \textbf{\Cref{chapter:background}} introduces the terminology and key concepts related to \glspl{DNN}, \gls{XAI}, and \gls{VA}; describes the families of deep architectures examined in this thesis; introduces the categorization of explanation methods; discusses the evaluation challenges in \gls{XAI}; and introduces the metrics used to evaluate the proposed methods.
        
        \item \textbf{\Cref{chapter:related}} reviews the literature on post-hoc \gls{XAI} methods for explaining \gls{DL}, focusing on techniques similar to those proposed in this thesis; reviews self-explainable DNNs;  analyzes the limitations of current methods; and discusses the relationships between existing approaches and the thesis techniques.

        \end{itemize}
    \item \textbf{\Cref{part:senn} \nameref{part:senn}} presents modules that can be added to existing neural networks to improve their explainability; presents a prototype-based design for self-explainable DNNs; introduces the novel family of self-explainable DNNs based on memory;
    
    \begin{itemize}
        \item \textbf{\Cref{chapter:pignn}} presents the design of a prototype-based layer for enhancing the interpretability of \glspl{GNN}; discusses the benefits in terms of interpretability; evaluates the performance and the explanations over several architectures; and discusses alternative design choices.
        \item \textbf{\Cref{chapter:senn}} introduces a design to improve the interpretability of recurrent models on sequential data; introduces a memory module to enhance the interpretability of black-box models on image data; introduces the unified mechanism of memory tracking to compute different types of explanations; evaluates the performance and the explanations of the proposed modules; discusses alternative design choices.
        
    \end{itemize}
    \item \textbf{\Cref{part:neurons} \nameref{part:neurons}} presents methods to inspect the knowledge learned by \glspl{DNN}; explore a method to enforce semantics in the latent representation of \glspl{GNN}; presents a technique to extract semantics from neurons activations.
    \begin{itemize}
        \item \textbf{\Cref{chapter:whitening}} presents a normalization for the graph neural networks and graph data domain to enforce latent representation to be aligned to molecular properties in the chemical domain and drug discovery; discusses the benefits in terms of interpretability; evaluates the performance and the explanations over several architectures, datasets, and layers; and discusses alternative design choices.
        \item \textbf{\Cref{chapter:compositional}} introduces a heuristic-guided algorithm to compute the semantics encoded in a wide spectrum of neurons' activations; proposes a set of metrics to describe the properties of explanations returned by algorithms that compute explanations for neurons' activations;  discusses the computational efficiency of the proposed algorithm; evaluates the explanations computed by the algorithm; discusses novel phenomena related to neurons activations; discusses alternative designs.
        
    \end{itemize}
    \item \textbf{\Cref{part:va} \nameref{part:va}} reviews and analyzes \gls{VA} systems that incorporate \gls{XAI} methods to help users understand \glspl{DNN}.
        \begin{itemize}
            \item \textbf{\Cref{chapter:va}} collects, reviews, and analyzes, from a \gls{XAI} perspective, the \gls{VA} systems that use XAI methods to improve the explainability of \glspl{DNN}; analyzes the potential and benefits of interactive systems; discusses limitations and future direction of the field.
        \end{itemize}
    \item \textbf{\Cref{part:conclusion} \nameref{part:conclusion}} summarizes the contributions of the thesis and discusses limitations and future research directions.
    \begin{itemize}
        \item \textbf{\Cref{chapter:conclusion}} summarizes the thesis's contribution; highlights the advantages and limitations of the proposed approaches; and discusses short and long-term future research directions that this thesis opened.
    \end{itemize}
\end{itemize}
\section{Related Publications}
Part of the thesis has been previously published in the following journal and conference articles:
\begin{itemize}
    \item \textit{Towards a fuller understanding of neurons with Clustered Compositional Explanations}. \\
        \textbf{Biagio La Rosa}, Leilani Gilpin, and Roberto Capobianco\\
        In Thirty-seventh Conference on Neural Information Processing Systems (NeurIPS 2023)
    \item \textit{Explainable AI in Drug Discovery: Self-interpretable Graph Neural Network for molecular property prediction using Concept Whitening}.\\
    Michela Proietti, Alessio Ragno, \textbf{Biagio La Rosa}, Rino Ragno, and Roberto Capobianco. \\
    Machine Learning (Journal), pp. 1–32 (2023)
    \item \textit{The State of The Art of Visual Analytics for eXplainable Deep Learning}.\\
\textbf{Biagio La Rosa}, Graziano Blasilli, Romain Bourqui, David Auber, Giuseppe Santucci, Roberto Capobianco, Enrico Bertini, Romain Giot, and Marco Angelini\\
Computer Graphic Forum (Journal)\\
Presented also at 25th EG Conference on Visualization (EuroVIS 2023)
\item \textit{A self-interpretable module for deep image classification on small data}.\\
\textbf{Biagio La Rosa}, Roberto Capobianco and Daniele Nardi.\\
Applied Intelligence (Journal) (2023)
\item \textit{Prototype-based Interpretable Graph Neural Networks}.\\
Alessio Ragno, \textbf{Biagio La Rosa}, and Roberto Capobianco\\
IEEE Transactions on Artificial Intelligence (Journal) (2022)
\item Detection Accuracy for Evaluating Compositional Explanations of Units.\\
Sayo M. Makinwa, \textbf{Biagio La Rosa}, Roberto Capobianco.\\
In Proceedings of AIxIA 2021 - Advances in Artificial Intelligence, pages 550–563.
Springer International Publishing, (2021)
\item \textit{A Discussion about Explainable Inference on Sequential Data via Memory-Tracking}.\\
\textbf{Biagio La Rosa}, Roberto Capobianco and Daniele Nardi.\\
Discussion Papers of AIxIA 2021
\item \textit{Explainable Inference on Sequential Data via Memory-Tracking}.\\
\textbf{Biagio La Rosa}, Roberto Capobianco and Daniele Nardi.\\
In Proceedings of the Twenty-Ninth International Joint Conference on Artificial Intelligence (IJCAI-20) (2020)
\end{itemize}

\part{Preliminaries}
\label{part:preliminaries}

\glslocalunset{ML}
\glslocalunset{AI}
\glslocalunset{DL}
\glslocalunset{DNN}
\glslocalunset{XAI}

\chapter{Background} 
\label{chapter:background}
This chapter serves as a foundational introduction necessary for comprehending the subsequent content of the thesis. It introduces the terminology and concepts used to describe the methods proposed in this thesis. 

The chapter is organized as follows: \Cref{sec:background_DL} introduces terminology and concepts related to \gls{DL} and the architectures used in this thesis; \Cref{sec:back_xai} introduces terminology, metrics, evaluation, and categorizations related to \gls{XAI}; finally, \Cref{sec:back_va} introduces terminology, concepts, and categorizations related to \gls{VA}. 

\section{Fundamentals of Deep Neural Networks}
\label{sec:background_DL}
This section describes the terminology used for the building blocks of a \gls{NN} and \gls{DL} and introduces the families of architectures examined in this thesis.

We begin the description by highlighting the objective of a \gls{NN}: learn how to solve a task based on data observations. A data observation comprises several elements termed \emph{features}. The collection of observations forms the \emph{training dataset}, which is then used to train the \gls{NN} in a process known as \emph{training process}. In supervised learning and classification tasks, the primary paradigms explored in this thesis, each observation is associated with a \emph{ground truth} label. The ground truth denotes the desired output and is used as feedback during the learning process.

 \def\layersep{1.5cm}
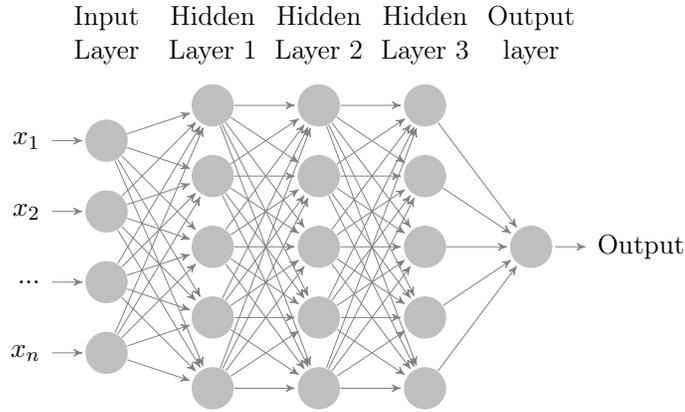
\begin{figure}[t]
\begin{center}
\begin{tikzpicture}[
   shorten >=1pt,->,
   draw=black!50,
    node distance=\layersep,
    every pin edge/.style={<-,shorten <=1pt},
    neuron/.style={circle,fill=black!25,minimum size=17pt,inner sep=0pt},
    input neuron/.style={neuron},
    output neuron/.style={neuron},
    hidden neuron/.style={neuron},
    annot/.style={text width=4em, text centered}
]

        \node[input neuron, pin=left:$x_{1}$ ] (I-1) at (0,-1) {};
	\node[input neuron, pin=left:$x_{2}$ ] (I-2) at (0,-2) {};
	\node[input neuron, pin=left:$...$ ] (I-3) at (0,-3) {};
	\node[input neuron, pin=left:$x_{n}$ ] (I-4) at (0,-4) {};
    \newcommand\Nhidden{3}

    \foreach \N in {1,...,\Nhidden} {
       \foreach \y in {1,...,5} {
          \path[yshift=0.5cm]
              node[hidden neuron] (H\N-\y) at (\N*\layersep,-\y cm) {};
           }
    \node[annot,above of=H\N-1, node distance=1cm] (hl\N) {Hidden Layer \N};
    }

    \node[output neuron,pin={[pin edge={->}]right:Output}, right of=H\Nhidden-3] (O) {};

    \foreach \source in {1,...,4}
        \foreach \dest in {1,...,5}
            \path (I-\source) edge (H1-\dest);

    \foreach [remember=\N as \lastN (initially 1)] \N in {2,...,\Nhidden}
       \foreach \source in {1,...,5}
           \foreach \dest in {1,...,5}
               \path (H\lastN-\source) edge (H\N-\dest);

    \foreach \source in {1,...,5}
        \path (H\Nhidden-\source) edge (O);


    \node[annot,left of=hl1] {Input Layer};
    \node[annot,right of=hl\Nhidden] {Output layer};
\end{tikzpicture}
\end{center}
\caption{An example of a 3-layer neural network. Each node in the hidden layers is 
connected with all the nodes of the previous layer and all the nodes of the next
layer.}
\label{fig:mlp}
\end{figure}
 A \gls{NN} constitutes a hierarchical mathematical model, composed of multiple layers of interconnected artificial neurons, which receives an observation and yields an output referred to as a \emph{prediction}. The layers of a \gls{NN} can be categorized based on their position within the architecture: input, hidden, and output layers (\Cref{fig:mlp}).
The \emph{input layer}, constituting the lowest layer in the hierarchy, feeds input features to the subsequent layers. A \emph{hidden layer} receives the output of neurons from the previous layers and outputs to the next layer. The outputs of a hidden layer collectively form the \emph{latent representation} of the sample. The manifold encompassing all possible latent representations of a given hidden layer is termed the \emph{latent space}. The \emph{output layer}, positioned at the apex of the hierarchy, accepts the output of the preceding layers as input and delivers the prediction. A network featuring at least two hidden layers is termed a \gls{DNN}.

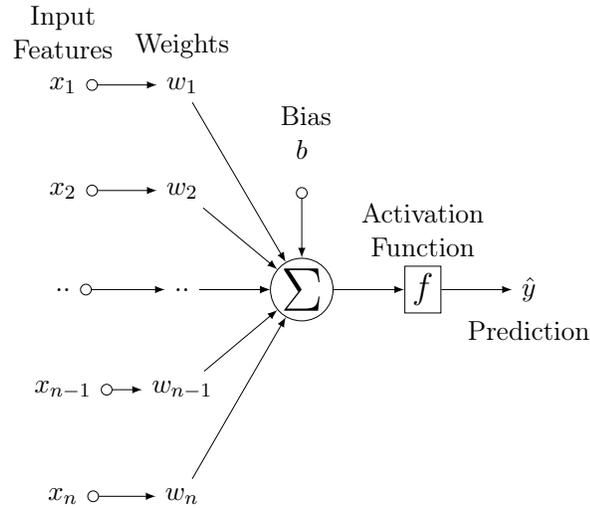
\begin{figure}[t]
\begin{center}
\begin{tikzpicture}[
init/.style={
  draw,
  circle,
  inner sep=0.5pt,
  font=\Huge,
  join = by -latex
},
squa/.style={
  draw,
  inner sep=3pt,
  font=\Large,
  join = by -latex
},
start chain=2,node distance=10mm
]
\begin{scope}[start chain=1]
	\node[on chain=1,label=above:{\parbox{4cm}{\centering Input \\ Features}}]    (x1) {$x_1$};
	\node[on chain=1,label=above:Weights,join=by o-latex]   (w1) {$w_1$};
\end{scope}
\node[on chain=2,below=of x1]  (x2) {$x_{2}$};
\node[on chain=2,join=by o-latex,below=of w1] (w2) {$w_2$};
\begin{scope}[start chain=3]
	\node[on chain=,below=of x2](x3) {$..$};
	\node[on chain=3,join=by o-latex,below=of w2] (w3) {$..$};
\end{scope}
\node[on chain=3,init] (sigma)  {$\displaystyle\Sigma$};
\node[on chain=3,squa,label=above:{\parbox{4cm}{\centering Activation \\ Function}}]    {$f$};
\node[on chain=3,label=below:Prediction,join=by -latex]  {$\hat{y}$};
\begin{scope}[start chain=5]
	\node[on chain=,below=of x3]  (x4) {$x_{n-1}$};
	\node[on chain=5,join=by o-latex,below=of w3] (w4) {$w_{n-1}$};
\end{scope}
\begin{scope}[start chain=7]
	\node[on chain=,below=of x4]  (x5) {$x_n$};
	\node[on chain=7,join=by o-latex,below=of w4] (w5) {$w_n$};
\end{scope}
\node[label=above:\parbox{0.6cm}{\centering Bias \\ $b$},above=of sigma]  (b) {};
\draw[o-latex,segment length=0.5pt] (b) -- (sigma);
\draw[-latex] (w1) -- (sigma);
\draw[-latex] (w2) -- (sigma);
\draw[-latex] (w3) -- (sigma);
\draw[-latex] (w4) -- (sigma);
\draw[-latex] (w5) -- (sigma);
\end{tikzpicture}
\end{center}
\caption{A simplified version of an artificial neuron. }
\label{fig:neuron}
\end{figure}
The neurons across layers are interconnected through edges termed \emph{weights}. Each weight is associated with a value. In its simplest form (\Cref{fig:neuron}), a neuron processes the outputs of neurons from the preceding layer, each multiplied by the corresponding weight connecting one of the previous neurons to the current one. Subsequently, an aggregation function (e.g., summation) combines all inputs. Finally, an activation function is applied to the aggregated value to compute the neuron output. The weight values are adjusted throughout the training process to achieve the desired output~\cite{Rosenblatt1958}.     
The mechanism for updating the weights, known as the \textit{error-correction learning rule}, can be expressed as:
\begin{equation}
w_{i,t+1} = w_{i,t} + \eta[y_{j,d} - y_{j,t}]x_{i,t}
\end{equation}
where $w_{i,t}$ denotes the weight $i$ at iteration $t$, $y_{j,d}$ represents the
desired output for the input $j$ and $y_{j,t}$ denotes the current neuron output. If the prediction is correct, then $y_{d,j} = y_{j,t}$ and the weights remain unaltered. The $\eta \in [0,1]$ parameter denotes the \textit{learning rate} controlling the magnitude of the update: a small value leads to gradual adjustments, averaging
the past inputs; large value facilitates rapid adaptation, albeit with lesser consideration for past errors.
In the case of neurons in hidden layers, where no pre-defined desired output exists (i.e., the ground truth), the backpropagation algorithm is employed to update their weights. This algorithm computes the error for each neuron $j$ as follows:
\begin{equation}
\delta_j^L = \phi'(a_j) \sum_k{w_{jk}}\delta^{L+1}_k
\end{equation}
where the superscript indicates the layer of the neuron, $\delta^{L+1}_k$ represents the error of neuron $k$ in the layer $L + 1$, connected to the current neuron via
weight $w_{jk}$, and $\phi'$ denotes the derivative of the activation function of the
node $j$.
The error is then multiplied by the input received by the neuron to obtain the gradient of the error with respect to the weight
$w_{ji}$:
\begin{equation}
\frac{\partial E_n}{\partial w_{ji}} = \delta_j^Lz_i^{L-1}
\end{equation}
The set of the partial derivatives for all the weights in the network forms the
gradient vector $\nabla E_n$, which is used to update the weights of the network:
\begin{equation}
w^{t+1} =w^t  - \eta \nabla E_n
\end{equation}
These equations serve as a general framework extended by the optimization algorithm called \emph{optimizers} such as  Adam~\cite{Kingma2014},  RMSProp, or AdaGrad~\cite{Duchi2010}. Each optimizer presents its own learning schedule, differing in how weights are updated at each step. For instance, Adam and RMSProp employ distinct learning rates for different parameters and adjust them during the training. The highly interconnected structure and the training process enable each neuron to specialize in detecting particular feature correlations, enabling the entire network to represent any function. 

The training process is divided into \emph{epochs}, where the entire training dataset is fed to the network to update weights. As the dataset may be too large to be stored in memory, it is usually divided into \emph{batches} of $m$ samples progressively fed during the epoch. At each step, the optimizer computes the \emph{loss} (i.e., error) associated with the prediction and updates the weights. Various loss functions can be employed based on the task.

In recent years, numerous pre-trained \glspl{DNN} have been available publicly. These models are trained on large corpora to learn fundamental features shared across several tasks. They can then serve as initial configurations for training models on smaller downstream tasks.  
The idea of training pre-trained models on downstream tasks starts from the observation that \textbf{most \glspl{DNN} can be represented as the composition of two functions, the feature extractor $f_{FE}$ and the classifier $f_{CL}$}:
\begin{align}
    f_{FE}~\mathpunct{:}&~\bm{x} \rightarrow \bm{h}\\
    f_{CL}~\mathpunct{:}&~\bm{h} \rightarrow \bm{y}\\
    f~\mathpunct{:}&~f_{CL}(f_{FE}(\bm{x}))
\end{align}
The feature extractor $f_{FE}$ transforms input $\bm{x}$ from features representation to a latent representation $\bm{h}$ capturing feature relations. The latent representation denotes the output of the last layer of the feature extractor. Then, this representation is fed to the classifier $f_{CL}$, which transforms this representation into a prediction. Therefore, since the common features shared across tasks are likely encoded into the feature extractor, the idea is to preserve the feature extractor's learned knowledge and change the classifier for the downstream task. We can distinguish between two forms of training for novel tasks: \emph{fine-tuning} and \emph{transfer learning}. In the former, the weights of the feature extractor remain \emph{frozen} and are not updated during training, with adjustments confined to the classifier's weights. Conversely, in transfer learning, all the parameters are updated during the new training phase and the feature extractor's weights are used to initialize the network's parameters. 

At the end of the training process, the network's capability is assessed by providing a set of samples unseen during training, referred to as the \emph{testing dataset}. The quality of the learning process is evaluated using a set of metrics that summarizes the network's performance on the testing dataset. This thesis uses two metrics to evaluate the performance of \glspl{DNN}: accuracy and ROC-AUC. 
\emph{Accuracy} is the ratio between the number of correctly predicted samples and the total number of samples in the dataset. It ranges from 0 (there are no correct predictions) to 1 (all predictions are correct). \emph{ROC-AUC} additionally takes into account the ratio between true positive (e.g., prediction is 1 when the ground truth is 1) and false positive (e.g., prediction is 0 when the ground truth is 1) and it is useful when the dataset is unbalanced in the number of samples per class.

\subsection{Architectures}
As mentioned in the previous section, neurons and layers can be interconnected in several ways, and multiple activation functions can be employed to compute the output of neurons. Based on the configuration schema of neurons and layers, we can distinguish different families of architectures. This section briefly describes the main layers and architectures used throughout the thesis: LSTM, attention, memory-augmented neural networks, convolutional neural networks, and graph neural networks.

\subsubsection{LSTM}
\label{sec:lstm}
Long-Short Term Memory (LSTM)~\cite{Hochreiter1997} networks are recurrent neural networks that deal with sequential data. Sequential data is a type of data where data points have a temporal dependency on other data points. A dataset of sequential data is a dataset where points chained by a temporal dependency are collected into a structure called a \emph{sequence}. Each item in the sequence is a \emph{timestep}.

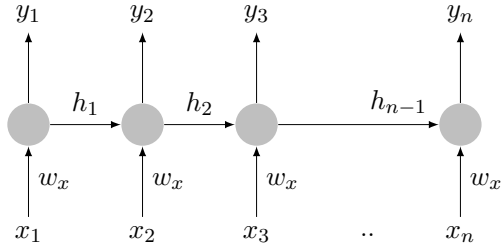
\begin{figure}[t]
\centering
\begin{tikzpicture}[
init/.style={
  draw,
  circle,
  inner sep=0.5pt,
  font=\Huge,
  join = by -latex
},
 neuron/.style={circle,fill=black!25,minimum size=17pt,inner sep=0pt}]
\node[neuron](s1){};
\node[neuron,right=of s1 ](s2){};
\node[neuron,right=of s2 ](s3){};
\node[right=of s3 ](s4){};
\node[neuron,right=of s4](final){};
\node[below=of s1 ](x1){$x_1$};
\node[below=of s2 ](x2){$x_2$};
\node[below=of s3 ](x3){$x_3$};
\node[right=of x3 ](x4){..};
\node[below=of final](xn){$x_n$};
\node[above=of s1 ](y1){$y_1$};
\node[above=of s2 ](y2){$y_2$};
\node[above=of s3 ](y3){$y_3$};
\node[above=of final ](yn){$y_n$};
\draw[-latex] (s1) --node[above](h1){$h_1$}  (s2);
\draw[-latex] (s2) -- node[above](h2){$h_2$}(s3);
\draw[-latex] (s3) -- node[above,near end](h3){$h_{n-1}$}(final);
\draw[-latex] (x1) -- node[right](w1){$w_x$} (s1);
\draw[-latex] (x2) -- node[right](w2){$w_x$}(s2);
\draw[-latex] (x3) -- node[right](w3){$w_x$}(s3);
\draw[-latex] (xn) -- node[right](wn){$w_x$}(final);
\draw[-latex] (s1) -- (y1);
\draw[-latex] (s2) -- (y2);
\draw[-latex] (s3) -- (y3);
\draw[-latex] (final) -- (yn);
\end{tikzpicture}
\caption{An unfolded view of an LSTM cell.}
\label{fig:rnn}
\end{figure}

In an LSTM, each neuron takes one timestep at a time as input and the previous neuron's output, yielding a prediction. Therefore, if the sequence includes $n$ timesteps, the LSTM yields $n$ predictions (\Cref{fig:rnn}).

In LSTMs, each neuron is a \emph{memory cell} with its state $s$. For each timestep, the neuron takes as input the current element $x_t$ in the sequence, the previous output $h_{t-i}$, and
the previous cell state $s_{t-1}$. The purpose of the cell state is to store, update, and carry information through the timesteps. The content of the cell state is managed by two gates: the forget and the input gate. The \emph{forget gate} $g_f$ controls which information to delete from the state; the \emph{input gate} $g_i$ controls which information to write into the state. Finally, a tanh function $\sigma_{tanh}$ computes the information to be stored $\hat{s_t}$ as:
\begin{equation}
    \hat{s_t} = \sigma_{tanh}(\bm{x}_t, h_{t-1})
\end{equation}
The next cell state is computed as the weighted sum between the current state and the information to be stored, weighted by the gates:
 \begin{equation}
 \bm{s}_t = g_f \bm{s}_{t-1} + g_i\hat{\bm{s}_t}
 \end{equation}
Finally, the output is obtained by multiplying the \emph{output gate} $g_o$, which controls the contribution of the cell state, and the candidate output, which is computed by applying the tanh function to the next cell state:
  \begin{equation}
 \bm{y}_t = g_o\sigma_{tanh}(\bm{h}_{t-1},\bm{x}_t)
 \end{equation}
 
The architecture and the memory cell allow LSTM to store information longer than classical recurrent networks, achieving better performance.
\subsubsection{Attention Layers}
Attention~\cite{Bahdanau2015} is a mechanism used to model and exploit long-range correlations between features. This mechanism combines three components: the queries $\bm{Q}$, the keys $\bm{K}$, and the values $\bm{V}$. These components can be either distinct vectors or linear projections of the same input~\cite{Vaswani2017}.
The process begins by comparing each query $\bm{q}$ to the keys and computing a score value for each pair:
\begin{equation}
    e_{\bm{q},\bm{k}_i} = f_{scoring}(\bm{q},\bm{k}_i)
\end{equation}
The score value is computed by a scoring function $f_{scoring}$ that changes based on the type of attention implemented. Then, a softmax function computes the attention weights based on the scores:
\begin{equation}
    w_{\bm{q},\bm{k}_i} = \sigma_{soft}(e_{\bm{q},\bm{k}_i})
\end{equation}
Finally, the layer outputs the weighted sum of the values, weighted by the attention weights:
\begin{equation}
    \bm{y} = attention(\bm{q},\bm{K}, \bm{V}) = \sum_{i}w_{\bm{q},\bm{k}_i}\bm{v}_{k_i}
\end{equation}

Several attention variants have been proposed by changing the scoring functions~\cite{Luong2015, Xu2017}, the projections~\cite{Beltagy2020}, or the applied steps. The attention layers are the fundamental blocks of the Transformer~\cite{Vaswani2017} architecture, which is the backbone of several state-of-the-art models.

\subsubsection{Memory Augmented Neural Networks}
\label{sec:mann}
\glspl{MANN} are a class of neural networks designed to overcome the limitation of classical recurrent architectures like LSTM. Indeed, LSTMs suffer from the vanishing gradient problem. When sequences are particularly long, gradients are multiplied repeatedly during backpropagation through the recurrent connections. This multiplication causes the gradients to get smaller and smaller until they disappear. This problem leads to a limited capability of memorization and exploitation of information from the earlier steps of the sequence. 

\glspl{MANN} mitigate this problem by employing an external memory for storing information for longer periods of time. A \glspl{MANN} is characterized by five elements: the controller, the memory, the write heads, the read heads, and the classifier.
The \emph{controller} is a \gls{DNN} that takes an input and returns an output, usually in the form of a latent representation. If the input is a sequence, then the output of the controller for each step is given by the following equation:
\begin{equation}
    \bm{h}_t = f(\bm{x}_t)
\end{equation}
The \emph{memory} is a matrix $\bm{M}$ of dimensions $N \times M$ that can be written or read by the network. Writing and reading operations are performed by functions called \emph{write heads} and \emph{read heads}. Most of the time, heads are based on attention mechanisms. Both of them usually depend on the output of the controller. In this way, the controller can learn how to use the memory. The output of writing operations is the updated memory:
\begin{equation}
    \bm{M}_{t+1} = wh(\bm{M}_{t}, \bm{h}_t)
\end{equation}
while the output of reading operations are readings $\bm{r}$ whose shape depends on the specific design:
\begin{equation}
    \bm{r}_t = rw(\bm{M}_{t}, \bm{h}_t)
\end{equation}
Finally, the \emph{classifier} yields the prediction by exploiting the readings from the memory and the controller output.
\begin{equation}
    \bm{y}_t = f_{CL}(\bm{h}_t, \bm{r}_t)
\end{equation}

\gls{MANN} designs can include all or parts of the described elements. Only the controller and the memory are common to all the \glspl{MANN}.

\subsubsection{Convolutional Neural Networks}
\label{sec:cnn}
\glspl{CNN} are networks originally designed to deal with image data characterized by the presence of \emph{convolutional layers} and \emph{pooling layers}.
The output of each neuron in a convolutional layer depends on three components: the filter, the receptive field, and the inputs it receives.
The receptive field is the set of neurons connected to the current neuron. While in most \glspl{DNN}, each neuron receives information from all the neurons of the previous layers, in the case of \glspl{CNN}, each neurons receive the outputs of only a subset of neurons in the previous layer. 
The filters are the weights that connect the neuron to its receptive field. In \glspl{CNN}, all the neurons in a layer share the same filters. 
Convolutional layers are designed to exploit spatial information. Indeed, neurons are organized in sequence, grids, or 3d volumes based on the dimension of the input. For 2d data, filters and receptive fields are grids and each neuron can be identified by the indices on the grid.
In mathematical terms, the output of a neuron placed in position $[i,j]$ is associated with a $d_r \times d_r$ receptive field and a filter $d_f \times d_f$ 
\begin{equation}
    h_{i,j}^l = \sigma(\sum_{a=0}^{d_f-1}\sum_{b=0}^{d_f-1} w_{ab}h_{(i+k)(j+b)^{l-1}})
\end{equation}
where $\sigma$ is a non-linear activation.

Generally, convolutional are interleaved with pooling layers, which reduce the dimensionality of the input they receive. The reduction consists of clustering a portion of the input and then combining the elements of each cluster into a single value. For example, a Max-Pooling layer yields the maximum values of each cluster as outputs. The clusters are computed using spatial information, for example, extracting multiple $3 \times 3$ squares of adjacent values in the matrix the layer receives.
Several designs of \gls{CNN} architectures that interleave pooling and convolutional layers have been proposed in the literature. Among them, the thesis include experiments on ResNet~\cite{He2016}, EfficientNet~\cite{Tan2019}, MobileNet~\cite{Sandler2018}, GoogLeNet~\cite{Szegedy2015}, DenseNet~\cite{Huang2017}, ShuffleNet~\cite{Zhang2018}, and WideResNet~\cite{Zagoruyko2016}.

\subsubsection{Graph Neural Networks}
\glspl{GNN} are networks built to work with graph data. Their central element is the message-passing paradigm used in their graph layers, where each node recursively updates its representations by aggregating the ones of its neighbors~\cite{Wu2021}.
How representations are updated and aggregated varies based on the layer's design. We report the aggregation and update functions used by the three architectures we use in this thesis: \emph{\glspl{GCN}}, \emph{\glspl{GAT}}, and  \emph{\glspl{GIN}}.

\glspl{GCN} aggregate latent representation of a node $v$ by averaging the latent representations of its neighbor nodes:
\begin{equation}
    \bm{h}^l = W \cdot (\{\bm{h}^{l-1}_u: u \in N(v)\})
\end{equation}
where $W$ is the weight matrix of the convolution, $N(v)$ is the set of neighbors of the node $v$ and $\bm{h}^{l-1}_u$ is the representation of the neighbor node $u$ in the previous layer. 

\glspl{GAT}~\cite{Velickovic2018} weight the convolutional aggregation via attention scores,
\begin{equation}
    \bm{h}^l = \sum(\{\bm{a}(u,v)\cdot W\cdot\bm{h}^{l-1}_u: u \in N(v)\})
\end{equation}
where $\bm{a}(u,v)$ is the attention weight between the nodes $u$ and $v$ and the other parameters are the same of GCN.

\glspl{GIN}~\cite{Xu2019} use learned parameters $\epsilon$ to weight the aggregation performed by a \gls{MLP}.
\begin{equation}
    \bm{h}^l = MLP((1+ \epsilon){h}^{l-1}_v + \sum_{u \in N(v) {h}^{l-1}_u}
\end{equation}

Node and graph classification are two popular tasks in \glspl{GNN}. In both cases, the input is usually a single graph. In \emph{node classification}, the task is to predict multiple labels, each associated with a node in the graph (e.g., predicting properties of the nodes). In \emph{graph classification}, the task is to predict a single output, usually the class or a general property of the whole graph. In the first case, the output of the layers described above is directly fed to a classifier. In the latter, the output of the layers is first fed to a readout layer, such as global pooling layers, and then to the classifier.

\section{Explainable AI}
\label{sec:back_xai}
By analyzing the architectures presented in the previous section, we can note that the introduced components and advancements between one architecture and another are designed to exploit novel and more complex relations encoded in the data, thereby improving the capability of the networks. However, the combination of these advancements in complexity, the opacity of the training process, and the usage of raw data make the behavior of these architectures opaque. To address this issue, the \gls{XAI} field aims to develop methods that can improve the explainability of artificial intelligence systems.

As in the case of the definition of intelligence, there is no universally accepted definition of explainability~\cite{Miller2019, Rudin2019, Gilpin2018}. In this thesis, we use explainability as a general term to refer to all the methods that \textit{``enable human users to understand, appropriately trust, and effectively manage the emerging generation of AI systems''}~\cite{Gunning2019}. \textbf{We use the terms explainability and interpretability interchangeably} throughout the text.

There are several ways to classify \gls*{XAI} methods proposed in the field. Here, we focus on categories tailored to \glspl{DNN} and this thesis. A first distinction separates \emph{post-hoc} approaches and \emph{self-interpretable \glspl*{DNN}} \cite{Zhang2021,Arrieta2020}: 

\begin{itemize}
\item \textbf{post-hoc} \textit{``methods target models that are not readily interpretable by design by resorting to diverse means to enhance their interpretability''}~\citet{Arrieta2020};
\item \textbf{self-explainable DNNs} are architectures that include components that can be used directly to ease the explanation of the results~\cite{Bahdanau2015,Chen2019, liu2019towards}.
\end{itemize}

A second distinction, introduced in the previous chapter, separates between \emph{intrinsic} and \emph{extrinsic} methods:
\begin{itemize}
    \item \textbf{intrinsic methods} aim to enhance the interpretability of deep models by looking at and leveraging the inner workings of the models; 
    \item \textbf{extrinsic methods} focus on generating explanations for \gls{DL} models by exploiting external means (e.g., gradients, surrogate models).
\end{itemize}

While often self-explainable DNNs are referred to as intrinsic methods and post-hoc as extrinsic ones, we prefer to keep the categories separated, given the proliferation of methods between these categories. 
For example, methods that look at attention patterns~\cite{Abnar2020} in Transformers exploit only the components of the models, and thus, they are intrinsic. However, Transformers cannot be considered self-explainable architectures. Therefore, a method of this kind could be just referred to as a post-hoc intrinsic method. 
 
A further distinction~\cite{Adadi2018} can be done by separating \emph{local} and \emph{global} methods:
\begin{itemize}
    \item \textbf{local} methods provide explanations that are valid for a limited set of data (usually a single prediction), and thus, their explanations do not generalize to other components;
    \item \textbf{global} methods explain a model's whole logic (usually approximating average outcomes).
\end{itemize}
Local methods are more faithful and precise in explaining single predictions but do not offer a guarantee of generalizability. On the other end, global methods are effective in providing an overview of the main learned relations but tend to be less precise and are less reliable in explaining individual components or predictions.

\subsection{Categorization}
\label{sec:metrics}
\gls{XAI} methods can also be split based on the type of explanations they return. While several \gls{XAI} taxonomies have been proposed in the literature to perform such a split, this section reports a simplified categorization~\cite{LaRosa2023survey} by grouping the types of methods discussed in the thesis. Each category is described in terms of the goals of its methods and explanation objects. 

\subsubsection{Feature attribution}
Feature attributions are scores assigned to each input feature or group of features representing the impact of that item on the network's decision process. Methods focusing on feature attributions provide answers to questions like ``Why does the model return this specific output?'' and ``What are the most relevant features exploited by the model to recognize objects in this class?''. Attributions can be either global or local. Global feature attribution measures the importance of features on average. Local feature attribution measures the importance of features exclusively for the current prediction. Scores are usually normalized and visualized as heat maps or numbers.

\subsubsection{Learned Features}
Learned features are features or concepts neurons, groups of neurons, or layers recognize. In the context of this thesis, a \textbf{concept} is intended as a set of semantically connected features annotated in a dataset (e.g., the concept of a wheel in a car). They address the question ``What has this component learned during the training process?'' and thus provide global post-hoc explanations. They can be visualized as samples exhibiting the learned knowledge or by exposing groups of features connected to such knowledge.

\subsubsection{Explanations by Example}
Explanations by examples are samples similar to the item that the user wants to explain and associated with the same prediction. Typically, the samples are extracted from the training dataset and share most features with the object to explain. Inspecting explanations by example helps users extract patterns exploited by the models. Explanations by examples can be applied to extract both global and local information. For example, when applied to the input of the model, explanations by example provide local information, highlighting features exploited by the model for the specific input. Conversely, these explanations provide (global) insights into the learned dynamics when applied to latent representations (e.g., as shown in \Cref{chapter:pignn}). 

\subsubsection{Counterfactuals}
\label{back:counterfactuals}
Counterfactuals are samples representing alternative configurations of input features. A counterfactual is a sample as close as possible to the input but associated with a different prediction. Samples can be either generated or extracted from a dataset. Methods of this category aim to find the minimum magnitude of meaningful edits needed on the current input sample to obtain a different prediction. These explanations are particularly useful for recourse and, most of the time, are local methods.

\subsection{Evaluation of XAI methods}
\label{sec:related_eval}

One of the open questions of the \gls{XAI} field regards the evaluation of explanation methods. Indeed, there is no ground truth as the real inner workings of the model are unknown. We can distinguish between three categories of evaluation: user-study, datasets, and proxy metrics.

\emph{User studies} involve users testing explanations in a real-world task and providing feedback. This form of evaluation was among the first to be explored. We identify three common workflows supporting user studies. The first consists of presenting users with multiple explanations and asking them to rank their preferences, which are then used to measure the quality of the tested methods. The second one consists of providing explanations to the users to assist them while solving a task, followed by collecting their feedback as a qualitative evaluation. Lastly, the third workflow permits users to interact with explanations and tasks at the same time. This task can be facilitated through the usage of interactive interfaces like dashboards, dialogue systems, and visual analytics systems. At the end of the interactive session, both quantitative and qualitative metrics and feedback can be collected. 

Evaluations based on user studies, and particularly the interactive ones, can be used to evaluate the usefulness of the explanations, their impact on specific tasks, and the preference regarding explanation visualization~\cite{DoshiVelez2017}. However, these evaluations are challenging to reproduce, causing problems in the comparison between methods proposed at different times. Moreover, user satisfaction is more influenced by their belief and mental models than the precision of the explanations in capturing the true behavior of the \gls{DNN}. Therefore, involving users should occur only after explanation methods have demonstrated a connection to the real decision process of the network. Otherwise, explanations can easily cause overtrust and mislead the user~\cite{Jin2023}.

In an effort to establish more reproducible environments and link the evaluations to the decision process, \emph{Dataset evaluations} propose to use datasets with available ground truth explanations~\cite{liu2021synthetic} to evaluate explanations. These evaluations draw inspiration from the common evaluation practices applied in machine learning and discussed in \Cref{sec:background_DL}. However, this form of evaluation is feasible only for a subset of methods and simplified tasks where the rules governing the task are known and there is only one unique way to solve the task. 

The third choice involves using a \emph{mathematical proxy} to compare explanation methods quantitatively. They are not intended as replacements for user evaluations but as a necessary prerequisite for selecting the appropriate set of explanation methods. Although several proxy metrics have been proposed to measure the quality of the explanations quantitatively in recent years, there is no standard global metric yet, and each explanation type and context has its own set of metrics.

\textbf{This thesis mainly uses proxy metrics to evaluate the proposed methods}. The metrics are chosen among each category's most recent and popular ones. When available (\Cref{chapter:senn}), we support the evaluation on toy datasets where ground truth can be retrieved and the models satisfy the appropriate requirements (e.g., perfect accuracy). We also discuss the interactive systems supporting user studies in \Cref{chapter:va}. The following paragraphs list the metrics used to evaluate the quality of the explanations for each category presented in the previous section.

\subsubsection{Feature Attributions.}

\begin{definition}
    The $\bm{fidelity+^{prob}}$~\cite{Yuan2022} metric measures the difference in probability of the original prediction and the prediction generated when the most relevant features are masked.
\end{definition}
\begin{equation}
    Fidelity+^{prob} = \frac{1}{N}{\sum_{i=1}^N(f(x_i)- f(\theta_1(x_i)))}
    \label{metric:fidelity_prob}
\end{equation}
where $\theta_1(x)$ is a function that mask $x$ by removing the most important features. Higher is better. This metric is equivalent to the deletion score proposed in \cite{Petsiuk2018rise}.

\begin{definition}
    The $\bm{fidelity+^{acc}}$~\cite{Yuan2022} metric measures the difference in the accuracy of the original prediction and the prediction generated when the most relevant features are masked.
\end{definition}
\begin{equation}
        Fidelity+^{acc} = \frac{1}{N}{\sum_{i=1}^N(\argmax{f(x_i)} = \argmax{f(\theta_1(x_i)}))}
        \label{metric:fidelity_acc}
    \end{equation}
    where $\theta_1(x)$ is a function that mask $x$ by removing the most important features. Higher is better. Fidelity can be also computed by using the ROC-AUC instead of the accuracy: in this case, the metric is denoted as $Fidelity+^{ROC-AUC}$.
    
\begin{definition}
     The $\bm{fidelity-^{acc}}$~\cite{Yuan2022} metric measures the difference in accuracy between the original prediction and the prediction generated when the least relevant features are masked.
\end{definition}
\begin{equation}
    Fidelity-^{acc} =\frac{1}{N}{\sum_{i=1}^N(\argmax{f(x_i)} = \argmax{f(\theta_2(x_i)}))}
    \label{metric:fidelity_neg_acc}
\end{equation}
where $\theta_2(x)$ is a function that mask $x$ by removing the least important features. Higher is better.

\subsubsection{Learned Features.}

\begin{definition}
    The \textbf{intra-concept similarity}~\cite{Chen2020} metric measures the mean pairwise cosine similarity between samples of the same concept.
\end{definition}
\begin{equation}
    d_{ii} = \frac{1}{n^2} \: \left( \sum\limits_{j=1}^{n} \sum\limits_{k=1}^{n} \: \frac{\textrm{\textbf{x}}_{ij} \cdot \textrm{\textbf{x}}_{ik}}{
    \lVert \textrm{\textbf{x}}_{ij} \rVert \; \lVert \textrm{\textbf{x}}_{ik}\rVert} \right)
\end{equation}
where $j$ is the index of the sample, $i$ is the index of the concept, $\bm{x}_{ij}$ is the latent representation for the sample $j$ of the concept $i$, and $n$ is the number of samples including the concept $i$.

\begin{definition}
    The \textbf{inter-concept similarity}~\cite{Chen2020} metric measures the mean pairwise cosine similarity between samples of two different concepts.
\end{definition}
\begin{equation}
    d_{pq} = \frac{1}{nm} \: \left( \sum\limits_{j=1}^{n} \sum\limits_{k=1}^{m} \: \frac{\textrm{\textbf{x}}_{pj} \cdot \textrm{\textbf{x}}_{qk}}{\lVert\textrm{\textbf{x}}_{pj}\rVert \; \lVert\textrm{\textbf{x}}_{qk}\rVert} \right)
\end{equation}
where $p$ and $q$ are the index of the two concepts, and $m$ and $n$ are the number of samples for concept $p$ and $q$, respectively.

\begin{definition}
    The \textbf{separability}~\cite{Chen2020} metric measures the separability of two concepts in a latent space. It is expressed as the ratio between intra and inter-concept similarities.
\end{definition}
\begin{equation}
    Sep_{ij} = \frac{d_{ij}}{\sqrt{d_{ii}d_{jj}}}
    \label{metric:separability}
    \end{equation}
A lower score is considered better.

\begin{definition}
   The \textbf{intersection over union (IoU)}~\cite{Bau2017} metric measures the alignment between a concept annotation and the firing rate of a latent representation.
\end{definition}
\begin{equation}
        IoU(L, \mathfrak{D} ) = \frac{\sum_{x \in \mathfrak{D}}|M^k(x,i) \cap C(x,c)|}{\sum_{x \in \mathfrak{D}}|M^k(x) \cup C(x,c)|}
        \label{metric:iou}
\end{equation}
 where $k$ represents the index of a neuron, $c$  represents a concept, $M^k(x,i)$ is a function 
 that returns the binary mask that indicates the parts of the input on which the neuron fires, and $C(x,c)$ corresponds to a function that returns a binary mask that represents which parts of the input are associated with the concept $c$.

\subsubsection{Explanations by Examples.}

\begin{definition}
    The \textbf{input non-representativeness}~\cite{Kenny2021} metric measures the L\textsubscript{1} distance in the logits obtained by feeding the current input and the explanation by example to the model. 
\end{definition}
\begin{equation}
    Inr = L^1(f(x), f(x_e))
    \label{eq:input_non_representativeness}
\end{equation}
A lower input non-representativeness means the explanation is of better quality.

\begin{definition}
    The \textbf{prediction non-representativeness}~\cite{Nguyen2020} metric measures the cross entropy loss between the current predicted class and the class predicted by using the explanation by example as input.
\end{definition}
\begin{equation}
    Pnr = CrossEntropy(y_x, y_{x_e})
    \label{eq:prediction_non_representativeness}
\end{equation}
 A lower prediction non-representativeness means the explanation is of better quality.

\subsubsection{Counterfactuals.}

\begin{definition}
    Given a predicted class $y_x$, a counterfactual class $y_c$, an autoencoder $f_{y_x}$ trained on reconstructing samples of the predicted class, and an autoencoder $f_{y_c}$ trained on reconstructing samples of the counterfactual class, the \textbf{IM1}~\cite{Looveren2021} metric measures the ratio between errors of the two autoencoders to reconstruct the counterfactual.  
\end{definition}
\begin{equation}
    \frac{\lVert x_{cf} - AE_{y_c}(x_{cf})\rVert } {\lVert x_{cf} - AE_{y_x}(x_{cf})\rVert } 
    \label{eq:IM1}
\end{equation}
 A low IM1 score means the counterfactual is closer to the counterfactual class data distribution than the input one and it is of a better quality. This is useful for methods that start from the input and perturb it to generate a counterfactual.

\begin{definition}
Given a predicted class $y_x$, a counterfactual class $y_c$, an autoencoder $f_{y_x}$ trained on reconstructing samples of the predicted class, and an autoencoder $f_{y_c}$ trained on reconstructing samples of the counterfactual class, \textbf{IIM1 score}~\cite{Looveren2021} measures the ratio between errors of the two autoencoders to reconstruct the counterfactual.     
\end{definition}
\begin{equation}
    \frac{\lVert x_{cf} - AE_{y_x}(x_{cf})\rVert }  {\lVert x_{cf} - AE_{y_c}(x_{cf})\rVert }   
    \label{eq:IIM1}
\end{equation}
 IIM1 metric~\cite{LaRosa2022} is the inverted version of the IM1 score. A low score means that the counterfactual is closer to the input's class data distribution than the counterfactual's. This is useful for methods that select counterfactuals from a pool of samples associated with different predictions.

\begin{definition}
 Given a predicted class $y_x$, an autoencoder $f_{y_x}$ trained on reconstructing samples of the predicted class, and an autoencoder $f_{all}$ trained on reconstructing samples of all the classes in the dataset, the    
 \textbf{IM2}~\cite{Looveren2021} metric measures the difference in the reconstruction errors of the counterfactual between the autoencoder trained on the input classes and the one trained on all the classes. 
\end{definition}
    \begin{equation}
        \frac{\lVert AE_{y_x}(x_{cf}) - AE_{all}(x_{cf})\rVert }{\lVert x_{cf}\rVert_1}
        \label{eq:IM2}
    \end{equation}
 A low IM2 score means the data distribution of the counterfactual class describes the counterfactuals as good as the distribution over all classes and the counterfactual can be considered interpretable. Historically, it is a score proposed to evaluate the usefulness of a generated counterfactual.

\section{Visual Analytics}
\label{sec:back_va}

The \gls{XAI} field is not the sole field that is actively working on supporting users to understand \glspl{DNN}. Indeed, according to \citet{Choo2018}, \gls{VA} systems are playing ``\textit{a critical role in enhancing the interpretability of deep learning models, and it is emerging as a promising research field}''. This section lists the main components used in the \Cref{chapter:va} to analyze the \gls{VA} systems aiming to help users understand \glspl{DNN} through \gls{XAI} methods.

We begin by enunciating the definition of \gls{VA}: it is the science of analytical reasoning supported by interactive visual interfaces~\cite{Thomas2006}. A \gls*{VA} system helps users synthesize and derive insights from data by detecting the expected relationships, discovering the unexpected ones, and communicating these findings to the human user for further actions~\cite{Kohlhammer2011}. 

The \gls*{VA} process~\cite{Keim2008, Sacha2014} can be divided into four components:
\begin{itemize}
    \item \textbf{data} are the starting point of the system and, in the context of this thesis, correspond to inputs, outputs, activations, explanations, etc.; 
    \item \textbf{model} applies transformations to the data and, in the context of this thesis, corresponds to the \gls{DNN}; 
    \item \textbf{visualization} is the interface between users and the system that allows the detection and discovery of relationships and insights on the data and model;
    \item \textbf{knowledge} is the user-driven component and ``\textit{consists in finding evidence for existing assumptions or learning new knowledge about the problem domain}''~\citet{LaRosa2023survey}
\end{itemize}

The user interacts with the \gls{VA} system through interfaces and interactions~\cite{Tominski20IVDA}. Interactions allow the user to analyze data, generate new visualizations, and steer the analyzed model~\cite{Endert2012, Hografer2022}. Interactions represent the means by which the users can achieve the desired goal and help to create a better mental model of the investigated problem. Given a trained model loaded in a \gls{VA} system, we can distinguish between three types of interactions~\cite{LaRosa2023survey}:
\begin{itemize}
\item \textbf{passive observations} involve only the navigation across data (samples, layers, explanations, etc.) in terms of filtering and selection;

\item \textbf{interactive input observations} involve the modification or creation of novel inputs and feeding them to the model to evaluate changes in the models' response;

\item \textbf{interactive model observations} involve the modification of the deep learning model components to evaluate changes in the models' response;
\end{itemize}

Finally, \gls{VA} systems are tailored to specific target users and tasks. \citet{Strobelt2018} propose the following classification for target users:
\begin{itemize}
\item \textbf{architects} are advanced \gls{DL} experts whose goal is developing new architectures or modifying the design of the current ones;

\item \textbf{trainers} are users (with sufficient background knowledge in \gls{DL}) whose goal is to apply existing architectures to new domains, tasks, and datasets; 

\item \textbf{end-users} have limited or no \gls{DL} knowledge and use already trained models in their specific application domain.
\end{itemize}

Note that end-users cannot use a system built for the other categories, while all the other categories of users can use systems designed for end-users.

\chapter{Related Work}
\label{chapter:related}

This chapter reviews existing methods for explaining \glspl{DNN}. 
By utilizing the categorization introduced in \Cref{sec:back_xai}, this chapter discusses post-hoc methods and self-explainable DNNs. Since the chapter cannot provide an exhaustive overview of the state-of-the-art methods for the whole XAI field, the review of post-hoc methods focuses attention on methods that have already been applied to \glspl{DNN}, are data agnostic, and related to the approaches proposed in the next chapters. The chapter is organized as follows: \Cref{sec:related_post} reviews post-hoc methods for deep learning, \Cref{sec:related_senn} reviews current self-explainable DNNs, and \cref{sec:related_related} discusses the relation between the reviewed approaches and the methods proposed throughout the thesis.

\section{Post-Hoc Methods}
\label{sec:related_post}

This section briefly describes the state of the art of post-hoc methods for the categories described in \Cref{sec:back_xai}. It focuses only on methods tailored for \glspl{DNN} or explicitly tested on them. For a deeper description of \gls{XAI} methods in machine learning, please refer to the recent surveys on the topic~\cite{Gilpin2018, Adadi2018, Guidotti2022}.

\subsection{Feature Attribution}
Feature attribution methods assign scores to every input feature based on their relevance to a set of predictions. It is one of the oldest and most extensively explored research areas in the \gls{XAI} field.

\citet{Gilpin2018} distinguish between methods based on perturbations, backpropagations, gradients, and surrogate models.
The concept behind perturbation-based methods is straightforward: modify the value of a feature (or a set of features), input the perturbed instance into the model, and collect the results. Scores are computed by iteratively repeating this process and measuring the deviation between the original prediction and the predictions of perturbed instances~\cite{Zeiler2014, zintgraf2017, Grau2023}. An optimization process can be used to guide the generation of meaningful perturbations~\cite{Fong2017, Ying2019}. 

Approaches utilizing surrogate models, such as LIME~\cite{Ribeiro2016} and SHAP~\cite{Lundberg2017}, also follow the idea of using perturbations. However, in this case, perturbations are not directly used to estimate attributions but serve as samples for training a local interpretable surrogate model. The surrogate model learns to mimic the black-box model's decision process in the samples' neighborhood to be explained. The weights learned by the surrogate model are then used as feature attribute scores. However, these approaches are slow, especially with a large number of features, as in the case of \glspl{DNN}. Additionally, due to the nonlinear nature of DNNs, the result is influenced by the number of features removed altogether at each iteration and the number of permutations.

Gradient-based methods offer a faster alternative to perturbation-based methods by considering the partial derivative of the target output as scores for feature attributions~\cite{Simonyan2014}. However, vanilla gradients produce noisy explanations. To address this problem, several works propose enhancements such as multiplicative terms to the gradients \cite{Shrikumar2017, Springenberg2015, Selvaraju2020, Chattopadhay2018}, the use of integrals and baselines~\cite{Sundararajan2017, Xu2020, Kapishnikov2021}, or smoothing functions~\cite{Smilkov2016, Bykov2022}. While popular for their simplicity, these methods tend to be loosely linked to the decision process~\cite{Adebayo2018} and suffer from the gradient-shattering effect.

Backpropagation methods~\cite{Zeiler2014, Bach2015,Shrikumar2017} and forward propagation methods~\cite{Grau2024} work similarly to gradient-based methods but employ custom rules and quantities instead of gradients. For instance, LRP~\cite{Bach2015} computes relevance for each neuron of the network during input parsing and then backpropagates the relevance and prediction to compute attribution scores. DeepLIFT~\cite{Shrikumar2017} extends this idea by considering baseline inputs as proposed in gradient-based methods.

In parallel with these approaches, there has been recent interest in computing scores for sets of semantically connected features (i.e., \emph{concepts}) rather than individual features. Particularly, TCAV~\cite{Kim2018} and its derivatives~\cite{ghorbani2019towards, Fang2020} investigate the impact of concepts on the decision process. These methods collect samples with and without the target concept, build a hyperplane to separate these samples, and use directional derivatives with respect to this hyperplane as attribution scores. The sensitivity to the collection of samples in terms of diversity and number of samples and the computational time required to probe for several concepts represent the major limitations of these approaches.

Finally, all the methods described thus far explain individual predictions. To obtain global explanations, researchers propose various methods that aggregate attribution scores of individual samples, either by averaging or clustering them~\cite{Lapuschkin2019, Ramamurthy2020,Salman2020}.

\subsection{Learned Features}

This category of methods focuses on extracting information regarding what individual components (e.g., layers, neurons) of neural networks have learned during the training process. Research in this area has explored two main directions~\cite{Casper2023, Sajjad2022, Gilpin2018}: feature synthesis and feature probing using external datasets.

Features synthesis involves generating synthetic explanations through iterative processes~\cite{Erhan2009, Olah2017} or using external models~\cite{Nguyen2016nips, Nguyen2017}. These processes are typically tailored to maximize (or minimize) an objective function related to the component to be explained. For instance, in the case of neurons, several approaches aim to generate synthetic inputs that maximize the activation of a specific neuron~\cite{Olah2017, Mahendran2015, Erhan2009, Simonyan2013}. The general process consists of generating this synthetic input by iteratively altering the features of a starting random input. However, vanilla maximization of the activation can yield abstract explanations that are difficult to interpret. To address this issue, several regularizers have been proposed to increase variance~\cite{Mahendran2015} or impose constraints on the generation process~\cite{Wang2018features,Nguyen2017, Olah2017, Fel2023, Yosinski2015}. 

Nonetheless, these methods encounter several challenges. For example, the stochastic nature of the process~\cite{Mordvintsev2018} may produce different explanations for the same neuron. Additionally, while neurons can recognize multiple features due to superposition~\cite{ bricken2023monosemanticity, elhage2022superposition}, these methods tend to converge towards one or a few concepts, overlooking the multifaceted nature of these components~\cite{Nguyen2016feat}. Furthermore, despite their popularity and the advancements to reduce abstractness,  humans often struggle to comprehend these explanations and instead prefer feature probing explanations~\cite{Borowski2021, Zimmermann2021}. 

Feature probing methods employ real samples to represent the learned features. In this category, the natural counterpart of activation maximization methods is to select samples from the dataset that maximize the neuron's activation~\cite{Casper2023}. In contrast to the activation maximization method, the selected samples are not abstract and naturally exhibit diversity. However, the connection between high activations and the explanation is weaker. Indeed, it is often unclear if the cause of the high activation is the whole sample or specific elements within it. While providing more samples can mitigate this issue, an excessive number of samples can diminish the usefulness of the explanation. 

\citet{Bau2017} propose addressing this issue using a concept-annotated dataset. In this case, the algorithm selects the concept that maximally activates a neuron, removing factors unrelated to the high activation. Similar works reverse the process by fixing a concept and searching for the neurons associated with the highest activations when the concept is included in the samples \cite{Dalvi2019, Durrani2020, Hennigen2020, Antverg2022, Na2019}. Like the generative approaches, associating a single concept with a neuron is inadequate due to superposition. Therefore, successive works \cite{Mu2020, Harth2022, Massidda2023, bykov2023labeling} propose associating logical formulas of concepts with neurons by finding the formula whose annotations are the most aligned with the highest activations through search algorithms.  

One weakness of feature probing approaches is their dependency on concept-annotated datasets. Indeed, the same neuron can be associated with two different explanations simply by changing the annotated dataset~\cite{Ramaswamy2022}. Moreover, the analysis of learned features is constrained by the annotated concepts and the quality of annotations. Recent works suggest replacing the concept-annotated dataset with captioning or multimodal models ~\cite{Hernandez2022, Oikarinen2023} to mitigate this dependency. However, these models require human annotations during training and are susceptible to the domain adaptation problem, thus shifting the dependency from dataset to model aspects.

\subsection{Explanations by examples}
Explanations by examples are samples similar to the current input and associated with the same prediction. Therefore, one of the first explored research directions was providing samples that approximate data prediction distribution as explanations. These approaches~\cite{Kim2016, Gurumoorthy2019} rely on clustering the data, identifying prototypes for each cluster, and using them as explanations. While effective for small tabular datasets, this approach falls short in providing descriptive prototypes for clusters in large datasets.

Conversely, post hoc methods for explanations by examples dealing with \glspl{DNN} and large datasets typically employ case-based reasoning approaches. These approaches utilize a proxy model, such as a K-NN classifier~\cite{Cover1967}, to explain the black-box by learning a mapping between them. Approaches of this category differ in the information provided to the proxy. The simplest approach is to directly use the training samples to train the proxy. However, since typically that not all features are equally important for a given model, several approaches propose weighting the features to yield more informative explanations~\cite{Park2004, Nugent2005}. For example,  a popular technique involves weighting each feature in the sample using feature attribution scores, thereby linking the explanations to the important part of the input~\cite{Adhikari2019, Kenny2019, Kenny2021}. Lastly, another viable approach is to use the activations of the last layers of a \gls{DNN} as samples to train the proxy~\cite{Papernot2018} or a meaningful subset of them \cite{Kenny2023features}.



\subsection{Counterfactuals}
Counterfactuals are samples similar to the current input but associated with a different prediction.
Finding counterfactuals for \glspl{DNN} dealing with non-tabular data is challenging. Indeed, \glspl{DNN} work with raw data, where there are no formal constraints on the position and values of the features, and the number of features involved is large. Therefore, it is challenging to generate plausible counterfactuals. Moreover, the decision process of these networks is extremely sensitive, and it is often possible to obtain a different prediction using adversarial attacks that change the value of a few features in a way that is not discernible to a human. While these modifications align with the definition of counterfactuals (\Cref{back:counterfactuals}), they are often meaningless and cannot be considered explanations. Conversely, counterfactual explanation methods aim to provide counterfactuals that are both plausible and meaningful.

The pioneer work of this category is the method proposed by \citet{Wachter2017}. This method generates counterfactuals by an iterative perturbation process guided by a loss function that minimizes the difference between the predictions on the perturbed instance, the desired outcome, and the L1 norm of the perturbations. Successive works propose alternative losses that take into account additional factors like the closeness of features~\cite{Laugel2018, Dhurandhar2018}, plausibility~\cite{Yang2020}, or the distance from a set of prototypes~\cite{Looveren2021}. One of the drawbacks of these methods is their latency due to the iterative process. One possible solution to mitigate such an issue is to use generative models~\cite{Liu2019counterfactuals, BarredoArrieta2020, Kenny2021counter} or genetic algorithms~\cite{Sharma2020}. However, since these procedures are black-boxes themselves, it is difficult to understand why a particular counterfactual has been selected as a good candidate. 

Another possibility is to remove features from the current input until the prediction is flipped ~\cite{Vermeire2022, Ramon2020} by guiding the process with feature attribution methods. However, feature attribution methods focus on the feature important for a specific instance and do not detect discriminative features that the model uses to discriminate among classes, limiting the applicability of these approaches. Connected to this line of research, some works aim at extracting the contrastive features~\cite{Wang2020counter, goyal19, Jung2022}, which are highly discriminant features for a class and uninformative for the others~\cite{Guidotti2022}. These methods can be considered complementary to methods that select counterfactuals from a dataset based on user-specified properties and classes~\cite{Poyiadzi2020}.

\section{Self-Explainable Deep Neural Networks}
\label{sec:related_senn}
While post-hoc methods represent the most popular tool for explaining \glspl{DNN}, recently, the field has observed a rising interest in the so-called self-explainable DNNs. These networks return explanations alongside their predictions or provide designs that can be easily inspected to provide explanations. The advantage of these methods is that explanations are fast to compute and are directly linked to the decision process of the networks. Moreover, they represent the natural next step for the progress in machine learning. Self-explainable DNNs can be divided into three categories: prototype-based, constraints-based, and attention-based.

\textbf{Prototype-based networks} have been introduced by \citet{Snell2017} to deal with few-shot classification. In their networks, prototypes are computed as the average of the learned embedding of a set of data points. However, since the prototypes are the average of multiple embeddings, they are not interpretable. The concept of prototypes has been merged with the concept of self-interpretable neural networks~\cite{Melis2018} in ProtoPNet~\cite{Chen2019}. In this case, the network learns a set of prototypes representing a \emph{part} of the input (i.e., a set of features semantically linked). Given an input, the network compares the input's latent representation against the learned prototypes and computes the prediction based on the similarity between the prototypes and the input representation. While in the networks proposed by \citet{Snell2017}, prototypes are not associated with semantic meaning, ProtoPNet enforces semantics by using losses and a projection phase that projects the prototypes to real training points. ProtoPNet has been recently extended to enforce diversity~\cite{Wang2021tesnet},  a better prototype organization~\cite{Hase2019,Rymarczyk2021, Rymarczyk2022, Nauta2021}, and semantic corrispondence~\cite{Nauta2023} of prototypes.  Networks inspired by the ProtoPNet design have also been proposed in reinforcement learning~\cite{Kenny2023, Ragodos2022}, sequence classification~\cite{Ming2019}, healthcare~\cite{Mohammadjafari2021},  and graph classification~\cite{Zhang2022}.

The interpretability is ensured by the fact that the prototypes are real samples and can be visualized. Moreover, the classification layer is easily interpreted since the predictions are based on a weighted average of the prototypes' activations. By comparing the input and the most activated prototypes, a user can extract insights about important features in the input, similar to feature attribution. Conversely, by extracting samples close in the latent space to the prototypes, a user can extract global explanations by example, both positive and negative~\cite{Singh2021} (i.e., this does not look like that prototype). 

\textbf{Constraint-based architectures} encourage the network to learn more interpretable representations in the form of disentangled or concept-aligned representations~\cite{Koh20, Varshneya2021, Chen2020}. 
\glspl{CBM}~\cite{Koh20} is an example of such architecture. Given a latent representation of a sample, \glspl{CBM} are trained to predict the presence or absence of a set of pre-defined concepts in the sample. The probabilities of all the concepts are then combined to classify the sample. The idea is that a user can look at the probabilities associated with the set of concepts to understand which ones are influencing the most the prediction. Moreover, the user can also intervene by changing the probability of a concept to the desired one. Several works extend \glspl{CBM} by improving the concept representations~\cite{havasi2022addressing}, the performance~\cite{Zarlenga2022, yuksekgonul2023posthoc}, and the induced bias~\cite{oikarinen2023labelfree, Yang2023}. 
Nonetheless, the induced bottleneck limits the expressiveness of the network, and thus their performance is usually lower than the black-box counterparts.

An alternative approach is to enforce the latent disentanglement without introducing the bottleneck~\cite{Varshneya2021, Chen2020, Subramanian2018, Losch2021}. This goal can be achieved either using additional semantic layers that project the latent representation in another more aligned subspace~\cite{Losch2021} or using a normalization layer, as proposed in Concept Whitening~\cite{Chen2020}.
Despite the progress and improvements in performance, several open problems are connected to the usage of these networks, such as concept leakage and dependencies related to the concept dataset or models. 

\textbf{Attention-based architectures} use attention modules to improve the performance and the interpretability of the decision process~\cite{Zheng2017, Li2021, Fu2017}. The idea is that since attention is a weighted sum of the vector representations, attention weights can be used directly as feature attribution scores as long as the vector representations are meaningful. Attention has been used to discover relations such as the coreference and syntax in natural language processing tasks~\cite{Tenney2019, Clark2019, Vig2019nlp} and modality relations in multimodal models~\cite{Chefer2023}. For architecture employing multiple attention layers, \citet{Abnar2020} propose a post-hoc intrinsic method to reconstruct the flow of attention weights along the network and compute feature attribution based on the flow. 

Several works in literature explored the conditions under which attention can be considered as a reliable proxy for explanations~\cite{Wiegreffe2019notnot}. For example, \citet{Wiegreffe2019} and \citet{Serrano2019} find altering the attention weights does not affect the predictions of some configurations of attention-based architectures, casting doubts on their reliability. In this case, the problem is caused by input dispersion, a phenomenon connected to the accumulation of information from different sources in a single vector representation. 

Finally, apart from these general categories, there are several other self-explainable DNNs tailored for specific tasks or domains~\cite{Dai2021, Gui2023} that combine concepts of the previous categories or propose novels one~\cite{Fu2017, Chen2020healthcare,rigotti2022attentionbased, Rayhan2023}, lacking however in generalizability.  

Overall, since the area of self-explainable DNNs is relatively recent and emerging, several open challenges exist to address. Among them, we can mention the limited generalizability since most of the approaches are tested only on specific families of architectures, the limited diversity among the approaches, the accuracy-explainability trade-off, and the bias induced by their design.

\section{Relations with Thesis Contribution}
\label{sec:related_related}
This thesis contributes to the research efforts in explaining \glspl{DNN} outlined in the previous section by:
(i) proposing novel designs for self-explainable DNNs;
(ii) advancing the understanding of features learned by \glspl{DNN}; and
(iii) discussing the integration of visual analytics and explanation methods to deliver explanations to users.

The proposed designs of self-explainable DNNs address issues related to limited generalizability and diversity, to the usage of custom training recipes, and the accuracy-explainability trade-off. 
While existing self-explainable DNNs are primarily applied to vision tasks and convolutional neural networks (\Cref{sec:related_senn}), our contributions extend to novel domains and architectures. 
Specifically, our proposed designs based on concepts and prototypes (\Cref{chapter:pignn} and \Cref{chapter:whitening}) relax common assumptions, like the availability of pre-trained models, and can be applied to various families of graph neural networks, an area often overlooked by state-of-the-art approaches. 
The generalization contribution is directed not only to the type of data and the families of covered architectures but also to the type of normalization layers, an unexplored area in the current literature.

While these designs contribute to the literature in terms of generalization, they reiterate prototypes and concept-based architectures and use custom training recipes to train the models, following the current literature. However, using recipes different from those used to train the black-box models can cause instability in the training process when applied to settings where the models use non-standard recipes. These issues are addressed by the proposed designs of self-explainable DNNs based on memory modules (\Cref{chapter:senn}). Exploiting memory modules for explainability purposes represents a novelty in the field, increasing the diversity of approaches. These designs can be directly applied to black-boxes models following their training settings. The combination of memory modules and attention mechanisms mitigates the problem of input dispersion, typical of attention-based models, and is more flexible than current popular designs since it can provide proxies for feature attribution, explanations by example, and, for the first time for a self-explainable DNNs, for local counterfactuals. 

The designs mentioned above introduce novel architectures replacing the current black-box models. Conversely, the third and fourth parts of the thesis contribute to the research about explaining black-box \glspl{DNN} without modifying them.

The \Cref{part:neurons} advances the state-of-the-art feature probing methods by investigating a wider spectrum of neuron behavior than the one investigated in past research (i.e., only the highest activations) and proposing metrics to evaluate such methods. Specifically, the proposed method overcomes the issues related to the computational complexity of the previous approaches~\cite{Mu2020} and provides explanations that better capture the superposition of neuron activations~\cite{elhage2022superposition}. Additionally, due to the wider analyzed spectrum, the thesis contributes to this area by shedding light on novel overlooked activation phenomena connected to the activations of popular \glspl{DNN}.

Finally, \Cref{part:va} contributes to the literature by reviewing visual analytics systems as a means to provide explanations computed by \gls{XAI} methods to the user. This effort targets exclusively \gls{VA} systems dealing with \glspl{DNN} and using \gls{XAI} methods to enhance their interpretability, and it represents the first effort in this direction. Indeed, current \gls{VA} literature~\cite{Alicioglu2021, Hohman2019} targets more general settings, and \gls{XAI} literature has limited awareness of these tools. Therefore, it is unclear whether the current VA solutions are effective and what the strengths and weaknesses of the current literature are. Both these contributions, together with increasing the awareness of the \gls{XAI} community and paving the steps for increasing the collaborations, are discussed in the \Cref{part:va}.

\part{Self-Explainable Neural Networks}
\label{part:senn}
\chapter{Prototype-based Interpretable Graph Neural Networks}
\label{chapter:pignn}

This chapter represents the entry point of the thesis contribution on self-explainable designs for \glspl{DNN}. As mentioned in \Cref{chap:intro}, these designs aim to enhance the interpretability of \glspl{DNN} and help developers to debug and improve their models. Contextually, these designs aim to reduce the typical performance trade-off of self-explainable DNNs and broaden the applicability of such methods. 
The idea shared by \textbf{all the techniques proposed in this part of the thesis is to insert the novel components into the \gls{DNN} without disrupting its structure and preserving most of its representation power}. Therefore, the proposed layers are placed in between the feature extractor $f_{FE}$ and the classifier $f_{CL}$ (\Cref{sec:background_DL}). 

As a first step, this chapter proposes a \textbf{self-explainable prototype-based layer for \glspl{GNN}} inspired by the design of prototype-based architectures~\cite{Chen2019} commonly used for image classification. Specifically, the resulting architecture, called \gls{PIGNN}, learns a set of prototypes representing nodes and uses the similarity between the node representations of the current input and the learned prototypes to perform both graph and node classification. 
This design enhances the interpretability of black-box \glspl{GNN}, supporting developers in understanding and improving their models. Indeed, this design supports the identification of pitfalls of the dataset or in the training setup as well as error analysis through explanations by examples and feature attributions.

The chapter is organized as follows: \Cref{sec:proto_design} describes the proposed prototype layer and how users can probe its behavior; \Cref{sec:proto_experiments} evaluates \gls{PIGNN} both in terms of performance and explanations quality; \Cref{sec:pignn_choices} discusses alternative design choices for PIGGN; finally, \Cref{sec:pignn_contributions} summarized the contributions of this chapter.
\begin{figure}[t]
    \centering
    \includegraphics[scale=0.25]{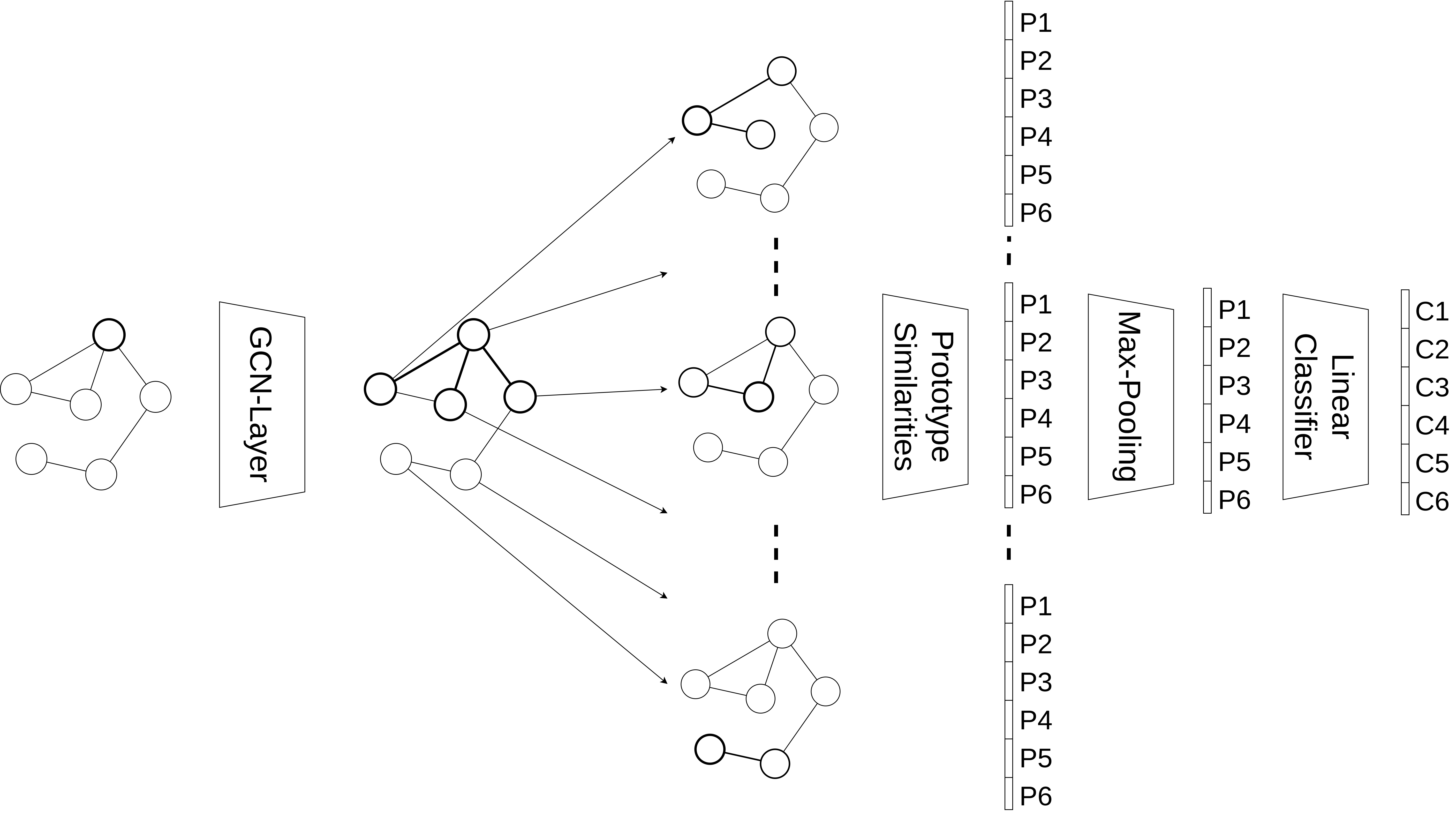}
    \caption{The architecture of PIGNN when the task is graph classification.}
    \label{fig:pignn}
\end{figure}
\section{Design}
\label{sec:proto_design}
As previously mentioned, the \textbf{\gls{PIGNN}} design introduces a layer after the feature extractor $f_{FE}$ to enhance interpretability. 
This layer includes a set of $m$ learned vectors $\bm{P} \in \mathbb{R}^{m \times emb} =\{p_j\}^m$, termed \emph{prototypes}. \textbf{Each prototype is optimized during training to represent a node embedding} of dimension $emb$.
Given an input $\bm{x}_i$, the feature extractor returns the latent representation $\bm{h} = f_{FE}(\bm{x}_i) \in \mathbb{R}^{n \times emb}$, where $n$ is the number of nodes in the graph input.
The latent representation is then compared against the full set of prototypes, producing a score that estimates their similarity:
\begin{equation}
    \bm{s} = Sim(\bm{h}, \bm{P}) \in \mathbb{R}^{n \times m}
\end{equation}
Each prototype is associated with $n$ scores, reflecting how strongly the semantic encoded in the prototype is represented in each node embedding. The layer employs global Max Pooling (\Cref{sec:background_DL}) to select the maximum score for each prototype:
\begin{equation}
    \bm{a}_i = max_{n \in Np}(\bm{s}) \in \mathbb{R}^{n \times m}
    \label{eq:proto_activation}
\end{equation}
The resulting scores can be regarded as how much the current graph input activates each prototype.
In the case of graph classification, the classifier takes the activation scores as input to make predictions (\Cref{fig:pignn}):
\begin{equation}
    \bm{y}_i = f_{CL}(\bm{a}_i)
\end{equation}
For node classification, the classifier directly utilizes the similarity scores as inputs, as each node is associated with a prediction: 
\begin{equation}
    \bm{y}_i = f_{CL}(\bm{s}_i)
\end{equation}

\subsection{Training Process.} \gls{PIGNN}, like all ProtoPNet~\cite{Chen2019} variants,  \textbf{requires a custom training recipe}. Each training epoch comprises three phases: weight optimization, prototype projection, and classifier optimization.\\

\paragraph{Weight Optimization.}
This phase adjusts the learned latent space of the features extractor to cluster relevant subgraphs around the prototypes and to separate prototypes from each other. To achieve the goal, the feature extractor and the prototype layer are optimized by minimizing the following loss~\cite{Wang2021tesnet}:
\begin{equation}
    L = L_{pred} + \lambda_1L_{cluster} +\lambda_2L_{orth}  - \lambda_3 L_{class\_separation} - \lambda_4L_{separation}
    \label{eq:loss_protopnet_complete}
\end{equation}
and $\lambda_1$, $\lambda_2$, $\lambda_3$, $\lambda_4$ are hyperparameters that balance the weight of the losses.
Here, $L_{pred}$ is the task loss used to optimize the model (e.g., cross-entropy between prediction and ground truth):

\begin{equation}
    L_{pred} =  \frac{1}{n}\sum_{i=1}^n CrossEntropy(f(x_i),y)
\end{equation}
$L_{cluster}$ encourages subgraphs important for similar predictions to cluster together:
\begin{equation}
    L_{cluster} = \frac{1}{n}\sum_{i=1}^n {min}_{j:\bm{p}_j\in P_{y_i}} min_{\bm{h}\in \text{khop\_subgraph}(x_i)} d(\bm{h},\bm{p_j}) 
    \label{eq:cluster_loss}
\end{equation}
where $khop\_subgraph(x\_i)$ is the set of all the subgraphs of the graph $x_i$ such that they are centered in one of the nodes in $x_i$ and include all the neighbors of that node at a maximum distance equal to $k$ and $d(\bm{h},\bm{p_j})$ is the distance between the latent representation and the prototype $\bm{p_j}$ computed as:
\begin{equation}
    d(\bm{h},\bm{p_j}) = \frac{\bm{h}^T\bm{p_j}}{\lVert \bm{h} \rVert}
\end{equation}
$L_{orth}$ enforces the orthonormality of prototypes, avoiding prototypes converging toward the same semantics:
\begin{equation}
    L_{orth} = \sum_{k=1}^K \lVert P_kP_k^T - \mathbb{I} \rVert
\end{equation}
$L_{class\_separation}$ encourages separation between prototypes associated with different classes:
\begin{equation}
    L_{class\_separation} = \frac{1}{\sqrt{2}}\sum_{k_1=1}^{K-1}\sum_{k_2=k_1+1}^{K} \lVert P_{k_1}P_{k_1}^T-P_{k_2}P_{k_2}^T \rVert
\end{equation}
Finally, $L_{separation}$ promotes differentiation between prototypes associated with other classes:
\begin{equation}
    L_{separation} =  \frac{1}{n}\sum_{i=1}^n {min}_{j:\bm{p}_j\notin P_{y_i}} min_{\bm{h}\in \text{khop\_subgraph}(x_i)}  d(\bm{h},\bm{p_j}) 
    \label{eq:loss_separation}
\end{equation}

Note that during this phase, the weights of the classifier $f_{CL}$ are frozen. The total loss is a generalization of the ones proposed in \citet{Chen2019}, which can be recovered by setting $L_{orth}$ and $L_{class\_separation}$ to zero.

\paragraph{Prototype Projection.}
Since the first phase can produce out-of-distribution or irrealistic prototypes, this phase maps each prototype to the closest subgraph in the training dataset:
\begin{equation}
    \bm{p}_j = \argmin_{\bm{h}\in FE_j} \lVert \bm{h}-\bm{p}_j \rVert
    \label{eq:proto_projection}
\end{equation}
where $FE_j$ is the set of node embeddings returned by the feature extractor for all the samples in the dataset of the same class of the prototype $\bm{p}_j$. This phase is crucial to keep the prototypes interpretable (i.e., associated with a clear semantic) and visualize them (\Cref{sec:proto_explanation_design}).

\paragraph{Classifier Optimization.}
This phase optimizes the classifier $f_{CL}$ while keeping the feature extractor and the prototype layer frozen. It aims to obtain a sparse connection between prototypes and output neurons, which is a desirable property for interpretability. To achieve the goal, we use a loss that penalizes the weights between prototypes associated with a class and the output neurons corresponding to all the other classes:
\begin{equation}
    L_{CL} =  L_{pred} + \lambda_5 + \sum_{k=1}^{nc}\sum_{j:\bm{p}_j\notin P_k}abs(w^{j,k}_{CL})
    \label{eq:proto_classifier_loss}
\end{equation}
where the weight $w^{j,k}$ connect the the prototype $\bm{p}_j$ to the output neuron corresponding to the class $k$.

\subsection{Explanations}
\label{sec:proto_explanation_design}
The decision process of the described \gls{PIGNN} can be probed in two ways. The first consists of understanding the semantics of the learned prototypes, shedding light on what the model deems important during the training process. The second leverages prototype activation to yield node attributions.
\textbf{To understand the semantics of the learned prototypes, we can use explanations by examples, considering prototypes as inputs}. As described in the previous section, during the second training stage, the prototype is projected into one of the subgraphs included in the dataset, becoming a point in the training distribution of node embeddings. Consequently, we can extract neighbors of each prototype using an algorithm like K-NN~\cite{Cover1967} over the node embeddings of dataset samples. Then, we can use and visualize the neighbors as explanations by examples, helping users understand the semantics encoded by the prototypes.

Regarding \textbf{node attribution}, note that, by design, a node embedding encodes information of the \emph{k-hop subgraph} originating from the node. This subgraph includes all the neighbors within the radius $k$, where $k$ denotes the number of layers before the considered embedding. The k-hop subgraph that excites the prototype the most is identified by selecting the node embedding with the highest similarity to the prototype (determined by \Cref{eq:proto_activation}).  Feature attribution for each node $i$ in the k-hop subgraph is then computed as the sum of similarities between its node embedding and each prototype associated with the prediction. This computation is weighted by the learned weights connecting the prototypes and the predicted class:
\begin{equation}
    e (x_i) = \sum_{p_j \in \bm{P}_c} w_{jc}s_{ij}
\end{equation}
where $x_i$ is one of the nodes in the k-hop subgraph, $c$ is the predicted class for the input $x$, $w_{jc} \in W_{CL}$ denotes the learned weight of the classifier connecting the prototype $p_j$ with the output neuron corresponding to the class $c$, and $\bm{P}_c$ is the set of prototypes associated with the class $c$.

This process can be applied to all nodes in the graph. However, given that graphs are typically compact structures and tasks often involve identifying specific substructures, we propose considering only the \textbf{k-hop subgraph of the current input corresponding to the most activated prototype}. This approach yields explanations that are both easy to understand and sparse, aligning them with the inherent characteristics of graph structures.

\section{Experiments}
\label{sec:proto_experiments}
\subsection{Performance}
This section evaluates the performance of popular \glspl{GNN} augmented with the proposed prototype layer. Specifically, we train \gls{GCN}~\cite{Kipf2017}, \gls{GAT}~\cite{Velickovic2018}, and \gls{GIN}~\cite{Xu2019} on various datasets: BA-shapes~\cite{Ying2019},  BA-Community~\cite{Ying2019}, MUTAG~\cite{Debnath1991}, BBBP~\cite{Martins2012}, BACE~\cite{Subramanian2016}, BA-2Motifs~\cite{Luo2020}, Tree-Grid~\cite{Ying2019}, Cora~\cite{McCallum2000}, Citeseer~\cite{Giles1998} and Pubmed~\cite{Sen2008}. BA-shapes, BA-Community, Tree-Grid, and  BA-2Motifs are syntenic datasets built for node classification (BA-shapes, BA-Community, Tree-Grid) and graph classification (BA-2Motifs). MUTAG, BBBP, and BACE are molecule datasets for graph classification, where the task is to classify molecules as active or inactive. Cora, Citeseer, and Pubmed are datasets for node classification, where nodes represent scientific publications, and the edges represent citations. \textbf{Differently from the standard procedures for prototype-based models in vision, we train these models from scratch} since there are no publicly available pre-trained \glspl{GNN} on large corpora. Additionally, we jointly optimize the feature extractor and the prototype layers during the same training process.
We adopt the setup of \citet{Ying2019}, training 3-layers \gls{GCN}, \gls{GAT}, and \gls{GIN} with 128 units each. Models are trained with and without prototype layers using 15 different seeds. Following \cite{Chen2019} and \cite{Wang2021tesnet}, we set the number of prototypes to ten per class, the value of $\lambda_1$ to 0.8, the value of $\lambda_2$ to 1e-7, 0.08, the value of $\lambda_3$ to 0.7, the value of $\lambda_4$ to 0.8, and the value of $\lambda_5$ to 1e-3.

\begin{table}[!b]
\centering
\caption{Avg. Accuracy and standard deviation over 15 runs of GNNs with and without prototype layers on synthetic datasets.}
\begin{tabular}{lrrrr}
\toprule
Model   & BA-2Motifs         &BA-Community  & BA-Shapes & Tree-Grid \\ 
\midrule
GCN                       &  \textbf{87.20 \footnotesize{$\pm$ 4.19}}                &  \textbf{97.34 \footnotesize{$\pm$ 0.14}}          & 91.23 \footnotesize{$\pm$ 1.08}        &  84.79 \footnotesize{$\pm$1.76}       \\
+ Proto     & 59.00 \footnotesize{$\pm$ 4.12}                & \textbf{97.31 \footnotesize{$\pm$ 0.29}}          &\textbf{96.30 \footnotesize{$\pm$ 0.42}}&  \textbf{97.17 \footnotesize{$\pm$ 0.59}}       \\
\midrule
GAT                   &  \textbf{51.20 \footnotesize{$\pm$ 2.32}}                &  \textbf{86.61 \footnotesize{$\pm$ 1.76}}          & \textbf{76.54 \footnotesize{$\pm$ 1.18}}        &  \textbf{58.49 \footnotesize{$\pm$ 0.00}}       \\
+ Proto               &  \textbf{50.10 \footnotesize{$\pm$ 3.08}}                &  \textbf{87.94 \footnotesize{$\pm$ 1.66}}          & 42.86 \footnotesize{$\pm$ 0.00}        &  \textbf{58.49 \footnotesize{$\pm$ 0.00}}       \\ 
\midrule
GIN                   &  \textbf{99.50 \footnotesize{$\pm$ 0.67}}       &  92.34 \footnotesize{$\pm$ 1.95}          & 89.79 \footnotesize{$\pm$ 3.89}        &  87.03 \footnotesize{$\pm$ 1.93}       \\
+ Prototypes                 &  \textbf{99.80 \footnotesize{$\pm$ 0.40}}       &  \textbf{98.36 \footnotesize{$\pm$ 0.34}} &\textbf{96.96 \footnotesize{$\pm$ 0.14}}& \textbf{98.67 \footnotesize{$\pm$ 0.28}}\\
\bottomrule
\end{tabular}
\label{tab:gnn_syntentic}
\end{table}

\begin{table}[!b]
\centering
\caption{Avg. Accuracy and standard deviation over 15 runs of GNNs with and without prototype layers on molecular datasets.}
\begin{tabular}{lrrr}
\toprule
Model      & BBBP         & MUTAG     & BACE    \\ 
\midrule
GCN                  & 85.75 \footnotesize{$\pm$ 0.61}           & \textbf{77.00 \footnotesize{$\pm$ 6.78}}        & 76.32 \footnotesize{$\pm$ 1.74}   \\

+ Proto            & \textbf{87.61 \footnotesize{$\pm$ 1.18}}  & \textbf{80.50 \footnotesize{$\pm$ 5.68}}        & \textbf{79.28 \footnotesize{$\pm$ 1.67}}         \\
\midrule
GAT                  & \textbf{87.32 \footnotesize{$\pm$ 1.20}}  & \textbf{77.50 \footnotesize{$\pm$ 8.14}}        & \textbf{78.22 \footnotesize{$\pm$ 2.66}}            \\
+ Proto           & \textbf{87.02 \footnotesize{$\pm$ 0.73}}  & \textbf{79.00 \footnotesize{$\pm$ 6.24}}        & \textbf{77.57 \footnotesize{$\pm$ 2.44}}             \\ 
\midrule
GIN                  & \textbf{85.61 \footnotesize{$\pm$ 1.56}}           & \textbf{82.50 \footnotesize{$\pm$ 7.50}}        & 75.40  \footnotesize{$\pm$ 1.45}            \\
+ Proto           & \textbf{87.17 \footnotesize{$\pm$ 1.05}}  &\textbf{85.50 \footnotesize{$\pm$ 3.50}}& \textbf{79.34 \footnotesize{$\pm$ 2.68}}    \\ 
\bottomrule
\end{tabular}

\label{tab:gnn_molecular}
\end{table}
We compare the black-box models against the same models augmented with a prototype layer in synthetic  (\Cref{tab:gnn_syntentic}) and molecular datasets (\Cref{tab:gnn_molecular}) to assess the validity of the augmentation. The results suggest that \textbf{adding the layer to the black-box model and jointly training them does not compromise performance}. In most configurations, models with prototype layers exhibit the same or higher performance than those without prototypes. Exceptions include \gls{GAT} on BA-Shapes and \gls{GCN} on BA-2motif, which can be addressed using a simplified loss (\Cref{sec:pignn_choices}). We observe that the performance gain depends on the particular combination of dataset, model, and tasks. The highest gap in accuracy is reached on node classification tasks and synthetic datasets and by the \gls{GIN} model. These results align with the ones reported by \citet{Wang2021tesnet} on image data for architectures augmented by prototypes and \citet{Xu2019} for \gls{GIN} superiority. They also suggest that the better the model, the higher the gap between the black-box and the augmented model. 

\begin{table}[b!]
\centering
\caption{Avg. Accuracy and standard deviation over 15 runs of GNNs with and without prototype layers on citation datasets.}
\begin{tabular}{lrrr}
\toprule
 & Citeseer     & Pubmed       & Cora         \\ 
 \midrule
GCN        & 73.60 \footnotesize{$\pm$ 0.35} & 74.52 \footnotesize{$\pm$ 0.54} & 79.09 \footnotesize{$\pm$ 0.81} \\
GCN + Proto   & 70.72 \footnotesize{$\pm$ 1.01} & \textbf{76.77 \footnotesize{$\pm$ 0.86}} & 78.65 \footnotesize{$\pm$ 0.80} \\
SEGNN      & \textbf{74.19 \footnotesize{$\pm$ 0.51}} & \textbf{76.92 \footnotesize{$\pm$ 0.30}} & \textbf{79.66 \footnotesize{$\pm$ 0.56}} \\
ProtGNN   & 53.07 \footnotesize{$\pm$ 1.98} & 75.93 \footnotesize{$\pm$ 1.14} & 59.73 \footnotesize{$\pm$ 1.95} \\
\bottomrule
\end{tabular}
\label{tab:proto_citing}
\end{table}
With the approach validated, we compare its performance against an alternative explainable \gls{GNN} based on prototypes (ProtGNN~\cite{Zhang2022}), another type of explainable network (SEGNN~\cite{Dai2021}), and the black-box models on Cora, Citeseer, and PubMed (\Cref{tab:proto_citing}). ProtGNN is a prototype-based model that computes prototypes as graph embedding and represents the alternative configuration of prototype-based layers for \gls{GNN}. SEGNN is an explainable \gls{GNN} that classifies nodes based on K-nearest nodes found on the basis of the sub-graphs similarity.

While SEGNN achieves the best performances, \gls{PIGNN} closely matches its performance on two over of three datasets and outperforms the alternative prototype-based design in all tasks. Lower prototype-based network performance against SEGNN suggests ten prototypes (\Cref{sec:pignn_choices}) may be insufficient to capture enough patterns in citation networks. However, in the next section, we observe that better-quality explanations of these networks somewhat mitigate this issue.
\subsection{Explanations}
\begin{figure}[t!]
  \begin{subfigure}{.45\linewidth}
    \centering
    \includegraphics[scale=0.18]{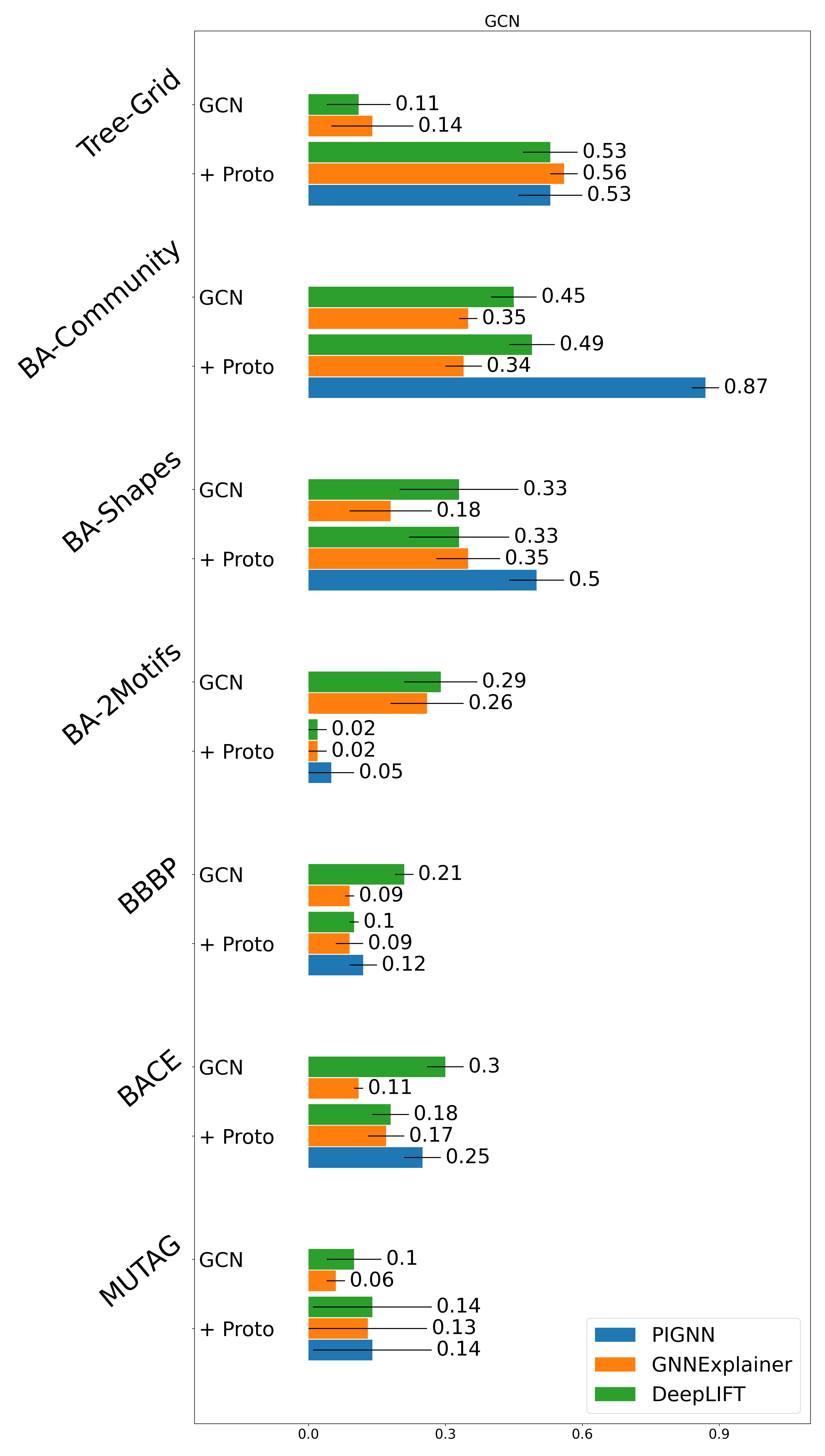}
    \caption{\label{fig:fidelity_gcn}}
  \end{subfigure}%
  \hfill
  \begin{subfigure}{.45\linewidth}
    \centering
    \includegraphics[scale=0.18]{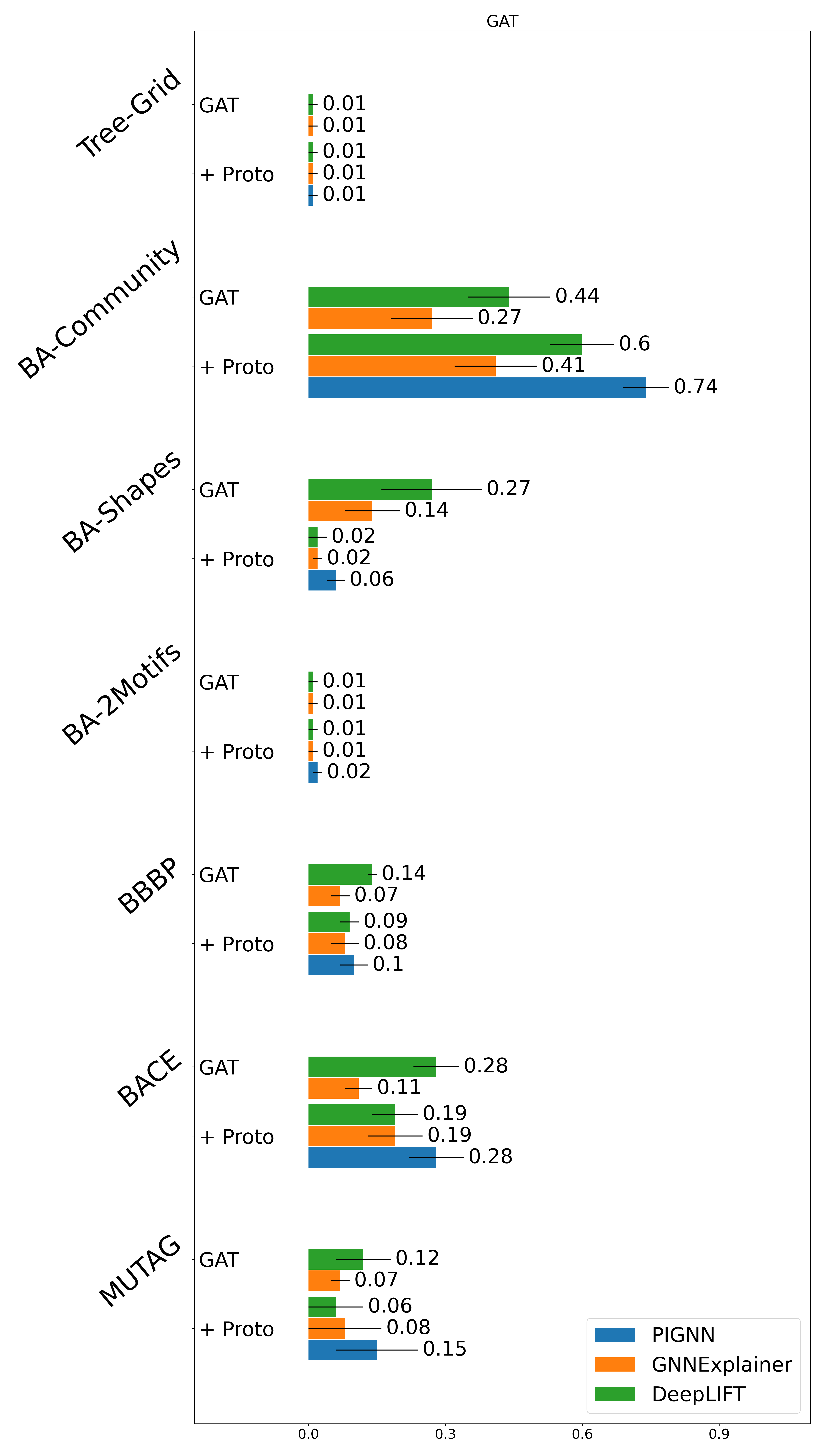}
    \caption{\label{fig:fidelity_gat}}
  \end{subfigure}
  \caption{Avg. Fidelity and standard deviation achieved by the intrinsic method, DeepLIFT, and GNNExplainer on GCN (a) and GAT(b) models across several datasets.}
\end{figure}
\begin{figure}[t!]
    \centering
    \includegraphics[scale=0.19]{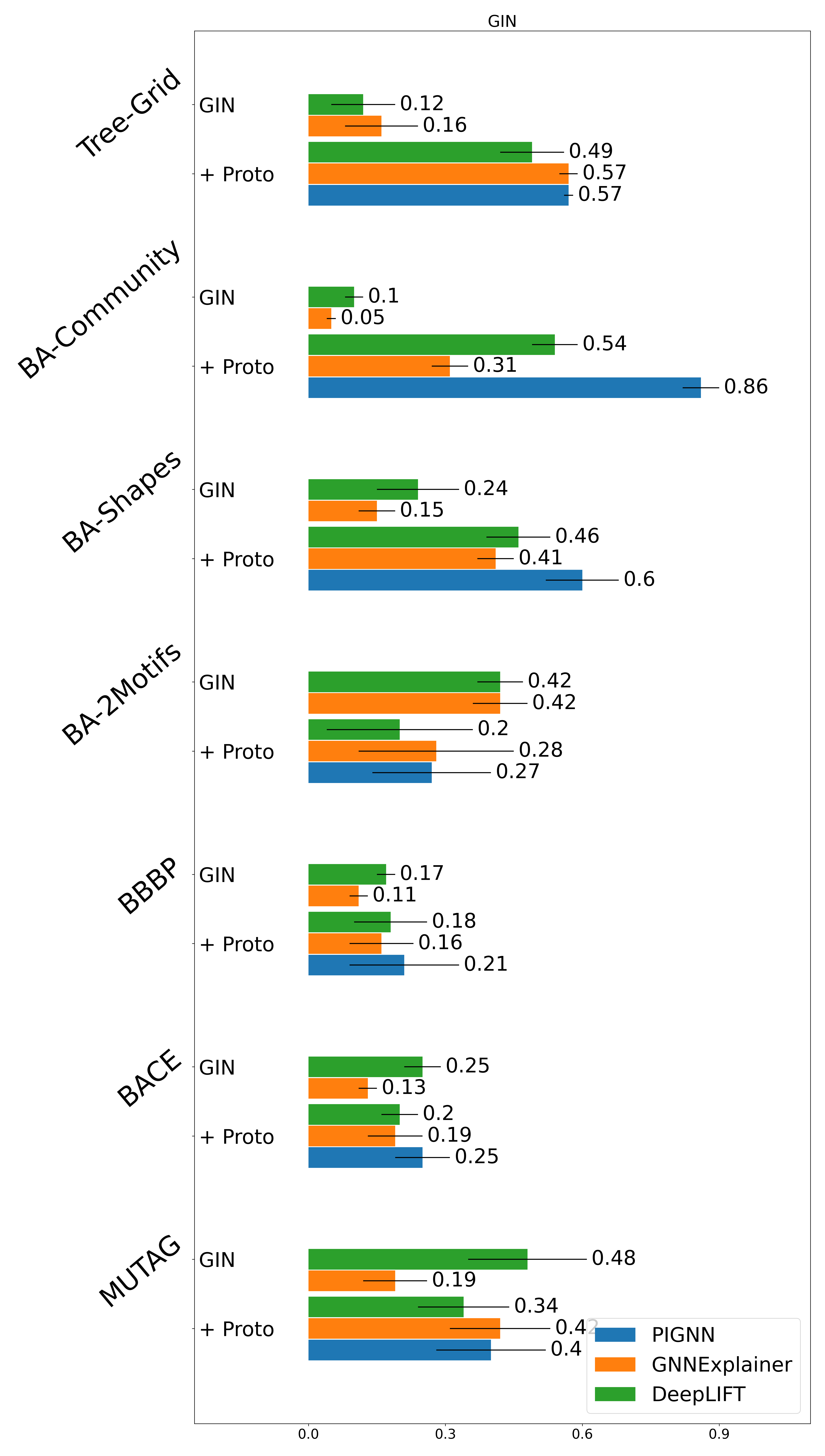}
    \caption{Avg. Fidelity and standard deviation reached by the intrinsic method, DeepLIFT, and GNNExplainer on GIN models across several datasets.}
    \label{fig:fidelity_gin}
\end{figure}

This section discusses and evaluates the explanations retrievable using \gls{PIGNN}. We begin by evaluating the quality of feature attributions computed using prototype activations in terms of $Fidelity+^{prob}$ (\Cref{metric:fidelity_prob}). We compare its scores against DeepLIFT~\cite{Shrikumar2017}, a popular model-agnostic post-hoc extrinsic method,  and GNNExplainer~\cite{Ying2019}, a recent feature attributions method tailored for GNNs. We compare the performance of both models with and without prototypes.

\Cref{fig:fidelity_gcn}, \Cref{fig:fidelity_gat}, and \Cref{fig:fidelity_gin} summarize the explanation quality of explanation computed on synthetic and molecular datasets over \gls{GAT}, \gls{GCN}, and \gls{GIN} models. Overall, \textbf{\gls{PIGNN}'s explanations exhibit better quality than extrinsic methods}. In only a few cases, the combination of black-box models and extrinsic methods outperforms \gls{PIGNN}. When this occurs, the gap is small, and the explanations require more computation time (up to 20x for GNNExplainer). The good quality is also confirmed in \Cref{tab:proto_exp_exp}, where \gls{PIGNN} outperforms SEGNN and ProtGNN in terms of both $Fidelity+^{prob}$ and $Fidelity+^{ROC-AUC}$ when masking the most important subgraph. Notably, SEGNN is the worst performer in terms of interpretability. We hypothesize that its low performance is due to its dense decision process, reliant on the similarity between k-hop subgraphs, and thus, able to recover the prediction using similarities among other nodes. Conversely, prototype-based networks rely on the bottleneck induced by the prototypes, making their decision process more sparse and sensitive to removing crucial subgraphs.

\begin{figure}[t!]
    \centering
    \includegraphics[scale=0.3]{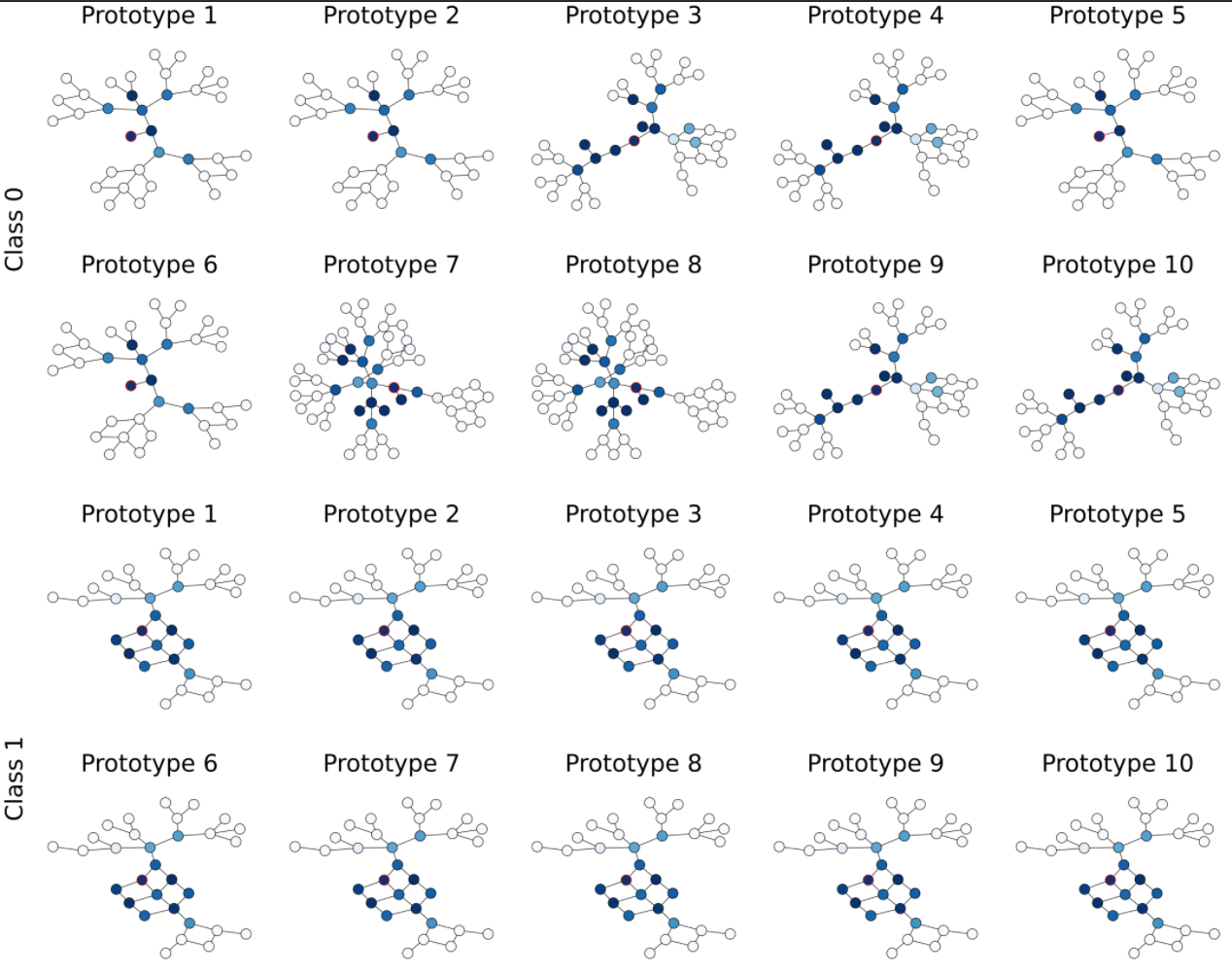}
    \caption{Sub-graphs associated with the prototypes learned by PIGNN based on GCN on
the Ba-2Motifs dataset. The most activated node is marked
with a red borderline. The blue shade encodes the importance of the node for the subgraph.}
    \label{fig:exp_prototypes}
\end{figure}
\begin{figure}[t!]
    \centering
    \includegraphics[scale=7]{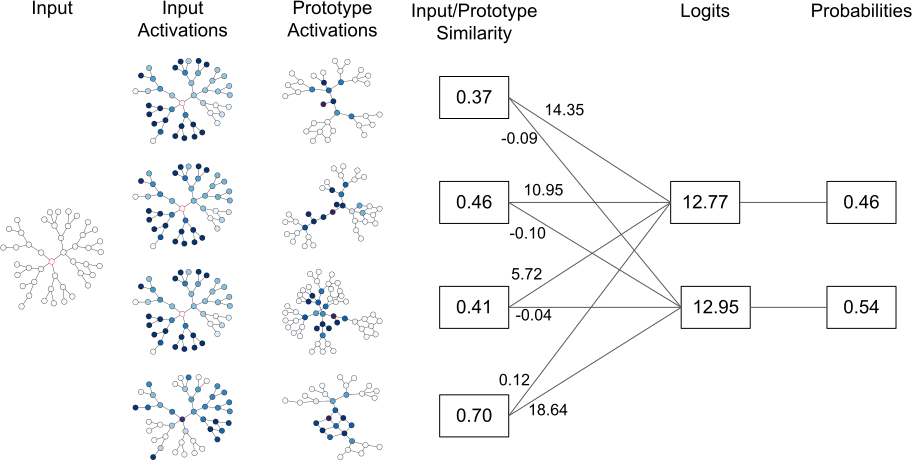}
    \caption{Visualization of explanation for a prediction. The red border indicates the analyzed node; the Input Activations column reports the similarity between the graph's nodes and the prototype displayed in the same row in the Prototype Activation column.}
    \label{fig:exp_prediction}
\end{figure}

\begin{table}[b!]
\centering
\caption{Explanations performances on big datasets. $Fidelity+^{prob}$ and $Fidelity+^{ROC-AUC}$ (\Cref{metric:fidelity_neg_acc}) are reported for the self-interpretable models and the extrinsic methods on the black-box GCN model.}
\label{tab:proto_exp_exp}

\begin{tabular}{lcccccc}
\toprule
    &        \multicolumn{3}{c} {$Fidelity+^{prob}$ (↑)}      &    \multicolumn{3}{c}{$Fidelity+^{ROC-AUC}$ (↓)}                       \\
   & Citeseer        & Pubmed         & Cora            & Citeseer        & Pubmed         & Cora            \\
\midrule
PIGNN         & \textbf{0.29 \footnotesize{$\pm$ 0.03}}  & \textbf{0.24 \footnotesize{$\pm$ 0.02}}  & \textbf{0.50 \footnotesize{$\pm$ 0.06}}  & \textbf{0.81 \footnotesize{$\pm$ 0.01}}           & \textbf{0.87 \footnotesize{$\pm$ 0.01}}           & 0.88 \footnotesize{$\pm$ 0.02} \\
SEGNN          & 0.03 \footnotesize{$\pm$ 0.00}           & 0.02 \footnotesize{$\pm$ 0.01}           & 0.10 \footnotesize{$\pm$ 0.01}           & 0.86 \footnotesize{$\pm$ 0.02}           & 0.90 \footnotesize{$\pm$ 0.03}           & 0.96 \footnotesize{$\pm$ 0.00} \\
ProtGNN       & 0.02 \footnotesize{$\pm$ 0.02}           & 0.06 \footnotesize{$\pm$ 0.02}           & 0.07 \footnotesize{$\pm$ 0.07}           & \textbf{0.81 \footnotesize{$\pm$ 0.03}}           & 0.89 \footnotesize{$\pm$ 0.02}           & \textbf{0.83 \footnotesize{$\pm$ 0.01}} \\
\bottomrule
\end{tabular}
\end{table}

Now, we describe how developers can \textbf{exploit the design and characteristics of \gls{PIGNN} to understand the model's behavior}. The examples presented here are drawn from a \gls{PIGNN} model based on \gls{GIN} and trained on the Tree-Grid dataset. Nevertheless, the same approach is applicable across different scenarios.

\Cref{fig:exp_prototypes} shows the globally most influential k-hop graphs used by the model for making predictions. We can note that the first two rows of the image display a diverse array of patterns associated with class 0, including various tree structures. Conversely, the lowest rows show duplications in the prototypes for class 1. By inspecting the dataset, a developer could verify that there is low variability in the patterns associated with class 1, most of which are associated with 3x3 grid patterns. These dataset's characteristics cause the convergence of \gls{PIGNN} towards the recognition of class 1 associated with the recognition of 3x3 grid patterns, captured by all the prototypes. Once that the problem has been identified, the developer can mitigate the issue, for example, by limiting the number of prototypes or increasing diversity in the dataset.

The visualization of prototypes, together with the similarity associated with them, can also be used to \textbf{understand individual predictions and perform error analysis}, as shown in \Cref{fig:exp_prediction}.
In this particular case, the model incorrectly predicts the class of a node as 1.  This misclassification is indicated by a high similarity score of 0.7 with the fourth prototype, which represents class 1. The analysis of the prototype's meaning (Prototype Activations column) reveals that the misclassification arises from the model detecting a grid pattern within a portion of the input graph, highlighted in the fourth row of the Input Activation column.

\section{Design Choices}
\label{sec:pignn_choices}
This section discusses alternative design choices for \gls{PIGNN} and experiments varying the values of hyperparameters.
\paragraph{Number of Prototypes.} 
\begin{figure}[t]
    \centering
    \includegraphics[scale=0.18]{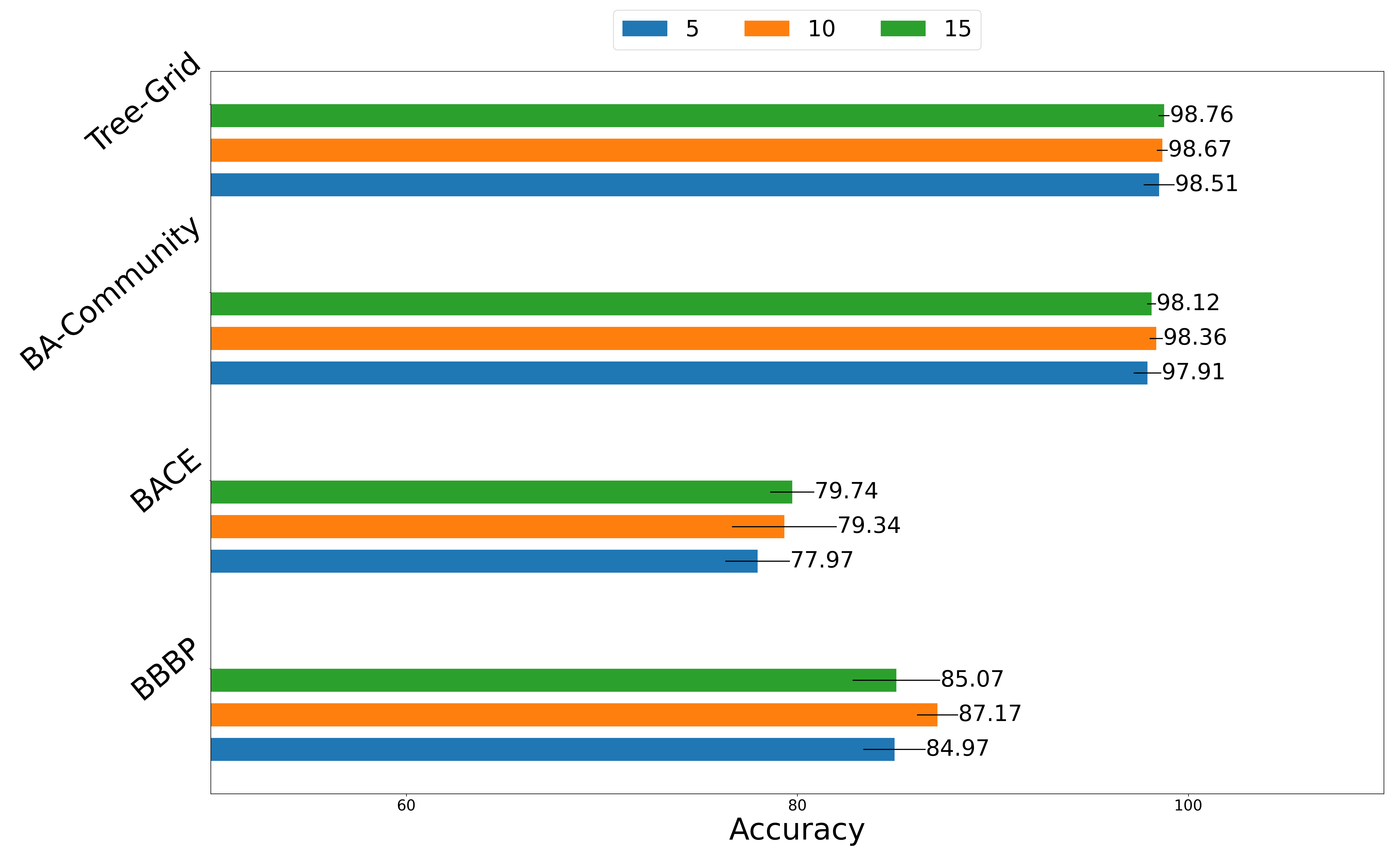}
    \caption{Accuracy of PIGNN when varying the number of prototypes.}
    \label{fig:number_proto}
\end{figure}
We begin the exploration of alternative design choices for \gls{PIGNN} by examining the impact of varying the number of prototypes in the model. As observed in the previous section, an excess of prototypes may lead to duplication despite incorporating the orthonormality loss. 
\Cref{fig:number_proto} depicts the performance of \gls{PIGNN} with 5, 10, and 15 prototypes across the BBBP, BACE, BA-Community, and Tree-Grid datasets.  We observe that there are no clear winners: while employing a larger number of prototypes generally seems beneficial, this trend does not hold true for BBBP, where using either 5 or 15 prototypes results in equivalent performance deterioration. Therefore, the number of prototypes is a hyperparameter to be tuned before training. Both domain-specific insights and prototype inspection, as shown in the previous section, can be used to guide the selection of the right number of prototypes.

\paragraph{Similarity Projection.}
\begin{table}[b!]
\caption{Fidelity comparison between projecting the similarities towards all the nodes in the graph and only towards the k-hop most important subgraph.}
\label{tab:khop}
\centering
\begin{tabular}{lllll}
\toprule
     & MUTAG & BACE & BBBP & Ba-2Motifs \\ 
\midrule
k-hop     & 0.35  \footnotesize{$\pm$ 0.10}    & 0.25  \footnotesize{$\pm$ 0.06}   & 0.18  \footnotesize{$\pm$ 0.07}   & 0.16  \footnotesize{$\pm$ 0.15}         \\
Full Nodes & 0.36  \footnotesize{$\pm$ 0.13}    & 0.24  \footnotesize{$\pm$ 0.06}   & 0.18  \footnotesize{$\pm$ 0.05}   & 0.16  \footnotesize{$\pm$ 0.15}         \\ 
\bottomrule
\end{tabular}
\end{table}
\begin{figure}[t]
\centering
     \begin{subfigure}{0.25\linewidth}
         \centering
         \includegraphics[width=\textwidth]{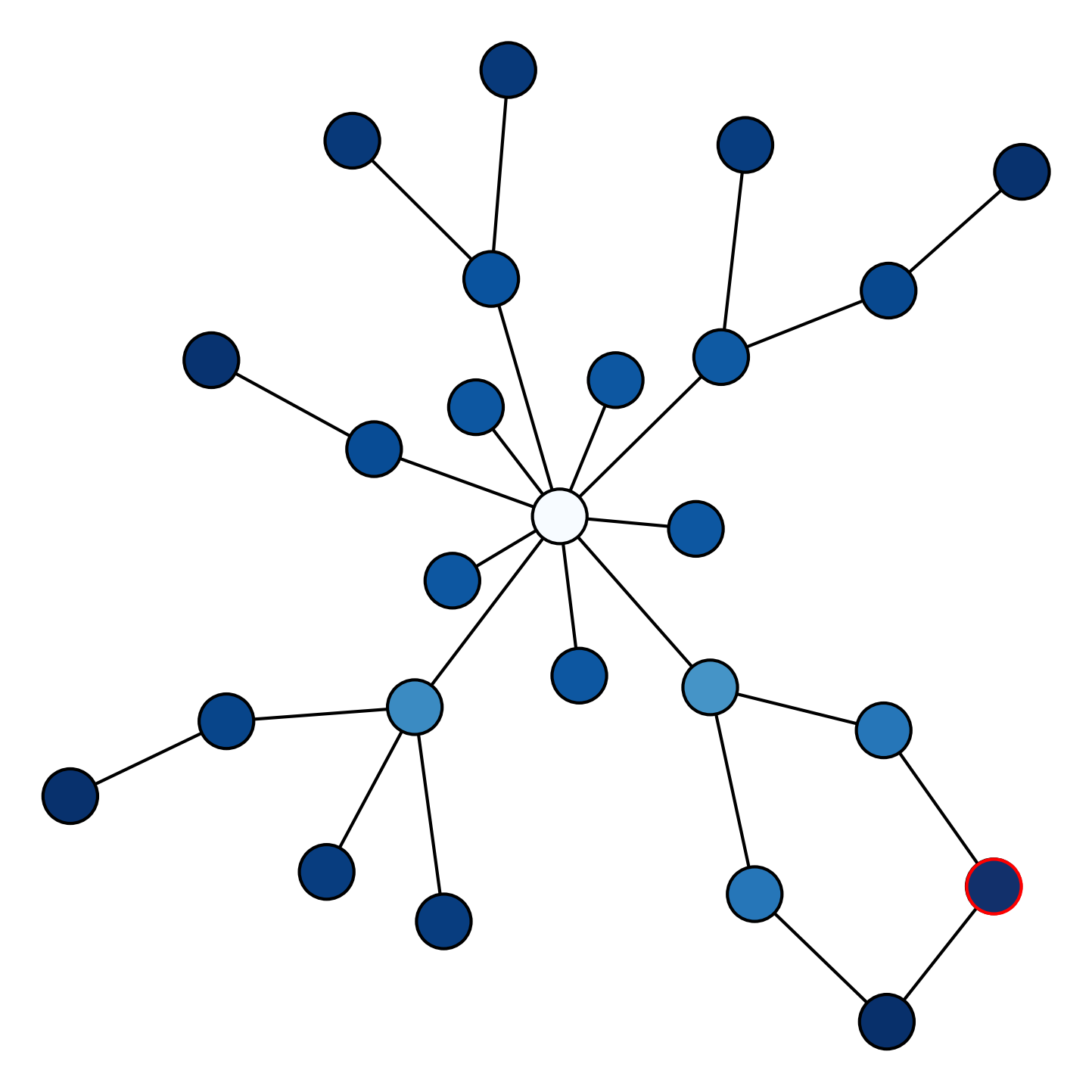}
         \caption{}
         \label{fig:pignn_fullnodes}
     \end{subfigure}
     \begin{subfigure}{0.25\linewidth}
         \centering
         \includegraphics[width=\textwidth]{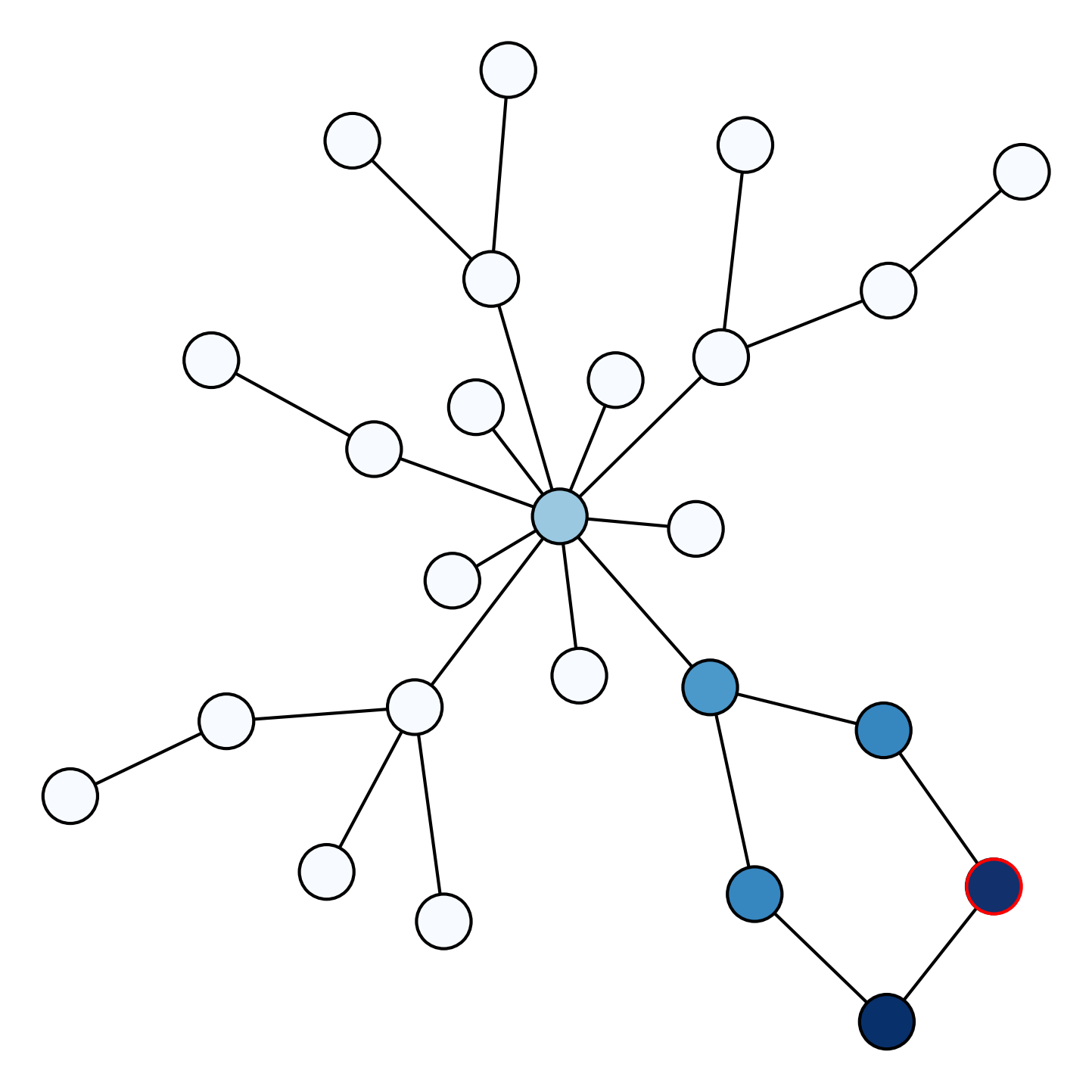}
         \caption{}
         \label{fig:pignn_khop}
     \end{subfigure}
        \caption{Visualization comparison between projecting the similarities towards all the nodes in the graph and only towards the k-hop most important subgraph.
        }
        \label{fig:visual_comp}
\end{figure}
A design choice about the explanations is whether to project the prototypes' similarities onto all the graph nodes or solely onto the k-hop most important subgraph, as proposed in \Cref{sec:proto_explanation_design}. \Cref{tab:khop} compares the fidelity of both approaches, revealing no discernible difference when considering only a subset of nodes. This observation is motivated by the nature of graph-related tasks, where often, identifying a single pattern suffices for graph or node classification. While interpretability scores remain unaffected, the visualization of the two approaches, as depicted in \Cref{fig:visual_comp}, differs significantly, with a sparse visualization offering enhanced comprehension and immediacy.

\paragraph{Loss Function.}
\begin{table}[!b]
    \centering
    \caption{Performance comparison between the full and the reduced loss applied to PIGNN.}
    \begin{tabular}{lrr}
    \toprule
    Dataset & Full Loss & Reduced Loss\\
    \midrule
    Citeseer & \textbf{70.72 \footnotesize{$\pm$ 1.01}} & \textbf{70.12 \footnotesize{$\pm$ 0.90}}\\
    Pubmed & \textbf{76.77 \footnotesize{$\pm$ 0.86}} & \textbf{76.38 \footnotesize{$\pm$ 1.07}}\\
    Cora & \textbf{78.65 \footnotesize{$\pm$ 0.80}}& \textbf{78.62 \footnotesize{$\pm$ 1.57}} \\
    BBBP & \textbf{87.17 \footnotesize{$\pm$ 1.05}} & 85.66 \footnotesize{$\pm$ 1.74}\\
    MUTAG & \textbf{85.50 \footnotesize{$\pm$ 3.50}} & \textbf{83.50 \footnotesize{$\pm$ 5.50}} \\
    BACE & \textbf{79.34 \footnotesize{$\pm$ 2.68}} & \textbf{80.86 \footnotesize{$\pm$ 3.12}}\\
    BA-2Motifs & \textbf{99.80 \footnotesize{$\pm$ 0.40}} & 98.90 \footnotesize{$\pm$ 1.14} \\
    BA-Community & \textbf{98.36 \footnotesize{$\pm$ 0.34}} & 96.96 \footnotesize{$\pm$ 0.14} \\ 
    BA-Shapes & \textbf{96.65 \footnotesize{$\pm$ 0.74}}  & 85.11 \footnotesize{$\pm$ 1.41} \\ 
     Tree-Grid & \textbf{98.67 \footnotesize{$\pm$ 0.28}} & 82.42 \footnotesize{$\pm$ 6.82} \\
    \bottomrule
    \end{tabular}
        \label{tab:proto_loss_bad}
\end{table}

\begin{table}[b!]
    \centering
    \caption{Examples of specific cases when the reduced loss can work better.}
    \begin{tabular}{llrr}
    \toprule
    Dataset & Model & Full Loss & Reduced Loss\\
    \midrule
    BA-2Motifs & GCN &  59.00 \footnotesize{$\pm$ 4.12}  &  \textbf{99.40 \footnotesize{$\pm$ 0.80}} \\
    BA-2Motifs & GAT &  50.10 \footnotesize{$\pm$ 3.08} &  \textbf{83.30 \footnotesize{$\pm$ 5.92}}\\
    \bottomrule
    \end{tabular}
    \label{tab:proto_loss_good}
\end{table}
\begin{table}[b!]
    \centering
    \caption{Interpretability scores achieved by PIGNN when using the full or reduced loss.}
    \begin{tabular}{lrrrr}
    \toprule
      &                \multicolumn{2}{c}{$Fidelity+^{prob}$ (↑)}     & \multicolumn{2}{c}{$Fidelity+^{ROC-AUC}$ (↓)} \\
    Dataset &  Full Loss & Reduced Loss &  Full Loss & Reduced Loss\\
    \midrule
    Citeseer & \textbf{0.29 \footnotesize{ $\pm$ 0.03}} &  \textbf{0.29 \footnotesize{ $\pm$  0.02}} &  \textbf{0.81 \footnotesize{ $\pm$  0.01}} &  0.86 \footnotesize{ $\pm$  0.01} \\
    Pubmed &  \textbf{0.24 \footnotesize{ $\pm$  0.02}}  &  0.20 \footnotesize{ $\pm$  0.01}  & \textbf{0.87 \footnotesize{ $\pm$  0.01}}  & 0.90 \footnotesize{ $\pm$  0.01}\\
    Cora & \textbf{0.50 \footnotesize{ $\pm$  0.06}} & 0.33 \footnotesize{ $\pm$  0.06} & \textbf{0.88 \footnotesize{ $\pm$  0.02}} & 0.94 \footnotesize{ $\pm$  0.00}\\
    \bottomrule
    \end{tabular}
    \label{tab:proto_xai_scores}
\end{table}
Another design choice is the one related to the selection of the loss function to be used during the first training phase. As mentioned earlier, an alternative loss function to the one outlined in \Cref{sec:proto_design} is proposed by \citet{Chen2019}, which omits the orthogonal and class separation losses. \Cref{tab:proto_loss_bad} illustrates that the full loss remains the optimal choice, as the reduced loss achieves comparable performance on only half of the datasets. However, specific scenarios, detailed in \Cref{tab:proto_loss_good}, demonstrate instances where the reduced loss outperforms the full one by a significant margin. In such cases, inherent characteristics of datasets like BA-2Motifs, such as dependency on a single motif, coupled with additional constraints imposed by the added losses, may cause the model to be stuck on a local minimum. A search over hyperparameters tailored to specific cases could mitigate this issue. Furthermore, as evidenced in \Cref{tab:proto_xai_scores}, in terms of interpretability, the full loss consistently demonstrates superiority in both $Fidelity+^{prob}$ and $Fidelity+^{ROC-AUC}$, thus suggesting that the added losses help the network to better separate the prototypes and thus improving their usefulness for interpretability purposes. Therefore, the full loss represents the best choice.

\paragraph{Transfer Learning}
\begin{figure}[t]
    \centering
    \includegraphics[scale=0.2]{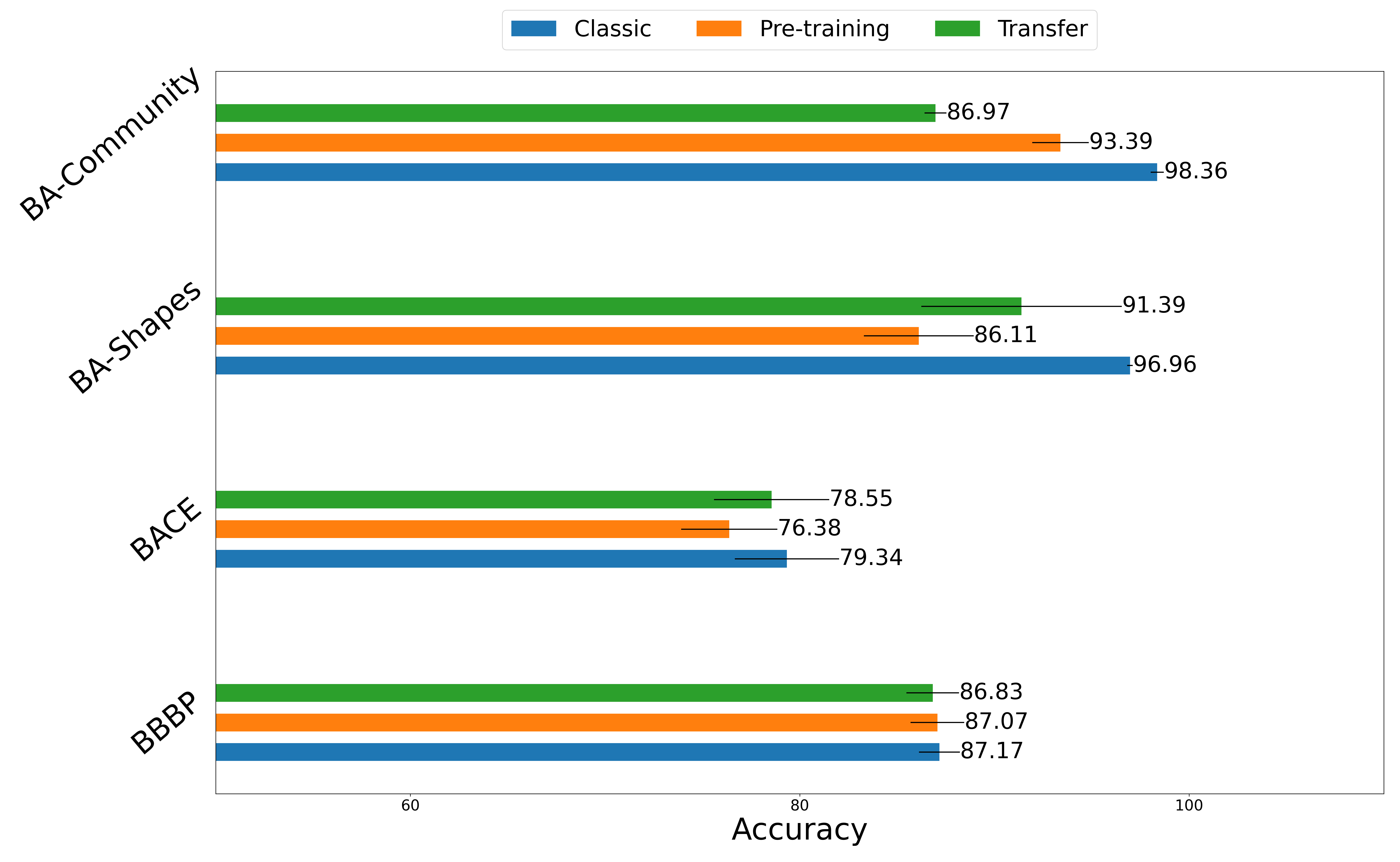}
    \caption{Avg. Accuracy and standard deviation of PIGNN when jointly trained with the feature extractor (Classic), trained after a pre-training on the same dataset (Pre-training), and fine-tuning a model trained in a different dataset via transfer learning (Transfer)}
    \label{fig:proto_pretraining}
\end{figure}
Finally, unlike other prototype-based architectures~\cite{Chen2019, Wang2021tesnet}, \gls{PIGNN} does not use pre-trained models and jointly trains both the feature extractor and the prototype layer simultaneously. Here, we explore the alternative scenario of transfer learning. It's worth noting that no publicly available pre-trained models on large graph datasets exist due to the disparate meanings of edges and nodes across different tasks. Therefore, we test transfer learning by pre-training the feature extractor in a dataset and then fine-tuning the whole model using the target dataset. Note that we select the first dataset among the ones sharing the same data type (i.e., molecular). We also consider the case where we pre-train the feature extractor and then fine-tune the same model augmented with the proposed layer on the same dataset. 

\Cref{fig:proto_pretraining} demonstrates that simultaneously training from scratch the feature encoder and the prototype layer yields superior results, outperforming the alternative in all considered tasks. The lower performance of transfer learning was expected, given that molecules from different datasets often belong to separate chemical spaces, thereby leading to the problem of the domain of applicability and lowering the general performance. On the other hand, the better performance and the large gap between simultaneous training and a pre-training phase are surprising and underscores the potential for further exploration of this phenomenon in different domains.

\section{Contributions}
\label{sec:pignn_contributions}

This chapter contributed to the research on self-explainable neural networks in the emerging field of \glspl{GNN}. Specifically, we investigated the application of prototype layers to graph data. We analyzed the differences and specific characteristics of the graph applications, along with the advantages and weaknesses associated with their utilization for enhancing the interpretability of \glspl{GNN}. 

Notably, the proposed approach preserves the performance of the black-box models while improving their interpretability in terms of fidelity with respect to their decision process. Unlike the image domain, prototype layers proved themselves to be more effective when jointly trained with the feature extractor. Moreover, they provide more sparse and easy-to-understand global and local explanations. This chapter showcased how they can be used to highlight the learned subgraphs in terms of prototypes, the most important subgraph for a specific prediction, and, more in general, to debug and improve \glspl{DNN} models. 
Averaging the quality of performance and explanations, \gls{PIGNN} represents an advancement with respect to the state-of-the-art self-explainable graph neural networks and a better choice than using black-box models coupled with extrinsic methods.

Alongside these strengths, we also analyzed some of the drawbacks of this approach. For example, the number of prototypes in prototype layers represents a parameter that is difficult to choose and impacts both the performance and quality of explanations, especially in unbalanced datasets. Additionally, despite the improvements in terms of joint training, the proposed layer still uses the custom training process of prototype-based architectures. Using these training recipes could cause instability in the training process when applied to settings where the black-box models are trained using non-standard procedures. This problem motivated us to investigate designs beyond the current self-explainable DNNs, described in the next chapter.
\chapter{Self-Explainable Memory-Augmented Deep Neural Networks}
\label{chapter:senn}
\glslocalunset{ML}
\glslocalunset{AI}
\glslocalunset{DL}
\glslocalunset{DNN}
\glslocalunset{XAI}

The previous chapter contributes to the literature in terms of generalization, thus following most of the established procedures and inheriting some of its limitations. These limitations could impact the applicability of the proposed techniques beyond the standard settings explored in the research.
To mitigate these limitations, this chapter takes a further step in exploring self-explainable deep neural networks and presents designs that augment black-box \glspl{DNN} with memory, thereby transforming them into explainable \glspl{MANN} (\Cref{sec:background_DL}).

While the augmentation of a \gls{DNN} with a memory module has already been explored to mitigate the issue of catastrophic forgetting~\cite{Graves2013, Graves2016, Franke2018} or the scarcity of data~\cite{Snell2017, Vinyals2016}, explicitly exploiting memory modules for explainability purposes represents a novelty in the field, increasing the diversity of approaches.

As in the previous chapter, the idea is to keep most of the \glspl{DNN}' structure intact. Therefore, we assume that the \gls{DNN} can be represented as the composition of two functions, the feature extractor $f_{FE}$ and the classifier $f_{CL}$, and place the novel memory modules after the feature extractor. Additionally, in this case, the designs aim to preserve the training recipe as much as possible.

Similar to other \glspl{DNN}, \glspl{MANN} are commonly regarded as black-box models. The opacity of a \gls{MANN} arises from three primary sources within its decision process: the controller, the memory, and the classifier. 
In a \gls{MANN}, the controller transforms the input representation of the current sample into a latent representation, which is then used to interact with the memory. Since the controller is often a black-box \gls{DNN}, the transformations applied to the input often lack interpretability, rendering the resultant latent representation devoid of semantics.

The memory can be challenging to interpret for various reasons. Firstly, the information stored within the memory may lack a clear link to discernible semantics. Secondly, the writing mechanism may merge new and old information, blurring the distinction between different instances. Lastly, the reading mechanism itself may be opaque, particularly when multiple readings are combined through non-linear operations.

Finally, the interpretability of the classifier depends on how the controller output and the readings are combined, in addition to the conventional factors influencing interpretability, such as the inputs and the design of the classifier itself.

Since \glspl{MANN} are more opaque than a \gls{DNN} alone, transforming a \gls{DNN} into a \gls{MANN} to improve its interpretability may seem counter-intuitive. Nevertheless, we propose \textbf{designs that exploit the advantages of \gls{MANN} and, at the same time, can effectively mitigate the challenges mentioned above via memory tracking operations}, thereby enhancing the interpretability of \glspl{DNN}. Specifically, our approaches treat the architectural components of the original black-box \gls{DNN} until the penultimate layer (feature extractor) as the controller of \gls{MANN} and place memory between the feature extractor and the classifier. Therefore, the feature extractor remains a black-box, and the proposed designs improve the interpretability of the last part of the network. 

These designs can be used to support developers in understanding and improving their models by supporting tasks like hypothesis verifications, bias, uncertainty detection, and more, in general, to better understand the decision process of these models. 


This chapter is split into three blocks: the first block provides an analysis of our proposed solution starting from the case of sequential data (\Cref{sec:sdnc}); the second block proposes a self-interpretable module for image classification (\Cref{sec:memorywrap}); the last block summarizes the results and contributions (\Cref{sec:emann_contributions}). 
\section{Simplified Differentiable Neural Computers}
\label{sec:sdnc}
This section proposes a first design of explainable \gls{MANN} able to work on sequential data. First, \Cref{sec:sdnc_design} describes the architectural design and how to retrieve explanations. Then, \Cref{sec:sdnc_experiments} evaluates the proposed architecture both in terms of performance and explanations. Finally, \Cref{sec:sdnc_design_choices} discusses several alternative design choices. 

\subsection{Design}
\label{sec:sdnc_design}
Building upon the design of \glspl{DNC}~\cite{Graves2016}, \textbf{\glspl{SDNC}}~\cite{LaRosa2020} are \glspl{MANN} fully differentiable in all their components. Given a sequential input $\bm{x}=[x_1; x_2; ...; x_n]$, the basic components of a \gls{SDNC} are the controller $f_{FE}$ and a memory $M$. The controller takes the input of the current step $t$ and returns an output $\bm{h_t}$ for each step. The memory is a matrix $M \in \mathbb{R}^{NR\times NC}$ of $NR$ row and $NC$ columns where the controller can read and write information into. Writings and readings within \glspl{SDNC} are executed using $NH_R$ read heads and $NH_W$ write heads. During the training process, the architecture learns a set of weights to transform the controller output into a set of parameters, subsequently utilized to interact with the memory:
\begin{itemize}
    \item $\bm{W}_e$ controls the information to be erased in memory;
    \item $\bm{W}_w^i$ manages the information to be written in memory by each write heads $i$;
    \item $\bm{W}_r^i$ regulates the information to be read in memory by each read heads $i$;
    \item $\bm{W}_y^i$ controls how to use the memory and controller information to produce the prediction.
\end{itemize}

First, the architecture normalizes the controller's output for the current step in a sequence: 
\begin{equation}
    \bm{h}_t = h_{i,k}^t = \frac{h_{i,k}^t-\mu_i}{\sqrt{\sigma_i^2+\epsilon}}
    \label{eq:layernorm}
\end{equation}
where $\sigma$ and $\mu$ are the mean and the variance across the controller's outputs, respectively.
Then, the model computes a series of vectors used for controlling readings and writings based on the set of parameters above-mentioned:
\begin{itemize}
    \item $\bm{e}  \in [0,1]^{NC} $, the erase vector representing the information to be erased in memory computed as $\sigma_{sig}(\bm{h}_t\bm{W}_e)$;
    \item $\bm{v} \in \mathbb{R}^{NC}$, the write vector representing the information to be written into memory, computed as $\bm{h}_t\bm{W}_w$;
    \item  $\bm{k} \in \mathbb{R}^{NH_R\times NC}$, a set of read keys, one for each read head;
    \item $\beta^i \in [1, \infty]^{NH_R}$, the read strengths of the read heads.
\end{itemize}

Readings and writing operations are performed using attention mechanisms, employing dynamic memory allocation for writing and content-based addressing for reading.



Dynamic memory allocation is a differentiable mechanism used to free memory. This mechanism updates the \emph{usage vector} $\bm{u}_t \in [0,1]^{NR}$, which tracks the usage of each location in memory. Specifically, the location usage is incremented upon being written in the preceding step:

\begin{equation}
    \bm{u}_t = \bm{u}_{t-1} + \bm{w}^w_{t-1} - \bm{u}_{t-1}\bm{w}^w_{t-1}
    \label{eq:usage_vector_sdnc}
\end{equation}
Based on the usage vector and a list $\bm{\phi}_t$ containing  memory location indices sorted by ascending usage, the architecture computes the \emph{allocation vector}:
\begin{equation}
    \bm{a}_t[\bm{\phi}_t[j]] = (1 - \bm{u}_t[\bm{\phi}_t[j]]) \prod_{i=1}^{j-1}\bm{u}_t[\bm{\phi}_t[i]]
    \label{eq:allocation_vector_sdnc}
\end{equation}
When the location usage is 1, the corresponding index in the allocation vector is set to 0, indicating that the memory location cannot be overwritten. Conversely, when the usage is 0, the location is available for writing operations and the allocation is set to 1. When the usage is between 0 and 1, the allocation is weighted by the position of the location in the sorted list: a higher position results in a lower allocation value. Subsequently, the write heads combine the allocation vector with the \emph{writing gate} to generate the \emph{write weightings}:
\begin{equation}
    \bm{w}_t^w = \bm{g}_t^w\bm{a}_t
    \label{eq:write_weights_sdnc}
\end{equation}
The \emph{writing gate} controls the extent to which information from the current step should be written. Therefore, if the gate is 0, no memory writing occurs during the current step. The architecture then updates each memory location by erasing information based on the \emph{erase vector}, writing the information contained in the \emph{write vector}, and controlling the information flow using the \emph{writing weightings}:
\begin{equation}
    \bm{M}_t = \bm{M}_{t-1}\circ(1 -  \bm{w}^w\bm{e}^T) + \bm{w}^w\bm{v}^T 
    \label{eq:writing_sdnc}
\end{equation}
Thus, a location is partially erased when both the erase vector and the write weightings do not contain a 0 in the index corresponding to that location. A location is written when neither the write weightings nor the write vector contain a 0 in the index corresponding to that location.

Read operations within the architecture rely on the \emph{content addressing} mechanism. Content addressing is determined by the cosine similarity between two vectors, as defined by the equation:
\begin{equation}
    D(\bm{v}_1, \bm{v}_2) = \frac{\bm{v}_1\cdot\bm{v}_2}{\lVert \bm{v}_1 \rVert \lVert\bm{v}_2 \rVert}
    \label{eq:cosine_similarity}
\end{equation}
In this context, the two vectors under consideration are the memory of the previous step $\bm{M}_{t-1}$ and the read keys computed based on the current output of the controller. Thus, the architecture calculated the read weightings as the softmax of the cosine similarity between the memory of the previous step and the read keys, weighted by the read strength:

\begin{equation}
    C(\bm{M}, \bm{k}, \beta)[i] = \frac{e^{D(\bm{k},\bm{M}[i])\beta}}{\sum_je^{D(\bm{k},\bm{M}[j])\beta}}
    \label{eq:content_addressing_sdnc}
\end{equation}

\begin{equation}
    \bm{w}_t^{r,i} = C(\bm{M}_{t-1},\bm{k}_t^{r,i},\beta_t^i)
    \label{eq:read_weights_sdnc}
\end{equation}
The read weightings are then used to compute the \emph{read vectors} as the weighted sum of the memory content:
\begin{equation}
    \bm{r}_t^i = \bm{M}^T_t\bm{w}_t^{r,i}
    \label{eq:read_vector_sdnc}
\end{equation}
Finally, the architecture concatenates the read vectors and the controller's output and sends them to the classifier to compute the prediction. To encourage the usage of memory during the inference process, a dropout is applied to the controller's output.
\begin{equation}
    y_t = W_y((\bm{h}_t\bm{p}_t) \oplus \bm{r}_t^1 \oplus \bm{r}_t^2 \oplus ... \oplus \bm{r}_t^{N_r})
    \label{eq:sdnc_output}
\end{equation}
\begin{equation}
    \label{eq:droput}
    \bm{p}_t = \sum_{i=0}^np_i\in [0,1] \sim Bernoulli(r_p) 
\end{equation}
where $r_p$ is the dropout rate that controls how many elements of the vectors should be set to zero on average.

Regarding the differences with \gls{DNC}, the latter includes more parameters than the \gls{SDNC}, a different write mechanism, a different read mechanism, and it includes a double recurrence. Specifically, \gls{DNC} includes two additional gates, a set of write strengths, and a set of write keys. The writing mechanism also considers the \emph{content weights} to compute the write weightings. Therefore, the architecture can write new information in a location containing similar content even if the usage vector is high. The double recurrence refers to concatenating the input with the read vectors of the previous steps. While these features are important to reach higher performance than the controller alone, they lower the interpretability of memory operations since they lead to overwriting operations or increase the opacity of the semantics encoded in a state. Finally, \glspl{DNC} also include the temporal linkage matrix for computing read weightings, which we removed following the same analysis of \citet{Franke2018}.

\subsubsection{Explanaining Via Memory Tracking}
\label{sec:sdnc_expla}
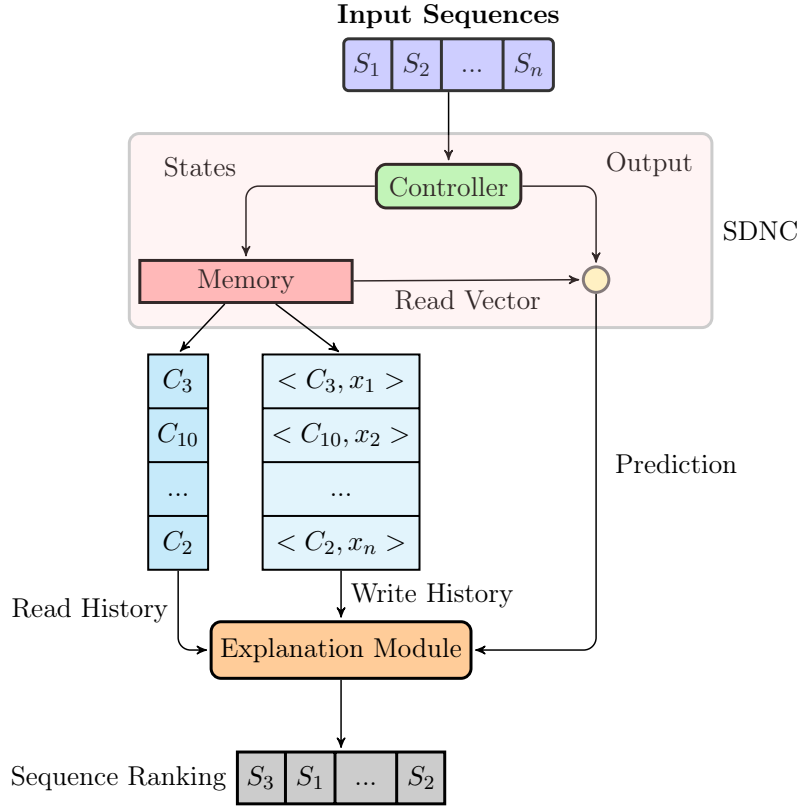
\begin{figure}[!t]
  \centering
  \begin{tikzpicture}%
    \node[rectangle split, rectangle split draw splits=true, rectangle split horizontal,
    rectangle split parts=4] (Sequence) [label=above:\textbf{Input Sequences},
    draw=black!100, very thick, rounded
    corners=2pt,
    fill=blue!20!white!120,opacity=0.8] {\strut $S_1$\nodepart{two}\strut $S_2$\nodepart{three}\strut ~...~~\nodepart{four}\strut $S_n$};
    \node[squarednode,inner sep=5pt, below=of Sequence, fill=green!30] (LSTM){Controller};
    \draw[post,rounded corners=5pt] (Sequence.five south) -- (LSTM);
    \node[below left=0.7cm and 0.3cm of LSTM, squarednode,rounded
    corners=0pt, fill=red!30,minimum width=2cm, text width=1.5cm, minimum width=3cm] (Memory) {Memory};
    \node[roundnode,below right=0.85cm and 0.9cm of LSTM, fill=yellow!30]  (concatLM) {};
    \draw[post,rounded corners=5pt] (LSTM) -| node[above right](out) {Output}(concatLM);
    \draw[post,rounded corners=5pt] (LSTM) -| node[above left](states) {States}(Memory);
    \draw[post,rounded corners=5pt] (Memory) -- node[below](rv){Read Vector}(concatLM);
    \node [label=right:{SDNC}, draw=black!100, very thick,
    rounded corners,
    fill=red!20,opacity=0.2,fit={(Memory) (concatLM) (LSTM)(states)(out)(rv) }] {};
    \node[below=of concatLM]  (Output) {};
    \node[below=of Output,label=right:Prediction]  (Dummy3) {}; 
    \node[vector=4, below right=0.7cm and -1.3cm of
    Memory,thick,fill=cyan!20!white!50](WH)
    {\strut $<C_3,x_1>$\nodepart{two}\strut $<C_{10},x_2>$\nodepart{three}\strut ...\nodepart{four}\strut $<C_2,x_n>$};
    \node[vector=4, thick, below left=0.7cm and -1cm of Memory,
    minimum height=2in,fill=cyan!20!white!100](RH)
    {\strut $C_3$\nodepart{two}\strut $C_{10}$\nodepart{three}\strut ...\nodepart{four}\strut $C_2$};
    \draw[post,rounded corners=5pt] (Memory) -- (RH.north);
    \draw[post,rounded corners=5pt] (Memory) -- (WH.north);
    \node[below=0.7cm of WH,squarednode, fill=orange!40, minimum height=0.8cm]  (Explanation) {Explanation Module}  ; 
    \draw[post, rounded corners=5pt] (RH) |- node[left, near start] {Read History}(Explanation);
    \draw[post, rounded corners=5pt] (WH) -- node[right]{Write History}(Explanation);
    \draw[post,rounded corners=5pt] (concatLM) |- (Explanation);
    \node[vector=4, below=of Explanation, rectangle split horizontal,
    label=left:Sequence Ranking, fill=black!20, very thick](Premises)
    {\strut $S_3$\nodepart{two}\strut $S_1$\nodepart{three}\strut ~...~~\nodepart{four}\strut $S_2$};
    \draw[post,rounded corners=5pt] (Explanation) --(Premises);
  \end{tikzpicture}
  \caption{Design of SDNC. The blue boxes represent the read-and-write history. 
    $<C_i,x_j>$ indicates that feature $x_j$ is stored in cell $C_i$. The ranking sorts the subsequence of the input based on their importance.}
  \label{fig:sdnc}
\end{figure}
As introduced in the previous section, the idea is to treat the architectural components of the controller as a black-box and improve the interpretability of the last part of the network by exploiting the memory layer. We can observe that information stored and read from the memory corresponds to the states and outputs of the black-box controller. Therefore, for interpretability purposes, the idea is to extract patterns in the sequence of outputs and connect these patterns to the final prediction. To achieve the goal, the proposed \textbf{memory tracking mechanism monitors both the reading and writing operations to derive explanations} (\Cref{fig:sdnc}). During the writing process, the tracking focuses on the writing weightings. The objective is to establish a mapping between steps and memory locations that were written during each step. As previously explained, a location is written if the value in the writing weightings corresponding to that location is zero. However, due to the differentiability of the operators, the weighting values of the are never exactly zero but rather close to it. Therefore, we relax the definition of written locations to introduce more flexibility. Let be $\mu_t^w$ the average value of the write weightings:

\begin{equation}
    \mu_t^w = \frac{\sum\bm{w}_t^w}{NR}
\end{equation}
Then the written locations at the step $t$ are represented by the vector  of indices $\bm{ind}_t^w$, computes as:
\begin{equation}
    \bm{ind}_t^w = \{j ~|~ \bm{w}_t^w[j] > \mu_t^w\}_{topk^w}
\end{equation}
where ${topk^w}$ is a hyperparameter and the notation $~_{topk^w}$ indicates that only the highest ${topk^w}$ locations are retained among those satisfying the condition. In other words, a location is considered written if its write weight is above the mean and is among the ${topk^w}$ highest values in $\bm{w}^w$.

Similarly, during the reading process, the tracking focuses on the read weightings. In this case, the goal is to establish a mapping between steps and locations that were read during each step. A location can be considered read if the value in the read weightings is greater than zero. However, a small value has little or no impact on the decision process. Therefore, we propose a more strict definition of read locations. The average value of the read weighting is defined as:
Let be $\mu_t^{r,i}$ the average value of the read weightings of the read head $i$:
\begin{equation}
    \mu_t^{r,i} = \frac{\sum\bm{w}_t^{r,i}}{NR}
\end{equation}
The read locations at the step $t$ are represented by the vector of indices $\bm{ind}_t^{r,i}$, computed as:

\begin{equation}
    \bm{ind}_t^{r,i} = \{j~|~ \bm{w}_t^{r,i}[j] > \mu_t^{r,i}\}_{topk^r}
\end{equation}
Here, ${topk^r}$ and $~_{topk^r}$ are parameters similar to those defined for writing operations but specific for reading operations. 
In summary, a location is considered read if its read weight is above the mean and is among the ${topk^r}$ highest values in $\bm{w}^{r,i}$.

Subsequently, the module ranks the features based on the frequency of readings. The ranking construction is flexible and can be adapted to tasks and needs related to explainability.




In the case of a multi-step decision process, a common scenario in \glspl{MANN}, the readings of the steps in the decision process are summed, and \textbf{attribution is assigned based on the frequency of readings}: 
\begin{equation}
    a(t_k:t_m) = \sum_{t_{f_0} < t_f < t_{f_n}}~\sum_{k < t < m}~\sum_{0<i<R}|{j :  j \in \bm{ind}_{t}^{w} \And j \in \bm{ind}_{t_f}^{r,i}}| 
    \label{eq:sdnc_memtrack}
\end{equation}
where $a(t_k:t_m)$ is the score assigned to the subsequence $[x_{k}:x_m]$, $t_{f_0}$ is the first step of the decision process, and $t_{f_n}$ is the last step of the decision process. The equation is the most general one, encompassing more specific cases like the feature-level attribution (setting $t_f = t_{f_0} = t_{f_n}$) and single-step decision processes (setting $t = k = m$). The explanation process is summarized in \Cref{fig:sdnc}.

\subsection{Experiments}
\label{sec:sdnc_experiments}
\subsubsection{Performance}
\gls{SDNC} is proposed as a network augmentation for a black-box model to enhance its interpretability. This section compares the performance achieved by the standalone black-box model and our proposed augmentation. 
While the design draws inspiration from \glspl{DNC}~\cite{Graves2016} and Advanced \glspl{DNC}~\cite{Franke2018}, \gls{SDNC} is not intended to enhance the controller's performance but rather to improve its interpretability. Specifically, we removed certain components of \gls{DNC} that might complicate interpretability while potentially boosting performance. Therefore, we expect \gls{DNC} to outperform \gls{SDNC}, and we do not consider \gls{DNC} and its variants as competitors.

 \begin{figure}[t]
 \centering
  \begin{tikzpicture}%
    \tikzset{
      wall/.style={line width=1mm},
    }
    \draw[wall, black] (-0.05,0.5)--(5.,0.5)--(5.0,1.5)--(6.0,1.5)--(6.0,-1.5)--(5.0,-1.5)--(5.0,-0.5)--(0.0,-0.5)--(0.0,0.5);
    \begin{scope}[on background layer]
      \draw[ultra thin,name path=0A] (0,0.5)--(0,-0.5);
      \draw[thick,name path=1A] (1,0.5)--(1,-0.5);
      \tikzfillbetween[of=0A and 1A]{gray, opacity=0.5};
      \draw[thick,name path=2A] (2,0.5)--(2,-0.5);
      \draw[thick,name path=3A] (3,0.5)--(3,-0.5);
      \tikzfillbetween[of=1A and 3A]{gray, opacity=0.5};
      \draw[thick,name path=4A] (4.,0.5)--(4.,-0.5);
      \tikzfillbetween[of=3A and 4A]{yellow, opacity=0.5};

      \draw[thick,name path=5A] (5,0.5)--(5,-0.5);
      \draw[thick,name path=6A] (6,0.5)--(6,-0.5);
      \tikzfillbetween[of=4A and 6A]{gray, opacity=0.5};
      
      \draw[thick, name path=UpA] (5,0.5)--(6,0.5);
      \draw[ name path=UpB] (5,1.5)--(6,1.5);
      \tikzfillbetween[of=UpA and UpB]{green, opacity=0.5};

      
      \draw[thick,name path=DownA] (5,-0.5)--(6,-0.5);
      \draw[ultra thin,name path=DownB] (5,-1.5)--(6,-1.5);
      \tikzfillbetween[of=DownA and DownB]{red, opacity=0.5};
    \end{scope}
    \node[shape=circle,draw,minimum size=8mm, fill=white!!100,inner sep=0pt] at (0.5,0){2};
    \node at (1.5,0){S};
    \node at (2.5,0){N};
    \node at (3.5,0){N};
    \node at (4.5,0){S};
    \node at (5.5,0){N};
    \node at (5.5,1){4};
    \node at (5.5,-1){-.1};
    \draw[->] (0.5,1.1)node[above] {Relevant Cell Indicator}--(0.5,0.6);
    \draw[->] (-0.6,0)node[above] {Start}--(-0.05,0);
    \draw[->] (4.4,1) node[left] {Success}--(4.95,1);
    \draw[->] (4.4,-1) node[left] {Fail}--(4.95,-1);
    \draw [->] (0.5,0-0.5) to [out=-30,in=-150] node[below]{Relevant Cell} (3.5,-0.5);
  \end{tikzpicture}
  \caption{Example of a 5-step modified T-maze, where the starting
    position (in white) indicates the relevant cell (in yellow).} \label{fig:tmaze}
\end{figure}
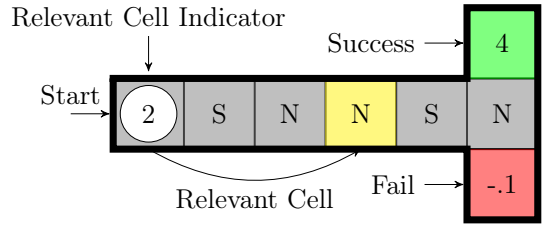
We evaluate the performance of \gls{SDNC} on a modified T-Maze task and the Cloze Story Test. For both tasks, we use a feedforward LSTM layer with 128 units as a black-box and as a controller of {SDNC}. The choice is motivated by the fact that this model has already been successfully tested in all the scenarios considered in this chapter: on T-Maze~\cite{Bakker2002}, on Story-Cloze test~\cite{Mihaylov2017}, and as a controller of \gls{DNC} variants~\cite{Franke2018, Graves2016}. 

T-maze~\cite{Bakker2002} is a non-Markovian discrete control task where an agent navigates a corridor of variable length from a starting position to a T-junction. The agent can move North, East, South, or West at each step. In our modified T-maze version, at each step, the agent observes a symbol indicating the direction to take at the junction to reach the goal. The starting position specifies the index of the step containing the right suggestion. All the other symbols are assigned randomly. The agent receives a reward of 4 for the correct action and a reward of ${-0.1}$ otherwise (see \Cref{fig:tmaze}). We cast the problem as a classification problem and train all the models on a dataset of 2000 samples, split following the rule of 80/20 between the train and test datasets. 

The Story Cloze Test~\cite{mostafazadeh2016} is a commonsense reasoning task to predict the correct ending of a story composed of four sentences (\emph{premises}). For this task, we embed each word using word2vec~\cite{Mikolov2013} and append three boolean flags to each vector, indicating whether the word is included in the story, the first ending, or the second one. We also include a zeroed query vector to request an answer from the model. The training dataset is split into training and validation data (10\% of the training data). 

We train the model for ten epochs in T-Maze and use an early stopping with patience equal to ten epochs in the Story Cloze Test. The model associated with the highest accuracy on the validation data is considered the final model. In the case of the Story Cloze Test, we adopt the configuration of \citet{Mihaylov2017} for the controller configuration, using the encoding of the last step of the story as the initial state of the encoding of both the possible endings. 
Regarding the memory parameters, in the Story Cloze Test, we fix the memory size to 512, the read heads number to 4 read heads, and the write heads number to 1. We apply the Adam optimizer using a starting learning rate of 1e-3 and a dropout rate equal to 0.2. Conversely, in T-Maze, we use the RMSprop optimizer using the default parameters of the Keras implementation\footnote{\url{https://www.tensorflow.org/api_docs/python/tf/keras/optimizers/experimental/RMSprop}}, and we fix the memory size to 50, and both the read heads and the write heads number to 1. 



\begin{figure}[t]
    \centering
    \includegraphics[scale=0.3]{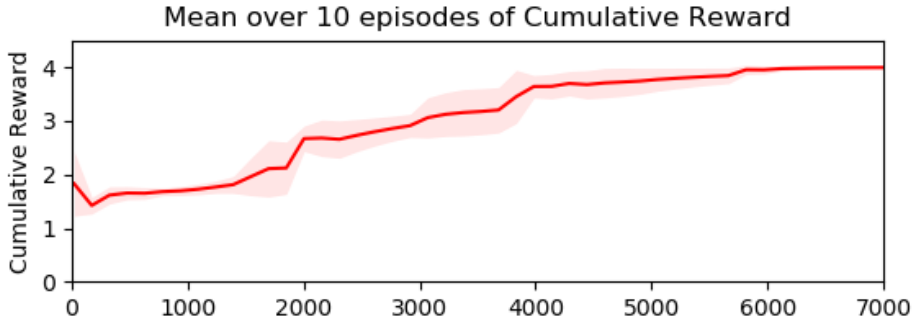}
    \caption{Cumulative reward and standard deviation of ten agents trained on the T-Maze problem.}
    \label{fig:tmazereward}
\end{figure}

As depicted in \Cref{fig:tmazereward} and \Cref{tab:clozeaccuracy}, the network effectively solves the task in the T-Maze case and \textbf{attains comparable generalization power} in terms of accuracy in the Story Cloze Test when compared to the controller alone. The black-box performance is retrieved from \citet{Mihaylov2017}. While the achieved performance is promising, it is important to acknowledge some weaknesses. Due to the increased parameters in \gls{SDNC}, the training process is slower (over 2x). Additionally, note that \textbf{\gls{SDNC} employs a distinct configuration} compared to the one of \citet{Mihaylov2017} with a lower learning rate, a smaller number of units, and a different batch size (1). Indeed, while the black-box benefits from large batch sizes~\cite{Mihaylov2017}, we observe that \gls{SDNC} performs optimally when trained using small batch sizes. Therefore, we expect that a user transitioning from a black-box to an \gls{SDNC} should conduct an optimization search over the controller configuration parameters to achieve comparable performance.

\begin{table}[b!]
    \caption{Average test Accuracy reached by the models in the testing Story Cloze Test dataset.}
    \centering
        \begin{tabular}{lr}
            \toprule
            Model&Test Acc. \% (Avg)\\
            \midrule
            Black-Box &0.72\\
            SDNC&0.72
            \\
            \bottomrule
        \end{tabular}
    \label{tab:clozeaccuracy}
\end{table}


\subsubsection{Explanations}
\label{sec:sdnc_expla_exp}
The main advantage carried on by the proposed \gls{SDNC} is on the interpretability side. Indeed, the controller is a black-box model and does not provide any way to inspect its decision process. Conversely, this section showcases and evaluates explanations obtained by applying the memory tracking mechanism (\Cref{sec:sdnc}) to \gls{SDNC}.

In the T-Maze problem, users could be interested in \textbf{verifying whether the models learned as intended} and leverage information from the relevant step corridor to solve the task. In this case, since the rules governing the dataset are known, we can retrieve the ground truth explanations (\Cref{sec:back_xai}), which correspond to the relevant step corridor. 
To achieve the goal, memory tracking specializes \Cref{eq:sdnc_memtrack} by collecting the two most read cells during the decision step. We consider two locations to also account for the first corridor step, which could be equally significant. To evaluate these explanations, we check the frequency (\emph{matching accuracy}) with which the memory locations containing that information rank among the first two positions in the memory tracking mechanism's output.

\begin{figure}[!t]
    \centering
    \includegraphics[scale=0.3]{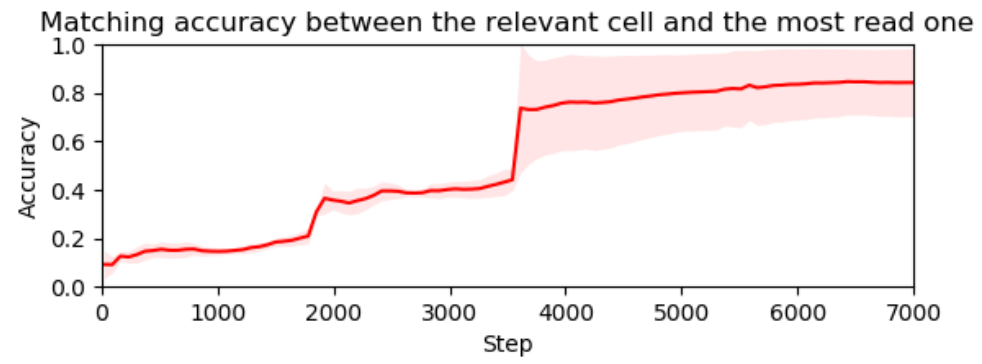}
    \caption{Matching accuracy score and standard deviation of ten agents trained on the T-Maze problem.}
    \label{fig:tmaze_matching}
\end{figure}

\Cref{fig:tmaze_matching} illustrates the agent's gradual improvement in the accuracy of its explanations during the learning process. Observing the similar behavior in the curves of \Cref{fig:tmaze_matching} and \Cref{fig:tmazereward}, we infer that, as expected, better detection of the relevant step corridor among the most read locations corresponds to improved task accuracy.

\begin{figure}[t]
	\centering
	\begin{tikzpicture}[every node/.style={inner sep=0,outer sep=0}]
	\node(out1){
		\resizebox{4in}{!}{
			\begin{tabular}{l}
			\best{Earl woke up early to make some coffee}. (48.3\%) He wanted to be alert
			for work\\ that day. (47.4\%) \worst{The aroma woke up all his
				roommates}. (0\%) They
			wanted to \\ make coffee too. (4.2\%)\\[-6pt]\\
			
			E1. \prediction{All of his roommates made coffee} (\textbf{CORRECT}) -- E2. All of his roommates\\
			were sick of coffee.\\
			
			\end{tabular}
	}};;
	\node [draw=black!100, very thick, rounded corners,fill=white!50,opacity=0.2,fit={  (out1) }]{};
	\node[below=0.01in of out1](out2){
		
		\resizebox{4in}{!}{
			\begin{tabular}{l}
			\worst{Samantha had recently purchased a used car}. (15.6\%) She loved
			everything about\\ the car except for the color. (30.3\%) \best{She took her
				car to her local paint shop}. (31\%)\\  She got it painted a bright pink
			color. (23\%)\\[-6pt]\\
			\prediction{E1. Samantha likes the color of her car now.} \textbf{(CORRECT)} --
			E2. Samantha thinks \\ her bus looks pretty now.\\
			\end{tabular}
			
	}};;
	\node [draw=black!100, very thick, rounded corners,fill=white!50,opacity=0.2,fit={  (out2) }]{};
	\node[below=0.01in of out2](out3){
		
		\resizebox{4in}{!}{
			\begin{tabular}{l}
			Tim didn't like school very much. (23.6\%) His teacher told
			him he had a test \\on Friday. (15\%)
			\worst{If he didn't pass this test, he could not go on the class
				trip}. (4.5\%) \\\best{Tim decided to play with his kites instead of
				study for the test}. (56.8\%)\\[-6pt]\\
			
			E1. Tim was unprepared and failed the test. -- \wrong{E2. Tim aced
				the test and passed with} \\ \wrong{flying colors.} \textbf{(WRONG)}\\
			\end{tabular}
			
	}};;
	\node [draw=black!100, very thick, rounded corners,fill=white!50,opacity=0.2,fit={  (out3) }]{};
	\node[below=0.01in of out3](out4){
		
		\resizebox{4in}{!}{
			\begin{tabular}{l}
			\best{Neil took a ferry to the island of Sicily}. (87.2\%)
			\worst{The wind blew his hair as he} \\ \worst{watched the waves}. (0\%)
			\worst{Soon it docked, and he stepped onto the island}. (0\%)\\
			It was so breathtakingly beautiful. (12.7\%)\\[-6pt]\\
			
			\prediction{E1. Neil enjoyed Sicily} \textbf{(CORRECT)} --  E2. Sicily was the worst place neil had\\ ever been.\\
			\end{tabular}
		}
	};;
	\node [draw=black!100, very thick, rounded corners,fill=white!50,opacity=0.2,fit={  (out4) }]{};
		
	\end{tikzpicture}
	\caption{Example outputs on the Story Cloze Test. A relevance score
		is associated with each premise. The blue color indicates the best premise, while the orange indicates the worst one.}
	\label{fig:sdnc_output}
\end{figure}
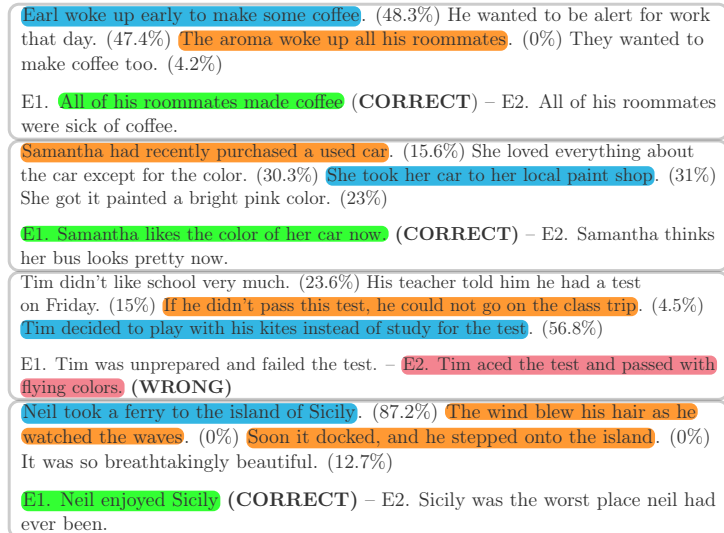

In the Story Cloze Test, we evaluate the memory-tracking mechanism's precision in \textbf{highlighting the story's most important subsequence} (premise) using the $Fidelity-^{acc}$ metric. In this case, the memory tracking mechanism, during each step of parsing the predicted ending, collects the ten most read locations. These locations are mapped to the stored words in memory, and a premise ranking is built based on the reading frequency, as explained in \Cref{eq:sdnc_memtrack}. Three kinds of premises are identified from the ranking: the best, the worst, and a random one. The \emph{best premise} is determined by the most frequently read words, the \emph{worst premise} by the least frequently read words, and the \emph{random premise} is arbitrarily chosen. \Cref{fig:sdnc_output} provides some examples of the computed explanations and their associated frequency. Each of these premises is then fed as isolated input to the network, and their predictions are collected. Intuitively, if a premise is important, the network should more consistently achieve the same prediction as when fed the full input, compared to a random premise.
\begin{table}[b!]
	\caption{Avg. fidelity over ten runs on the
		Story Cloze Test when feeding the model with only a random, the
		best and the worst premise.}
	\centering
	
		\begin{tabular}[t]{rccc}
			\toprule
			\multicolumn{4}{c}{Avg. $Fidelity-^{acc}$}\\
			\midrule
			Premise&Train&Dev&Test\\
			\midrule
			\quad Random& 0.57& 0.60& 0.56\\
			\quad Best&0.66&0.73& 0.64\\
			\quad Worst&0.51&0.53& 0.54\\
			\bottomrule
		\end{tabular}

	\label{tab:premise_cloze}
\end{table}
\Cref{tab:premise_cloze} validates the quality of the approach, indicating that the best premise achieves higher fidelity than the random premise on the training, development, and testing datasets. Additionally, the worst premise attains the lowest fidelity, reinforcing the reliability of the ranking returned by the memory tracking mechanism.

To ensure that the elements used for explanations are connected to the decision process, two additional tests are performed:  parameter randomization and data randomization tests. 
These tests evaluate the link between learned parameters, labels, and explanations~\cite{Adebayo2018} starting from the observation that \textbf{good explanations should be linked to the learned parameters and the input-label mapping}. 
The parameter randomization test checks the fidelity of the best and worst premises on an untrained network, while the data randomization test shuffles the labels in the training dataset, trains a model by using it, and evaluates the fidelity of explanations on the test dataset.
\begin{table}[b]
        \caption{Accuracy and $Fidelity-^{acc}$ when the model is untrained and then labels are shuffled.}
    \centering
    
    \begin{tabular}{llrr}
            \toprule
			&&\multicolumn{2}{c}{Avg. $Fidelity-^{acc}$}\\
         Randomization Test & Accuracy & Best & Worst \\
			\midrule
         
         Parameters & 0.51 & 0.56 & 0.55\\
         Data & 0.59 & 0.52 & 62.8 \\
         \bottomrule
    \end{tabular}

    \label{tab:randomization}
\end{table}

\Cref{tab:randomization} shows that the best and the worst fidelity are nearly the same in the parameter randomization test, implying that the method
cannot differentiate between a good and a bad premise in both cases. This result indicates that memory tracking correctly depends on the model's learned
parameters. The results of the data randomization test are similar, with the best premise accuracy failing to predict the original outcomes, confirming the link with the learned mapping. Moreover, we observe that the worst premise outperforms the best one. The cause of such behavior is an attempt of
the model to exploit the changed labels. Indeed, when a label is altered, the model should select the premise with the minimum similarity with the memory output to predict the changed label. This behavior is rewarded by some of the correctly labeled examples due to the overabundant nature of the task.

\begin{figure}
    \centering
    \includegraphics[scale=0.4]{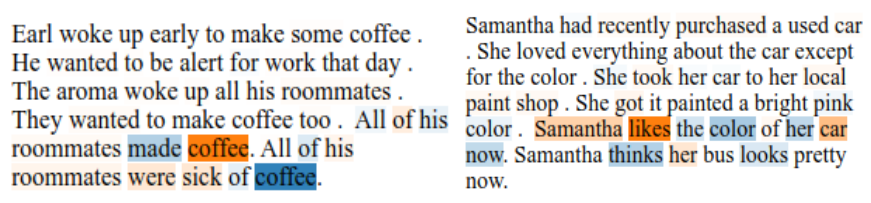}
    \caption{Output of LIME~\cite{Ribeiro2016} applied to a couple of samples from the Story Cloze Test dataset.}
    \label{fig:lime}
\end{figure}
The explanations retrieved for the two benchmarks underscore the \textbf{flexibility of memory-tracking} to adapt to the user query. For comparison, consider \Cref{fig:sdnc_output} and \Cref{fig:lime}, which presents the explanations returned by LIME~\cite{Ribeiro2016} for a couple of samples in the Story Cloze Test. It is easy to see that LIME focuses on words in the endings, indicating which words influence the prediction towards the first ending (blue) or the second one (orange) if removed. A similar output can be achieved by considering the readings of the endings' words, instead of the premises ones, using the memory tracking mechanism (i.e., the same approach used in the T-Maze task). Conversely, adapting LIME to extract which premises are important is challenging and requires a redesign of LIME or a modification at the model level. Similar to other model-agnostic methods, LIME treats the model as a black-box, and it simply analyzes the relation between modified inputs and the outputs. Consequently, guiding the explanation process to answer specific user questions becomes difficult. While \citet{Ribeiro2016} suggest using super-pixels permutations to extract feature attribution for a group of features (such as the premises), finding thousands of permutations of premises that preserve the semantics of the story is not trivial for sequential and text data. The last alternative to assign scores to the premises is to aggregate attributions at the word level into a sentence-level one. However, as mentioned earlier, since there is no way to impose constraints on the input, the highest-scored words are those in the endings, while all the words of the premises receive low associated weights, making the entire process unreliable. Conversely, in the \gls{SDNC} case, it is sufficient to modify the indices of \Cref{eq:sdnc_memtrack} to switch from one case to another, making these adaptations trivial and more preferable than post-hoc extrinsic methods.

While these results are encouraging and represent a big step forward with respect to the black-box controller, we note two weaknesses related to explanations based on \gls{SDNC}. The first is related to the skip connection that connects the controller to the classifier. While the scores are reliable on average, as highlighted by the fidelity and the additional tests, there could be cases where the classifier ignores the memory output and performs the classification based only on the controller output. These cases represent a current limitation but can be detected by using external means (like the gradients) to inspect the contribution of each element in the read vectors, thus combining the advantages of both approaches. The second limitation regards the input dispersion. Indeed, we observe that applying memory tracking for individual feature attribution (i.e., assigning scores to individual features) is less effective. In these cases, the highest scores can be associated with later steps, like punctuation at the end of the sentence. The nature of the controller itself causes this problem since LSTMs store information on their memory cells, and thus, consecutive steps can have similar information~\cite{LaRosa2021}. This problem is also the cause of the variance observable in \Cref{fig:tmaze_matching}. \Cref{sec:memorywrap} addresses some of these problems by proposing a different memory-based design.

\subsection{Design Choices}
\label{sec:sdnc_design_choices}
This section delves into the hyperparameters and design choices of \glspl{SDNC}, highlighting their connection to the interpretability of the results.

\paragraph{Memory Size.}
The first parameter to consider is the size of the memory. The settings used in the previous examples include a large memory size to minimize overwriting the same locations. Indeed, given the design of \gls{SDNC}, the network is forced to utilize empty locations as long as possible. Once the memory is full, \gls{SDNC} selects the least used locations as candidates for the writing operations. As the only way to increase the usage of a location is by writing on it, the least used locations are those written during the first steps of the sequence. Future work could explore adjusting the design of \gls{SDNC} to improve this mechanism. A simple alternative could be to increase the usage of read locations, but this choice is task-dependent since there are several tasks where a location could be useful only at the end (e.g., T-Maze), and, thus, it should be read only at the end of the task. Therefore, we suggest using a memory big enough to avoid overwritings. This mechanism represents one of the directions that future work could investigate to improve the design of \gls{SDNC}. 
Concerning the number of read heads and write heads, the reader is encouraged to refer to the \gls{DNC} paper~\cite{Graves2016}. There are no differences between single and multiple heads in terms of interpretability. The only aspect to consider is that using more write heads implies more writing, and thus, a larger memory is needed to avoid overwriting.

\begin{table}[b]
	\caption{Avg. fidelity obtained on the
	development Story Cloze Test dataset using
	different $topk^r$ values.}
	\begin{center}
		\begin{tabular}{rccccc}
			\toprule
			\multicolumn{6}{c}{Avg. $Fidelity-^{acc}$}\\
                \midrule
			&\multicolumn{5}{c}{Number of locations}\\
			Premise&1&5&10&25& $|\bm{ind}_t^{r,i}|$\\
			\midrule
			Best&\bf{0.76}&\bf{0.76}&0.73&0.71&0.40\\
			Worst&0.60&0.60&0.53&\bf{0.40}&0.49\\
			\bottomrule
		\end{tabular}
	\end{center}
	\label{tab:cloze_threshold}
\end{table}
\paragraph{Number of Locations for explanations.}
Another important hyperparameter is ${topk^r}$, which controls how many locations to consider for each step. This hyperparameter should tuned in a validation set or selected based on the task. For example, in the T-Maze problem, the number is set to two based on the desired behavior to probe and the problem's structure. Conversely, in the case of the Story Cloze Test task, we perform a grid search over the values $\{1, 5, 10, 25\, > \mu_t^{r,i}$, as shown in \Cref{tab:cloze_threshold}.

When considering a smaller number of read locations for each step, both the best and worst premise increase their fidelity. In contrast, when considering a larger number of read locations, both premises lower their scores. Therefore, ${topk^r}$ can be selected based on the objective and the desired explanations. If we are interested only in the precision of the most important premise, then fewer steps are the best choice. Conversely, for the entire ranking, a larger number of steps increases the gap between indices.

Note that the number of locations changes the type of promoted readings. For example, in the case of \Cref{tab:cloze_threshold}, if a few locations are selected for each step, at the end of the ending parsing, it is unlikely that the read words are associated with all the premises. Often, they are associated with only one or two premises, making the choice of the worst premise nearly random. However, these few selected locations are associated with the highest content weights and are highly influential; thus, the identification of the best premise is reliable. Conversely, if a large number of locations are selected for each step, all the premises will be represented, and the choice of the worst premise becomes more reliable. Instead, the choice of the best premise is influenced by the noise of the multiple readings, which reward words read most often rather than words read more constantly. Therefore, a premise read multiple times but in a single step can be scored as higher than one read more constantly during the parsing but using few locations per step. 
\section{Memory Wrap}
\label{sec:memorywrap}
The previous section described a first attempt at using memory modules for enhancing the interpretability of \glspl{DNN}. \gls{SDNC} is designed to deal with data sequence since \glspl{MANN} have been historically applied in this domain. Being able to keep most of the training process intact, the \gls{SDNC} design represents a step forward with respect to the prototype layers proposed in \Cref{chapter:pignn}. However, we observed issues related to the controller dealing with data sequence.

Starting from this analysis, here we further develop the employment of memory modules for enhancing the interpretability of \glspl{DNN}. The goal is to take a step forward to mitigate the issues highlighted in the previous section and investigate memory modules and the memory-tracking mechanism beyond the data sequence domain.

In particular, this section investigates the augmentation of popular \glspl{DNN} with memory layers in a non-traditional domain for \glspl{MANN},  the vision domain. In this context, the few available memory-based architectures~\cite{Vinyals2016, Snell2017} usually target settings where few data are available for training (e.g., few-shot learning), and none of them is designed for interpretability. 

To deal with such a domain, we propose Memory Wrap, \textbf{a memory-based layer that can be attached to any \gls{DNN} and that does not require modifications in the training recipe of the augmented \gls{DNN}}. Memory Wrap is designed to preserve the performance of the black-box model and avoid the problem of input dispersion (\Cref{sec:related_senn}) by assigning a precise semantic to each memory slot. Additionally, its structure can be used by users to extract explanations by examples and counterfactuals. These explanations are useful for inspecting the decision process and detecting biased and uncertain predictions.

This section is organized as follows: \Cref{sec:mw_design} describes the architectural design of Memory Wrap and how to retrieve explanations; \Cref{sec:mw_experiments} evaluates the proposed module both in terms of performance and explanations; and \Cref{sec:mw_design_choices} discusses alternative design choices for the module. 
\subsection{Design}
\label{sec:mw_design}
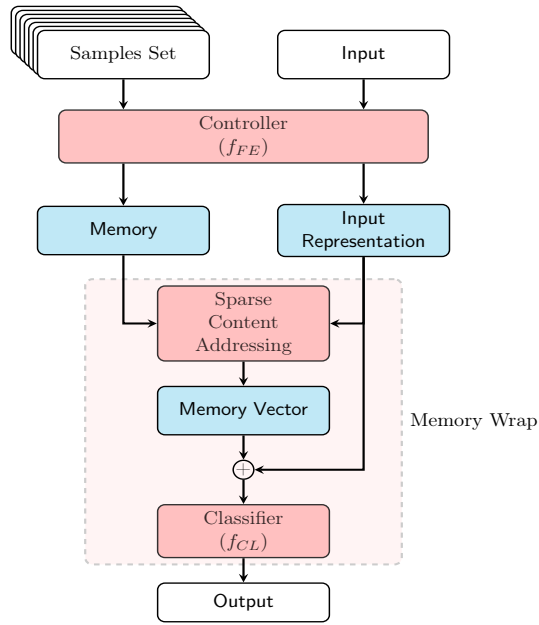
\begin{figure}[t]
  \centering
  \resizebox{2.9in}{3.3in}{
\begin{tikzpicture}
    \node(dummy1){};
    \node [left=0.5cm of dummy1, supportset, fill=white!30] (ss) {Samples Set}; 
    \node [input, right= 0.5cm of dummy1, fill=white!30] (input) {Input};
    \node[architecture,below=1cm of dummy1,minimum width=7cm] (encoder) {Controller\\($f_{FE}$)};
    \node[below=1.25cm of encoder](dummy2){};
    \node [left=0.5cm of dummy2, input] (encoding_ss) {Memory}; 
    \node [input, right=0.5 of dummy2] (encoding_input) {Input\\Representation};
    \node[below=0.5cm of dummy2](dummy3){};
    \node[architecture,below=1cm of dummy2,rounded corners, thick, fill=red!30](attention){Sparse\\Content\\Addressing};
    \node[below=0.5cm of attention,input](sv){Memory Vector};
    \node[below=0.5cm of sv,circle,inner sep=0pt,minimum size=1pt, draw, thick](concat){+};
    \node[below=0.5cm of concat,architecture](mlp){Classifier\\($f_{CL}$)};
    \node[below=0.5cm of mlp,input,minimum height=0.8cm,fill=white!20](output){Output};
    \begin{scope}[on background layer]
      \node [minimum width=6cm,label=right:{Memory Wrap}, draw=black!100, very thick,
      rounded corners, dashed, fill=red!20,opacity=0.2,fit={(mlp) (concat) (attention)}] {};
    \end{scope}
    
    \draw[arrow]             (ss.south) -- (ss.south|-encoder.north);
    \draw[arrow]             (input.south) -| (input.south|-encoder.north);
    \draw[arrow]             (ss.south|-encoder.south) -- (encoding_ss.north);
    \draw[arrow]             (input.south|-encoder.south) -- (encoding_input.north);
    \draw[arrow]             (encoding_input.south) |- (attention.east);
    \draw[arrow]             (encoding_ss.south) |- (attention.west);
    \draw[arrow]             (attention.south) -| (sv);
    \draw[arrow]             (sv.south) -- (concat.north);
    \draw[arrow]             (encoding_input.south) |- (concat);
    \draw[arrow]             (concat) -- (mlp.north);
    \draw[arrow]             (mlp.south) -- (output.north);
 \end{tikzpicture}
  }
 \caption{Architecture of a deep neural network augmented with Memory Wrap.}
  \label{fig:teasermemorywrap}
\end{figure}

Similarly to all the approaches discussed previously, \textbf{Memory Wrap}~\cite{LaRosa2022} can be placed after the feature extractor $f_{FE}$, which acts as a controller (see \Cref{fig:teasermemorywrap}).

The module, denoted as $f_{MW}$, receives two inputs: the latent representation of the current input $\bm{x}_i$, and the latent representations of a set of $n$ sample $S_j=\{\bm{x}_1, \bm{x}_2, ..., \bm{x}_n\}$. This set of samples serves as the memory $\bm{M}$ of Memory Wrap. 

Therefore, the output of the augmented \gls{DNN} can be expressed as:
\begin{equation}
\begin{split}
    y & = f(\bm{x_i},\bm{S_j}) \\
      & = f_{MW}(f_{FE}(\bm{x_i}),f_{FE}(\bm{S_j})
\end{split}
\end{equation}

In contrast to the dynamic memory discussed in the previous section, where the controller decides at each step which information to store based on later outcomes, Memory Wrap is designed for data independence. Each sample in $\bm{S_j}$ is treated as independent, free from temporal dependencies. Hence, the controller does not learn how to write the memory but can learn which information to read from it. 

To fill the memory, an external function $g$ is responsible for extracting samples from the dataset based on specific objectives and writing their latent representation into $\bm{M}$. After the writing operation, each memory row contains the latent representation generated by the controller for one of the samples in input. Optionally, a set of linear layers can project data into different manifolds and perform writing based on multi-head attention, similarly to \Cref{sec:sdnc}.

Conversely, for reading operations, the module employs a sparse content addressing mechanism to assign zero probability to irrelevant input tokens. The content addressing weights are computed, as in \Cref{eq:cosine_similarity} (\Cref{sec:sdnc}), thus, as the cosine similarity between the memory rows and the latent representation of the current input.

However, in this case, the mechanism applies the sparsemax function as a replacement for the softmax function. This replacement reduces noise in the decision process and improves interpretability, as highlighted by recent literature~\cite{Martins2016, Malaviya2018}.
Therefore, the content weights are computed as follows:
\begin{equation}
    \mathbf{w}_c = C(\bm{x}, \bm{M}) = Sparsemax(D(f_{FE}(\bm{x}),\bm{M}))
    \label{eq:sparse_content_addressing}
\end{equation}
The sparsemax is a sparse differentiable function proposed by \citet{Martins16} that can be computed by using the algorithm proposed by \citet{Peters2019}. This sorting-based algorithm finds the probability distribution satisfying the following equation:
\begin{equation}
    \label{eq:sparseobjective}
    \phi_2 (x_j) = \underset{\mathbf{p}\in\Delta^{n-1}}{\arg\min}\;   \mathbf{p}^T\mathbf{x}+\mathbf{H}^{t}_{\alpha}(\mathbf{p}) 
\end{equation}
where $\Delta^{n-1}$ is the probability simplex, $\alpha \in [0,2]$ is a hyperparameter controlling the smoothness of the function, and $\mathbf{H}^T_{\alpha}$ is the
Tsallis entropy~\cite{Tsallis1988}: 
\begin{equation}
\label{eq:tsallis}
\mathbf{H}^{t}_{\alpha}(\mathbf{p}) =  
\begin{cases}
     \frac{1}{\alpha(\alpha-1)}\sum_{j}(p_j - p_j^{\alpha}) & \alpha \neq 1\\
      -\sum_{j}p_j~\text{log}~p_j & \alpha = 1\\
\end{cases} 
\end{equation}
For $\alpha = 2$ (i.e., the value used for sparsemax), the Tsallis entropy reduces to the well-known Gibbs-Boltzmann-Shannon entropy. 

To compute weights, we can use the following equation, which returns the solution to the system: 
\begin{equation}
    \phi_2 (x_j) = ReLU([(\alpha-1)\mathbf{x}-\tau\mathbf{1}]^{\frac{1}{\alpha-1}})
\end{equation}
where $\tau$ is the
Lagrange multiplier corresponding to the $\sum_i{p_i=1}$
constraint. For further details about the algorithm and the proof of the derivation, please refer to the work of \citet{Martins16}.

Since the readings are based on the content addressing mechanism, the network learns to read (i.e., select) samples from memory similar to the current input, aiding the classifier in the current prediction. The retrieval of similar memory samples and the combination with the input enables the network to overcome missing or noisy information in the current input. Moreover, the employment of sparse mechanisms and the disjoint encoding mitigates the issue of input dispersion.


After the computation of the content weights, Memory Wrap computes the memory vector $\mathbf{v}_{M}$ as the weighted sum of the memory $\bm{M}$, where the weights are set equal to the sparse content weights computed in the previous step.

\begin{equation}
  \mathbf{v}_{M} = \mathbf{M}^T\mathbf{w}_c.
\end{equation}
Since the module employs the sparsemax, a few rows of the memory are represented inside $\mathbf{v}_{M}$. Finally, the classifier $f_{CL}$ takes the concatenation of the memory vector and the input latent representation as input and returns the final prediction (see \Cref{fig:teasermemorywrap}):

\begin{equation}
    y = f_{CL}(\mathbf{v}_M \oplus f_{FE}(\bm{x}_i))
\end{equation}
The idea is that by using the similarity and the sparsemax together, strong features of the target class will be more represented than features of other classes, thus helping the network in the decision process.

\subsubsection{Explaining Via Memory Tracking}
\label{sec:memory_tracking_memorywrap}
In \Cref{sec:sdnc}, the memory contains states connected to the parsing of input features. Hence, the memory tracking mechanism results in feature attribution explanations. Conversely, in this case,  the memory contains different samples, and the memory tracking mechanism computes example-based explanations: explanations by examples and counterfactuals. 

\begin{figure}[t!]
    \centering
    \includegraphics[scale=0.4]{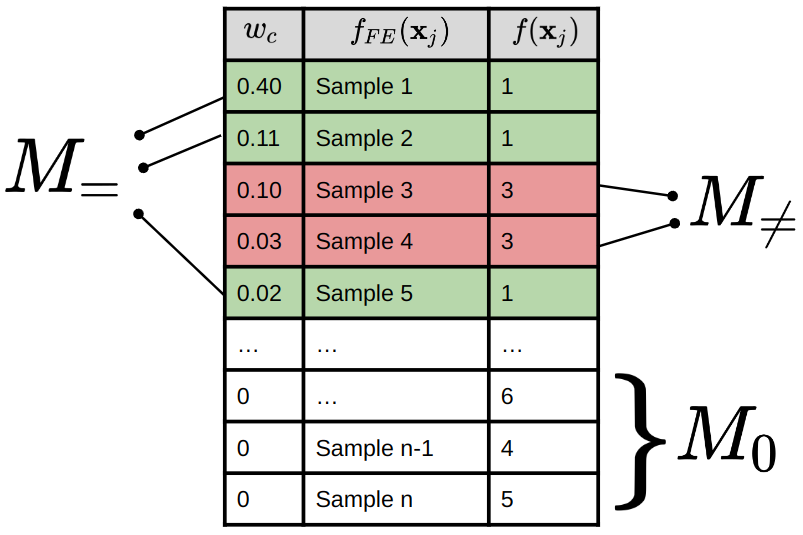}
    \caption{Illustration of a snapshot of the memory of Memory Wrap. Each row can be associated with one of the three sets $\bm{M_0}$, $\bm{M_=}$, and $\bm{M_{\neq}}$ based on its associated content weight $w_c$ and prediction. }
    \label{fig:memory_memorywrap}
\end{figure}

Since Memory Wrap uses the sparsemax, we can identify three disjoint sets of rows in the memory $\bm{M}$ (\Cref{fig:memory_memorywrap}). Each row $j$ in $\bm{M}$ corresponds to the latent representation generated by the controller when fed with the sample $x_j$. First, we can isolate the samples that have no impact on the decision process because they are associated with content weights equal to 0. Consequently, these samples are not represented in the read vector and are collected in the set $\bm{M_0}$. Since these samples are associated with zero weights, the neural network deems them different from the current input, and thus, they do not represent good candidates for explanations. Hence, the set $\bm{M_0}$ is disregarded when searching for potential candidates for explanations. Note that this set exists due to the employment of the sparsemax function: any softer version would produce only the other two sets.

To identify the other two sets, each memory sample associated with a content weight greater than zero is fed to the network, and the corresponding predictions are gathered. Among these samples, a distinction is made between those associated with the same prediction as the current input, collected in the set $\bm{M}_{=}$, and those associated with a different prediction, collected in the set $\bm{M}_{\neq}$. In mathematical terms, $\bm{M}_{=}$ and $\bm{M}_{\neq}$ can be defined as:
\begin{equation}
    \bm{M}_{=} = \{f_{FE}(\mathbf{x}_j) \mid f(\mathbf{x}_i,S_j) = f(\mathbf{x}_j,S_k) : \mathbf{x}_j \in S_j \wedge S_j \neq S_k\}
\end{equation}

\begin{equation}
        \bm{M}_{\neq} = \{f_{FE}(\mathbf{x}_j) \mid f(\mathbf{x}_i,S_j) \neq f(\mathbf{x}_j,S_k) : \mathbf{x}_j \in S_j \wedge S_j \neq S_k\}
\end{equation}
and clearly, $\bm{M}_{=}$, $\bm{M}_{\neq}$, and  $\bm{M}_{0}$ satisfy the following constraints:
\begin{equation}
    \bm{M}_{=} \cup \bm{M}_{\neq} \cup \bm{M}_{0} = \bm{M}
\end{equation}
\begin{equation}
    \bm{M}_{=} \cap \bm{M}_{\neq} \cap \bm{M}_{0} = \emptyset
\end{equation}
 $\bm{M}_{=}$ and $\bm{M}_{\neq}$ inherently represent good sets of candidates for explanations by examples and counterfactuals, respectively. More precisely, the best candidate for being used as an explanation by example is the sample in $\bm{M}_{=}$ associated with the highest weight:
 \begin{equation}
     \bm{X}_e = \{\bm{x}_k | f_{FE}(\mathbf{x}_k)\in \bm{M}_{=} \wedge \forall \bm{x}_j \in \bm{M}_{=} : w_j \leq w_k\}
 \end{equation} 
Intuitively, the sample selected $\bm{x}_e \in \bm{X}_e$ represents a good explanation by example for three reasons. Firstly, it is highly activated, actively contributing to the current prediction. Secondly, a high content weight indicates similarity between the encoding of the input $\bm{x}_i$  and the encoding of $\bm{x}_e$; thus, the network treats them similarly. Lastly, it is predicted in the same class as the current input, and thus, the common features between the explanation by example and the current input can inform the user about the features exploited by the network for that prediction.

 Similarly,  the best candidate for being used as a counterfactual is the sample in $\bm{M}_{\neq}$ associated with the highest weight:
 \begin{equation}
     \bm{X}_c = \{\bm{x}_k | f_{FE}(\mathbf{x}_k)\in \bm{M}_{\neq} \wedge \forall \bm{x}_j \in \bm{M}_{\neq} : w_j \leq w_k\}
 \end{equation} 
In this case, the sample selected $\bm{x}_c \in \bm{X}_c$ represents a good counterfactual explanation because it is associated with a high content weight and predicted in a different class. A different prediction indicates that despite the high similarity between the two samples, the network yields different predictions. Therefore, examining the differences between the samples can help users discern the necessary changes to achieve a different prediction. This information aligns with the criteria of a good counterfactual (\Cref{sec:back_xai}). Finally, note that $\bm{x}_c \in \bm{X}_c$ is partially included in
the memory vector, and frequently, the counterfactual class is the second or third predicted class. Therefore, its inspection can help users understand the uncertainties of the neural network. 

In the rare case of multiple samples associated with the highest weights (i.e., the cardinality of $\bm{X}_c$ or $\bm{X}_e$ greater than 0), several criteria can be employed to resolve the tie. For instance, the samples closer in the input space can be selected, or multiple explanations can be provided for the same input.
\subsection{Experiments}
\label{sec:mw_experiments}
\subsubsection{Performance}
We test Memory Wrap on image classification tasks on several popular \glspl{DNN}. Namely, we train from scratch and augment ResNet18~\cite{He2016}, EfficientNetB0~\cite{Tan2019},
MobileNet-v2~\cite{Sandler2018}, GoogLeNet~\cite{Szegedy2015},
DenseNet~\cite{Huang2017}, ShuffleNet~\cite{Zhang2018}, WideResNet
28x10~\cite{Zagoruyko2016}, and ViT~\cite{Dosovitskiy2021} with Memory Wrap and compare their performance against the networks trained without the module on the Street View House Number (SVHN)~\cite{Netzer2011}, CINIC10~\cite{Darlow2018} and CIFAR10~\cite{Krizhevsky2009} datasets. To train the models, we follow the training setup of the respective papers, repositories, or datasets \textbf{without any modification}.

Specifically, we follow the setup of  Huang et al.~\cite{Huang2017} for SVHN and CIFAR10 and the suggested procedure for training for CINIC10 for all the models but WideResNet~\cite{Zagoruyko2016} and  ViT~\cite{Dosovitskiy2021}, for which we follow the alternative setup described in their papers. The procedure for all the other models consists of training the models using the Stochastic Gradient Descent (SGD) algorithm and decreasing the learning rate by a factor of 10 after 50\% and 75\% of epochs. In all the cases, we initialize the learning rate to 1e-1. We apply a data augmentation based on random horizontal flips and train the networks for 40 epochs in the SVHN dataset and 300 epochs in CIFAR10 and CINIC10 (200 when the controller is ViT). Regarding the parameters of Memory Wrap, we use a memory of 100 randomly extracted samples from the dataset (i.e., the $g$ function is the random selection) and, as a classifier, a multi-layer perception with one hidden layer containing a number of units equal to double the dimension of the input. Note that, ideally, each sample should be associated with a different memory set. However, supporting this type of mapping would require too much space, even for small batch sizes. Indeed, if $m$ is the size of memory and $n$ is the dimension of the batch, then the new input would contain $m \times n$ samples in place of $n$. To reduce the memory footprint, we provide a shared memory set for each batch, thus reducing the space required to $m+n$. 

Since \glspl{MANN}~\cite{Vinyals2016, Snell2017} used in the vision domain usually target settings where few data are available for training, we test four different sizes for the training data: 1000, 2000, 5000 samples, and the entire dataset. The sets correspond to the 1\%, 2\%, 5\%, and 100\% of the labeled samples in CINIC10 and 2\%, 4\% 10\%, and 100\% of the labeled samples in the other datasets.

We begin the evaluation by comparing the networks augmented with Memory Wrap against networks without the module (std), Matching Networks~\cite{Vinyals2016}, Prototypical Networks~\cite{Snell2017}, K-NN~\cite{Cover1967}, sparse K-NN, and an ablated version of Memory Wrap when trained on the above-mentioned datasets' subsets. These competitors cover several alternatives regarding memory usage, interpretable decision processes, architectures for small data settings, and module designs.

Specifically, Prototypical Networks are memory-based networks that compute prototypes for each class as the mean of the samples of that class in the memory set, and then they use the distance between the prototypes and the input to compute the prediction.\\
Matching Networks are memory-based networks that classify inputs based on a weighted linear combination of the labels in the memory set, where the weights are the distances between its samples and the current input computed over an embedding learned by a recurrent network. \\
K-NN~\cite{Cover1967} is a popular baseline in these settings~\cite{Vinyals2016}, and it is an interpretable model since the prediction of each input is based on the mode of the labels of its $k$ nearest neighbors (i.e., the closest in the latent space).\\
Sparse K-NN replaces the selection of the $k$ parameter with the usage of sparsemax. In this case, the label is chosen based on the mode of the labels of samples in memory associated with a weight greater than zero.\\
Finally, the ablated version of Memory Wrap (MemOnly) uses only the read vector as input to the classifier, removing the skip connection from the controller. Therefore, this network uses the input only to compute the sparse content-based attention weights and compute the read vectors. 

We use MobileNet, ResNet18, and EfficientNet trained on SVHN as the largest testbed to filter the best competitors for the full experiments. In this test, we use three different sizes for the training data: 1000, 2000, and 5000 samples. We collect the competitors' performance on the official testing dataset over 15 different experiments and analyze the average accuracy.

\begin{figure}
    \centering
    \includegraphics[scale=0.5]{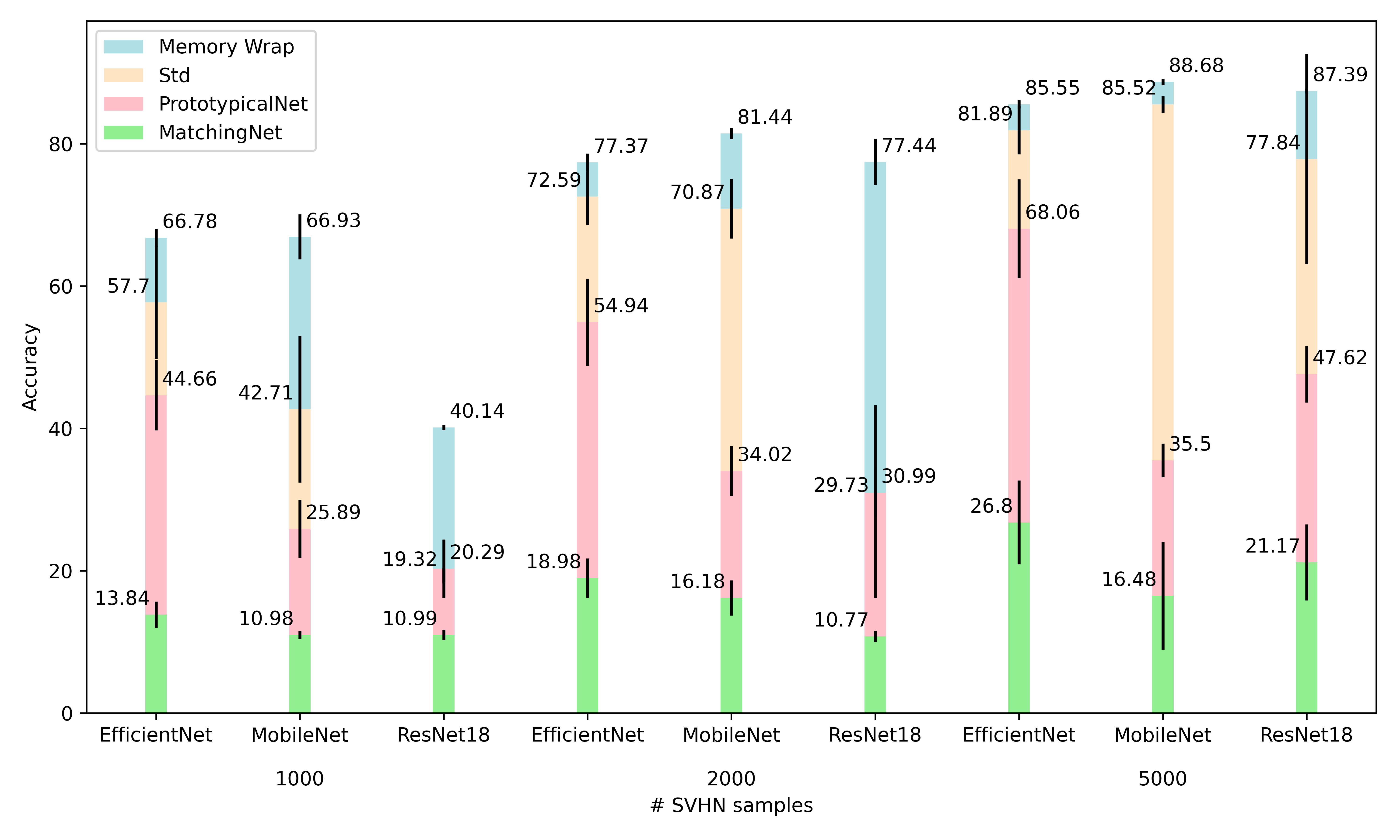}
    \caption{Performance comparison between memory-based modules.}
    \label{fig:mw_performance_memory}
\end{figure}
\begin{figure}
    \centering
    \includegraphics[scale=0.33]{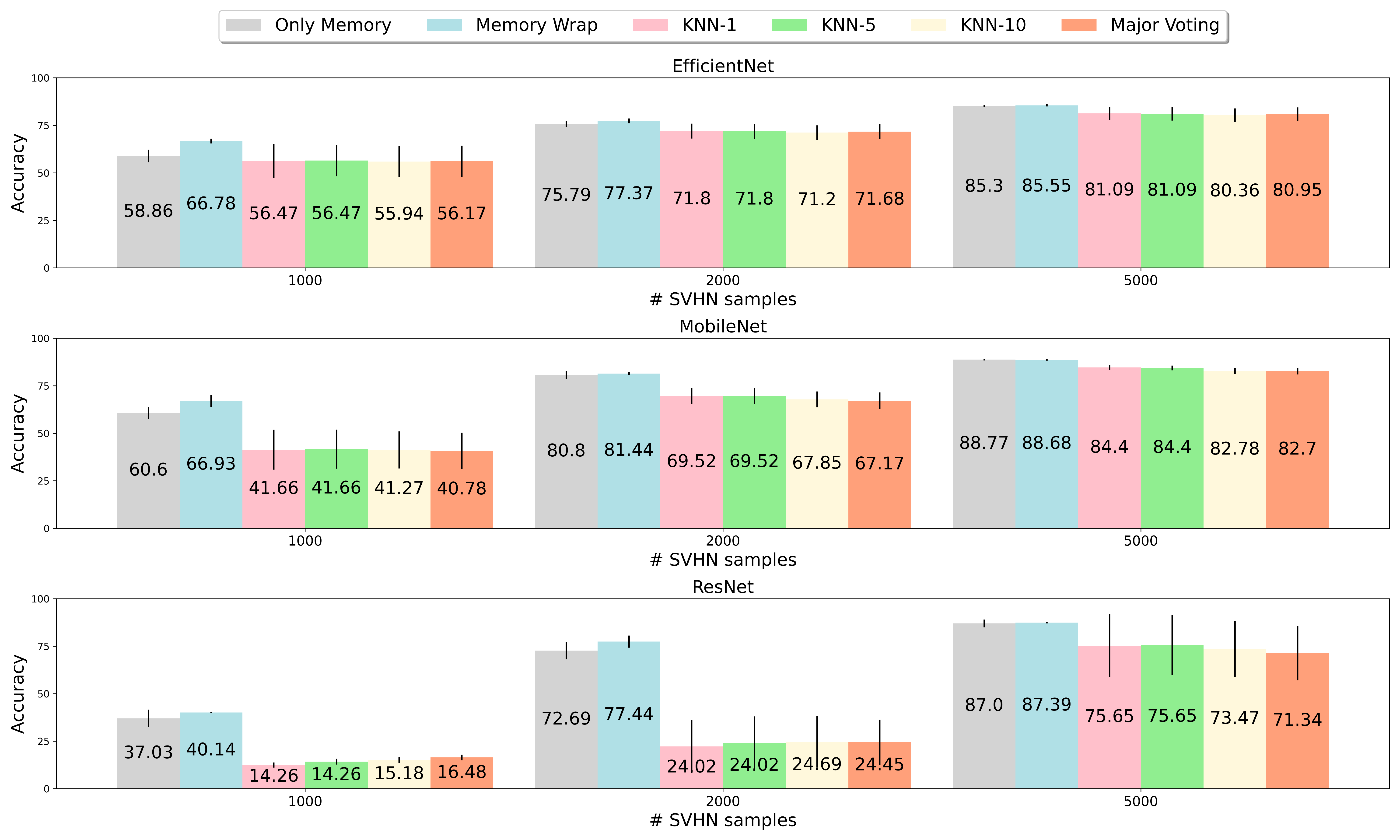}
    \caption{Performance comparison between k-NN variants and Memory Wrap variants.}
    \label{fig:mw_performance_knn}
\end{figure}

In \Cref{fig:mw_performance_memory} and \Cref{fig:mw_performance_knn}, we observe that \textbf{Memory Wrap reaches the highest performance across all the configurations against both the memory competitors and the interpretable ones in small data settings}. In these settings, the interpretable models are more performant than the memory ones. Indeed, the K-NN baseline reaches comparable performance with respect to the black-box model alone (std), dropping its performance of just 1-2\% but gaining a more interpretable decision process. This result aligns with recent findings about this strong baseline~\cite{Vinyals2016}. Note that while the best $k$ to be used as a hyperparameter of this baseline varies across the configurations, the results across them are similar most of the time, and thus, the choice of $k$ has no big impact.

Matching Networks and Prototypical Networks reach the lowest performance across all the configurations. We argue that the cause of low performance resides in how they encode and use the memory samples, tailored to the few-shot learning scenarios. For example, Matching Networks embed both the input and memory using recurrent networks (LSTM). When trained on small datasets, the resulting embeddings capture a small portion of the semantics. Thus, they are less effective since the embeddings of the memory set can be misaligned with respect to the input one. Prototypical Networks mitigate this problem by encoding both the input and memory set using the controller in the same way as Memory Wrap. However, the fact that the output of Prototypical Networks is based on all the samples in memory (i.e., by computing the average encoding) can create instability. Indeed, the presence of outliers in memory, a common scenario when samples are not selected a priori, can produce a large shift in the prediction. 

Memory Wrap combines the positive qualities of all the competitors. Like Prototypical Networks, Memory Wrap learns how to use the memory during the training process and encodes both the input and the memory samples using the same controller. Like K-NNs, it selects a subset of similar samples from memory, and thus, only a small number of samples in memory can have an impact on the decision process, avoiding the problem of outliers. The difference with K-NN is that the latter is trained independently from the black-box model, and thus, the performance of the black-box model represents an upper bound. Conversely, Memory Wrap jointly learns the controller and how to use its encodings to exploit the memory mechanisms, improving both components at the same time.
These characteristics allow Memory Wrap to reach an improvement in performance and data efficiency between $2.5\times$ (ResNet) and $1.5\times$ (EfficientNet and MobileNet) over the black-box models. Regarding the ablated version, we can observe that it performs better than the black-box models. However, the information carried on by the skip connection in the full Memory Wrap is crucial to achieving the best absolute performance, likely due to the access of shortcuts in it (e.g., the presence of rare features). 

\begin{figure*}[th]
    \centering
    \includegraphics[scale=0.27]{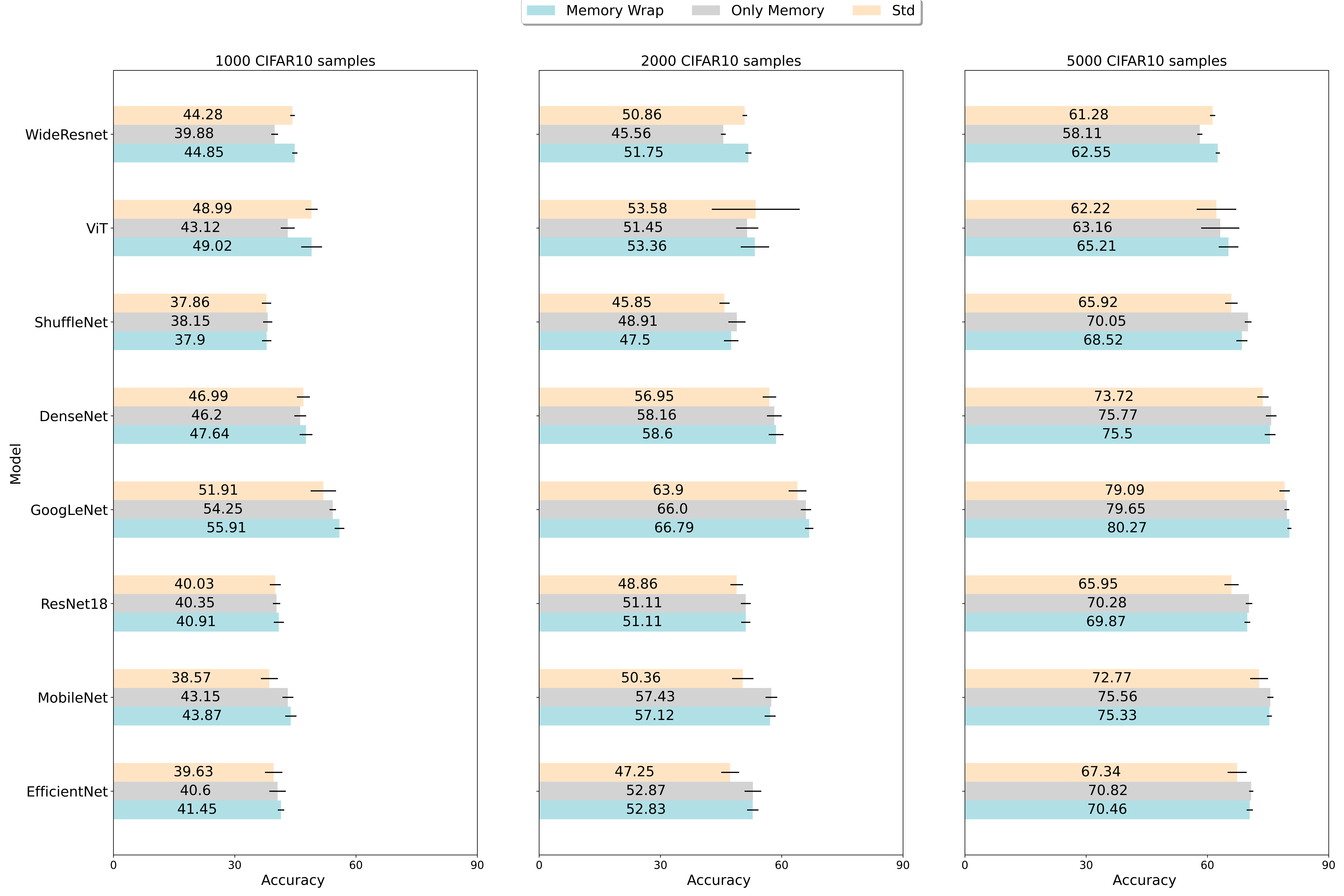}
    \caption{Avg. accuracy and standard deviation over 15 runs of the black-box model, the full Memory Wrap, and its ablated version on subsets of CIFAR10.}
    \label{fig:cifar}
\end{figure*}
\begin{figure*}[th]
    \centering
    \includegraphics[scale=0.27]{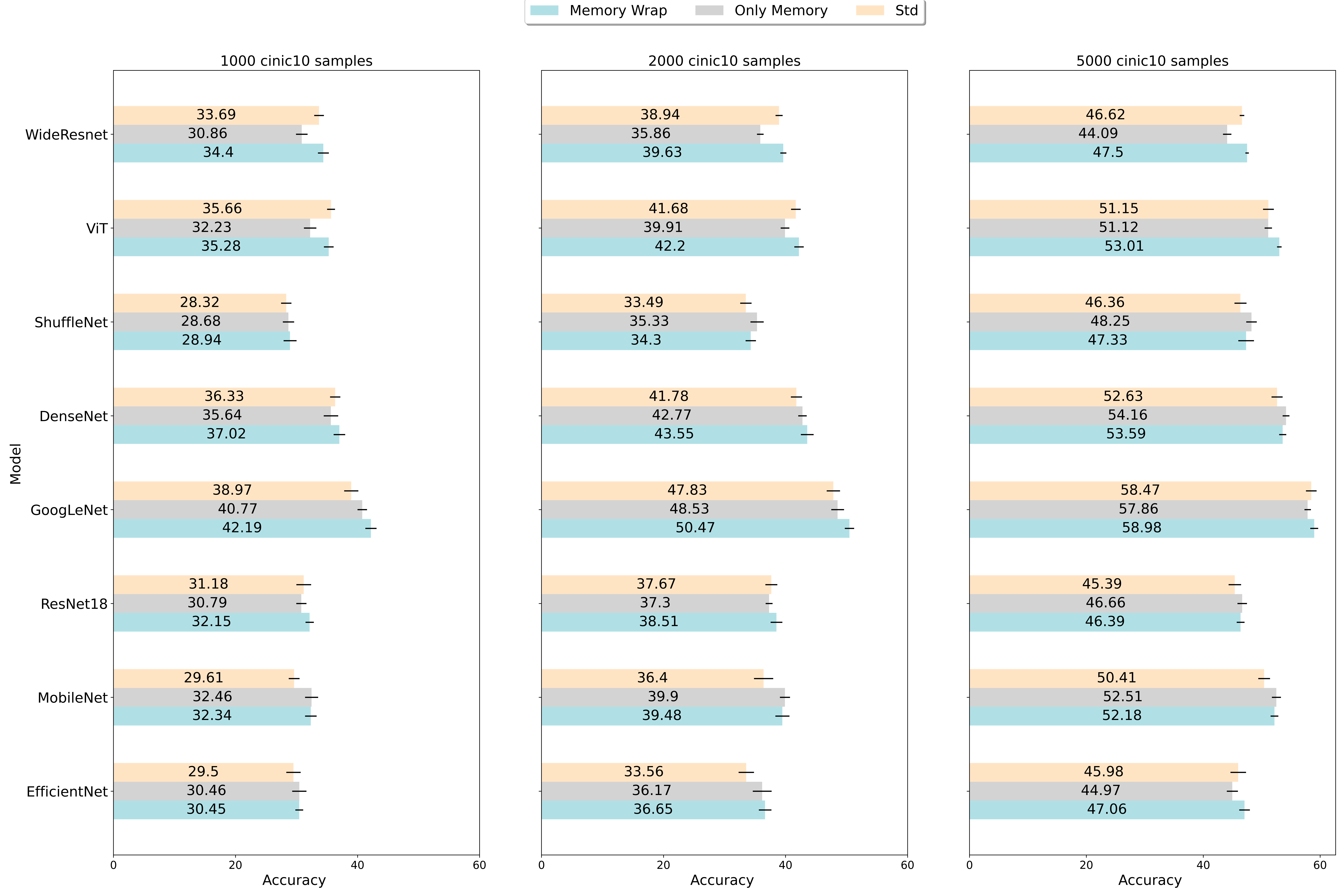}
    \caption{Avg. accuracy and standard deviation over 15 runs of the black-box model, the full Memory Wrap, and its ablated version on subsets of CINIC10.}
    \label{fig:cinic10}
\end{figure*}

From the previous results, we select the three best competitors (Memory Wrap, the black-box model, and the ablated Memory Wrap) and evaluate their performance on CIFAR10 (\Cref{fig:cifar}), CINIC10 (\Cref{fig:cinic10}), and the rest of the selected \glspl{DNN}. We observe that Memory Wrap achieves the best performance across all the models, confirming the results of the previous case. However, here we can note two differences: the ablated version reaches lower performance than the black-box models in ViT and WideResNet, and the gap between Memory Wrap and the black-box models is shorter on average than the gaps reported in SVHN. Indeed, CIFAR10 and CINIC10 represent a less structured dataset than SVHN. In CIFAR10 and CINIC10, classes share several features, like colors, and there is a high variability among the samples of the same class. For example, a ``stealth airplane'' has completely different colors and shapes from commercial airplanes. These characteristics lower the benefits of exploiting the similarity in memory since samples retrieved by similarity can be associated with other classes. In these cases, the skip connection of Memory Wrap becomes crucial, allowing the model to tolerate wrong retrieval from memory better than the ablated version. 

Finally, \Cref{table:complete} reports the performance of Memory Wrap when it is trained using the full dataset. We observe that Memory Wrap achieves, on average, the same performance as the black-box models. Therefore, \textbf{in the full dataset scenario, there is no gain in performance when applying Memory Wrap}. However, in contrast with black-box models, Memory Wrap provides ways to interpret its behavior, which we discuss in the next paragraph, thus representing a useful way to augment the black-box models also in these cases.
\begin{table}[b]
      \setlength{\tabcolsep}{4pt}
        \centering
                
             \caption{Avg. accuracy and standard deviation of the baselines and Memory Wrap when the training datasets are the whole SVHN, CIFAR10 and CINIC10 datasets.}
        \label{table:complete}
        \scalebox{0.75}{
        \begin{tabular}{@{}lrlll@{}p{\footnotesize	}}
          \toprule
          &\multicolumn{3}{c}{Full Datasets Avg. Accuracy \%}\\
          \midrule
          Encoder&Model&SVHN&CIFAR10&CINIC10\\
          \midrule
           EfficientNet& Black-Box &  94.39\footnotesize	{ $\pm$ 0.24}  & \textbf{88.13\footnotesize	{ $\pm$ 0.38}} & \textbf{77.31\footnotesize	{ $\pm$ 0.35}}\\
           & Memory Wrap & \textbf{94.67\footnotesize	{ $\pm$ 0.16}} & \textbf{88.05\footnotesize	{ $\pm$ 0.20}} & \textbf{77.34\footnotesize	{ $\pm$ 0.27}}\\
          \midrule
           MobileNet&Black-Box & \textbf{95.95\footnotesize	{ $\pm$ 0.09}}  & \textbf{88.78\footnotesize	{ $\pm$ 0.41}} & \textbf{78.97\footnotesize	{ $\pm$ 0.31}} \\
           & Memory Wrap &  95.63\footnotesize	{ $\pm$ 0.08}  & \textbf{88.49\footnotesize	{ $\pm$ 0.32}}  & \textbf{79.05\footnotesize	{ $\pm$ 0.15}}\\
          \midrule
           ResNet18 & Black-Box & \textbf{95.70\footnotesize	{ $\pm$ 0.10}} & \textbf{91.94\footnotesize	{ $\pm$ 0.19}} & \textbf{82.05\footnotesize	{ $\pm$ 0.25}} \\
           & Memory Wrap &  95.49\footnotesize	{ $\pm$ 0.11}  & 91.49\footnotesize	{ $\pm$ 0.17}  & \textbf{82.04\footnotesize	{ $\pm$ 0.16}}\\
          \bottomrule
        \end{tabular}
        
        }
      \end{table}

\subsubsection{Explanations}
This paragraph evaluates and showcases the explanations extracted by analyzing the Memory Wrap structure. All the experiments and figures use MobileNet as a controller. However, the findings also hold for all the other black-box models.

\begin{table}[b]
    \centering
    
            \caption{Avg. and standard deviation of the matching accuracy reached over 15 runs by the
    sample in the memory set with the highest sparse content-based attention weight (Top), the example with the lowest weight but greater than zero (Bottom) and a random sample (Random)}
    \label{table:expaccuracy}
    \scalebox{0.75}{
    \begin{tabular}{@{}lllll@{}}
      \toprule
      \multicolumn{4}{c}{Prediction matching accuracy \%}\\
      \midrule
     Example&1000&2000&5000\\
     \midrule
    &\multicolumn{3}{c}{SVHN}\\
        Top & \textbf{84.24\footnotesize	{ $\pm$ 1.22}} &  \textbf{90.59\footnotesize	{ $\pm$ 0.52}}&  \textbf{94.47\footnotesize	{ $\pm$ 0.22}}\\
        Bottom & \textbf{46.46\footnotesize	{ $\pm$ 1.77}}&  \textbf{57.39\footnotesize	{ $\pm$ 1.09}}&  \textbf{69.94\footnotesize	{ $\pm$ 1.37}}\\
       
        Random & 11.76\footnotesize	{ $\pm$ 0.30}& 11.66\footnotesize	{ $\pm$ 0.17}&  11.71\footnotesize	{ $\pm$ 0.13} \\
      \midrule
          &\multicolumn{3}{c}{CIFAR10}\\
       Top &
      \textbf{82.04\footnotesize	{ $\pm$ 1.14}}& \textbf{87.75\footnotesize	{ $\pm$ 0.72}}&  \textbf{91.76\footnotesize	{ $\pm$ 0.22}}\\
       Bottom & \textbf{46.01\footnotesize	{ $\pm$ 1.92}} & \textbf{60.10\footnotesize	{ $\pm$ 1.29}}&  \textbf{69.94\footnotesize	{ $\pm$ 0.82}}\\
       Random &10.22\footnotesize	{ $\pm$ 0.28} & 10.23\footnotesize	{ $\pm$ 0.20}&  9.80\footnotesize	{ $\pm$ 0.43}\\
      \midrule
    &\multicolumn{3}{c}{CINIC10}\\
      Top &   \textbf{76.31\footnotesize	{ $\pm$ 0.73}}& \textbf{78.50\footnotesize	{ $\pm$ 0.50}}&  \textbf{78.45\footnotesize	{ $\pm$ 0.62}}\\
       Bottom&  \textbf{37.01\footnotesize	{ $\pm$ 1.12}}& \textbf{41.34\footnotesize	{ $\pm$ 0.73}}&  \textbf{37.55\footnotesize	{ $\pm$ 1.49}}\\
       Random &  10.47\footnotesize	{ $\pm$ 0.19}& 10.30\footnotesize	{ $\pm$ 0.11}&  10.16\footnotesize	{ $\pm$ 0.12}\\
      \bottomrule
    \end{tabular}
    }
\end{table}

Following the evaluation conducted in \Cref{sec:sdnc_expla_exp}, we begin our evaluation of the explanations linked to the Memory Wrap design by investigating the reliability of the ranking returned by the sparse content-based attention weights. To examine the ranking, we select the sample $\in M_{ce}$ with the highest weight (\textit{Top}), the sample $\in M_{ce}$ with the lowest weight (\textit{Bottom}), and a random sample from memory. Then, we feed these samples as input to the network and collect their predictions. Finally, we measure the \emph{prediction matching accuracy} by counting how frequently these predictions correspond to the prediction associated with the current input.

\Cref{table:expaccuracy} shows that the matching accuracy achieved by the sample with the highest weight is consistently higher than that of the randomly selected sample and the sample with the lowest weight. The random selection is the worst performer because it chooses a random example from samples of all classes. Given that the dataset has ten classes, the probability of selecting the correct class is approximately $10\%$. Conversely, the sample with the lowest weight actively contributes to the decision-making process, and as it resembles the current input, it may belong to the same class or a different class that could be misclassified into the same class. Furthermore, we note that the higher the model's accuracy, the higher the bottom sample's matching accuracy. This observation is consistent with the finding that when Memory Wrap achieves high accuracy, the memory samples associated with a weight greater than zero are typically all samples from the same class, thereby enhancing the matching accuracy. Collectively, these results confirm the \textbf{reliability of the sparse content-based attention weights ranking and the quality of the sample with the highest weight as a good proxy for prediction}.

Once the reliability of the ranking has been proved, we can assess the quality of explanations by examples and counterfactuals extracted using the method outlined in \Cref{sec:memory_tracking_memorywrap}.  The explanations by examples are evaluated using the \emph{input non-representativeness}~\cite{Kenny2021} (\Cref{eq:input_non_representativeness}) and the \emph{prediction non-representativeness}~\cite{Nguyen2020} (\Cref{eq:prediction_non_representativeness}) metrics. The counterfactuals are evaluated using IM1 (\Cref{eq:IM1}), IM2~\cite{Looveren2021}(\Cref{eq:IM2}), and IIM1 (\Cref{eq:IIM1}) scores. Competitors include Memory Wrap, the random selection, and the posthoc methods CHP~\cite{Kenny2021} and KNN*~\cite{Papernot2018} for explanations by examples and a method based on prototoypes~\cite{Looveren2021} for counterfactuals. These methods are chosen for their publicly available implementation, recent relevance, and compatibility with Memory Wrap's assumptions (i.e., they have been tested on image data and do not require additional knowledge). Note that the post-hoc methods require more computation time than Memory Wrap and that Memory Wrap is not optimized during the training process to select the best possible explanations, but the explanations are a side product of its design.

\begin{figure}[t!]
    \centering
    \includegraphics[scale=0.4]{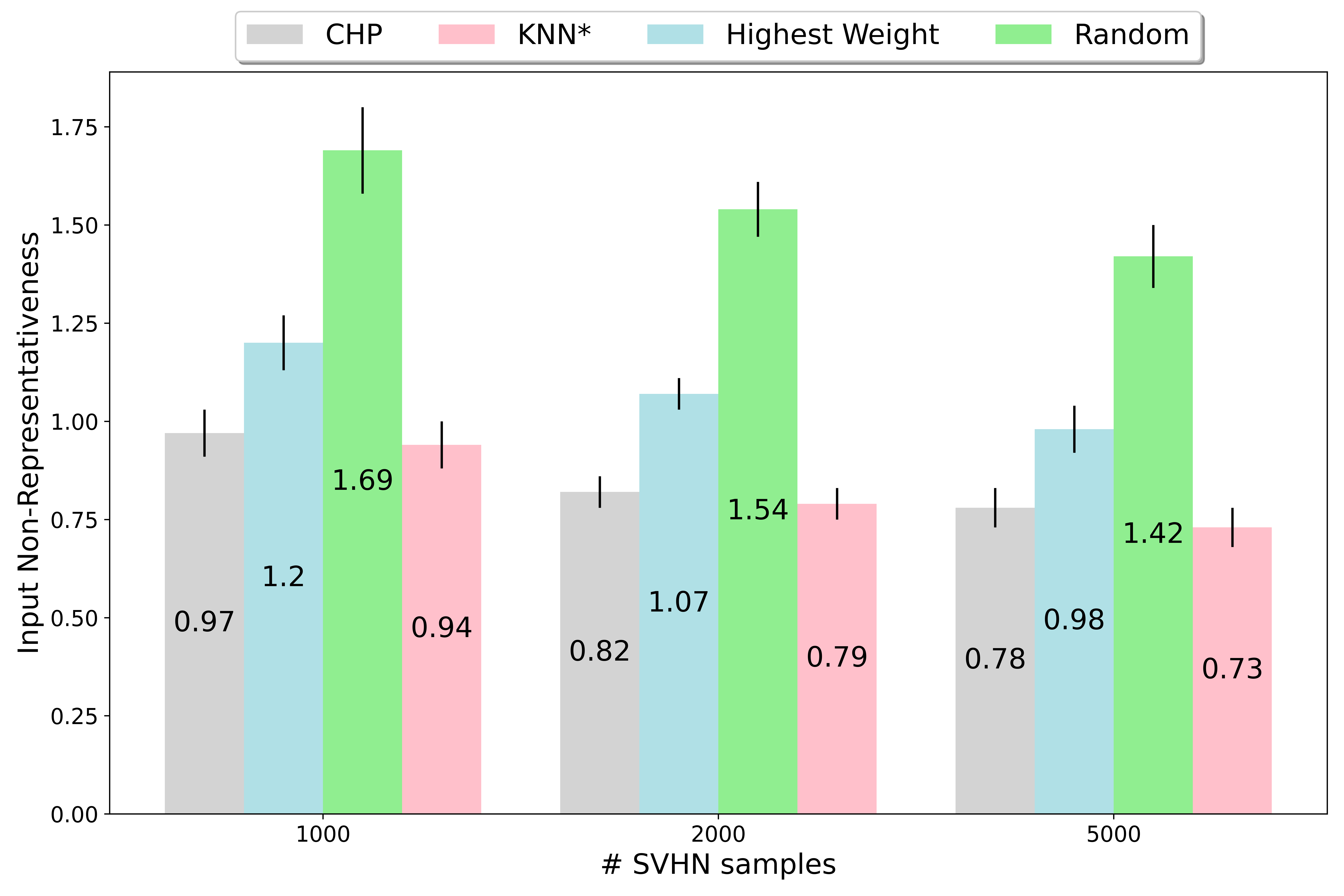}
    \caption{Avg. Input representativeness scores and standard deviation reached by CHP (gray), KNN* (pink), the random baseline (green), and Memory Wrap (blue). Lower is better.}
    \label{fig:expbyexaplotlog}
\end{figure}

\begin{figure}[t!]
    \centering
    \includegraphics[scale=0.4]{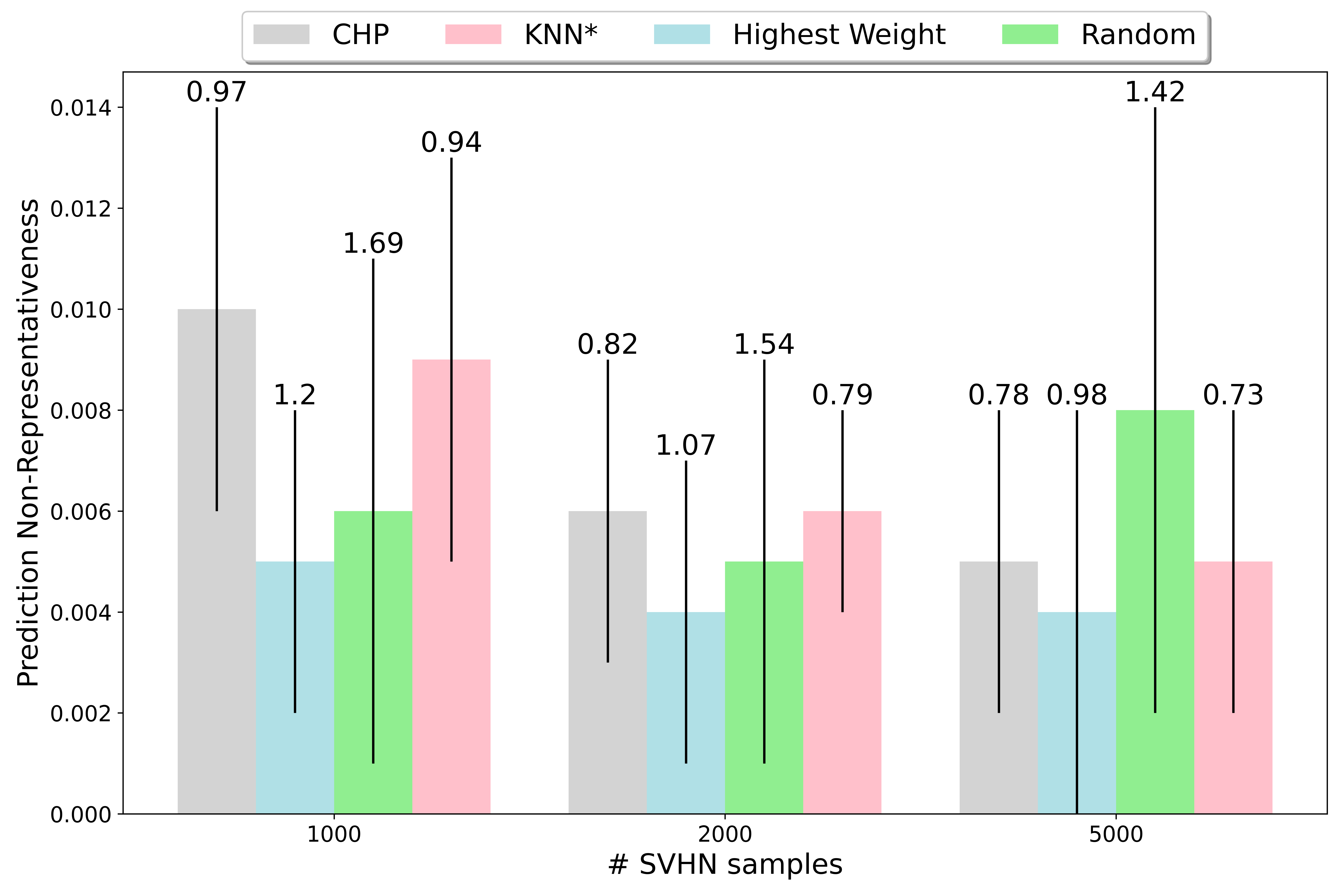}
    \caption{Avg. Prediction representativeness scores and standard deviation reached by  CHP (gray), KNN* (pink), the random baseline (green), and Memory Wrap (blue). Lower is better.}
    \label{fig:expbyexaplotpred}
\end{figure}

\Cref{fig:expbyexaplotlog} and \Cref{fig:expbyexaplotpred} show that \textbf{explanations by example selected by Memory Wrap are better than the competitors} in terms of prediction non-representativeness and worse than them in terms of input non-representativeness.
While the lower performance in input non-representativeness is expected since competitors are designed to optimize this score, the discrepancy in non-representativeness prediction deserves further analysis. This behavior can be explained by analyzing the cases when the network is uncertain, causing the logits of several classes to be close. Since, by design, the considered post-hoc methods select samples with low L\textsubscript{1} distance, they pick other uncertain samples, leading to low average input non-representativeness. As the samples selected by the post-hoc methods are challenging samples for the network, the network may predict them in different classes, especially if trained on the lowest settings, increasing the prediction non-representativeness, which is based solely on the predicted classes. In Memory Wrap, the uncertainty is usually encoded in the counterfactuals included in the set $M_{\neq}$. Therefore, the sample selected as an explanation by example tends to be a strong prototype of the predicted class, resulting in low prediction non-representativeness and higher input non-representativeness.

\Cref{table:countercomparsion} compares counterfactuals in terms of IM1, IM2~\cite{Looveren2021}, and IIM1 scores~\cite{LaRosa2022}. Since the post-hoc method generates counterfactuals, it scores lower in both IM1 and IM2. Indeed, Memory Wrap selects real samples by design, keeping the distribution identical to the training one. Even when it selects edge cases (e.g., confusing images), their distribution is usually closer to one of the counterfactual classes than one of the input classes. More interestingly, Memory Wrap reaches worse scores in IIM1, but the gap is not large, despite the fact Memory Wrap has limited access to the possible counterfactuals due to the random selection of the memory set and the sparsemax, which filter out most of the candidate counterfactuals. 
\begin{table}[b!]
    \caption{Avg. IM1, IM2, and IIM1 scores on the SVHN dataset reached by Memory Wrap and a prototype-based method. Lower is better.}
    \label{table:countercomparsion}
      \centering
\begin{tabular}{@{}rlll@{}}
\toprule
\multicolumn{4}{c}{Counterfactuals Scores Avg.}\\
\midrule
 &\multicolumn{3}{c}{Samples}\\
Score&Dataset&MemoryWrap&Proto~\cite{Looveren2021}\\
\midrule
      & 1000 & 0.991 & 0.994   \\
 IM1      & 2000 & 0.985 &  0.997  \\
       & 5000 & 0.985 &  0.990  \\
\midrule
               & 1000 & 1.703 &  2.623 \\
 IM2(x10)      & 2000 & 1.674 &  2.707  \\
       & 5000 &  1.698 &  2.770 \\
\midrule
               & 1000 & 1.014 & -   \\
 IIM1      & 2000 & 1.015 &  -  \\
       &  5000 & 1.013 &  -  \\
\bottomrule
\end{tabular}

\end{table}

Now, we describe how users can exploit the design and characteristics of Memory Wrap to gain insights into its decision process.

First of all, as described in \Cref{sec:mw_design}, the \textbf{presence of counterfactuals can be used as a means to detect uncertainty} in the decision process. This property is tested on MobileNet and SVHN by measuring the Person correlation~\cite{Stigler1989} between the correctness of the prediction, the number of counterfactuals in $M_{\neq}$ and the index of the counterfactual in the list of memory samples sorted by their content weights.
\Cref{table:pearson} confirms this hypothesis, revealing a positive correlation between the position of the highest-rated counterfactual and the prediction correctness and a negative correlation between counterfactual count and correctness. Thus, an increase in both the number and rating of counterfactuals in memory is a marker of increased uncertainty in the network's prediction. These results align with the findings in \Cref{table:counteraccuracy}, where predictions featuring the highest rated counterfactual in memory reach a substantially lower accuracy than predictions with no counterfactuals in memory. Consequently, the absence of counterfactuals or their index can offer valuable insights into the reliability of the current prediction. 
\begin{table}[b]
    \centering
              \caption{Avg. and standard deviation of the Pearson product-moment correlation coefficients between the correctness of the predictions, the number of counterfactuals (number), and the position of the first counterfactual (position).}
     \label{table:pearson}

    \begin{tabular}{@{}crrrr@{}}
      \toprule
      \multicolumn{4}{c}{Avg. Pearson Coefficient}\\
      \midrule
      Model   & 1000&2000& 5000\\
      \midrule
      Position &  0.39 \footnotesize{ $\pm$ 0.02} & 0.40 \footnotesize{ $\pm$ 0.01} & 0.41 \footnotesize{$\pm$ 0.01} \\
      Number & -0.42 \footnotesize	{ $\pm$ 0.01} & -0.45 \footnotesize{ $\pm$ 0.01} & -0.46 \footnotesize{ $\pm$ 0.01}  \\
      \bottomrule
      \end{tabular}

\end{table}

\begin{table}[b!]
    \centering
              \caption{Accuracy reached by the model on SVHN when the sample
    with the highest weight in the memory set is a counterfactual (Top Counter) and when there are no counterfactuals at all (No Counter).}
     \label{table:counteraccuracy}
 \scalebox{0.85}{
    \begin{tabular}{@{}rllll@{}}
      \toprule
      \multicolumn{4}{c}{Avg. Accuracy \% (Coverage\%)}\\
      \midrule
      Counter   & 1000&2000& 5000\\
      \midrule
      Top& 35.45 \footnotesize{(19.32)} & 38.29 \footnotesize{(12.06)} & 39.79 \footnotesize{(7.56)} \\
     No & 95.20 \footnotesize{(18.73)} & 97.00 \footnotesize{(36.47)} & 97.84 \footnotesize{(56.88)}  \\

      \bottomrule
      \end{tabular}
      }

\end{table}


\begin{figure}[t]
  \begin{subfigure}{.4\linewidth}
    \centering
    \includegraphics[width=.9\linewidth]{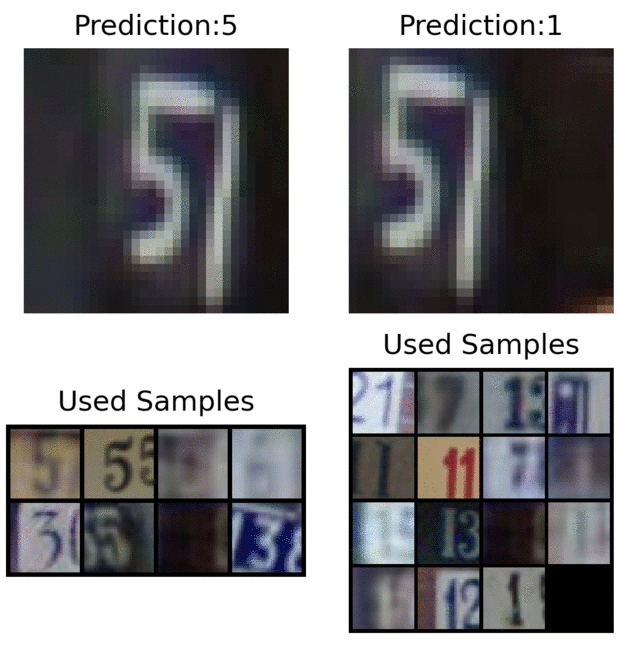}
    \caption{\label{fig:mem_svhn}}
  \end{subfigure}%
  \hfill
  \begin{subfigure}{.43\linewidth}
    \centering
    \includegraphics[width=.9\linewidth]{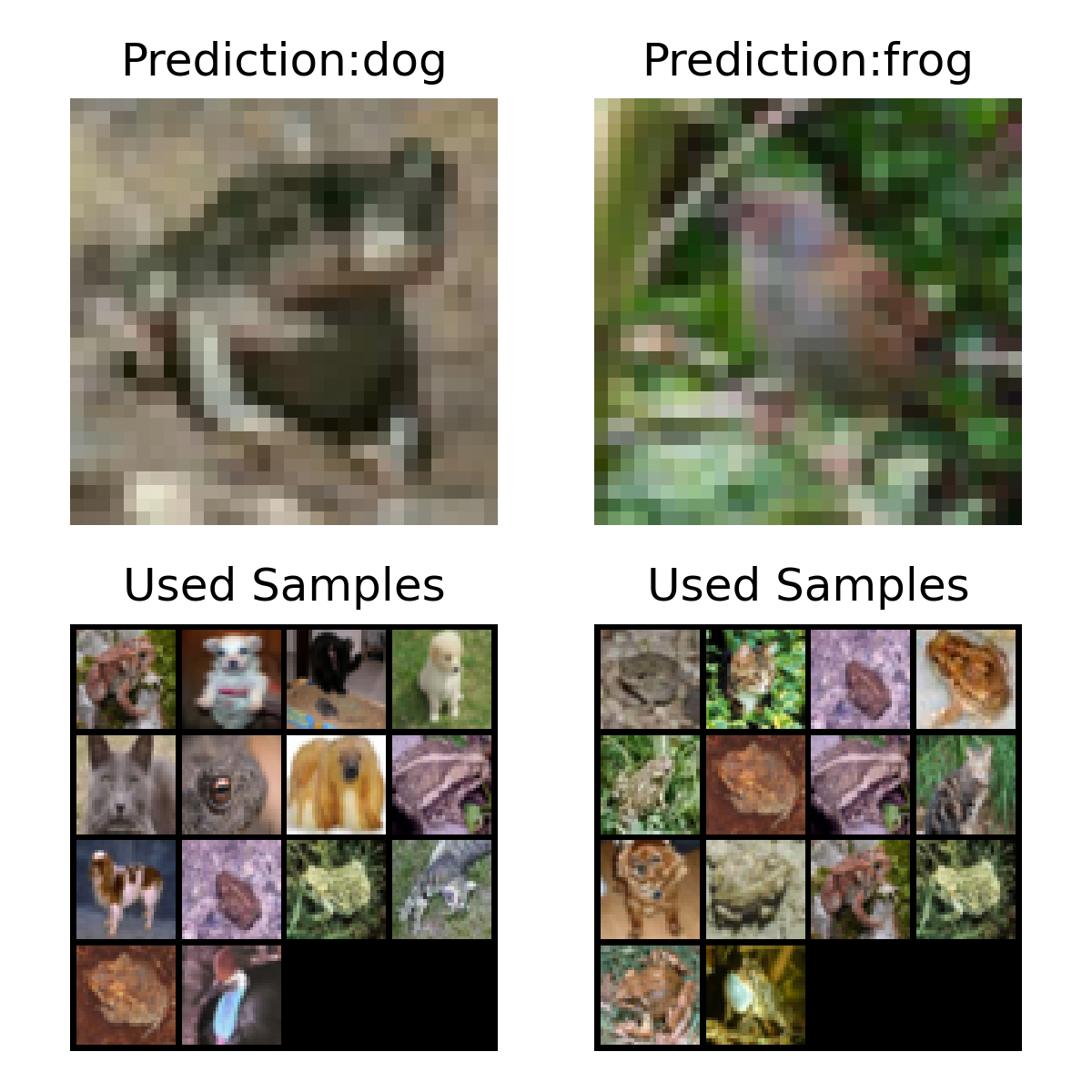}
    \caption{\label{fig:mem_cifar}}
  \end{subfigure}
  \caption{Inputs (first rows) from SVHN (a) and CIFAR10 (b), their associated predictions, and an overview of
  the samples in the memory set that have an active influence on the decision process (second row).}
  \label{fig:mem_example}
\end{figure}
\Cref{fig:mem_svhn} and \Cref{fig:mem_cifar} show a another couple of examples from SVHN and CIFAR10. Two horizontally shifted input images are fed with the same memory samples in the first case ( \Cref{fig:mem_svhn}). Despite their similarity, Memory Wrap selects different samples from the memory for each input. We can analyze the samples associated with a positive weight to \textbf{understand the decision process}. The left image is associated with samples containing images of ``5'' and ``3''. On the other hand, the right image is associated with samples containing images of ``7'' and ``1''. Despite their high similarity, mo samples have been associated with both images, indicating the network correctly focuses on the center image while disregarding other objects. Furthermore, the inspection of the memory sample reveals that the shapes have a bigger impact than the background colors in these decisions, which is evident from the low rank of samples with dark backgrounds. Conversely, in \Cref{fig:mem_cifar}, colors and background exert a bigger influence, particularly in the right image (i.e., the bird predicted as a frog): the majority of memory samples include a green background or brown animals in the center.

\begin{figure}[t!]
    \centering
    \includegraphics[scale=0.5]{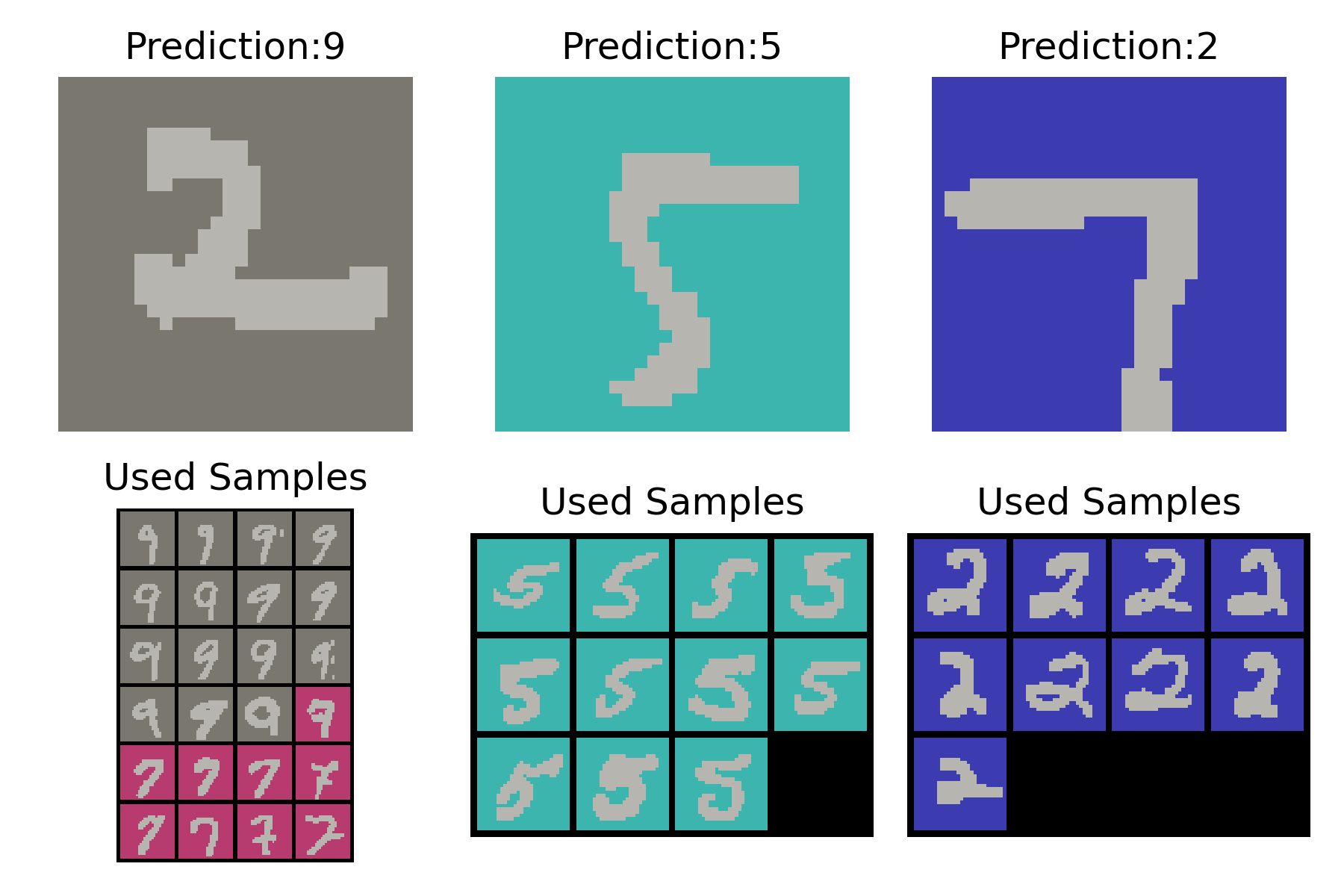}
    \caption{An example of a model biased towards the background. The figure shows how the inspection of the memory samples used by the network makes clear the reasons behind the bad performance of the network. }
    \label{fig:cmnist}
\end{figure}

Finally, memory inspection can be used by developers to \textbf{detect biased decisions}. For example, \Cref{fig:cmnist} shows some examples of predictions of a biased EfficientNetB0 model trained on CMNIST~\cite{pmlr-v119-bahng20a}, a dataset where we introduce a strong bias that correlates the background of images with their labels at training time. In this case, it is easy to see that the inspection and visualization of the memory samples help in detecting the bias. Indeed, the samples associated with a weight greater than zero show how the model uses the background to classify the digits: in all the cases, the highest-rated samples are images with the same color background, although the digit is different, thus exposing the learned bias. 




\subsection{Design choices}
\label{sec:mw_design_choices}
In the following paragraphs, we explore alternative design choices for Memory Wrap. 
\paragraph{g function.}
   \begin{table}[b]
      \setlength{\tabcolsep}{4pt}
        \centering
               \caption{Avg. accuracy and standard deviation over five runs Memory Wrap using different selection mechanisms to select samples in the memory set.}
                    
        \label{table:selection}
        \scalebox{0.75}{
        \begin{tabular}{@{}lrlll@{}}
          \toprule
          &\multicolumn{3}{c}{SVHN Avg. Accuracy \%}\\
          \midrule
          Encoder&Distance&1000&2000&5000\\
          \midrule
           EfficientNet& Random & 67.22\footnotesize	{ $\pm$ 3.47}  & 77.16\footnotesize	{ $\pm$ 0.97} & 85.32\footnotesize	{ $\pm$ 0.90}\\
           & Balanced  &   65.80\footnotesize	{ $\pm$ 2.74} & 77.19\footnotesize	{ $\pm$ 1.27} & 85.80\footnotesize	{ $\pm$ 0.41}\\
         & Replay-Last  &   66.53\footnotesize	{ $\pm$ 1.80} & 76.36\footnotesize	{  $\pm$ 1.09} & 85.80\footnotesize	{ $\pm$ 0.49}\\
          & Replay-Last5  &   66.32\footnotesize	{ $\pm$ 1.70} & 77.56\footnotesize	{ $\pm$ 1.57} & 85.50\footnotesize	{ $\pm$ 0.54}\\
          \midrule
           MobileNet&Random   & \textbf{68.34\footnotesize	{ $\pm$ 1.40}} &81.14\footnotesize	{ $\pm$ 0.69} & \textbf{88.36\footnotesize	{ $\pm$ 0.09}} \\
           & Balanced &   \textbf{67.19\footnotesize	{ $\pm$ 1.77}} & 81.45\footnotesize	{ $\pm$ 1.30} & \textbf{88.36\footnotesize	{ $\pm$ 0.68}}\\
          & Replay-Last  &   \textbf{66.92\footnotesize	{ $\pm$ 1.95}} & 81.17\footnotesize	{ $\pm$ 1.35} & \textbf{88.51\footnotesize	{ $\pm$ 0.29}}\\
          & Replay-Last5  &   65.93\footnotesize	{ $\pm$ 1.98} & 82.09\footnotesize	{ $\pm$ 0.93} & 88.19\footnotesize	{ $\pm$ 0.26}\\
          \midrule
           ResNet18 & Random &  40.35\footnotesize	{ $\pm$ 9.31} & 74.24\footnotesize	{ $\pm$ 2.70} & 87.29\footnotesize	{ $\pm$ 0.39}\\
           & Balanced  &   40.74\footnotesize	{ $\pm$ 8.98} & 75.07\footnotesize	{ $\pm$ 4.20} & \textbf{87.43\footnotesize	{ $\pm$ 0.25}}\\
            & Replay-Last  &   38.75\footnotesize	{ $\pm$ 11.90} & \textbf{78.56\footnotesize	{ $\pm$ 1.84}} & 87.35\footnotesize	{ $\pm$ 0.85}\\
          & Replay-Last5  &   38.73\footnotesize	{ $\pm$ 14.25} & \textbf{79.55\footnotesize	{ $\pm$ 1.16}} & \textbf{87.70\footnotesize	{ $\pm$ 0.34}}\\
          \bottomrule
        \end{tabular}
        }
      \end{table}
We begin with the selection of the $g$ function responsible for choosing the memory samples from the datasets. To be feasible, the $g$ function should be sufficiently fast to avoid slowing down the training process. Consequently, we exclude functions based on the similarity to a specific input (e.g., selecting the top-k most similar samples in the training set). Instead, we test several random selection methods: pure random, balanced random, from the previous batch, and from the previous $k$ batches. The \emph{pure random} competitor selects $m$ training samples randomly from the dataset; the \emph{balanced random} competitor works similarly to the pure random one but ensures that each class has $\frac{m}{num\_classes}$ samples in memory on average; the third competitor selects $m$ samples from the last batch, assuming the batch size is larger than $m$; the last competitor selects $m$ samples from the pool of samples from the last $k$ batches. Note that the last two competitors apply the selection at training time but not at testing time since, at testing time, no learning occurs, and the benefits of selecting the previous batches diminish.

\Cref{table:selection} presents the performance across the SVHN dataset and three different DNNs. No clear winners emerge, with all competitors achieving almost identical performance, especially when applied to EfficientNet. Consequently, they all converge towards random selection. This phenomenon can be explained by analyzing the potential advantages of the competitors. The balanced random competitor selects a fair amount of samples for each class. However, due to sparsemax, Memory Wrap selects a few samples to be used during the inference process. Thus, providing more samples for each class does not enhance the quality of the memory unless samples similar to the current input are chosen. Since the balanced competitor is random and the difference in the number of samples per class is minimal compared to the pure random competitor, the advantages of 1-2 additional samples per class are nullified by the design of Memory Wrap itself. Moreover, the rare cases where few or no samples are provided for a specific class can be considered random noise useful to regularize the training process. Regarding the competitors using the last batches, the potential advantage lies in feeding samples that the network is able to recognize since the network's weights have just been updated to recognize them better. However, while this advantage can play a role during the first epochs of the training process, it diminishes as soon as the network improves its performance. Additionally, note that the similarity is computed with the current input, which is novel with respect to recent samples on which the weights have been updated. Therefore, when the current input is not correctly recognized, the advantage becomes a weakness of these selection mechanisms.

Regarding the testing time, the selection of memory samples is flexible and can be adapted to the application context. We argue that the random choice serves a robust baseline and, on average, is more resistant to adversarial and privacy attacks, the memory samples are in-distribution with respect to the training distribution, and the randomness ensures diversity. However, in contexts where the user can access perfect representative prototypes satisfying the requested objective (e.g., fairness, diversities, etc.) for each class, a fixed selection could be employed without any loss of precision.

\paragraph{Memory Size.}
\begin{table}[!b]
    \centering
      \caption{Avg. accuracy and standard deviation over five runs of Memory Wrap trained using a variable number of samples in memory, when the training dataset is a subset of SVHN.}
      \label{table:impactSVHN}
  \scalebox{0.9}{
    \begin{tabular}{@{}lrlll@{}}
      \toprule
      \multicolumn{5}{c}{Reduced SVHN Avg. Accuracy\%}\\
      \midrule
      &&\multicolumn{3}{c}{Samples}\\
      Encoder&Memory&1000&2000&5000\\
      \midrule
      EfficientNet  & 20 & 64.95\footnotesize	{ $\pm$ 2.46} & 75.80\footnotesize	{ $\pm$ 1.17}    &   84.86\footnotesize	{ $\pm$ 0.99}   \\
      & 100  & 67.16\footnotesize	{ $\pm$ 1.33}  & 77.02\footnotesize	{ $\pm$ 2.20} & 85.82\footnotesize	{ $\pm$ 0.45}\\
      & 300  & 66.70\footnotesize	{ $\pm$ 1.58}  & 77.97\footnotesize	{ $\pm$ 1.34}  &   85.37\footnotesize	{ $\pm$ 0.68} \\
      &  500 & 66.76\footnotesize	{ $\pm$ 0.98} & 77.67\footnotesize	{ $\pm$ 1.17}  & 85.25\footnotesize	{ $\pm$ 1.02}\\
      \midrule
      MobileNet & 20  & 63.42\footnotesize	{ $\pm$ 2.46} & 80.92\footnotesize	{ $\pm$ 1.42}  &  88.33\footnotesize	{ $\pm$ 0.36} \\
      & 100  & 68.31\footnotesize	{ $\pm$ 1.53}  & 81.28\footnotesize	{ $\pm$ 0.69}  &  88.47\footnotesize	{ $\pm$ 0.10} \\
      &  300 & 65.08\footnotesize	{ $\pm$ 0.30}  & 82.05\footnotesize	{ $\pm$ 0.75}  & 88.93\footnotesize	{ $\pm$ 0.37} \\
      &  500 & 69.88\footnotesize	{ $\pm$ 1.76}  & 80.92\footnotesize	{ $\pm$ 1.74} & 88.61\footnotesize	{ $\pm$ 0.32} \\
      \midrule
      ResNet18 &  20 & 39.32\footnotesize	{ $\pm$ 7.21}  & 72.54\footnotesize	{ $\pm$ 3.03}  & 87.30\footnotesize	{ $\pm$ 0.41} \\
      & 100  & 40.38\footnotesize	{ $\pm$ 9.32}  & 74.36\footnotesize	{ $\pm$ 2.69}  &  87.39\footnotesize	{ $\pm$ 0.45} \\
      & 300 & 44.42\footnotesize	{ $\pm$ 10.97}  & 74.63\footnotesize	{ $\pm$ 3.28}  & 87.75\footnotesize	{ $\pm$ 0.62} \\
      &  500 & 40.59\footnotesize	{ $\pm$ 12.27}  & 76.97\footnotesize	{ $\pm$ 2.48}  &  87.55\footnotesize	{ $\pm$ 0.35} \\
      \bottomrule
    \end{tabular}
}
  \end{table}
Linked to the $g$ function, a second design choice concerns the \textbf{number of samples} stored in memory. This number is influenced by the number of classes in the dataset, the memory footprint, and the training time. Given fixed hardware and a minimum number of samples per class, the memory size grows according to the number of classes. A dataset with 1000 classes (e.g., ImageNet) requires a memory of 10000 samples to provide ten samples per class, which is a challenging requirement for most common workstations. Conversely, a binary dataset requires a small memory, and the footprint is negligible. Moreover, the larger the memory, the longer the training time due to the processing time of the samples. \Cref{table:impactSVHN} reveals that providing fewer than ten samples per class negatively impacts the performance. Providing more than ten samples is beneficial for some configurations. However, since the gain tends to diminish progressively and the impact on training time tends to increase, the number of samples becomes a hyperparameter to tune for Memory Wrap as a trade-off between them.

\paragraph{Similarity.}
  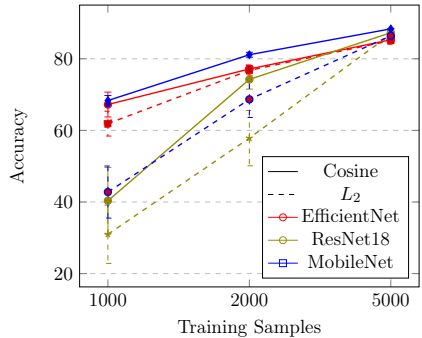
\begin{figure}[t]
\centering
\scalebox{0.7}{
\begin{tikzpicture}
\begin{axis}[
       xlabel={Training Samples},
    ylabel={Accuracy},
    xtick={1,2,3},
    xticklabels={1000,2000,5000},
    ytick={0,20,40,60,80,100},
    ymajorgrids=true,
    grid style=dashed,
  legend entries={Cosine,
  $L_2$,
                EfficientNet,
                ResNet18,
                MobileNet},
legend pos = south east
]
]
\addlegendimage{no markers,black}
\addlegendimage{dashed,black}
\addlegendimage{red, mark=o}
\addlegendimage{olive, mark=o}
\addlegendimage{blue, mark=square}
\addplot+[red,thick,error bars/.cd,y dir=both, y explicit]
    coordinates {
    (1,67.22) +- (0,3.47)
    (2,77.16) +- (0,0.97)
    (3,85.32) +- (0,0.90)
    };

\addplot+[red,dashed,thick,error bars/.cd,y dir=both, y explicit]
    coordinates {
    (1,61.89) +- (0,3.47)
    (2,76.76) +- (0,1.53)
    (3,85.09) +- (0,0.44)
    };

\addplot+[olive,thick,error bars/.cd,y dir=both, y explicit]
    coordinates {
    (1,40.35) +- (0,9.31)
    (2,74.24) +- (0,2.70)
    (3,87.29) +- (0,0.39)
    };

\addplot+[olive,dashed,thick,error bars/.cd,y dir=both, y explicit]
    coordinates {
    (1,30.98) +- (0,8.14)
    (2,57.84) +- (0,7.72)
    (3,86.87) +- (0,1.03)
    };

\addplot+[blue,thick,error bars/.cd,y dir=both, y explicit]
    coordinates {
    (1,68.34) +- (0,1.40)
    (2,81.14) +- (0,0.69)
    (3,88.36) +- (0,0.09)
    };

\addplot+[blue,dashed,thick,error bars/.cd,y dir=both, y explicit]
    coordinates {
    (1,42.78) +- (0,7.28)
    (2,68.67) +- (0,5.06)
    (3,86.43) +- (0,0.87)
    };

\end{axis}
\end{tikzpicture}
}
\caption{Comparison between different encoders trained using Cosine Similarity (solid lines) and $L_2$ distance (dashed lines) on a subset SVHN dataset.}
\label{fig:distance}
\end{figure}

Finally, \Cref{fig:distance} compares $L_2$ \textbf{distance measures} and the cosine similarity when used as a similarity for the content addressing mechanism. We observe that, as noted by previous studies ~\cite{Graves2016, Vinyals2016}, the cosine similarity outperforms $L_2$ in almost all the configurations,  representing the optimal choice for memory operations in neural networks.

\section{Contributions}
\label{sec:emann_contributions}
This chapter delved into the design of self-explainable DNNs based on memory. Specifically, it introduced two distinct designs of \glspl{MANN}. The first design is fully differentiable and capable of employing memory for both readings and writings. The second is restricted to reading from the memory. We analyzed the advantages and weaknesses associated with their utilization for enhancing the interpretability of \glspl{DNN}.

One of the strongest advantages lies in the fact that the augmentation based on memory improves or maintains the performance of the black-box model alone. These results align with the one obtained in the previous chapter (\Cref{chapter:pignn}). Consequently, the trade-off of opting for a more interpretable model is moved from the performance aspect to considerations related to the training time and the memory footprint.
These methods are also model-agnostic, denoting their agnosticism concerning the models they augment. Indeed, we applied their augmentation to three classes of architectures: recurrent, convolutional, and attention-based. Memory Wrap is also data-agnostic since it can be potentially applied to sequential data or any other kind of data as long as they are divided into separate samples. Additionally, we demonstrated their high flexibility in the types of explanations that can be derived, spanning from single and group feature attributions to explanations by example and counterfactuals. Alongside these strengths, we also analyzed some of the drawbacks of these approaches, such as the number of hyperparameters to tune in the case of fully differentiable architectures, their impact on the explanation process, the large footprint required by \glspl{MANN}, and the reliance on the skip connection of the controller, which remains a source of opacity in the decision process. 

This chapter closes the \Cref{part:senn} focused on the description of the designs that add novel layers to \glspl{DNN} to enhance their interpretability. Specifically, the contribution of this part includes a novel prototypes-based layer for GNNs and the introduction of a novel family of self-explainable DNNs: explainable memory-based architectures. The proposed designs preserve or even improve the performance of the black-box models to which they attach these layers, thus nullifying the typical performance trade-off of self-explainable DNNs. 

More importantly, we proved and showcased multiple kinds of intrinsic explanations that can be retrieved by inspecting the proposed designs and their components. The extracted explanations reach similar or better scores than the extrinsic methods in most of the considered metrics and appear more flexible than them. We show how retrieved explanations can help developers to understand better the models they are developing in verifying the correct behavior, detecting and understanding errors and biased predictions, or detecting reliable and unreliable predictions.

However, all these designs require a modification of the architecture, introducing a bottleneck, and training from scratch of the model. Moreover, these designs introduce novel hyperparameters that must be tuned. Conversely, the next part of the thesis focuses on techniques that can applied to trained networks and require small or no modifications at all.

\part{Explaining Latent Representations}
\label{part:neurons}
\chapter{Graph Concept Whitening}
\label{chapter:whitening}

In the previous part, the outlined techniques aimed to integrate novel modules between the feature extractor and the classifier, with the primary goal of enhancing interpretability while preserving the performance of black-box models to the greatest extent possible. These modules strive to simplify the understanding of the final computation stages and are jointly trained with the black-box feature extractor. Conversely, techniques described in this part focus on pre-trained models and how to improve the interpretation of the learned knowledge (i.e., the latent space). 

As a first step, this chapter presents \gls{GCW}~\cite{Proietti2023}. This normalization technique aims to ease the interpretation of the semantics encoded in the latent representations learned by the feature extractor,  subsequently used by the classifier to generate predictions. \gls{GCW} includes a few modifications of the feature extractor and a small alignment process, aiming as the previous part to preserve most of the black-box structure to not lower the performance. \gls{GCW} is inspired and adapts the concept whitening technique~\cite{Chen2019} to \glspl{GNN} and the chemical domain. In particular, it encourages the components of the latent representations to represent molecular properties, providing a useful entry point for network inspection and interpretability. The goal of such alignment is to support users of this domain on the tasks of molecular property prediction and drug discovery. 

The chapter is organized as follows: \Cref{sec:whitening_design} describes \gls{GCW} and how users can understand the learned behavior of the network; \Cref{sec:whitening_experiments} evaluates \gls{GCW} both in terms of performance and explanations quality; \Cref{sec:cw_choices} discusses alternative design choices; finally, \Cref{sec:cw_contributions} summarizes the contributions of this chapter.

\section{Design}
\label{sec:whitening_design}
The primary objective of \textbf{\gls{GCW}}~\cite{Proietti2023} is to facilitate the interpretation of the latent space of the \gls{GNN} by normalizing the latent space and ensuring that each dimension in the node embeddings represents a molecular property. The inspection of these dimensions aids the users in understanding the semantics encoded within the latent space.

Let $C=\{c_1, c_2..., c_k\}$ denote a predefined set of concepts (i.e., in our case, molecular properties). For each concept $c_j$, we can define a subset of the training dataset $X_{c_j}$ comprising only samples that exhibit that concept. Let $\bm{h}_i^l$ represent the latent representation of the sample $i$ returned by the layer $l$. The layer's goal is to align the $j$-th dimension of $\bm{h}$ to the concept $c_j$ and ensure that samples in $X_{c_j}$ exhibit larger values of $h_j$ compared to other samples in the training dataset.

To achieve this goal, the layer applies a normalization that centers around the mean, decorrelates, and rotates the latent space to ensure that samples $\in X_{c_j}$ related to the concept $c_j$ are highly activated along the $j$-th axis.\\
To decorrelate and standardize the data, the layer applies the following function to a set of latent representation $\bm{H} \in \mathbb{R}^{n\times d}$ of $n$ samples:
\begin{equation}
    g(\bm{H}) = W(\bm{H}-\mu1)
\end{equation}
where $\mu$ is the sample mean:
\begin{equation}
    \mu = \frac{1}{n}\sum_{i=\mathfrak{1}}^n \bm{H}_i
\end{equation}
The matrix $W$ of the previous equation is the whitening matrix, and it is defined such that:
\begin{equation}
    W^TW = \frac{1}{\frac{1}{n}(H-\mu1^T)(H-\mu1^T)^T}
\end{equation}
This matrix can be computed using several algorithms. Here, we follow \citet{Chen2020} and use the zero-phase component analysis~\cite{Huang2018}:
\begin{equation}
    \bm{H} = \bm{D} \bm{\Lambda}^{-\frac{1}{2}} \bm{D}^T
\end{equation}
where $\Lambda \in \mathbb{R}^{d\times d}$ is the eigenvalue diagonal matrix and $D \in \mathbb{R}^{d\times d}$ is the eigenvector matrix. Finally, to rotate the latent representation, the layer multiplies the $H$ matrix with an orthogonal matrix $Q \in \mathbb{R}^{d \times d}$ where each column $q_j$ represents the $j^{th}$ axis.

\begin{equation}
    \argmax_{\bm{q}_1,\bm{q}_2,...,\bm{q}_k} \sum_{j=1}^k \frac{1}{n_j} \bm{q}_j^Tg(H_{c_j} \mathfrak{1})
\end{equation}
subject to the constraint $Q^TQ=I_D$. Here, $H_{c_j}$ is the matrix containing the latent representations of $X_{c_j}$

\subsection{Alignment Process} 
The alignment process is applied as a fine-tuning phase and optimizes both the performance and alignment between the presence of the concepts and the axes. The alignment is achieved by updating the rotation matrix $Q$ via stochastic gradient descent. Each epoch of the alignment process of a model incorporating \gls{GCW} consists of two phases: \emph{accuracy optimization} and \emph{concept alignment}.
During accuracy optimization, the objective is to minimize the loss associated with classification accuracy (e.g., the cross-entropy loss):
\begin{equation}
    L_y = \sum_{i=1}^n CrossEntropy(f(x_i),y_i)
\end{equation}
In the second phase, the layer focuses on aligning the presence of concepts with the activation of axes. Specifically, it employs the following concept alignment loss:
\begin{equation}
    \argmax_{\bm{q}_1,\bm{q}_2,...,\bm{q}_k} \sum_{j=1}^k \frac{1}{n_j} \sum_{x_i^{c_j}\in X_{c_j}} \bm{q}_j^Tg(H^l)
\end{equation}
such that $Q^{T}Q=I_d$.

During the first phase, Q remains fixed and all the other parameters are updated. Conversely, during the second phase, the alignment process optimizes Q while freezing all other parameters. Specifically, the orthogonal matrix Q is updated using the Cayley transform:
\begin{equation}
    \bm{Q}^{(t+1)} = (I + \frac{\eta}{2}\bm{A})^{-1} \: \bm{Q}^{(t+1)} = (I + \frac{\eta}{2}\bm{A}) \bm{Q}^{(t)}
    \label{cayeley}
\end{equation}
where $\bm{A} = \bm{G}(\bm{Q}^{(t)})^T \: - \: \bm{Q}^{(t)}\bm{G}^T$ represents a skew-symmetric matrix, $\bm{G}$ is the gradient of the loss function and $\eta$ is the learning rate.
Since $\bm{G}$ requires a single scalar to compute the gradient, the concept activations must be projected into a scalar. \gls{GCW} uses the max function, selecting the maximum value over the concept activation. However, various alternatives can be explored during the hyperparameters tuning (\Cref{sec:cw_choices}).

\subsection{Explanations}
\label{sec:cw_expla}
\gls{GCW} is designed to improve the interpretation of the latent space activations. 
Specifically, the inspection of the activations of a concept $c_k$ can reveal the influence of that concept in the parsing of the input in the given layer. Conversely, the inspection of the full representation can help users understand the amount of semantics captured by the model until that point. Moreover, since \gls{GCW} is a normalization layer, we can apply the gradient or post-hoc techniques to project the concept's importance onto the input space. This feature is particularly useful in the case of abstract concepts, where there is no fixed direct correspondence between the concept and a set of features. This is a common scenario in the case of graph data and the chemical domain. Moreover, following the trace of importance across the layers, it is possible to inspect circuits~\cite{Olah2020} of how the network combines elements during its decision process.

\section{Experiments}
\label{sec:whitening_experiments}
\subsection{Concept Selection}
As described in the previous section, \gls{GCW} requires specifying the concepts to which the latent space must align. Since we deal with molecular datasets, in this case, we select concepts expressing known relevant molecular properties for each of the studied tasks. A concept is considered present in the current input when the corresponding property value is above or below a given threshold. 

The selected concepts are:
\begin{itemize}
    \item Quantitative Estimate of Drug-likeness~\cite{Bickerton2012} (QED), which expresses the similarity to the properties of oral drugs;
    \item TPSA~\cite{Palm1997}, which expresses the correlation with passive molecular transport through membranes;
    \item logarithm of the octanol-water Partition coefficient (logP), which estimates the molecules' hydro-lipophilicity balance~\cite{Kujawski2012};
    \item the number of nitrogen and oxygen atoms (NOCount);
    \item the number of heteroatoms ($\#$ Heteroatoms);
    \item molecular weight (mol\_weight), which is the sum of the atomic weight values of its atoms;
    \item number of H-Bond Acceptors (HBA);
    \item number of H-Bond Donors (HBD).
\end{itemize}

\begin{table}[b]
\centering
\caption{Avg. Accuracy and ROC-AUC over 15 runs on ClinTox when adding one concept at a time for the GCW layers. We highlight the best results of combined accuracy and ROC-AUC.}
\label{tab:clintox_concepts}
\begin{tabular}{lcc}
\toprule
Architecture & Accuracy & ROC-AUC\\
\midrule
Baseline  & 0.94 \footnotesize{$\pm$ 0.01} & 0.86 \footnotesize{$\pm$ 0.08}\\
GCW - QED  & \textbf{0.95 \footnotesize{$\pm$ 0.02}} & \textbf{0.95 \footnotesize{$\pm$ 0.04}}\\
GCW - TPSA  & 0.94 \footnotesize{$\pm$ 0.02} & 0.93 \footnotesize{$\pm$ 0.04}\\
GCW - mol\_weight  & \textbf{0.95 \footnotesize{$\pm$ 0.02}} &  \textbf{0.93 \footnotesize{$\pm$ 0.08}}\\
GCW - HBA  & \textbf{0.95 \footnotesize{$\pm$ 0.01}} & \textbf{0.95 \footnotesize{$\pm$ 0.02}}\\
GCW - HBD  & 0.94 \footnotesize{$\pm$ 0.01} & 0.93 \footnotesize{$\pm$ 0.08}\\
GCW - logP  & \textbf{0.95 \footnotesize{$\pm$ 0.01}} & \textbf{0.93 \footnotesize{$\pm$ 0.04}}\\
GCW - NOCount  & 0.95 \footnotesize{$\pm$ 0.02} & 0.91 \footnotesize{$\pm$ 0.09}\\
GCW - \# Heteroatoms  & \textbf{0.95 \footnotesize{$\pm$ 0.02}} & \textbf{0.95 \footnotesize{$\pm$ 0.03}}\\
\toprule
\end{tabular}
\end{table}

\subsection{Performance}
We start by comparing the black-box models and the same models fine-tuned by replacing Batch Normalization layers with \gls{GCW} layers. As black-boxes, we evaluate \gls{GCN}, \gls{GAT}, and \gls{GIN}. We test these models on BBBP, BACE, ClinTox, and HIV datasets. These datasets are popular chemical datasets for studying molecular properties and drug design. For BBBP, BACE, and HIV datasets, concepts are selected based on established literature~\cite{Sakiyama2021, Subramanian2016, Sirois2005, Kiralj2003}. We use QED, TPSA, LogP, NoCount, and  n\_heteroatoms for BBBP; QED, TPSA, mol\_weight, HBA, and HBD for BACE; and QED, nDoubleBonds, nO, HBA, HBD, and LogP for HIV. For ClinTox, we choose the best-performing concepts among those used for BBBP and BACE: QED, mol\_weight, HBA, and n\_heteroatoms (\Cref{tab:clintox_concepts}). Threshold values are determined based on Lipinski’s rule of five~\cite{Lipinski2001} for molecular weight, logP, HBA, and HBD, while other thresholds are set as the mean of the property across each dataset.
All the models include 128 units per layer. In \gls{GAT}, we set the number of attention heads to 2. Models are trained on BBBP and HIV using a batch size of 64, a learning rate of $1e-3$, and a weight decay of $5e-4$. Models are trained on BACE and Clintox using a batch size equal to 128, a learning rate equal to $1e-2$, and without weight decay. The fine-tuning lasts a maximum of 50 epochs with an early stopping with a patience of five epochs. All parameters have been fixed after a grid search over the hyperparameter space. Note that the fine-tuning phase is longer than the commonly used in the vision domain~\cite{Chen2020}. Black-box models are pre-trained using an early stopping with a patience of 20 epochs. We run the models from 15 different initializations and compute the average ROC-AUC and its standard deviation.

\begin{table}[b]
\centering
\caption{Avg. ROC-AUC and standard deviation over 15 runs before and after adding GCW to the models.}
\label{tab:GCW_performance}
\begin{tabular}{lcccc}
\toprule%
Model & BBBP & BACE & ClinTox & HIV \\
\midrule
GCN                  & 0.88 \footnotesize{$\pm$ 0.03}            & 0.86 \footnotesize{$\pm$ 0.02}            & 0.86 \footnotesize{$\pm$ 0.08}            & \textbf{0.79 \footnotesize{$\pm$ 0.03}} \\
+ GCW                 & \textbf{0.91 \footnotesize{$\pm$ 0.02}}   & \textbf{0.92 \footnotesize{$\pm$ 0.02}}   &  \textbf{0.93\footnotesize{$\pm$ 0.05}}   &  \textbf{0.79 \footnotesize{$\pm$ 0.02}} \\
\midrule
GAT                  & 0.89 \footnotesize{$\pm$ 0.03}            & 0.86 \footnotesize{$\pm$ 0.02}            & 0.84\footnotesize{$\pm$ 0.08}             & \textbf{0.80 \footnotesize{$\pm$ 0.03}} \\
+ GCW                 & \textbf{0.97 \footnotesize{$\pm$ 0.02}}   & \textbf{0.93 \footnotesize{$\pm$ 0.01}}   & \textbf{0.88\footnotesize{$\pm$ 0.05}}    & \textbf{0.80 \footnotesize{$\pm$ 0.02}} \\
\midrule
GIN                  & \textbf{0.89 \footnotesize{$\pm$ 0.02}}   & 0.87 \footnotesize{$\pm$ 0.03}            & \textbf{0.75\footnotesize{$\pm$ 0.11}}    & \textbf{0.78 \footnotesize{$\pm$ 0.03}}\\
+ GCW                 & \textbf{0.91 \footnotesize{$\pm$ 0.02}}   & \textbf{0.91 \footnotesize{$\pm$ 0.02}}   & \textbf{0.74\footnotesize{$\pm$ 0.11}}    & \textbf{0.79 \footnotesize{$\pm$ 0.02}}\\
\midrule
Random forest   & 0.79 \footnotesize{$\pm$ 0.01} & 0.79 \footnotesize{$\pm$ 0.01} & 0.74 \footnotesize{$\pm$ 0.01} & 0.65 \footnotesize{$\pm$ 0.01} \\
MLP             & 0.87 \footnotesize{$\pm$ 0.01} & 0.84 \footnotesize{$\pm$ 0.01} & 0.75 \footnotesize{$\pm$ 0.05} & 0.84 \footnotesize{$\pm$ 0.01}\\
MPNN            & 0.88 \footnotesize{$\pm$ 0.02} & 0.70 \footnotesize{$\pm$ 0.03} & 0.72 \footnotesize{$\pm$ 0.02} & 0.75 \footnotesize{$\pm$ 0.01}\\
\bottomrule
\end{tabular}
\end{table}

\Cref{tab:GCW_performance} reports the ROC-AUC of the black-box models, the same models fine-tuned by using \gls{GCW}, and three commonly used baselines in these datasets: random forest, \gls{MLP}, and message-passing neural networks (MPNN)~\cite{Gilmer17}. The results demonstrate the performance benefits of fine-tuning models using \gls{GCW}. Specifically,  \textbf{models fine-tuned with \gls{GCW} outperform the black-box variants} in all the configurations and the baselines in all but in the HIV dataset. In this dataset, \gls{MLP} reaches the best performance against all the competitors. The improvement of the \gls{GCW} models can be attributed to the ability of the \gls{GCW} layers to encourage node embeddings to represent class-specific information, aiding discrimination among classes. These results align with the ones reported by \citet{Chen2020}. The only difference is in the number of epochs needed to reach comparable results. While they fine-tune the models for just one epoch, \glspl{GNN} require up to 50 epochs to align the representations and reach comparable or superior performance.

\subsection{Explanations}
\begin{figure}[t]
    \centering
    \includegraphics[scale=0.7]{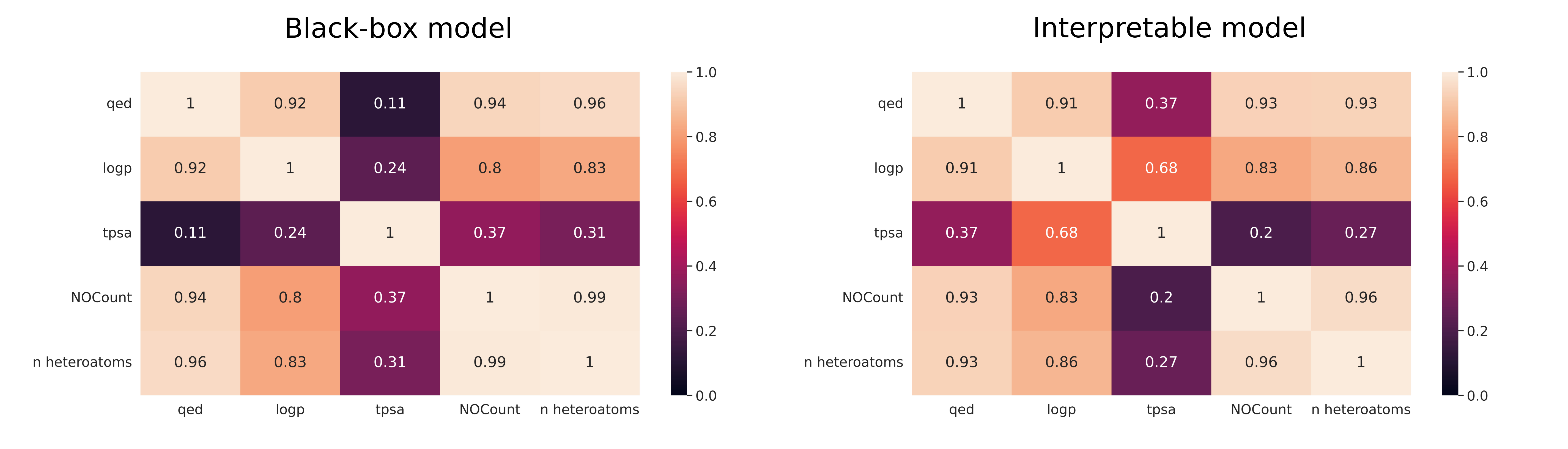}
    \caption{Separability score of the latent space of a black-box model and its interpretable version. Lower is better.}
    \label{fig:separability_cw}
\end{figure}
In this section, we investigate and showcase the interpretability improvements of models employing \gls{GCW} layers. We use \gls{GAT} trained on BBBP as the example model. However, similar results can be obtained for other models.

We begin the investigation by comparing the separability (\Cref{metric:separability}) in the latent space exposed by the black-box models and the models incorporating \gls{GCW}. \Cref{fig:separability_cw} shows that \textbf{models with \gls{GCW} layers achieve superior separability}. In particular, note that molecular properties that are not similar, like NoCount and TPSA, exhibit better separation in in the interpretable model than in the black-box model (score 0.2 vs 0.37). Conversely, chemically related properties (e.g., TPSA and LogP) are similar in the representation (score 0.68). These results confirm a better disentanglement induced by \gls{GCW} useful for improving the interpretability of the axes.

To quantify the improvement in interpretability, we examine the impact of the interpretable design on other types of explanations, such as feature attribution toward the predicted class. To measure the impact, we report in \Cref{tab:cw_fidelity} the $fidelity+^{prob}$ of the feature attribution computed by GNNExplainer~\cite{Ying2019} before and after the fine-tuning phase. The results demonstrate that incorporating \gls{GCW} layers does not compromise the interpretability scores compared to when applied on black-boxes. While performance is comparable, the mean fidelity is usually higher and the standard deviation lower, suggesting \textbf{more stable and robust explanations}.
\begin{table}[b!]
\centering
\caption{Avg. and standard deviation of the Fidelity+ score of feature attributions computed on both black-boxes and models employing GCW layers.}
\label{tab:cw_fidelity}
\begin{tabular}{lcccc}
\toprule%
Model & BBBP & BACE & ClinTox & HIV \\
\midrule
Black-box  & 0.24 \footnotesize{$\pm$ 0.41} &  0.45 \footnotesize{$\pm$ 0.45} & -0.12 \footnotesize{$\pm$ 0.25} & 0.00 \footnotesize{$\pm$ 0.01} \\
+GCW  & 0.31 \footnotesize{$\pm$ 0.38} & 0.45 \footnotesize{$\pm$ 0.41} & -0.06 \footnotesize{$\pm$ 0.12} & 0.02 \footnotesize{$\pm$ 0.19} \\
\bottomrule
\end{tabular}
\end{table}

\begin{figure}[t]
    \centering
    \includegraphics[scale=0.4]{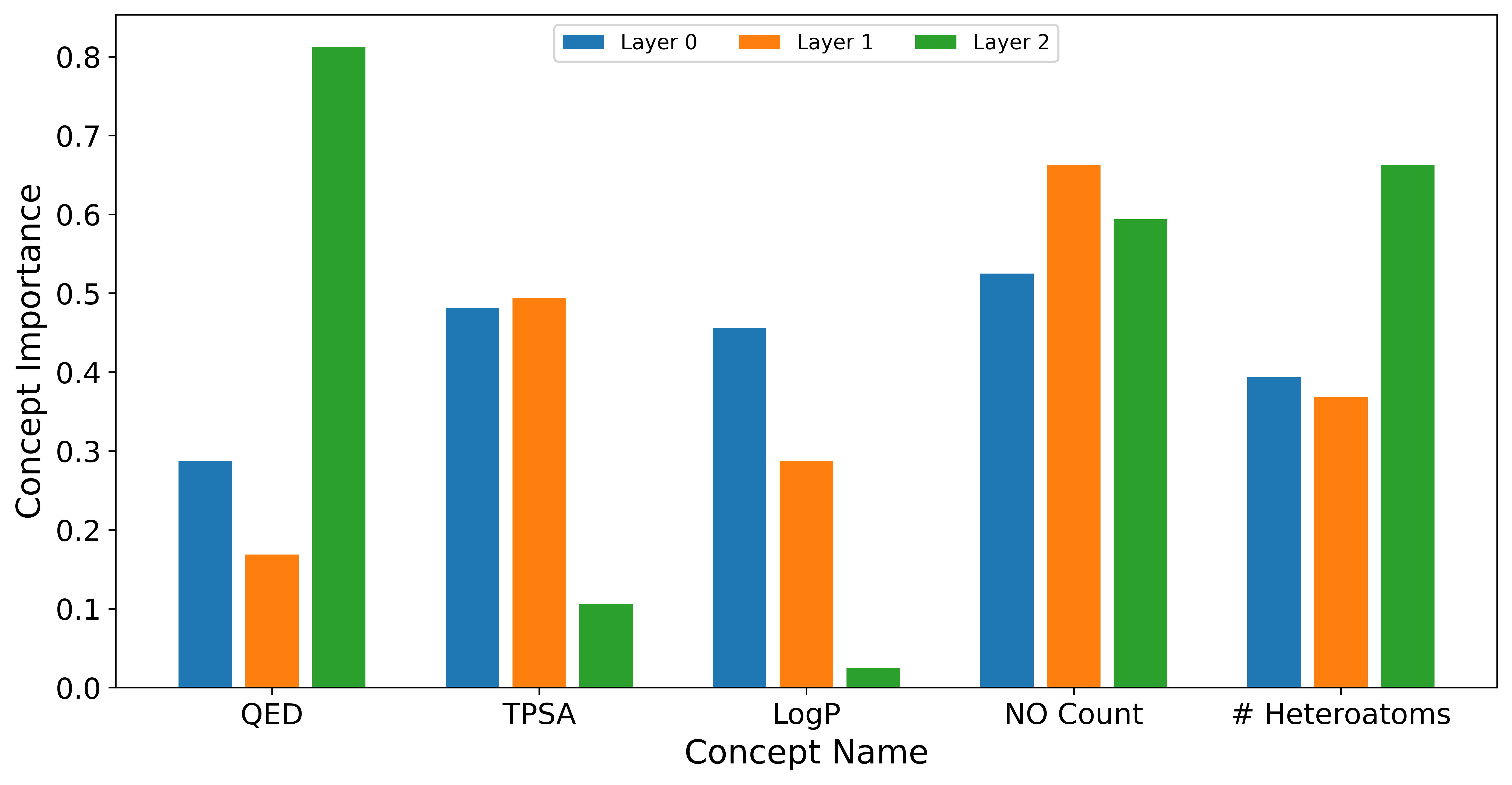}
    \caption{Avg. importance scores assigned to five concepts in the layers of a GAT model trained on BBBP.}
    \label{fig:layer_importance}
\end{figure}
Once the interpretability improvements have been validated, we showcase how the concept-aligned latent space can be used to study molecular properties and help users in \textbf{drug discovery}.
The first example explores the flow of learned knowledge across the network. In particular, we compute the concepts' importance across layers by using the positive directional derivative of the gradient cumulated on the concept axis. As shown in \Cref{fig:layer_importance}, some properties are more important in lower layers than higher ones (e.g., TPSA and logP), while others exhibit the opposite trend (e.g., QED and \#Heteroatoms). This phenomenon resembles the behavior observed for CNN on the vision domain, where lower layers detect shapes and colors, while higher layers detect the composition of concepts. Moreover, lower layers exhibit an almost uniform importance across different concepts; thus, the network uses information from all the concepts at this stage. Conversely, higher layers exhibit significant differentiation, and the network uses only a subset of concepts. We also highlight that the most important concepts in the last hidden layer, which are then used by the classifier to perform the prediction, are QED, \#Heteroatoms, and NOCount. These findings align with the literature on drug discovery and the ones of \citet{Sakiyama2021}. 
\begin{figure}[t]
    \centering
    \includegraphics[scale=0.08]{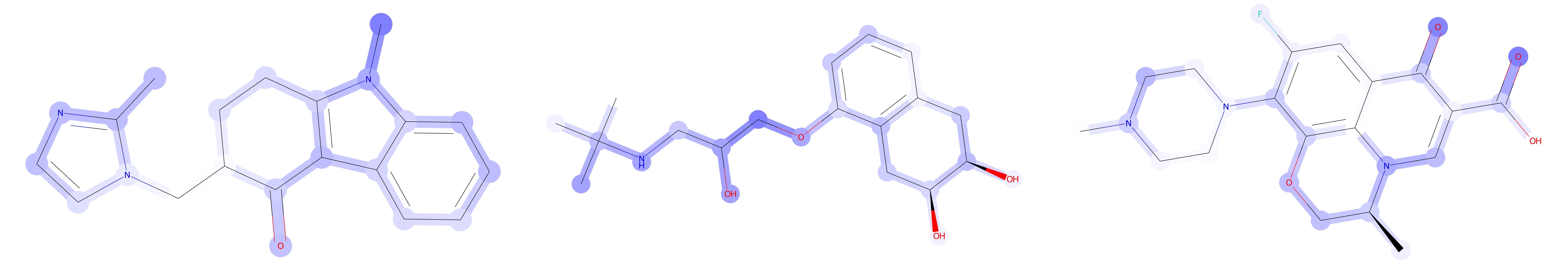}
    \caption{Feature attribution for the concept \#Heteroatoms in the BBBP dataset.}
    \label{fig:heteroatoms}
\end{figure}

The second example shows how it is possible to use the concept axes to \textbf{discover atom relationships that can be exploited for the optimization of the molecules’ activities} to design novel drugs. As anticipated in \Cref{sec:cw_expla}, it is possible to project concept activation onto input features and understand the influence of features on input parsing. \Cref{fig:heteroatoms} showcase an application of this procedure to the \#Heteroatoms concept. The figure displays three samples that mostly activate the concept's axis. Notably, oxygen and nitrogen atoms receive the largest attribution. These results confirm the quality of the learned alignment as oxygen and nitrogen are heteroatoms and decisive in determining the molecule's class.

\section{Design Choices}
\label{sec:cw_choices}
This section delves into alternative choices for \gls{GCW}, specifically focusing on the activation mode and the layers to replace.
\paragraph{Activation Modes.} In the previous sections, we assumed the maximum value over the node embeddings as the activation score. Here, we explore alternative activation modes: \emph{mean},  \emph{pos\_mean},  \emph{topk\_pool}, and  \emph{weighted\_topk\_pool}. The mean mode computes the mean over the node embedding. The pos\_mean computes the mean over only the positive value of the node embedding. Topk\_pool computes the mean over the subgraph node embeddings extracted using the recently proposed topk\_pool operator~\cite{Gao2019}. Finally, the weighted\_topk\_pool activation pool computes the mean as in topk\_pool but weighs the contribution of each component by the projection score returned by the topk\_pool operator~\cite{Gao2019}.
\begin{table}[!b]
\begin{center}
\caption{Avg. ROC-AUC and standard deviation reached by models including GCW using different activation modes.}
\label{tab:activation_modes}
\begin{tabular}{lcccc}
\toprule
Activation Mode & BBBP & BACE & ClinTox & HIV \\
\midrule
GCN          \\                      
$\:-$ max                  & \textbf{0.91 \footnotesize{$\pm$ 0.02}}   & \textbf{0.92 \footnotesize{$\pm$ 0.02}}   &  \textbf{0.93\footnotesize{$\pm$ 0.05}}   &  \textbf{0.79 \footnotesize{$\pm$ 0.02}} \\
$\:-$ mean                 & 0.88 \footnotesize{$\pm$ 0.03}            & \textbf{0.92 \footnotesize{$\pm$ 0.03}}   & \textbf{0.95 \footnotesize{$\pm$ 0.03}}   & 0.72\footnotesize{$\pm$ 0.02} \\
$\:-$ pos\_mean            & \textbf{0.89 \footnotesize{$\pm$ 0.03}}   & \textbf{0.92 \footnotesize{$\pm$ 0.02}}   &  \textbf{0.95 \footnotesize{$\pm$ 0.02}}  & \textbf{0.77 \footnotesize{$\pm$ 0.02}} \\
$\:-$ topk\_pool           & 0.87 \footnotesize{$\pm$ 0.03}            & \textbf{0.92 \footnotesize{$\pm$ 0.01}}   &  0.92\footnotesize{$\pm$ 0.08}            &  0.70 \footnotesize{$\pm$ 0.02} \\
$\:-$ weighted\_topk\_pool & 0.81 \footnotesize{$\pm$ 0.06}            & \textbf{0.93 \footnotesize{$\pm$ 0.02}}   &  0.92 \footnotesize{$\pm$ 0.10}           & 0.69 \footnotesize{$\pm$ 0.03}  \\
\midrule
GAT      \\                          
$\:-$ max                  & \textbf{0.97 \footnotesize{$\pm$ 0.02}}   & \textbf{0.93 \footnotesize{$\pm$ 0.01}}   & \textbf{0.88\footnotesize{$\pm$ 0.05}}    & \textbf{0.80 \footnotesize{$\pm$ 0.02}} \\
$\:-$ mean                 & 0.92 \footnotesize{$\pm$ 0.05}            & \textbf{0.93 \footnotesize{$\pm$ 0.02}}   & \textbf{0.90 \footnotesize{$\pm$ 0.03}}   & 0.75 \footnotesize{$\pm$ 0.02} \\
$\:-$ pos\_mean            & 0.91 \footnotesize{$\pm$ 0.02}            & 0.91 \footnotesize{$\pm$ 0.02}            & \textbf{0.88\footnotesize{$\pm$ 0.04}}    & \textbf{0.80 \footnotesize{$\pm$ 0.02}} \\
$\:-$ topk\_pool           & 0.92 \footnotesize{$\pm$ 0.03}            & \textbf{0.94 \footnotesize{$\pm$ 0.02}}   & \textbf{0.90\footnotesize{$\pm$ 0.04}}    & 0.76 \footnotesize{$\pm$ 0.02} \\
$\:-$ weighted\_topk\_pool & 0.89 \footnotesize{$\pm$ 0.03}            & \textbf{0.93 \footnotesize{$\pm$ 0.02}}   & \textbf{0.90\footnotesize{$\pm$ 0.04}}    & 0.74 \footnotesize{$\pm$ 0.04} \\
\midrule
GIN      \\                           
$\:-$ max                  & \textbf{0.91} \footnotesize{$\pm$ 0.02}   & \textbf{0.91 \footnotesize{$\pm$ 0.02}}   & \textbf{0.74\footnotesize{$\pm$ 0.11}}    & \textbf{0.79 \footnotesize{$\pm$ 0.02}}\\
$\:-$ mean                 & 0.88 \footnotesize{$\pm$ 0.03}            & \textbf{0.92 \footnotesize{$\pm$ 0.02}}   & \textbf{0.70\footnotesize{$\pm$ 0.15}}    & 0.71\footnotesize{$\pm$ 0.02} \\
$\:-$ pos\_mean            & \textbf{0.90 \footnotesize{$\pm$ 0.02}}   & \textbf{0.92 \footnotesize{$\pm$ 0.02}}   & \textbf{0.69\footnotesize{$\pm$ 0.12}}    & \textbf{0.79 \footnotesize{$\pm$ 0.02}} \\
$\:-$ topk\_pool           & 0.84 \footnotesize{$\pm$ 0.04}            & 0.82 \footnotesize{$\pm$ 0.02}            & \textbf{0.73\footnotesize{$\pm$ 0.13}}    & 0.70 \footnotesize{$\pm$ 0.03} \\
$\:-$ weighted\_topk\_pool & 0.85 \footnotesize{$\pm$ 0.03}            & \textbf{0.92 \footnotesize{$\pm$ 0.02}}   & \textbf{0.77 \footnotesize{$\pm$ 0.08}}   & 0.71 \footnotesize{$\pm$ 0.03} \\
\bottomrule
\end{tabular}
\end{center}
\end{table}

From \Cref{tab:activation_modes}, we can observe that the max activation mode is the best across several datasets and models. More in detail, in the BBBP and HIV datasets, this activation mode outperforms all the others by large margins, except for the pos\_mean mode where the gap is lower. Conversely, in BACE and ClinTox all the activation modes achieve comparable performance, with weighted\_topk\_pool even reaching the maximum possible mean in one case (i.e., in the ClinTox dataset). 

\paragraph{Normalization Layers.}
\begin{table}[!b]
\begin{center}

\caption{Avg. ROC-AUC and standard deviation of models when GCW replaces several types of normalization layers.}
\label{tab:normalization}
\begin{tabular}{lcccc}
\toprule%
Normalization type & BBBP & BACE & ClinTox & HIV \\
\midrule
BatchNorm       & 0.89 \footnotesize{$\pm$ 0.03}           & 0.86 \footnotesize{$\pm$ 0.02}          & 0.86 \footnotesize{$\pm$ 0.08}                   & \textbf{0.80 \footnotesize{$\pm$ 0.03}}\\
$\:-$ GCW       & \textbf{0.97 \footnotesize{$\pm$ 0.02}}  & \textbf{0.93 \footnotesize{$\pm$ 0.01}} & \textbf{0.93 \footnotesize{$\pm$ 0.05}}          & \textbf{0.80 \footnotesize{$\pm$ 0.02}}\\
LayerNorm       & \textbf{0.88 \footnotesize{$\pm$ 0.03}}  & \textbf{0.88 \footnotesize{$\pm$ 0.03}} & \textbf{0.83 \footnotesize{$\pm$ 0.03}}          & 0.71 \footnotesize{$\pm$ 0.04} \\
$\:-$ GCW       & \textbf{0.88 \footnotesize{$\pm$ 0.03}}  & \textbf{0.86 \footnotesize{$\pm$ 0.03}} & 0.71 \footnotesize{$\pm$ 0.09}                   & \textbf{0.81 \footnotesize{$\pm$ 0.02}} \\
InstanceNorm    & 0.60 \footnotesize{$\pm$ 0.12}           & 0.69 \footnotesize{$\pm$ 0.07}          & 0.58 \footnotesize{$\pm$ 0.06}                   & 0.61 \footnotesize{$\pm$ 0.05}\\
$\:-$ GCW       & \textbf{0.92 \footnotesize{$\pm$ 0.03}}  & \textbf{0.94 \footnotesize{$\pm$ 0.02}} & \textbf{0.92 \footnotesize{$\pm$ 0.05}}          & \textbf{0.81 \footnotesize{$\pm$ 0.02}} \\
GraphNorm       & 0.59 \footnotesize{$\pm$ 0.20}           & 0.76 \footnotesize{$\pm$ 0.05}          & 0.56 \footnotesize{$\pm$ 0.09}                   & 0.60 \footnotesize{$\pm$ 0.05}\\
$\:-$ GCW       & \textbf{0.89 \footnotesize{$\pm$ 0.03}}  & \textbf{0.92 \footnotesize{$\pm$ 0.02}} & \textbf{0.82 \footnotesize{$\pm$ 0.10}}          & \textbf{0.80 \footnotesize{$\pm$ 0.02}} \\
\bottomrule
\end{tabular}
\end{center}
\end{table}
Here, we test the efficacy of \gls{GCW} when used to replace different types of normalization layers. We train from scratch the best-performing architectures of the previous experiments, replacing their Batch Normalization layers with Layer Normalization, Instance Normalization, and Graph Normalization layers. Then, we fine-tune the trained model, replacing the normalization layers with \gls{GCW}, and collect the ROC-AUC before and after the fine-tuning. \Cref{tab:normalization} suggests that \gls{GCW} can be used as a replacement for all the types of normalization layers since, most of the time, it reaches comparable or better performance than the original normalization. Note the significant gap when \gls{GCW} replaces the Instance Normalization and Graph Normalization layers. We hypothesize that the gap is due to a mismatch between these normalizations and the backbone architectures. In these cases, \gls{GCW} recovers the performance since it is a variant of batch normalization layers, and thus, it is more similar to the one used in the design of the backbone models.
\section{Contributions}
\label{sec:cw_contributions}
This chapter contributed to the research on making the interpretation of latent space easier by investigating the application of concept whitening to graph data and in the chemical domain. The chapter analyzed the differences and specific characteristics of the graph applications, along with the advantages and weaknesses associated with their utilization for enhancing the interpretability of \glspl{GNN}. 

Notably, the approach preserves the performance of the black-box models while improving their interpretability in terms of the separability of the latent space and explanation fidelity when used in combination with post-hoc methods. With \gls{GCW}, we take a step forward with respect to the approaches described in the previous chapters. Instead of adding elements to the network, this approach modifies only the normalization layers to improve the interpretability of the network. We discussed how this modification can be performed in the fine-tuning stage without loss of accuracy and how to inspect the decision process of the network via the inspection of the latent space. In particular, we showed how a disentangled latent space can help chemists in molecular activity prediction and drug discovery tasks.

Alongside these strengths, \gls{GCW} presents some drawbacks. For example, the selection of concepts and the dependence on the prior design of the architecture (i.e., it performs better when replacing BatchNormalization layers) represent limitations to the applicability of the approach compared to all the previous approaches, which are, instead, model-agnostic. Moreover, it still requires a new training phase and a modification of the architecture, even if it is small. These limitations will be addressed in the next chapter, where we will present a post-training algorithm applicable to any DNN.


\chapter{Clustered Compositional Explanations}
\label{chapter:compositional}

This chapter shares the same motivation as the previous chapter, elucidating the semantics encoded in the learned latent representations. While Graph Concept Whitening proposes modifying the architecture to enforce the semantics, the technique described in this chapter aims to extract the semantics encoded in the components of the latent representation of any pre-trained black-box model without any modification. 

This chapter focuses on the family of methods that investigate what kind of information is recognized by neurons by probing them using a concept dataset~\cite{Bau2017, Mu2020, Massidda2023} (\Cref{sec:related_post}). The idea is to probe each neuron and select, as an explanation, the label whose annotations are aligned the most with its activations.  The current literature~\cite{Bau2017, Mu2020, Hernandez2022, Massidda2023} focuses only on the exceptionally high activations due to the computational challenges associated with considering wider ranges. Conversely, this chapter describes an \textbf{algorithm that combines heuristics and clustering to extract the composition of concepts recognized by neurons at several intervals of activations}, mitigating the computational challenges connected to their exploration and the superposition issue (\Cref{sec:back_xai}). By using this algorithm, it is possible to extract novel findings about the phenomena related to models' activations, helping us to understand these black-boxes better.

The chapter is organized as follows: \Cref{sec:neurons_design} describes the proposed algorithm and its heuristic;
\Cref{sec:neurons_metrics} introduces and summarizes metrics to use to evaluate the explanations' quality; \Cref{sec:neurons_experiments} evaluates the algorithm in terms of efficiency and explanations quality and sheds light on phenomena related to neurons activations; \Cref{sec:neurons_choices} discusses design choices for the proposed algorithm; finally, \Cref{sec:pignn_contributions} summarized the contributions of this chapter.

\section{Algorithm Design}
\label{sec:neurons_design}

\begin{figure}
    \centering
    \includegraphics[scale=0.4]{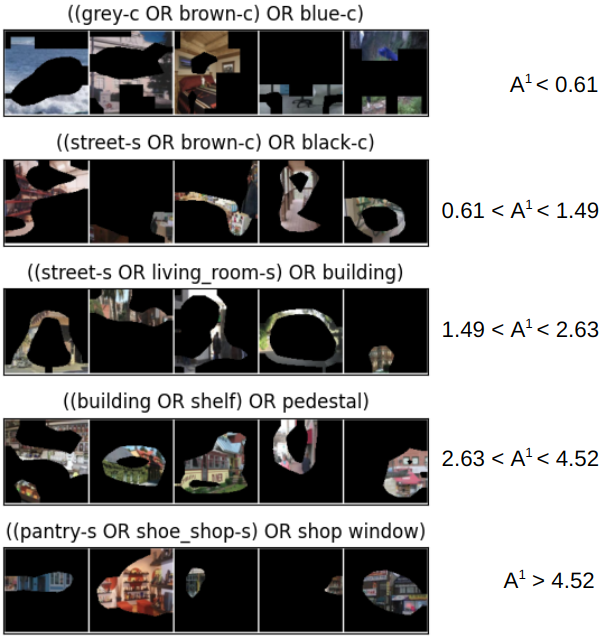}
    \caption{Explanations in terms of recognized concepts and sample examples of different activation ranges for unit \#1.}
    \label{fig:neurons_splash}
\end{figure}
The proposed algorithm, named \textbf{Clustered Compositional Explanations algorithm}, aims to describe a wider spectrum of neuron activations than current methods. To achieve the goal, the algorithm combines clustering algorithms with a beam search guided by heuristics. The idea is to cluster the activations of a neuron, compute explanations for each cluster, and provide a summary of the explanations to the user (\Cref{fig:neurons_splash}). 

Specifically, let $\mathfrak{D}$ be a dataset where each sample can include concept annotations. Each annotation is represented by a function  $S(x, L)$ that returns a binary mask for a sample $x$ where only the features associated with a concept labeled as $L$ are set to one. Given the matrix $A^k(x)$, which represent  the activation map of the $k$ neuron when the input is the sample $x$;  the algorithm computes a set of activation thresholds:
\begin{equation}
    \mathrm{T} = \{ [min(Cls),max(Cls)],  \forall Cls \in  Clustering(A^k(\mathfrak{D}))\}
\end{equation} 
where $Clustering(A^k(\mathfrak{D}))$ returns a set of $n_{cls}$ clusters including non-zero activations of the $k$ neuron. For simplicity, we assume the clusters are disjoint so that two values are enough to identify a cluster precisely. Therefore, each element in $\mathrm{T}$ is a pair of threshold $[\tau_i,\tau_{j}]$ where $\tau_i$ is the minimum value of activations included in the given cluster and $\tau_j$ is the maximum value. However, the approach can also be applied to overlapping clusters as long as they can be identified by a set of thresholds.

For each pair of thresholds (or equivalently for each cluster), the algorithms compute the matrix $M^k_{[\tau_i,\tau_j]}(\mathfrak{D})$, which is a binary mask of the activations $A^k(\mathfrak{D})$ obtained by setting to 0 all the values lower than $\tau_i$ and greater than $\tau_j$.

At this point, the algorithm assigns a label to each cluster by extracting the one whose annotations overlap the most with the matrix $M^k_{[\tau_i,\tau_j]}(\mathfrak{D})$
\begin{equation}
    L^{best} = \{ \operatorname*{arg\,max}_{L \in \mathfrak{L}^n} IoU(L, \tau_i,\tau_{j}, \mathfrak{D} ),  \forall~[\tau_i,\tau_{j}] \in \mathrm{T}\}
    \label{eq:objective_cce}
\end{equation}
where $\mathfrak{L}^l$ is the set of logical connections arity $l$ that can be built between concepts $L$ from the concept dataset $\mathfrak{D}$;

In this equation, $IoU(L, \tau_i,\tau_{j}, \mathfrak{D} )$ refer to \Cref{metric:iou}, adapted to the context so that:
\begin{equation}
    IoU(L, \tau_1,\tau_2, \mathfrak{D} ) = \frac{\sum_{x \in \mathfrak{D}}|M_{[\tau_1,\tau_2]}(x) \cap S(x,L)|}{\sum_{x \in \mathfrak{D}}|M_{[\tau_1,\tau_2]}(x) \cup S(x,L)|}
\label{eq:iou}
\end{equation}
For each cluster, the label can be extracted by using the \gls{CoEx} algorithm~\cite{Mu2020}.

This algorithm assigns logical formulas of arity $n$ to a range of activations by performing a beam search of size $b$ over the most promising logical connections of atomic concepts. At each step $i$, only the $b$ best labels of arity $i$ are used as bases for computing labels of arity $i+1$. The first beam is selected among the labels associated with the best scores computed by the \gls{NetDissect} algorithm~\cite{Bau2017}. \gls{NetDissect} optimizes Equation (\ref{eq:objective_cce}) by performing an exhaustive search over the space of all the atomic concepts in the dataset. 

However, by design, the vanilla beam search implemented by the \gls{CoEx} algorithm would require $n_{cls}\times (n-1) \times b$  times the computation time of \gls{NetDissect} to compute the labels for all the clusters in a single neuron, where $b$ and $(n-1)$ are the width and deepness of beam search tree, respectively. This computational time is practically unfeasible for most of the current deep neural networks, thus impeding the analysis of a wider spectrum of activations. 

To mitigate this issue, we propose the \emph{Min-Max Extension per Sample Heuristic} (\textbf{MMESH}) to guide the beam search and cut the search space. 
The purpose of the heuristic is to estimate the IoU score and sort the search space based on the estimation. If the heuristic is admissible, and thus, it never underestimates the true IoU, then it is possible to cut the search space at each step of the beam search by removing all labels whose estimated IoU score is below the best score found so far. Obviously, the time employed to estimate the scores and to sort the search space must be lower than the time needed to explore the full search space. In the considered scenario, the bottleneck is represented by the repeated multiplication between large matrices (as large as the dataset), and a good heuristic should avoid performing them as much as possible.

To achieve this goal, \textbf{MMESH exploits the coordinates that identify the smallest and the largest possible extension of the concepts in each dataset sample}. Given a sample $x$, \emph{the smallest possible extension corresponds to the longest contiguous segment that includes only concept features}. For example, in the case of 1D data, it corresponds to the shortest subsequence that includes only concept features, and the coordinates are the sequence's first and last index. In the image data, the smallest possible extension corresponds to the largest rectangle inscribed inside the polygon defined by the segmentation mask associated with the concept, and the coordinates are the ones of the top left and bottom right corners of the rectangle. Conversely, \emph{the largest possible extension corresponds to the shortest contiguous segment that includes all the features associated with the concept}. In this case, for 1D data, it corresponds to the subsequence that starts from the first occurrence of the concept's features and ends in the last occurrence. For image data, the largest possible extension corresponds to the bounding box covering the entire concept's mask, and the coordinates are the ones of the box's top left and bottom right corners. 

\begin{figure}[t!]
    \centering
    \includegraphics[scale=0.3]{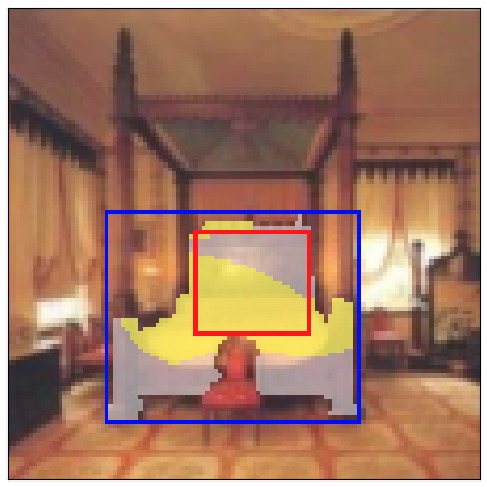}
    \caption{Visualization of information used by MMESH: the size of the concept mask (white), size of the intersection (yellow), minimum extension (red bounding box), maximum extension (blue bounding box).}
    \label{fig:heuristic_example}
\end{figure}

Given a sample $x$, a neuron $k$, and a label $L \in \mathfrak{L}^{i}$, MMESH estimates the IoU score by combining the two quantities described above, the size of the label mask on the sample $S(x, L)$, and the size of the intersection between the label's terms mask and the neuron's activation on the sample $IMS(x, t) \forall t \in L$. The last quantity can be computed during the execution of the previous beam. \Cref{fig:heuristic_example} visualizes the information used by the heuristic.

MMESH supports the estimation of logical formulas connected by OR, AND, and AND NOT operators, which are the most used operators~\cite{Mu2020, Makinwa2022, Massidda2023, Harth2022} for compositional explanations. To estimate the IoU score, MMESH must estimate both the intersection $\widehat{I}$ and the union $\widehat{U}$. Expanding $\widehat{U}$, we can rewrite Equation (\ref{eq:iou}) as:
\begin{equation}
\begin{split}
    IoU(L, \tau_1,\tau_2, \mathfrak{D} ) &  = \frac{\widehat{I}}{\widehat{U}} =   \frac{\sum_{x \in \mathfrak{D}}\widehat{I_x}}{\sum_{x \in \mathfrak{D}}\widehat{U_x}} =     \\
    & = \frac{\widehat{I_x}}{\sum_{x \in \mathfrak{D}} |M_{[\tau_1,\tau_2]}(x)| + \sum_{x \in \mathfrak{D}}|\widehat{S(x,L})| - \widehat{I_x}}
\end{split} 
\label{eq:iouextended}
\end{equation}
Since $|M_{[\tau_1,\tau_2]}(x)|$ is unique, it is shared among all the labels. Therefore, it can be computed one time per cluster, and it does not need to be estimated. 
As previously mentioned, the heuristic needs to be admissible in order to be used effectively. Therefore, the estimation of the numerator $\widehat{I}$ must be an overestimation, and the estimation of the denominator $\widehat{I}$ must be an underestimation. Hence, the heuristic assumes the best-case scenario for each logical operator for the estimation of the numerator since it must be an overestimation:
\begin{numcases} {\widehat{I}_x=}
min(|IMS_{[\tau_1,\tau_2]}(x, L_{\leftarrow})|+ |IMS_{[\tau_1,\tau_2]}(x, L_{\rightarrow})|, |M_{[\tau_1,\tau_2]}(x)| ) & $op=OR$  \label{eq:esti_ix_OR}\\
min(|IMS_{[\tau_1,\tau_2]}(x, L_{\leftarrow})|, |IMS_{[\tau_1,\tau_2]}(x, L_{\rightarrow})|) & 
$op=AND$ \label{eq:esti_ix_AND}\\
min(|IMS_{[\tau_1,\tau_2]}(x, L_{\leftarrow})|, |M_{[\tau_1,\tau_2]}(x)| - |IMS_{[\tau_1,\tau_2]}(x, L_{\rightarrow})|)  &  
$op=AND~NOT$ \label{eq:esti_ix_AND_NOT}
\end{numcases}

Where $op$ is the logical connector of the formula, $L_{\leftarrow} \in \mathfrak{L}^{l-1}$ and $L_{\rightarrow} \in \mathfrak{L}^1$ denote the left side and the right side of a label of arity $i$ obtained by adding an atomic term to the label at each step, respectively. Equation (\ref{eq:esti_ix_OR}) corresponds to assuming disjoint labels (i.e., masks). Conversely,  Equation (\ref{eq:esti_ix_AND}) and Equation (\ref{eq:esti_ix_AND_NOT}) assume fully overlapping masks.

Conversely, in the case of the denominator, since the estimation must be an underestimation, the heuristic assumes the worst-case scenario for each operator:

\begin{numcases} {\widehat{S(x,L)}=}
max(|S(x,L_{\leftarrow})|, |S(x,L_{\rightarrow})|, \widehat{S(x,L_{\leftarrow} \cup L_{\rightarrow}}), \widehat{I_x}) , & $op=OR$  \label{eq:esti_sx_OR}\\
max(MinOver(L), \widehat{I_x})& 
$op=AND$ \label{eq:esti_sx_AND}\\
max(|S(x, L_{\leftarrow})| - MaxOver(L), \widehat{I_x})  & 
$op=AND~NOT$ \label{eq:esti_sx_ANDNOT}
\end{numcases}

where $MaxOver(L)$  returns the maximum possible overlap between the largest possible extensions of the formula's terms, $MinOver(L)$ returns the  minimum possible overlap between the smallest possible extensions of the formula's terms, and 
    $\widehat{S(x, L_{\leftarrow} \cup L_{\rightarrow}}) = |S(x, L_{\leftarrow})|+ |S(x, L_{\rightarrow})| - MaxOver(L))$.
In this case, Equation (\ref{eq:esti_sx_OR}) corresponds to assuming fully overlapping masks between the labels connected by the operator. Conversely,  Equation (\ref{eq:esti_sx_AND}) and Equation (\ref{eq:esti_sx_ANDNOT}) assume disjoint masks.

\subsection{Proof of Admissibility of MMESH}
\label{appx:proof}
This section proves that \textbf{MMESH is an admissible heuristic} (i.e., it never underestimates the true IoU score).
As previously explained, in order to be admissible, $\widehat{I_x}$ and $\widehat{S(x,L)}$ must satisfy the following equations: 
\begin{numcases} {}
|\widehat{I_x}| \ge |I_x|  \label{eq:i_constraint} \\
0 \le |\widehat{S(x,L)}| - |\widehat{I_x}| \le |S(x,L)| - I
\label{eq:label_constraint}
\end{numcases}
$\forall x \in \mathfrak{D}$.

We prove the two equations by proving they hold in the two extreme cases: fully overlapping between the labels connected by the operators and disjoint masks. As a consequence, they will also hold for partially overlapping masks.

\subsubsection{Proof of Equation (\ref{eq:i_constraint})}
\paragraph{Disjoint Masks.}
In this case, the real intersection is given by the following quantities:
\begin{numcases} {I_x =}
|IMS_{[\tau_1,\tau_2]}(x, L_{\leftarrow})|+|IMS_{[\tau_1,\tau_2]}(x, L_{\rightarrow})| &  $op = OR$ \label{eq:I2_or}\\
0 &  $op = AND$  \label{eq:I2_and}\\
|IMS_{[\tau_1,\tau_2]}(x, L_{\leftarrow})| &  $op = AND~NOT$  
\label{eq:I2_andnot}
\end{numcases}



We can prove Equation (\ref{eq:i_constraint}) by observing that: 
\begin{itemize}
    \item eq. (\ref{eq:I2_or}) $\le$ eq. (\ref{eq:esti_ix_OR})  because the intersections $IMS_{[\tau_1,\tau_2]}(x, \cdot)$ are disjoint subsets of the matrix $M_{[\tau_1,\tau_2]}(x)$ and, thus, eq. (\ref{eq:I2_or}) = eq. (\ref{eq:esti_ix_OR});
    \item eq. (\ref{eq:I2_and}) $\le$ eq. (\ref{eq:esti_ix_AND}) due to the non-negativity property of the cardinality;
    \item eq. (\ref{eq:I2_andnot}) $\le$ eq. (\ref{eq:esti_ix_AND_NOT}) holds because the presence of the minimum operator and the fact that the mask $M_{[\tau_1,\tau_2]}(x) - IMS_{[\tau_1,\tau_2]}(x, L_{\rightarrow})$ contains $IMS_{[\tau_1,\tau_2]}(x, L_{\leftarrow})$ since $L_{\leftarrow}$ and $L_{\rightarrow}$ are associated with disjoint masks.
\end{itemize} 
\paragraph{Fully Overlapping Masks.}
In this case, the real intersection is given by the following quantities:


\begin{numcases} {I_x =}
max(|IMS_{[\tau_1,\tau_2]}(x, L_{\leftarrow})|, |IMS_{[\tau_1,\tau_2]}(x, L_{\rightarrow})|) & $op = OR$
 \label{eq:I3_or}\\
min(|IMS_{[\tau_1,\tau_2]}(x, L_{\leftarrow})|, |IMS_{[\tau_1,\tau_2]}(x, L_{\rightarrow})|) & 
$op=AND$ \label{eq:I3_and}\\
min(|IMS_{[\tau_1,\tau_2]}(x, L_{\leftarrow})|, |M_{[\tau_1,\tau_2]}(x)| - |IMS_{[\tau_1,\tau_2]}(x, L_{\rightarrow})|) & 
$op=AND~NOT$ \label{eq:I3_andnot}
\end{numcases}

\subsubsection{Proof of Equation (\ref{eq:label_constraint})}
To prove Equation (\ref{eq:label_constraint}), it is sufficient to prove that the following equation holds for both disjoint and overlapping masks:
\begin{equation}
    \widehat{S(x,L)} \le S(x,L) 
    \label{eq:first_proof}
\end{equation}
Indeed, once the previous equation is proved, then also the constraint $\widehat{S(x, L)} - \widehat{I_x} \le S(x, L) - I_x$ holds because $\widehat{I_x} \ge I_x$ as shown before. Moreover, $\widehat{S(x,L)} - \widehat{I_x} \ge 0$ is already satisfied by $\widehat{S(x,L)}$ because all the cases include the max operator and $\widehat{I_x}$ as a term.

We prove the equation by proving it holds in the two extreme cases: fully overlapping between the labels connected by the operators and disjoint masks. As a consequence, it will also hold for partially overlapping masks.

\paragraph{Disjoint Masks.}
In this case, the real joint label's mask is given by:
\begin{numcases} {S(x,L)=}
|S(x,L_{\leftarrow})| + |S(x,L_{\rightarrow})| & $op=OR$  \label{eq:seg1_or}\\
0& 
$op=AND$\label{eq:seg1_and}\\
|S(x, L_{\leftarrow})|  & 
$op=AND~NOT\label{eq:seg1_andnot}$
\end{numcases}


We can prove Equation (\ref{eq:first_proof}) by observing that:
\begin{itemize}
    \item eq. (\ref{eq:esti_sx_OR}) $\le$ eq. (\ref{eq:seg1_or}) because the maximum between two cardinalities is lower than their sum and $MaxOver(L) \ge 0$;
    \item eq. (\ref{eq:esti_sx_AND}) $\le$ eq. (\ref{eq:seg1_and}) because the masks are disjoint and thus $MinOver(L) = 0$ and $\widehat{S(x,L)} = S(x,L)$;
    \item eq. (\ref{eq:esti_sx_ANDNOT}) $\le$ eq. (\ref{eq:seg1_andnot}) because $|\widehat{I_x}| \le |S(x, L_{\leftarrow})|$ since the intersection is a subset of the whole set, and $MaxOver(L)$ returns a positive number, implying that $ |S(x, L_{\leftarrow})| - MaxOver(L) \le |S(x, L_{\leftarrow})|$.
\end{itemize}

\paragraph{Fully Overlapping Masks.}
In this case, the real joint label's mask is given by:
\begin{numcases} {S(x,L)=}
max(|S(x,L_{\leftarrow})|,|S(x,L_{\rightarrow})|) & $op=OR$  \label{eq:seg2_or}\\
min(|S(x,L_{\leftarrow})|,|S(x,L_{\rightarrow})|)& 
$op=AND$\label{eq:seg2_and}\\
max(|S(x,L_{\leftarrow})|-|S(x,L_{\rightarrow})|, 0)  & 
$op=AND~NOT\label{eq:seg2_andnot}$
\end{numcases}

We can prove Equation (\ref{eq:first_proof}) by observing that:
\begin{itemize}
    \item eq. (\ref{eq:esti_sx_OR}) $\le$ eq. (\ref{eq:seg2_or}) because $\widehat{S(x, L_{\leftarrow} \cup L_{\rightarrow})}$ cannot be larger than individual masks due to the overlap;
    \item eq. (\ref{eq:esti_sx_AND}) $\le$ eq. (\ref{eq:seg2_and}) because, in the case of fully overlapping masks, $MinOver(L)$ returns a subset of the smallest label's mask and, thus, $MinOver(L) \le min(|S(x, L_{\leftarrow})|,|S(x, L_{\rightarrow})|)$;
    \item eq. (\ref{eq:esti_sx_ANDNOT}) $\le$ eq. (\ref{eq:seg2_andnot}) due to the non-negativity of the cardinality.
\end{itemize}

Therefore, since the heuristic satisfies Equation (\ref{eq:i_constraint}) and Equation (\ref{eq:label_constraint}), the heuristic is admissible and returns the optimal formula inside the beam.
\section{Metrics}
\label{sec:neurons_metrics}
This section introduces our \textbf{set of metrics to evaluate the qualities of logical explanations associated with neurons}. While the IoU metric has been only reframed to be compatible with threshold values, all the other metrics are proposed by this thesis.

\paragraph{Intersection Over Union} measures the alignment between labels' annotations and activation map in the given activation range $(\tau_1,\tau_2)$.  A higher IoU means the algorithm can better capture the pre-existent alignment~\cite{Mu2020}.
\begin{equation}
    IoU(L, \tau_1,\tau_2, \mathfrak{D} ) = \frac{\sum_{x \in \mathfrak{D}}|M_{[\tau_1,\tau_2]}(x) \cap S(x,L)|}{\sum_{x \in \mathfrak{D}}|M_{[\tau_1,\tau_2]}(x) \cup S(x,L)|}
\end{equation}

\paragraph{Detection Accuracy} measures the bijection from the overlapping set between label annotations and neuron activation inside the activation range to the set of label annotations ~\cite{Makinwa2022}. In other words, it measures the quantity of annotations recognized at the given activation range. Higher is better.
\begin{equation}
    DetAcc(L, \tau_1,\tau_2, \mathfrak{D} ) = \frac{\sum_{x \in \mathfrak{D}}| M_{[\tau_1,\tau_2]}(x) \cap S(x,L)|}{\sum_{x \in \mathfrak{D}}|(S(x,L)|}
\end{equation}

\paragraph{Samples Coverage} measures the bijection from the set of samples where there is an overlap between label annotations and neuron activation inside the activation range to the set of samples associated with label annotations. In other words, it is the percentage of candidate samples recognized at the given activation range. Higher is better.
\begin{equation}
    SampleCov(L, \tau_1,\tau_2, \mathfrak{D} ) = \frac{|\{x \in \mathfrak{D}: |M_{[\tau_1,\tau_2]}(x) \cap S(x,L)|>0\}|}{|\{x \in \mathfrak{D}: |(S(x,L)|>0\}|}
\end{equation}

\paragraph{Activation Coverage} measures the bijection from the overlapping set between label annotations and neuron activation inside the activation range to the set of neuron activations within the range. In other words, measures the label $L$ exclusivity for the activation range $(\tau_1,\tau_2)$. Higher is better.

\begin{equation}
    ActCov(L, \tau_1,\tau_2, \mathfrak{D} ) = \frac{\sum_{x \in \mathfrak{D}}| M_{[\tau_1,\tau_2]}(x) \cap S(x,L)|}{\sum_{x \in \mathfrak{D}}|M_{[\tau_1,\tau_2]}(x)|}
\end{equation}

\paragraph{Explanation Coverage} measures the bijection from the set of samples where there is an overlap between label annotations and neuron activation inside the activation range to the set of samples associated with label annotations. In other words, it measures the correlation's strength between the label's presence and the neuron activation. Higher is better.

 
\begin{equation}
    ExplCov(L, \tau_1,\tau_2, \mathfrak{D} ) = 
    \frac{|\{x \in \mathfrak{D}: |M_{[\tau_1,\tau_2]}(x) \cap S(x,L)|>0\}|}{|\{x \in \mathfrak{D}: |M_{[\tau_1,\tau_2]}(x)|>0\}|}
\end{equation}

\paragraph{Label Masking} measures the correlation between the explanation label $L$ and the activation range $[\tau_1,\tau_2]$. Given a function $\theta{(x, L)}$ that masks all the features in $x$ but the ones corresponding to the label $L$, the score corresponds to the cosine similarity between activation of neuron $k$ obtained by feeding the full and the masked input. Higher is better.
\begin{equation}
    LabMask(L, \tau_1,\tau_2, \mathfrak{D} ) = \frac{\sum_{x \in \mathfrak{D}} CosSim(M^k_{[\tau_1,\tau_2]}(x)A^k{(\theta{(x,L)})}, M^k_{[\tau_1,\tau_2]}(x)A^k(x))}{|\{x \in \mathfrak{D}: |M_{[\tau_1,\tau_2]}(x)|>0\}|}
\end{equation}

\section{Experiments}
\label{sec:neurons_experiments}
\subsection{Motivation}
\label{appx:threshold}
The section begins by showcasing the limitations of the current state-of-the-art algorithms, which prompted the development of our algorithm. Existing methods, such as \gls{NetDissect} and \gls{CoEx}, focus on understanding a neuron's recognition capabilities by examining its highest activations, typically within the 0.005 percentile of activations. The objective function that they optimize is the same as \Cref{eq:objective_cce}, but they investigate only the range $T = {[\tau^{top},\infty]}$ and NetDissect limits the explanations to the one in $L \in \mathfrak{L}^1$.

\begin{figure}[t]
    \centering
     \includegraphics[scale=0.4]{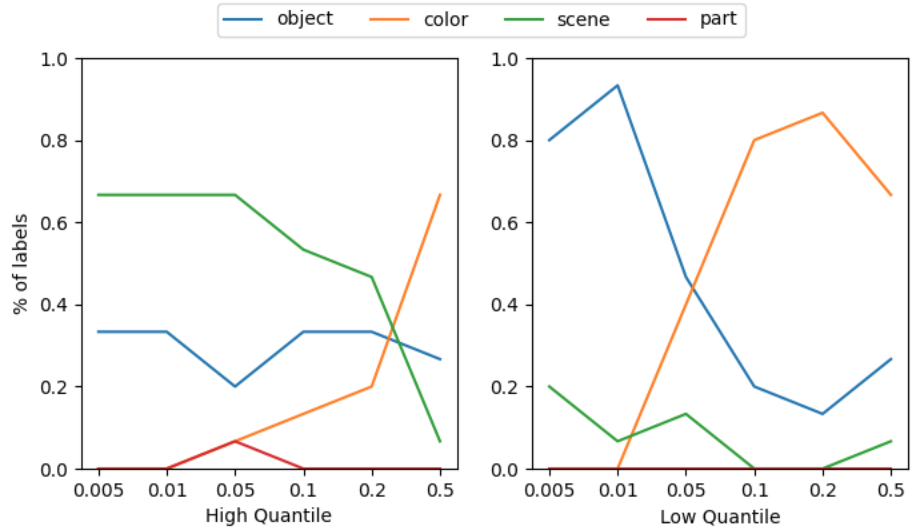}
    \caption{Percentage of labels' category associated with the labels returned by CoEx over different activation ranges. Left: ranges from quantile to infinite. Right: range from 0 to quantile. Results are computed over 100 randomly extracted units.}
    \label{fig:threshold_impact_appx}
\end{figure}

\Cref{fig:threshold_impact_appx} presents the results obtained by varying the percentile to different values. Specifically, we explore values ranging from the highest to the lowest [0.005, 0.01, 0.02, 0.05, 0.1, 0.2, 0 .5] percentile of neuron activations from 100 randomly extracted units in a ResNet18 model. For each run, we collect the labels associated with each unit by the \gls{CoEx} algorithm. Then, we map each label to its category based on the Pascal dataset~\cite{Everingham2009} annotations and record the percentage of labels for each category. The results are shown in \Cref{fig:threshold_impact_appx}. 

It is evident that the \textbf{distribution of labels varies with changes in the percentile}. For instance, when considering the highest activations, a larger quantile rewards colors while penalizing scene labels. Conversely, when considering the lowest activations, a larger quantile penalizes object labels. Notably, the distribution differs when considering the highest or lowest activations for the same quantile value. These findings suggest that neurons may recognize different concepts at different \emph{activation levels}, indicating that solely considering exceptionally high activations provides a partial view. This limited perspective can lead to issues when comparing the interpretability of latent space, as they may overlook concepts to which the neurons are aligned (e.g., as in the previous chapter).

\subsection{Performance}
\begin{table}[b]
\centering
\caption{Avg. and standard deviation of steps required to reach the optimal solution within the beam.}
\label{tab:performance}
\begin{tabular}{lr}
    \toprule
     Algorithm & Steps \\
     \midrule
     NoHeuristics &   39656 {\scriptsize $\pm$ 12659} \\
         
     MMESH &  129  {\scriptsize $\pm$ 712}\\
     \bottomrule

\end{tabular}
\end{table}
We begin the analysis of the proposed heuristic by comparing the computation steps needed to find the most aligned label against those required by \gls{CoEx}, which does not use heuristics. In this scenario, we use ResNet18~\cite{He2016} as the model to explain and Ade20k~\cite{zhou2017scene, zhou2019semantic} as the concept dataset. The results (\Cref{tab:performance}) are computed over 100 randomly selected units and follow the settings of ~\citet{Mu2020}. The results show that \textbf{MMESH is faster than CoEx by two orders of magnitude}. This improvement corresponds to transitioning from a computation time of 60 minutes per unit (\gls{CoEx}) to less than two minutes (our algorithm)\footnote{Timing collected using a workstation powered by an NVIDIA GeForce RTX-3090 graphic card.}. We expect that this enhancement will be more substantial in more complex settings, supporting larger datasets in terms of samples and concepts. Thus, for a reasonably low number of clusters, our algorithm can compute composition explanations for clusters within the same time frame (or less) taken by \gls{CoEx} to compute compositional explanations for a single cluster.
\subsection{Explanations}
\begin{table}[!b]
    \centering
    \caption{Avg. and Std Dev. across metrics reached by NetDissect, CoEx, and Clustered Compositional Explanations using Ade20K as a concept dataset.}
    
    \begin{tabular}{ccrrrrrr}
        \toprule
        & Cluster & IoU & ExplCov & SampleCov & ActCov &DetAcc&LabMask\\
         \midrule
                   & & & & ResNet   &  &&\\
         NetDissect & &
         0.05 \footnotesize{$\pm$ 0.03}&
         0.14 \footnotesize{$\pm$ 0.20}&
         0.58 \footnotesize{$\pm$ 0.27}&
         0.14 \footnotesize{$\pm$ 0.17}&
         0.15 \footnotesize{$\pm$ 0.15}&
         0.62\footnotesize{$\pm$ 0.26}\\
         CoEx & &
         0.08 \footnotesize{$\pm$ 0.03}&
         0.14 \footnotesize{$\pm$ 0.15}&
         0.53 \footnotesize{$\pm$ 0.25}&
         0.15 \footnotesize{$\pm$ 0.10}&
         0.18 \footnotesize{$\pm$ 0.10}&
         0.57\footnotesize{$\pm$ 0.22}\\
         Our Avg. & &
         0.15 \footnotesize{$\pm$ 0.07}&
         0.60 \footnotesize{$\pm$ 0.36}&
         0.69 \footnotesize{$\pm$ 0.24}&
         0.37 \footnotesize{$\pm$ 0.17}&
         0.20 \footnotesize{$\pm$ 0.09}&
         0.61\footnotesize{$\pm$ 0.16}\\
              Our &  1 & 
              0.27 \footnotesize{$\pm$ 0.02} &
              0.99 \footnotesize{$\pm$ 0.00}& 
              0.96 \footnotesize{$\pm$ 0.02}& 
              0.54 \footnotesize{$\pm$ 0.02}& 
              0.34 \footnotesize{$\pm$ 0.03}&
              0.64\footnotesize{$\pm$ 0.06}\\
             Our & 2 & 
             0.15 \footnotesize{$\pm$ 0.02} &
             0.91 \footnotesize{$\pm$ 0.16}&
             0.80 \footnotesize{$\pm$ 0.06}&
             0.49 \footnotesize{$\pm$ 0.09}&
             0.18 \footnotesize{$\pm$ 0.03}&
             0.68\footnotesize{$\pm$ 0.07}\\
            Our & 3 &
            0.12 \footnotesize{$\pm$ 0.03} &
            0.56 \footnotesize{$\pm$ 0.27}&
            0.64 \footnotesize{$\pm$ 0.15}&
            0.33 \footnotesize{$\pm$ 0.10}&
            0.16 \footnotesize{$\pm$ 0.05}&
            0.62\footnotesize{$\pm$ 0.14}\\
            Our & 4 & 0.10 \footnotesize{$\pm$ 0.04} &
            0.35 \footnotesize{$\pm$ 0.19}&
            0.52 \footnotesize{$\pm$ 0.22}&
            0.25 \footnotesize{$\pm$ 0.12}&
            0.15 \footnotesize{$\pm$ 0.07}&
            0.53\footnotesize{$\pm$ 0.19}\\
          Our & 5 & 0.09 \footnotesize{$\pm$ 0.05} &
          0.21 \footnotesize{$\pm$ 0.18}& 
          0.53 \footnotesize{$\pm$ 0.26}& 
          0.22 \footnotesize{$\pm$ 0.15}& 
          0.18 \footnotesize{$\pm$ 0.09}&
          0.58\footnotesize{$\pm$ 0.23}\\
          \midrule
          &  &  & & AlexNet  &&\\
         NetDissect & &
         0.03 \footnotesize{$\pm$ 0.02}&
         0.25 \footnotesize{$\pm$ 0.34}&
         0.32 \footnotesize{$\pm$ 0.21}&
         0.15 \footnotesize{$\pm$ 0.19}&
         0.13 \footnotesize{$\pm$ 0.10}&
         0.27 \footnotesize{$\pm$ 0.21}\\
         CoEx &  &
         0.05 \footnotesize{$\pm$ 0.02}&
         0.24 \footnotesize{$\pm$ 0.27}&
         0.22 \footnotesize{$\pm$ 0.16}&
         0.14 \footnotesize{$\pm$ 0.11}&
         0.11 \footnotesize{$\pm$ 0.07}&
         0.23 \footnotesize{$\pm$ 0.18}\\
                   Our Avg. & &
         0.12 \footnotesize{$\pm$ 0.07}&
         0.65 \footnotesize{$\pm$ 0.34}&
         0.67 \footnotesize{$\pm$ 0.28}&
         0.33 \footnotesize{$\pm$ 0.18}&
         0.17 \footnotesize{$\pm$ 0.08}&
         0.48 \footnotesize{$\pm$ 0.17}\\
             Our  & 1 & 
              0.25 \footnotesize{$\pm$ 0.04} &
              0.99 \footnotesize{$\pm$ 0.00}& 
              0.91 \footnotesize{$\pm$ 0.09}& 
              0.57 \footnotesize{$\pm$ 0.03}& 
              0.30 \footnotesize{$\pm$ 0.06}&
              0.55 \footnotesize{$\pm$ 0.11}\\
             Our  &  2 & 
             0.12 \footnotesize{$\pm$ 0.03} &
             0.88 \footnotesize{$\pm$ 0.20}&
             0.76 \footnotesize{$\pm$ 0.16}&
             0.43 \footnotesize{$\pm$ 0.13}&
             0.15 \footnotesize{$\pm$ 0.03}&
             0.60 \footnotesize{$\pm$ 0.13}\\
             Our  & 3 &
            0.10 \footnotesize{$\pm$ 0.03} &
            0.63 \footnotesize{$\pm$ 0.25}&
            0.56 \footnotesize{$\pm$ 0.16}&
            0.29 \footnotesize{$\pm$ 0.09}&
            0.14 \footnotesize{$\pm$ 0.04}&
            0.54 \footnotesize{$\pm$ 0.13}\\
            Our  & 4 & 
            0.08 \footnotesize{$\pm$ 0.03} &
            0.47 \footnotesize{$\pm$ 0.25}&
            0.37 \footnotesize{$\pm$ 0.15}&
            0.23 \footnotesize{$\pm$ 0.10}&
            0.12 \footnotesize{$\pm$ 0.03}&
            0.42 \footnotesize{$\pm$ 0.13}\\
          Our  & 5 & 
          0.06 \footnotesize{$\pm$ 0.02} &
          0.27 \footnotesize{$\pm$ 0.25}& 
          0.23 \footnotesize{$\pm$ 0.10}& 
          0.14 \footnotesize{$\pm$ 0.10}& 
          0.13 \footnotesize{$\pm$ 0.06}&
          0.27 \footnotesize{$\pm$ 0.14}\\

                   \midrule
          & & &  & DenseNet &  &&\\
                  NetDissect & &
         0.05 \footnotesize{$\pm$ 0.03}&
         0.05 \footnotesize{$\pm$ 0.04}&
         0.52 \footnotesize{$\pm$ 0.30}&
         0.07 \footnotesize{$\pm$ 0.05}&
         0.15 \footnotesize{$\pm$ 0.09}&
         0.40 \footnotesize{$\pm$ 0.26}\\
         CoEx & &
         0.06 \footnotesize{$\pm$ 0.03}&
         0.09 \footnotesize{$\pm$ 0.10}&
         0.51 \footnotesize{$\pm$ 0.27}&
         0.11 \footnotesize{$\pm$ 0.06}&
         0.13 \footnotesize{$\pm$ 0.07}&
         0.42 \footnotesize{$\pm$ 0.25}\\
         Our Avg. &  &
         0.20 \footnotesize{$\pm$ 0.08}&
         0.77 \footnotesize{$\pm$ 0.30}&
         0.83 \footnotesize{$\pm$ 0.20}&
         0.43 \footnotesize{$\pm$ 0.15}&
         0.27 \footnotesize{$\pm$ 0.10}&
         0.52 \footnotesize{$\pm$ 0.27}\\
              Our & 1 & 
              0.14 \footnotesize{$\pm$ 0.05} &
              0.63 \footnotesize{$\pm$ 0.27}& 
              0.74 \footnotesize{$\pm$ 0.12}& 
              0.35 \footnotesize{$\pm$ 0.13}& 
              0.20 \footnotesize{$\pm$ 0.07}&
              0.61 \footnotesize{$\pm$ 0.27}\\
             Our & 2 & 
             0.26 \footnotesize{$\pm$ 0.06} &
             0.96 \footnotesize{$\pm$ 0.09}&
             0.96 \footnotesize{$\pm$ 0.07}&
             0.53 \footnotesize{$\pm$ 0.06}&
             0.35 \footnotesize{$\pm$ 0.08}&
             0.48 \footnotesize{$\pm$ 0.25}\\
            Our & 3 &
            0.26 \footnotesize{$\pm$ 0.03} &
            0.99 \footnotesize{$\pm$ 0.01}&
            0.98 \footnotesize{$\pm$ 0.04}&
            0.55 \footnotesize{$\pm$ 0.02}&
            0.33 \footnotesize{$\pm$ 0.05}&
            0.38 \footnotesize{$\pm$ 0.20}\\
            Our & 4 & 
            0.20 \footnotesize{$\pm$ 0.07} &
            0.84 \footnotesize{$\pm$ 0.23}&
            0.85 \footnotesize{$\pm$ 0.21}&
            0.44 \footnotesize{$\pm$ 0.13}&
            0.27 \footnotesize{$\pm$ 0.10}&
            0.57 \footnotesize{$\pm$ 0.27}\\
          Our & 5 & 
          0.11 \footnotesize{$\pm$ 0.05} &
          0.42 \footnotesize{$\pm$ 0.28}& 
          0.62 \footnotesize{$\pm$ 0.21}& 
          0.28 \footnotesize{$\pm$ 0.14}& 
          0.18 \footnotesize{$\pm$ 0.08}&
          0.56 \footnotesize{$\pm$ 0.30}\\
         \bottomrule
    \end{tabular}
    \label{tab:expla_ade20k}
\end{table}
\begin{table}[!b]
    \centering
    \caption{Avg. and Std Dev. across metrics reached by NetDissect, CoEx, and Clustered Compositional Explanations on ResNet. Concepts are retrieved from the Pascal dataset.}
    
    \begin{tabular}{ccrrrrrr}
        \toprule
        & Cluster & IoU & ExplCov & SampleCov & ActCov &DetAcc&LabMask\\
         \midrule
         NetDissect & &
         0.05 \footnotesize{$\pm$ 0.05}&
         0.32 \footnotesize{$\pm$ 0.24}&
         0.25 \footnotesize{$\pm$ 0.17}&
         0.23 \footnotesize{$\pm$ 0.20}&
         0.08 \footnotesize{$\pm$ 0.07}&
         0.28 \footnotesize{$\pm$ 0.17}\\
         CoEx &  &
         0.09 \footnotesize{$\pm$ 0.05}&
         0.31 \footnotesize{$\pm$ 0.22}&
         0.22 \footnotesize{$\pm$ 0.15}&
         0.19 \footnotesize{$\pm$ 0.17}&
         0.09 \footnotesize{$\pm$ 0.07}&
         0.26 \footnotesize{$\pm$ 0.15}\\
         Our Avg. & &
         0.15 \footnotesize{$\pm$ 0.05}&
         0.68 \footnotesize{$\pm$ 0.31}&
         0.64 \footnotesize{$\pm$ 0.26}&
         0.39 \footnotesize{$\pm$ 0.18}&
         0.20 \footnotesize{$\pm$ 0.10}&
         0.55 \footnotesize{$\pm$ 0.17}\\
              Our & 1 & 
              0.28 \footnotesize{$\pm$ 0.02} &
              0.99 \footnotesize{$\pm$ 0.00}& 
              0.97 \footnotesize{$\pm$ 0.01}& 
              0.56 \footnotesize{$\pm$ 0.02}& 
              0.35 \footnotesize{$\pm$ 0.02}&
              0.63 \footnotesize{$\pm$ 0.08}\\
             Our & 2 & 
             0.15 \footnotesize{$\pm$ 0.01} &
             0.99 \footnotesize{$\pm$ 0.01}&
             0.84 \footnotesize{$\pm$ 0.03}&
             0.57 \footnotesize{$\pm$ 0.04}&
             0.17 \footnotesize{$\pm$ 0.02}&
             0.68 \footnotesize{$\pm$ 0.08}\\
            Our & 3 &
            0.11 \footnotesize{$\pm$ 0.02} &
            0.61 \footnotesize{$\pm$ 0.27}&
            0.60 \footnotesize{$\pm$ 0.09}&
            0.28 \footnotesize{$\pm$ 0.10}&
            0.17 \footnotesize{$\pm$ 0.05}&
            0.59 \footnotesize{$\pm$ 0.10}\\
            Our & 4 & 
            0.10 \footnotesize{$\pm$ 0.06} &
            0.42 \footnotesize{$\pm$ 0.15}&
            0.49 \footnotesize{$\pm$ 0.16}&
            0.27 \footnotesize{$\pm$ 0.09}&
            0.18 \footnotesize{$\pm$ 0.09}&
            0.53 \footnotesize{$\pm$ 0.15}\\
          Our & 5 & 
          0.09 \footnotesize{$\pm$ 0.06} &
          0.40 \footnotesize{$\pm$ 0.21}& 
          0.30 \footnotesize{$\pm$ 0.16}& 
          0.28 \footnotesize{$\pm$ 0.20}& 
          0.12 \footnotesize{$\pm$ 0.08}&
          0.34 \footnotesize{$\pm$ 0.15}\\
         \bottomrule
    \end{tabular}
    \label{tab:pascal}
\end{table}
\begin{table}[!b]
    \centering
    \caption{Avg. and Std Dev. of the desired qualities over the labels returned by NetDissect, CoEx, and Clustered Compositional Explanations (our) applied to ResNet pre-trained on ImageNet. Results are computed for 50 randomly extracted units.}
    \begin{tabular}{ccrrrrrr}
        \toprule
        & Cluster & IoU & ExplCov & SampleCov & ActCov &DetAcc&LabMask\\
         \midrule
          & & &  & ResNet18 &  &&\\
         NetDissect & &
           0.04 \footnotesize{$\pm$ 0.02}&
           0.13 \footnotesize{$\pm$ 0.17}&
           0.36 \footnotesize{$\pm$ 0.27}&
            0.10 \footnotesize{$\pm$ 0.12}&
            0.12 \footnotesize{$\pm$ 0.11}&
              0.37\footnotesize{$\pm$ 0.24}\\
         CoEx & &
           0.05 \footnotesize{$\pm$ 0.03}&
           0.13 \footnotesize{$\pm$ 0.16}&
           0.28 \footnotesize{$\pm$ 0.20}&
            0.12 \footnotesize{$\pm$ 0.08}&
            0.12 \footnotesize{$\pm$ 0.08}&
              0.32\footnotesize{$\pm$ 0.19}\\
         Our Avg. & &
            0.13 \footnotesize{$\pm$ 0.08}&
           0.65 \footnotesize{$\pm$ 0.37}&
           0.65 \footnotesize{$\pm$ 0.29}&
           0.36 \footnotesize{$\pm$ 0.19}&
            0.18 \footnotesize{$\pm$ 0.11}&
              0.56\footnotesize{$\pm$ 0.18}\\
              Our & 1 & 
            0.28 \footnotesize{$\pm$ 0.02}&
           0.99 \footnotesize{$\pm$ 0.01}&
           0.98 \footnotesize{$\pm$ 0.02}&
           0.54 \footnotesize{$\pm$ 0.02}&
            0.36 \footnotesize{$\pm$ 0.03}&
              0.64\footnotesize{$\pm$ 0.06}\\
             Our & 2 & 
        0.15 \footnotesize{$\pm$ 0.03}&
           0.95 \footnotesize{$\pm$ 0.12}&
           0.89 \footnotesize{$\pm$ 0.06}&
           0.53 \footnotesize{$\pm$ 0.07}&
            0.18 \footnotesize{$\pm$ 0.04}&
              0.73\footnotesize{$\pm$ 0.08}\\
            Our & 3 &
            0.13 \footnotesize{$\pm$ 0.05}&
           0.79 \footnotesize{$\pm$ 0.23}&
           0.66 \footnotesize{$\pm$ 0.11}&
           0.39 \footnotesize{$\pm$ 0.12}&
            0.10 \footnotesize{$\pm$ 0.03}&
              0.63\footnotesize{$\pm$ 0.11}\\
            Our & 4 & 
            0.07 \footnotesize{$\pm$ 0.03}&
           0.32 \footnotesize{$\pm$ 0.21}&
           0.42 \footnotesize{$\pm$ 0.18}&
           0.18 \footnotesize{$\pm$ 0.10}&
            0.12 \footnotesize{$\pm$ 0.06}&
              0.48\footnotesize{$\pm$ 0.15}\\
          Our & 5 & 
          0.06 \footnotesize{$\pm$ 0.04}&
           0.19 \footnotesize{$\pm$ 0.16}&
           0.29 \footnotesize{$\pm$ 0.19}&
           0.15 \footnotesize{$\pm$ 0.11}&
            0.12 \footnotesize{$\pm$ 0.08}&
             0.34\footnotesize{$\pm$ 0.18}\\
       \midrule
        & & &  & VGG-16  &  &&\\
         NetDissect &  &
          0.03 \footnotesize{$\pm$ 0.02}&
           0.13 \footnotesize{$\pm$ 0.18}&
           0.35 \footnotesize{$\pm$ 0.26}&
            0.11 \footnotesize{$\pm$ 0.11}&
            0.11 \footnotesize{$\pm$ 0.11}&
             0.39\footnotesize{$\pm$ 0.25}\\
         CoEx & &
           0.04 \footnotesize{$\pm$ 0.03}&
           0.10 \footnotesize{$\pm$ 0.10}&
           0.28 \footnotesize{$\pm$ 0.19}&
            0.10 \footnotesize{$\pm$ 0.07}&
            0.10 \footnotesize{$\pm$ 0.09}&
            0.33\footnotesize{$\pm$ 0.18}\\
         Our Avg. & &
            0.08 \footnotesize{$\pm$ 0.06}&
           0.38 \footnotesize{$\pm$ 0.34}&
           0.45 \footnotesize{$\pm$ 0.26}&
           0.22 \footnotesize{$\pm$ 0.17}&
            0.11 \footnotesize{$\pm$ 0.07}&
            0.36\footnotesize{$\pm$ 0.16}\\
              Our & 1 & 
          0.16 \footnotesize{$\pm$ 0.06}&
           0.92 \footnotesize{$\pm$ 0.16}&
           0.78 \footnotesize{$\pm$ 0.16}&
            0.48 \footnotesize{$\pm$ 0.13}&
            0.20 \footnotesize{$\pm$ 0.07}&
            0.38\footnotesize{$\pm$ 0.13}\\
             Our & 2 & 
             0.07 \footnotesize{$\pm$ 0.03}&
           0.43 \footnotesize{$\pm$ 0.24}&
           0.56 \footnotesize{$\pm$ 0.20}&
           0.20 \footnotesize{$\pm$ 0.10}&
            0.10 \footnotesize{$\pm$ 0.05}&
            0.43\footnotesize{$\pm$ 0.17}\\
            Our & 3 &
           0.06 \footnotesize{$\pm$ 0.03}&
           0.29 \footnotesize{$\pm$ 0.21}&
           0.40 \footnotesize{$\pm$ 0.18}&
           0.18 \footnotesize{$\pm$ 0.11}&
            0.08 \footnotesize{$\pm$ 0.04}&
            0.39\footnotesize{$\pm$ 0.15}\\
            Our & 4 & 
           0.08 \footnotesize{$\pm$ 0.04}&
           0.16 \footnotesize{$\pm$ 0.15}&
           0.28 \footnotesize{$\pm$ 0.16}&
           0.13 \footnotesize{$\pm$ 0.10}&
            0.08 \footnotesize{$\pm$ 0.04}&
            0.33\footnotesize{$\pm$ 0.15}\\
          Our & 5 & 
           0.05 \footnotesize{$\pm$ 0.04}&
           0.10 \footnotesize{$\pm$ 0.12}&
           0.24 \footnotesize{$\pm$ 0.18}&
            0.12 \footnotesize{$\pm$ 0.10}&
            0.08 \footnotesize{$\pm$ 0.07}&
            0.28\footnotesize{$\pm$ 0.17}\\
         \bottomrule
    \end{tabular}
    \label{tab:vggimagenet}
\end{table}

This section evaluates the explanations retrieved by the Clustered Compositional Explanation and analyzes phenomena related to activations emerging after the application of our algorithm to the latent space of popular \glspl{DNN}.  Following the literature on compositional explanations~\cite{Massidda2023, Makinwa2022, Mu2020}, we set the maximum explanation's length to three. The first experiment compares our algorithm against \gls{NetDissect} and \gls{CoEx} using the metrics described in \Cref{sec:neurons_metrics}. This experiment test all the competitors in ResNet18~\cite{He2016},  DenseNet~\cite{Huang2017} and AlexNet~\cite{Krizhevsky2012} pre-trained on the Place365 dataset~\cite{Zhou2018} and  ResNet18~\cite{He2016} and VGG16~\cite{Simonyan2014} pre-pretrained on ImageNet. As a concept dataset, we use combinations of the models on both Ade20k~\cite{zhou2017scene, zhou2019semantic} and the Pascal dataset~\cite{Everingham2009}. In all experiments, we focus on explaining the last layers of the networks following the setup of \citet{Mu2020}. However, the results can be easily extended to any layer of a \gls{DNN}. Finally, we employ K-Means as a clustering algorithm and fix the number of clusters to five.

\Cref{tab:expla_ade20k} and \Cref{tab:pascal} report the scores for individual clusters and their averages across different models and datasets, where the activation range orders clusters (i.e., Cluster 1 corresponds to the range including the lowest activations). 
Inspecting the mere scores, \gls{NetDissect} exhibit similar performance \gls{CoEx}.  \gls{CoEx} demonstrates a slight superiority in IoU, while \gls{NetDissect} is superior in Sample Coverage. Overall,  the \textbf{Clustered Compositional Explanations algorithm assigns labels of better quality} across all considered clusters, models, and datasets. Therefore, there is no rationale related to explanation quality to justify disregarding the full spectrum of neuron activations when explaining neuron behavior.

\begin{table}[b]
    \centering
    \caption{Percentage of predictions changed when the activations of a given cluster are masked.}
    \begin{tabular}{cr}
        \toprule
        & Prediction Change \%\\
         \midrule
         CoEx  & 
           12.9  \\
         \midrule
              Cluster 1  & 
            9.7 \\
             Cluster 2  & 
             12.90 \\
            Cluster 3  & 
             12.29\\
            Cluster 4  & 
             11.62\\
          Cluster 5 & 
             13.79\\
         \bottomrule
    \end{tabular}

    \label{tab:importance}
\end{table}
However, another reason to avoid considering the full spectrum could be that the network exclusively uses the highest activation to process specific concepts. To verify if this is the case, we masked the activations within a given range of the last layer of the feature extractor. Intuitively, if the network relies solely on the highest activations to recognize a concept, masking these activations should produce a more significant shift in predictions than masking lower activations. \Cref{tab:importance} demonstrates that this is not the case and changes in predictions are similar between using the narrow range of  \gls{CoEx} and the ranges used by the clusters of our algorithm. These results align with the essence of \glspl{DNN}, designed to \textbf{leverage all activations of a neuron}, regardless of their magnitude.

\begin{table}[!b]
    \centering
        \caption{Percentage of unspecialized and weakly specialized activation ranges in ResNet18 (ReLU Yes) and DenseNet161(ReLU No).}
    \begin{tabular}{ccccccc}
        \toprule
         & ReLU & Cluster 1 & Cluster 2 & Cluster 3 & Cluster 4 & Cluster 5 \\
         \midrule
        Unspecialized & Yes & 0.93 & 0.37 &  0.01 & 0.00 & 0.00 \\
         & No & 0.05 & 0.60&  0.95 & 0.30& 0.05\\
        Weakly Specialized & Yes &  0.00 & 0.03&  0.01 & 0.01& 0.00\\
         & No & 0.05& 0.12 & 0.02 & 0.17 & 0.02\\
         \bottomrule
    \end{tabular}

    \label{table:specialization}
\end{table}
Interestingly, \Cref{tab:importance} indicates that the labels associated with the lowest cluster have the least impact on the decision process. In \Cref{tab:expla_ade20k}, this cluster is associated with Explanation and Sample coverage close to one, suggesting that the neurons fire in almost all samples containing the cluster's label and rarely if the label is not included. By inspecting the labels associated with these clusters, the labels often refer to combinations of colors and background concepts (e.g., Sky OR Blue), with low variability across neurons, as most neurons are associated with the same labels. The same behavior is observed in the third cluster in DenseNet and, to a lesser extent, in the adjacent clusters. We apply the Clustered Compositional Explanation algorithm to random initialized networks to understand the convergence towards the same labels. Surprisingly, all the clusters in these networks are associated with labels similar to those in the lowest clusters of the trained network. Therefore, these labels represent unspecialized activations, which are activations that cannot be linked to specific behaviors and are less important than the others. We term these labels as \emph{default labels}.

Further investigation into activation-related phenomena reveals insights into unspecialized and weakly specialized activations. By analyzing labels across clusters, we compute the percentage of default labels. We define \emph{unspecialized activations} as those for which the whole compositional explanation is a default label. Conversely, activations where only the last part of the label converges on a different concept are defined as \emph{weakly specialized activations}. \Cref{table:specialization} shows that \textbf{Cluster 1 in ReLU networks and  Cluster 3 in non-ReLU networks include unspecialized activations}. These clusters correspond to activations closer to 0. Indeed, ReLU layers return non-negative activations; thus, the lowest cluster is the one that includes activations close to zero. Conversely, non-ReLU can return negative activations, which are captured by the lowest clusters, and activations close to zero are captured by middle clusters. The percentage of unspecialized activations decreases when moving away from zero. \textbf{Weakly specialized units are rare}, especially in ReLU networks. The increment in percentage in adjacent clusters may be due to clustering algorithm artifacts (i.e., spurious outliers in a cluster). This phenomenon can be further investigated in future work by using a cleaning process on the clusters or developing clustering algorithms tailored to the task.

\begin{figure}
    \centering
    \includegraphics[scale=0.4]{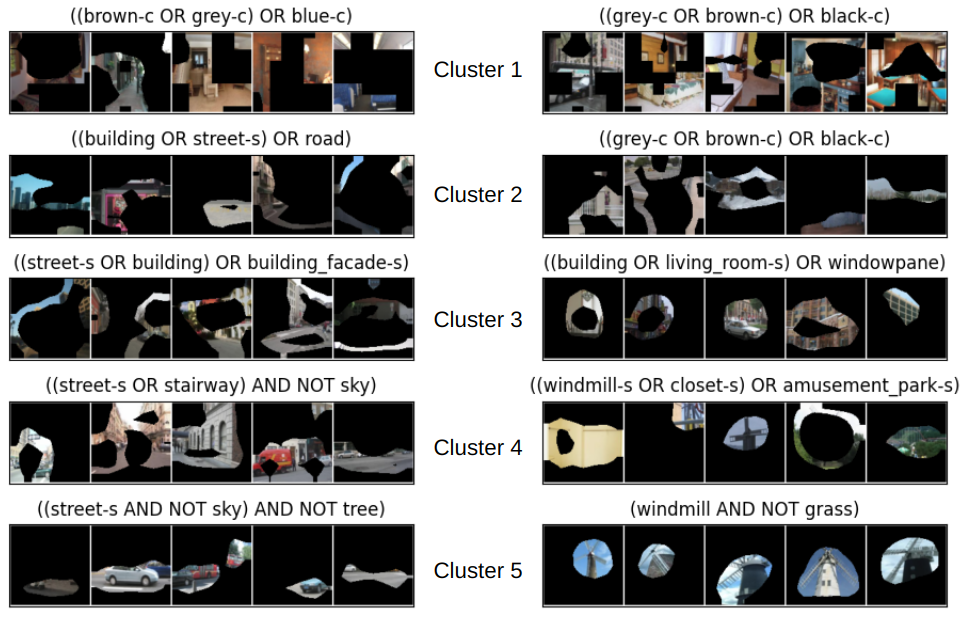}
    \caption{Examples of specialization (left) and polysemy (right). Neuron \#368 (right) recognizes unrelated concepts among different activation ranges (windmill, closet, amusement park).}
    \label{fig:examples}
\end{figure}

In ReLU layers, \textbf{we also observe progressive specialization of activations} with lower activations recognizing general concepts like scenes and buildings, middle activations combining general concepts with more complex ones, and the highest activations focusing on specific and smaller objects (\Cref{fig:examples}). This phenomenon also appears spatially aligned, with lower activations detecting background and side objects, while the highest ones focus on smaller foreground objects.
The specialization property highlighted by ~\citet{Mu2020} can be considered related to progressive specialization. \Cref{fig:examples} (left) shows an example of such specialization, where the neuron recognizes streets in more and more specific contexts.

Regarding individual clusters, we can analyze the performance of Cluster 5, which contains the highest activations. Therefore, this cluster also includes the activation range used by NetDissect and \gls{CoEx}. Comparing its scores against \gls{CoEx}, we observe that slightly increasing the range size does not lower the quality of explanations, as Cluster 5 outperforms \gls{CoEx} or reaches the same performance in all scores. 

Lastly, similarly to \citet{Mu2020}, we investigate the polysemy of activations, generalizing the polysemy to the clusters. A label is monosemic if all the concepts involved in the label refer to similar or related concepts. For example, the label ((chair AND table) OR kitchen) is monosemic since its atomic concepts are related. Conversely, a label is polysemic if it involves unrelated concepts. Therefore, we define a neuron as monosemic if all its clusters are associated with monosemic labels, excluding the default ones. Conversely, a neuron is polysemic if at least one of its clusters is associated with polysemic labels. In our manual inspection, \textbf{15\% of the neurons meet the criterion to be monosemic}. In these cases, neurons are referred to as \textbf{highly specialized}. The percentage is much lower than the one found by \citet{Mu2020}, who consider only the highest activations (\%85). This difference underscores the necessity of considering a wider spectrum of activations when studying their recognition capability.

\subsection{Metrics}
In this section, we investigate the feasibility of exploiting the aforementioned metrics in the optimization process, with a specific focus on non-sample-based metrics.

We begin our discussion with IoU, which has been the metric optimized by all algorithms discussed in this chapter to identify the most aligned concepts. Hence, the previously reported results correspond to those obtained by solely optimizing this metric. One limitation of IoU, as highlighted in previous literature~\cite{Mu2020, Makinwa2022}, is that the IoU can be artificially increased by increasing the formula's length. Common fixes for this issue include presetting the length for all neurons before applying the algorithm or introducing a term to penalize lengthy explanations~\cite{Mu2020}.

Concept masking involves multiple iterations of feeding the entire dataset to the architecture. Consequently, optimizing this metric during the beam process poses computational time challenges.

In the case of optimizing for activation coverage, given the fixed denominator, only the numerator - the intersection between labels and activations - matters. In this case, the intersection disproportionally favors larger concepts covering the entire activation, even if the covered portion is a tiny fraction of the label annotation. Consequently, at first sight, scene concepts covering the whole image are expected to be associated with the neurons. However, scene concepts do not cover the full dataset, resulting in several zeros in the equation. Therefore, the algorithm converges towards default rules composed of colors since they are the most frequent ones and the largest among the most frequent.

\begin{table}[!b]
    \centering
    \caption{Avg. and Std Dev. of the scores obtained by optimizing the algorithm using Detection Accuracy. Underlined results are those that reach lower scores compared to when using the IoU score. Bold results are those that reach higher scores.}
    
    \begin{tabular}{crrrrrr}
        \toprule
        & IoU & ExplCov & SampleCov & ActCov &DetAcc&LabMask\\
         \midrule
         Our & 
         \underline{0.01 \footnotesize{$\pm$ 0.03}}&
         \underline{0.01 \footnotesize{$\pm$ 0.03}}&
         0.74 \footnotesize{$\pm$ 0.26}&
         \underline{0.01 \footnotesize{$\pm$ 0.04}}&
         \textbf{0.68 \footnotesize{$\pm$ 0.23}}&
         0.68\footnotesize{$\pm$ 0.21}\\
         \midrule
              Cluster 1 & 
              \underline{0.00 \footnotesize{$\pm$ 0.00}} &
              \underline{0.00 \footnotesize{$\pm$ 0.00}}& 
              0.90 \footnotesize{$\pm$ 0.19}& 
              \underline{0.00 \footnotesize{$\pm$ 0.00}}& 
              \textbf{0.89 \footnotesize{$\pm$ 0.12}}&
              0.60\footnotesize{$\pm$ 0.17}\\
             Cluster 2 & 
             \underline{0.00 \footnotesize{$\pm$ 0.00}} &
             \underline{0.00 \footnotesize{$\pm$ 0.00}}&
             0.82 \footnotesize{$\pm$ 0.22}&
             \underline{0.00 \footnotesize{$\pm$ 0.00}}&
             \textbf{0.74 \footnotesize{$\pm$ 0.21}}&
             0.72\footnotesize{$\pm$ 0.23}\\
            Cluster 3 &
            \underline{0.00 \footnotesize{$\pm$ 0.00}} &
            \underline{0.00 \footnotesize{$\pm$ 0.00}}&
            0.69 \footnotesize{$\pm$ 0.26}&
            \underline{0.00 \footnotesize{$\pm$ 0.00}}&
            \textbf{0.68 \footnotesize{$\pm$ 0.22}}&
            \textbf{0.70\footnotesize{$\pm$ 0.23}}\\
            Cluster 4 & \underline{0.00 \footnotesize{$\pm$ 0.00}} &
            \underline{0.00 \footnotesize{$\pm$ 0.01}}&
            \textbf{0.65 \footnotesize{$\pm$ 0.27}}&
            \underline{0.00 \footnotesize{$\pm$ 0.00}}&
            \textbf{0.53 \footnotesize{$\pm$ 0.20}}&
            \textbf{0.69\footnotesize{$\pm$ 0.21}}\\
          Cluster 5 & 
          \underline{0.03 \footnotesize{$\pm$ 0.06}} &
          \underline{0.03 \footnotesize{$\pm$ 0.07}}& 
          \textbf{0.64 \footnotesize{$\pm$ 0.24}}& 
          \underline{0.03 \footnotesize{$\pm$ 0.08}}& 
          \textbf{0.54 \footnotesize{$\pm$ 0.19}}&
          \textbf{0.68\footnotesize{$\pm$ 0.22}}\\
         \bottomrule
    \end{tabular}
    \label{tab:activation_coverage}
\end{table}
\Cref{tab:activation_coverage} shows the results using Detection Accuracy as an optimization metric. While improvements in terms of Sample Coverage and Detection accuracy are expected, it is interesting to note that Concept Masking also benefits from this metric. Conversely, the IoU score and Activation Coverage drop down to zero,  indicating a potential correlation between these metrics and an opposite behavior to the Detection Accuracy.

\textbf{These results underscore the absence of a straightforward method in the optimization process to artificially inflate the overall score using novel metrics} and reinforce the notion that multiple metrics should be used at the same time to have a fuller understanding of the quality of these approaches.

\section{Design Choices}
\label{sec:neurons_choices}
This section delves into design choices connected to the implementation of our proposed algorithm.
\begin{table}[!b]
    \centering
    \caption{Avg. and Std Dev. of the proposed metrics when using a variable number of clusters. Results are computed for 50 randomly extracted units.}
    
    \begin{tabular}{crrrrrr}
        \toprule
        Clusters& IoU & ExplCov & SampleCov & ActCov &DetAcc&LabMask\\
         \midrule
         5 & 
         0.15 \footnotesize{$\pm$ 0.37}&
         0.60 \footnotesize{$\pm$ 0.23}&
         0.70 \footnotesize{$\pm$ 0.17}&
         0.34 \footnotesize{$\pm$ 0.11}&
         0.21 \footnotesize{$\pm$ 0.10}&
         0.55 \footnotesize{$\pm$ 0.22}\\
         10 & 
         0.09 \footnotesize{$\pm$ 0.05}&
         0.48 \footnotesize{$\pm$ 0.35}&
         0.68 \footnotesize{$\pm$ 0.23}&
         0.30 \footnotesize{$\pm$ 0.16}&
         0.13 \footnotesize{$\pm$ 0.06}&
         0.58 \footnotesize{$\pm$ 0.17} \\
         15 & 
         0.07 \footnotesize{$\pm$ 0.04}&
         0.44 \footnotesize{$\pm$ 0.36}&
         0.66 \footnotesize{$\pm$ 0.21}&
         0.25 \footnotesize{$\pm$ 0.16}&
         0.09 \footnotesize{$\pm$ 0.05}&
         0.56 \footnotesize{$\pm$ 0.18} \\
         \bottomrule
    \end{tabular}
    \label{tab:numclusters}
\end{table}
\begin{table}[b]
\centering
\caption{Avg. and standard deviation of visited states per unit. Results are computed for 100 randomly extracted units.}
\label{tab:heuristics}
\begin{tabular}{lllr}
    \toprule
     Heuristic & \multicolumn{2}{c}{Info from} &Visited Labels \\
        & NetDissect & Dataset &  \\
     \midrule
     
     Areas & - & \checkmark& 23602 {\scriptsize $\pm$ 3420} \\
    
     CFH & \checkmark & \checkmark& 5990 {\scriptsize $\pm$ 3066}\\
     
     MMESH & \checkmark & \checkmark & 129  {\scriptsize $\pm$ 712}\\
     \bottomrule

\end{tabular}

\end{table}
\paragraph{Number of Clusters.}
One critical design choice is the number of clusters per neuron. Here, we evaluate the quality of explanations when the number of clusters ranges from 5 to 15. \Cref{tab:numclusters} suggests that increasing the number of clusters does not yield better explanations. Moreover, by inspecting the labels associated with the clusters, most labels (75\%) are repeated across adjacent clusters when the number of clusters is high. Note also that computing explanations for 15 clusters requires more computation time ($\times 3$) than computing explanations for five clusters. Therefore, there is a trade-off between computational time and novel coverage. While these results suggest that a lower number of clusters is the best option, we hypothesize that the iterative procedure of K-means can also influence them. Therefore, developing a clustering algorithm tailored to the task could potentially flip these results by finding that a higher number of clusters leads to better explanations.

\paragraph{Clustering Algorithm.}
Connected to the choice of the number of clusters, another design choice regards the clustering algorithm. The algorithm can be chosen based on the dataset's size and the activation vectors' size. In the settings reported in this thesis, the dataset and the activations vectors are so large that the K-Means algorithm is the only viable option among the clustering algorithms publicly available. Indeed, to compute the clusters, more sophisticated algorithms would require either an amount of time greater than the one saved by our heuristics or too much memory space to be feasible in standard workstations. 

\paragraph{Heuristics.}
The third choice is about the heuristic used to speed up the beam search. Here, we evaluate two alternative heuristics designed to use less information for estimating the IoU score: the \emph{Areas} and the \emph{Coordinates-Free}  heuristics.

The \textbf{Areas} heuristic uses only the information about the mask size of the terms composing the current label:
\begin{equation}
    \widehat{IoU}(L, \tau_1,\tau_2, \mathfrak{D} )  = \frac{\widehat{I_x}}{\sum_{x \in \mathfrak{D}} |M_{[\tau_1,\tau_2]}(x)|  - \widehat{I_x}}
\label{eq:areas}
\end{equation}
where  
\begin{numcases} {\widehat{I}_x=}
min(|S_{[\tau_1,\tau_2]}(x, L_{\leftarrow})|+ |S_{[\tau_1,\tau_2]}(x, L_{\rightarrow})|, |M_{[\tau_1,\tau_2]}(x)| ) & $op=OR$  \\
min(|S_{[\tau_1,\tau_2]}(x, L_{\leftarrow})|, |S_{[\tau_1,\tau_2]}(x, L_{\rightarrow})|) & 
$op=AND$\\
min(|S_{[\tau_1,\tau_2]}(x, L_{\leftarrow})|, size(x) - |S_{[\tau_1,\tau_2]}(x, L_{\rightarrow})|)  & 
$op=AND~NOT$
\end{numcases}


The \textbf{Coordinates-Free Heuristic} (\emph{CFH}) avoids the estimation of $\widehat{S(L,x)}$ by setting it to 0, thus saving the computation time needed to compute the maximum and minimum possible extension:
\begin{equation}
    IoU(L, \tau_1,\tau_2, \mathfrak{D} )  = \frac{\widehat{I_x}}{\sum_{x \in \mathfrak{D}} |M_{[\tau_1,\tau_2]}(x)|  - \widehat{I_x}}
\label{eq:cfh}
\end{equation}
where  $\widehat{I}_x$ is defined as in MMESH.

\Cref{tab:heuristics} compares the three competitors based on the number of visited labels before finding the optimal label. The reported results are the average across 100 randomly selected units. Labels are computed  using the standard settings of \gls{CoEx} (i.e., by fixing  $T = {[\tau^{top},\infty]}$). Not surprisingly, MMESH achieves the target label much faster than the other heuristics. However, in scenarios where computing the coordinates of the maximum and minimum extension is prohibitively expensive, CFH serves as a viable alternative, as it is one order of magnitude faster than not using heuristics at all. Regarding the Areas heuristics, while it performs worst among those tested, it offers a key advantage: it does not require running the first step of the algorithm (i.e., NetDissect), as it does not use the information about the minimum possible intersection. Consequently, it can be employed within NetDissect to speed up the search process when the label search space is large. 

\section{Contributions}
\label{sec:neurons_contributions}
This chapter contributes and end the part of this thesis dedicated to the research on explaining the latent space learned by \glspl{DNN}.
It introduced a method applicable to any \gls{DNN} for analyzing alignment between latent representations and a predefined set of concepts. Specifically, in contrast with previous literature, it studies the full spectrum of activations by combining clustering, heuristics, and beam search.

The chapter discussed the gain in terms of the efficiency of the algorithm with respect to the state-of-the-art ones and how analyzing a wider spectrum of activations does not lower the quality of explanations and instead provides a fuller picture of the neurons' behavior. The chapter also sheds light on novel phenomena connected to neuron activations, such as the unspecialized activations around zero and the progressive specialization. Additionally, this chapter contributes to the research on metrics for explanation methods by proposing novel metrics for evaluating approaches that explain latent representations. 
Alongside these strengths, the chapter also analyzed some of the drawbacks of these approaches, like the challenges related to the usage of clustering algorithms.

Finally, this chapter ends the part of the thesis devoted to the introduction of novel methods for explaining deep learning. We began with the introduction of a generalization of a known architecture (\Cref{chapter:pignn}) to a specific domain and settings, and subsequently, we progressively discussed more generalized approaches aimed at mitigating previous issues and expanding the potential applications. This latest algorithm, being model-independent and not necessitating modification or additional training, signifies the apex of such efforts. Therefore, the next chapter describes a more general analysis of the XAI field. Then, the final chapter discusses the limitations of the proposed approaches and possible future directions, including the ones discussed in this chapter.
\part{Bringing Explanations to the User: Visual Analytics}
\label{part:va}
\chapter{Explaining Deep Learning in Visual Analytics Systems}
\label{chapter:va}

In the preceding chapters, this thesis explored various mechanisms to explain the behavior of \glspl{DNN}. We showcased several usage scenarios where explanations are employed to assist users. In those instances, the explanations are static, target the model developer, and necessitate the presence of an \gls{XAI} expert. 

This chapter discusses a promising direction for delivering explanations to users and allowing them to interact with models and explanations. Specifically, it discusses \gls{VA} systems that employ explanation methods to understand \glspl{DNN}. The \gls{VA} community has recently developed intuitive and interactive user interfaces to aid users in understanding machine learning models. Unlike \gls{XAI} research, \gls{VA} systems are typically designed for specific tasks and target users. The benefits and novelty of these methods rely more on the functionalities offered to the user and the tasks they can solve rather than the development of specific \gls{XAI} methods. In particular, \gls{VA} excels in ease of exploration, hypothesis verification, and discovery when dealing with large datasets. 

A natural question arises: can \gls{VA} and \gls{XAI} complement each other by combining their strengths and integrating \gls{XAI} methods into \gls{VA} systems? This chapter evaluates the current state of integration between the two disciplines by reviewing and analyzing 67 papers proposing \gls{VA} systems that incorporate \gls{XAI} methods for understanding \glspl{DNN}, which is the topic of this thesis. 

The chapter is organized as follows: \Cref{sec:va_overview} provides a brief overview of visual analytics systems employing explanation methods; \Cref{sec:va_analysis} analyzes strengths and weaknesses of the current integration between \gls{XAI} and \gls{VA}; \Cref{sec:va_future} discusses challenges and next steps needed in the field; and \Cref{sec:va_contributions} summarized the contributions of this chapter. 

\section{Overview}
\label{sec:va_overview}
This section provides an overview of how \gls{VA} systems use explanation methods to enhance the interpretability of \glspl{DNN}. Specifically, this section analyzes the support provided for each of the explanation categories presented in \Cref{sec:back_xai}. Additionally, it discusses and presents a novel \gls{VA}-specific category: model behavior.
For each category, this section describes both the main visualization and interaction techniques used in the systems and the tasks the users can accomplish thanks to the integration of these \gls{XAI} methods into the system. While some of these tasks could be achieved by using \gls{XAI} methods alone, \gls{VA} systems expedite the analysis process through broader exploration and provide tools to the users without the need for \gls{DL} or \gls{XAI} expert involved in the process.

\subsection{Feature Attribution}
\begin{figure}[t]
    \centering
    \includegraphics[scale=1.2]{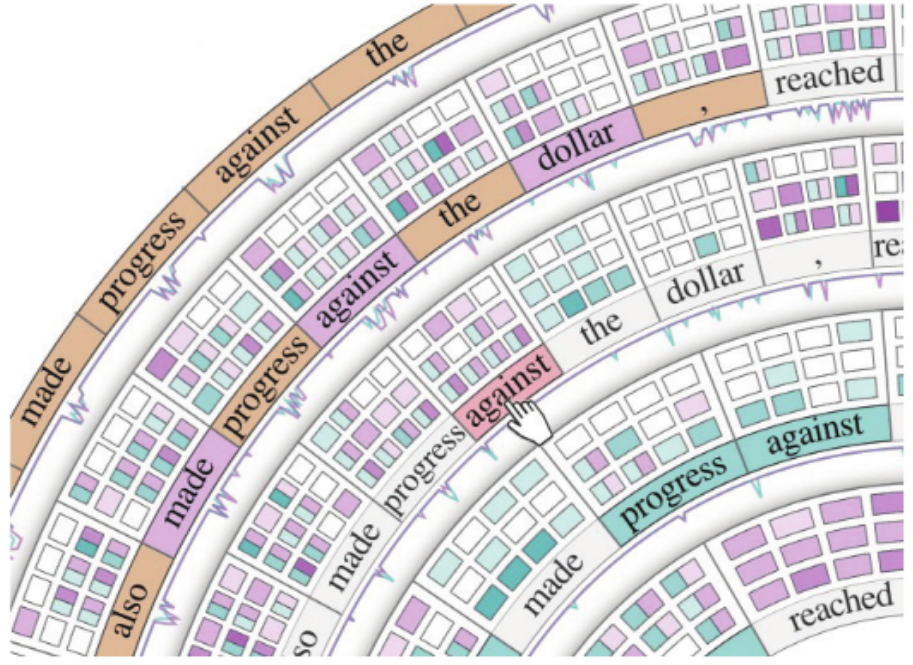}
    \caption{Example of complex visualization in VA systems of attention flow.}
    \label{fig:attentionflow}
\end{figure}
Unsurprisingly, feature attribution is the most popular category for \gls{VA} systems to enhance interpretability. For individual explanations and standard settings,  \gls{VA} systems generally employ common visualization techniques also used in \gls{XAI} papers, such as heatmaps~\cite{VanDenBrandt2020, Han2021, Huang2021, Zahavy2016, Hilton2020, Wang2019DeepVid, Strezoski2017, Chan2020, Jaunet2020, Chawla2020, Chae2017, Skrlj2021, Ji2021} and matrices~\cite{Wang2018rnn, DeRose2021, Jaunet2021, Park2019,
Liu2019nlize}. However, for more advanced settings, \gls{VA} exposes its full strength by proposing novel and more complex visualizations. For example, the system proposed by \citet{DeRose2021} visualizes Transformers' layers in a radial layout, helping users keep track of the flow of attention weights across multiple layers simultaneously (\Cref{fig:attentionflow}). Users can interact with the interface by rotating the radiants, filtering layers, heads, and input tokens. This kind of support is common to the systems supporting the analysis of the most complex architectures~\cite{Wang2021dodrio, DeRose2021, Skrlj2021}.

Regarding the interactions, sorting~\cite{Chawla2020, Ming2020, Kwon2019, Vig2019, Park2019} and filtering~\cite{Wang2018rnn, Dong2020, Jaunet2021, Ji2021} are the most basic and commonly supported for this explanation category. These two interactions are particularly useful for global feature attribution, where filtering can aid users in detecting features relevant to specific subsets of the data~\cite{Chawla2020, Ji2021, Zhao2020, Park2019, Hazarika2019, VanDenBrandt2020, Kwon2019, Shen2020, Ji2021}. 

Advanced interactions for this category include real-time what-if analysis and steering the model training by using attributions.
In the former case, local explanations can guide features removal (e.g., by brushing over an image) and monitor the changes in predictions~\cite{Han2021, Huang2021, Kwon2019} or in the decision process when the input is modified~\cite{Park2021, Jaunet2021, Strobelt2019, Liu2019nlize}. In the latter case, users can impose constraints on the training process by specifying the desired values in feature attributions. For example, some systems support the user in specifying the desired attention values in self-explainable DNNs~\cite{Ming2020, Kwon2019}. Then, the model is optimized to align attention values closer to the specified ones. 

Papers proposing these systems demonstrate that combining visualizations and iterative interactions with feature attribution helps users on several tasks, such as assessing the reliability \cite{VanDenBrandt2020, Park2021} of predictions, discovering bias in the network's decision process \cite{Vig2019, Jaunet2021}, and correct failures \cite{Chawla2020, Hilton2020, Wang2019DeepVid}. Moreover, in domains where domain experts are involved, these systems lead to the discovery of new knowledge in terms of interactions between factors~\cite{VanDenBrandt2020} and pitfalls in the design of the \glspl{DNN} themselves \cite{Han2021}. 

\subsection{Learned Features}
Several systems use learned features methods for helping users understand the knowledge learned by \glspl{DNN}. In this case, one of the major benefits of \gls{VA} systems lies in the extensive supported exploration and the summarization capability.

While the visualization of learned features for a specific component is mostly aligned with the current literature in \gls{XAI}, using real or generated patches in the vision domain~\cite{Yosinski2015, Zeng2017, Hohman2020, Das2020, Jia2019, Zhao2020, Park2022, rathore_topoact_2021, Liu2018, Yosinski2015, Hohman2020, Ma2021, Das2020, Hohman2017, rathore_topoact_2021, Strezoski2017} and word clouds in the text data domain~\cite{Yosinski2015, Hohman2020, Ma2021, Das2020, Hohman2017, rathore_topoact_2021, Strezoski2017}, the exploration and simultaneous visualization of learned features for entire parts of the network require more complex visualizations and interactions. 

\begin{figure}[t]
    \centering
    \includegraphics[scale=0.5]{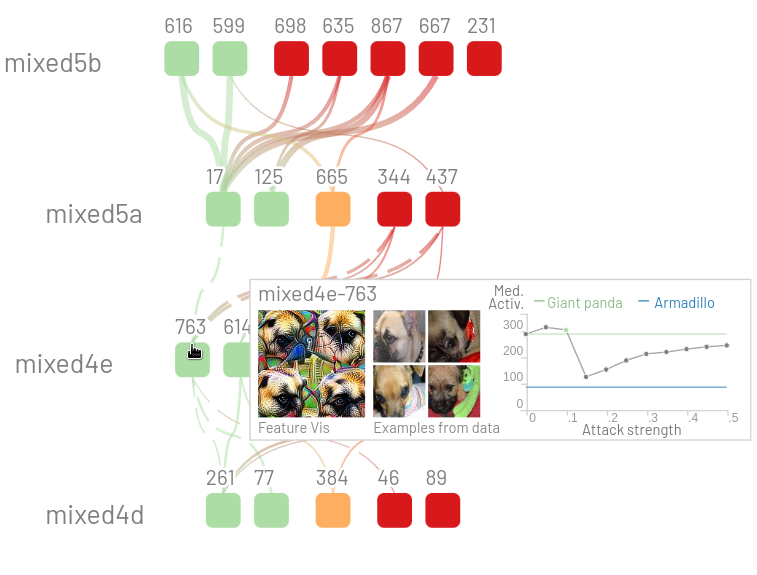}
    \caption{Example of abstraction in VA system.}
    \label{fig:bluff}
\end{figure}
\gls{VA} solutions address this challenge by providing abstractions summarizing the behavior of macro-components and then enabling the user to explore individual components within the macro ones through interactions. Abstractions are provided by aggregating activations, based on their average or similarity~\cite{Hohman2020, Ji2021}, using clustering algorithm~\cite{Zhong2017, Park2022, Liu2017}, or based on the importance of the individual components for the task~\cite{Das2020}. 

For example, \Cref{fig:bluff} shows an abstraction provided by BLUFF~\cite{Das2020}, which visualizes, for each layer, the neurons whose activations diverge the most when under an adversarial attack. This abstraction allows users to focus on the most problematic neurons and then explore the rest of the architecture. Abstractions are visualized in the form of Sankey diagrams \cite{Hohman2020, Das2020}, graphs \cite{Liu2018, rathore_topoact_2021} or scatter plots using dimensionality reduction techniques~\cite{Vandermaaten08,Pezzotti2016}.

In systems employing learned features techniques, interactions are mainly designed to ease the exploration of the networks. Zoom-in~\cite{Hohman2020, Park2022} and filtering~\cite{Hohman2020, Park2022} are the standard interactions available in most of the analyzed systems. More advanced interactions include the definition of custom clusters \cite{Liu2017, Zhong2017}, switching between facets \cite{Liu2017, Hohman2020}, and filtering the granularity of the visualization or the modification of the importance criterion~\cite{Das2020}.

Papers proposing \gls{VA} systems employing learned features techniques demonstrate their efficacy in supporting users in the analysis of how low-level features are aggregated into high-level features~\cite{Liu2017, Hohman2020, Das2020, Park2022, Ji2021, Liu2018}, in the diagnosis of the training process~\cite{Zhong2017, Pezzotti2018}, in the comparison between models or configurations~\cite{Ma2021}, and to validate that the learned knowledge is aligned to the expected one~\cite{Liu2017, Ji2021}.

\subsection{Explanations by Examples and Counterfactuals}
Explanations by examples and counterfactuals are rarely integrated into \gls{VA} systems, and when employed, they typically serve as complementary tools for other techniques. Therefore, they represent categories that need further research. This section jointly analyzes these two categories because half of the analyzed papers support both simultaneously, suggesting they are often considered complementary.

The most common visualizations employ a list of inputs enriched by metadata such as the similarity scores, predictions~\cite{Chan2020}, highlights of common features, or highlights of differences between the explanation and the input~\cite{Hoover2020, Chan2020, Strobelt2018, Chan2020}. Average statistics are usually summarized in bar charts \cite{Cheng2021} or tables~\cite{Wexler2019, Cheng2021}.  These visualizations align with the visualizations used in \gls{XAI} for this category~\cite{Kenny2023features}. When the explanations are retrieved using similarity on patterns (e.g., attention~\cite{Ji2021} or features~\cite{He2020}), these patterns are usually visualized and highlighted along with the explanations by examples.

\begin{figure}[t]
    \centering
    \includegraphics[scale=0.5]{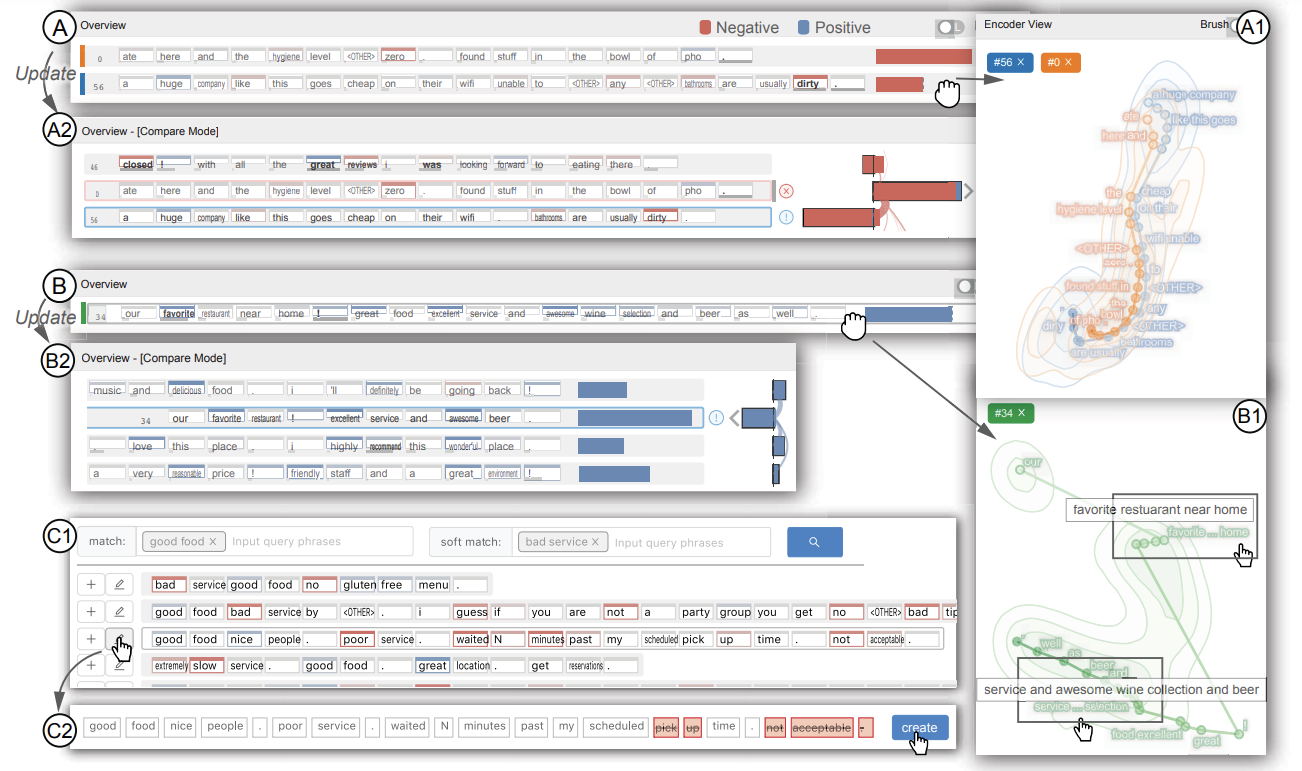}
    \caption{Example of interface to steer a model to add, delete, and revise prototypes. All edits are revertable and traceable in the editing history. Image from \citet{Ming2020} © 2019 IEEE}
    \label{fig:protosteer}
\end{figure}

Regarding interactions, they are mainly based on sorting mechanisms~\cite{Cheng2021, Liu2019nlize}, reducing or increasing the number of explanations~\cite{Ji2021, Cheng2021}, and selecting input features that should be shared by the explanation by example~\cite{Strobelt2018}. A special case is represented by self-explainable \glspl*{DNN} based on prototypes, where explanations by examples are used to extract the semantics encoded in prototypes, similarly to the process presented in \Cref{chapter:pignn}. In these cases, \citet{Ming2020} propose to use interactions to steer the model training. The system allows the user to specify a set of desired neighbors (i.e., explanations by examples), and these samples are used to generate new prototypes that include the desired samples as neighbors (\Cref{fig:protosteer}).

\gls*{VA} systems utilize both these types of explanations to enable users to understand individual predicitons~\cite{Strobelt2019} and suggest edits to the user~\cite{Strobelt2019, Wexler2019, Wright2021, Liu2019nlize, wang_hyppersteer_2020}. Explanations by examples are used to estimate the meaning of latent vectors~\cite{Strobelt2019, Hoover2020, Strobelt2018}, improve the representativeness of a prototype representation~\cite{Ming2020}, and keep track of the decision process layer by layer~\cite{Hoover2020, Strobelt2019}. Conversely, several systems use counterfactuals to verify hypotheses~\cite{Cheng2021, Strobelt2018}, or explore alternative scenarios in reinforcement learning~\cite{Mishra2021}

\subsection{Model Behavior}
Model behavior is a category specific to \gls*{VA} systems and aims at explaining the \gls{DNN} by facilitating the discovery of patterns in terms of predictions or values of the model components and linking them to specific behaviors. By design, these methods are inherently post-hoc and global, combining pattern mining, clustering,  interactions, and sometimes explanations of other categories~\cite{Wang2019, Jaunet2020}. They are especially useful when other explanation methods fail to provide a comprehensive understanding if used alone, or classical \gls{XAI} methods are not applicable.

The goal of this category is to improve the simulability of a network by providing answers to questions such as ``How does the model react in a given situation?''. This category is strongly linked to intrinsic methods of \gls{XAI} since the patterns often consist of the values of internals (e.g., neuron activations, attention weights).  The methods of this category are also loosely related to the \gls{XAI} research on circuits~\cite{Olah2020} and pathways discovery~\cite{meng2023massediting}.

One of the distinctive characteristics of \gls{VA} systems supporting this category of explanations is the extensive use of linked views. In this case, several views (i.e., parts of the interface) are connected, and a modification of an element in a view causes a change in the visualization of the other connected views. This feature facilitates the detection of patterns from multiple points of view simultaneously. Patterns can be visualized as lists of neurons~\cite{Park2021}, tables \cite{Wang2021a}, partial dependency plots \cite{wang_hyppersteer_2020} or summarized in a more compact visualization, like decision trees~\cite{Becker2020,Jia2019} and scatter plots~\cite{Zahavy2016}.

\begin{figure}[t]
    \centering
    \includegraphics[scale=0.58]{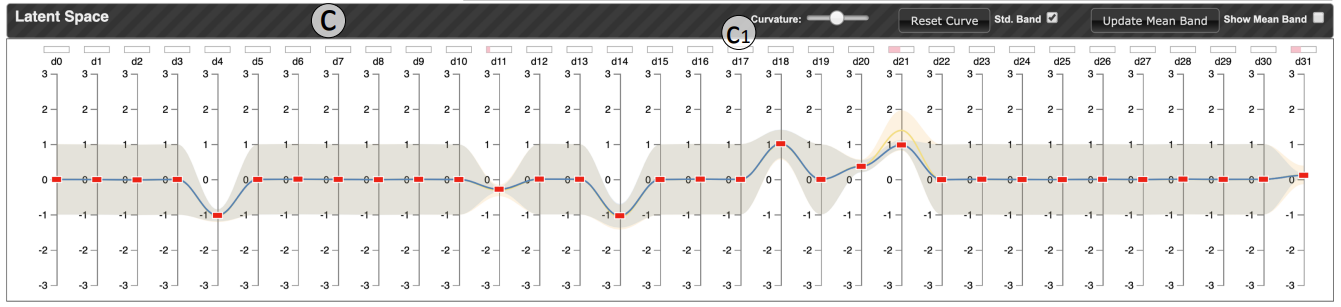}
    \caption{View to interact with a latent representation. The user can assign values to dimensions of the latent space in the parallel coordinate plot, and the system will adjust the other views accordingly. Image from \citet{Wang2020scan} ©2020 IEEE}
    \label{fig:scanviz}
\end{figure}

Interactions are the core components of these systems since the explanations are human-driven. The most popular interactions include the selection of subsets of interest~\cite{Shen2020, Bilal2018, Kahng2018}, zooming functionalities~\cite{Wang2021drl, Jin2020va, Jaunet2020},  sorting mechanisms, and definition of patterns~\cite{Wang2019}.
Several systems provide support for \emph{interactive input observations} and \emph{interactive model observations}. For example, \citet{Wang2020scan} (\Cref{fig:scanviz}) and \citet{Strobelt2018} use Parallel Coordinates Plots to represent input features or layer activations and let the user modify the values selected in each axis in real time to visualize how the network changes its behavior accordingly.

Papers proposing \gls{VA} systems supporting this category successfully help users to extract policies from deep reinforcement learning~\cite{Wang2019}, approximate the decision process of models, layer, and neurons~\cite{Zahavy2016, Park2021, Wang2019, Jaunet2020, Wang2021a, Jia2019, Becker2020}, analyze errors ~\cite{Liu2018, Jin2020va}, and formulate and refine hypotheses about the semantics associated with the latent spaces~\cite{Shen2020, Wang2021drl, Wang2020scan, Strobelt2018}.

\section{Analysis}
\label{sec:va_analysis}
This section analyzes the considered \gls{VA} systems by discussing the target users and domains of these systems, the kind of interactions supported, the evaluation of their usefulness, and their state of integration with \gls{DL} and \gls{XAI} fields. 
\paragraph{Target Users.}
\begin{figure}[t]
    \centering
    \includegraphics[scale=0.8]{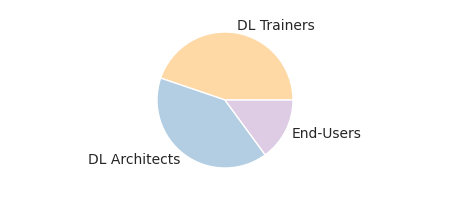}
    \caption{Distribution of target users in the analyzed VA systems}
    \label{fig:users}
\end{figure}
We begin the analysis by examining the target users of the collected systems. \Cref{fig:users} shows the distribution of target users in the analyzed VA systems, revealing that \gls{DL} architects and trainers are the primary focus. This is not surprising since \gls{DL} experts (i.e., individuals with a background in \gls{DL}) are also the predominant target users of \gls{XAI} methods, particularly when used for debugging, which is one of the current main applications of \gls{XAI} techniques. 
Conversely, only 14\% of systems target end-users without knowledge about \gls{DL}. This distribution can be considered an indication of the maturity of the \gls{XAI} field: the field seems still confided mainly in research applications.

\begin{figure}[t]
    \centering
    \includegraphics[scale=0.2]{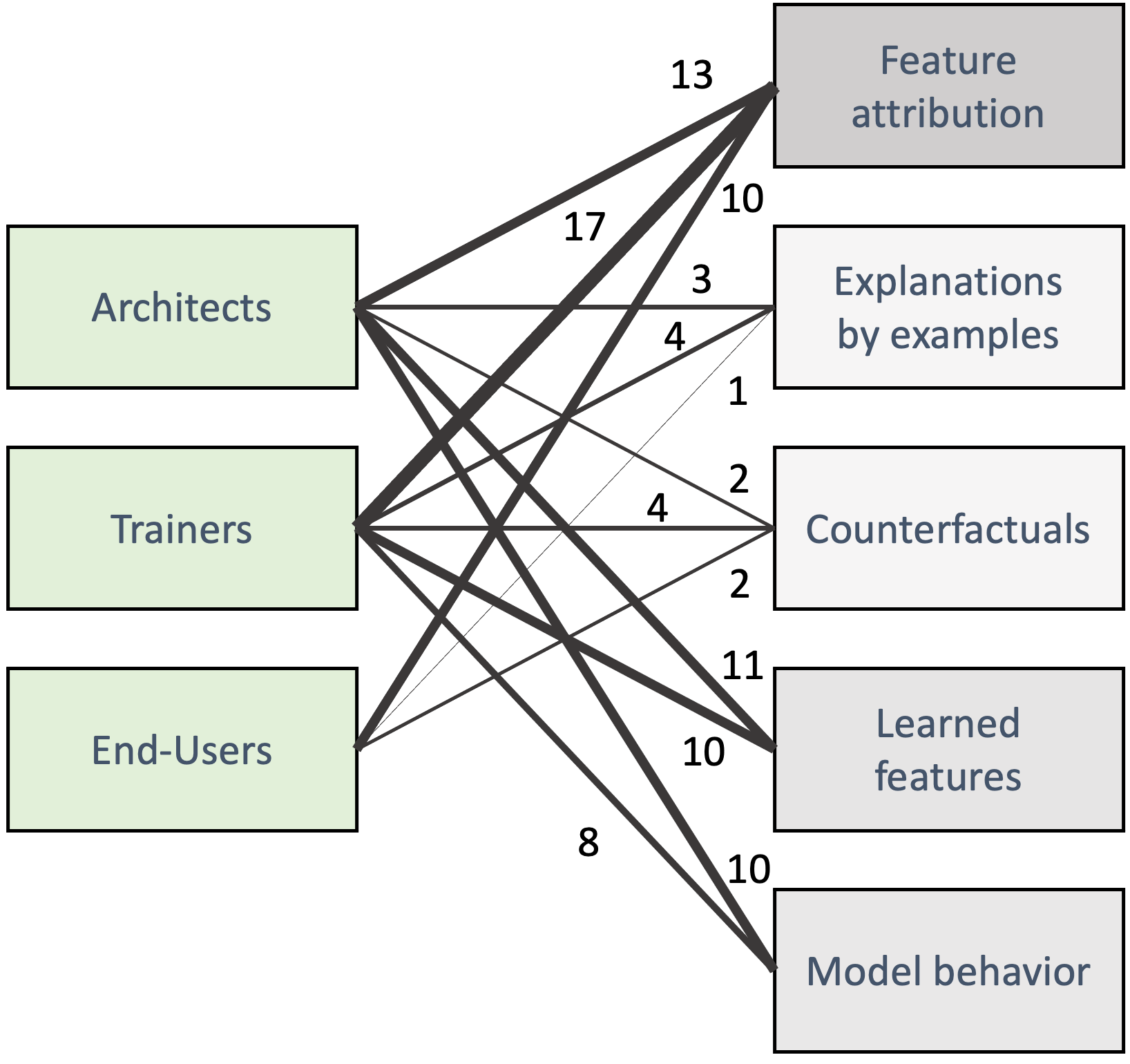}
    \caption{Summary of user support for different XAI categories. The edge width is proportional to the number of systems connecting users and categories. The opacity of XAI categories is proportional to the total number of contributions.}
    \label{fig:users_categories}
\end{figure}
The analysis of the categories supported by systems targeting end-users (\Cref{fig:users_categories}) reveals a reliance on feature attribution methods, with limited support for explanations by examples and counterfactuals and no support for model behavior and learned features categories. These findings suggest that the latter explanation types may not yet be sufficiently mature, and further research is needed in this direction. For instance, inspecting and navigating the network's neurons to understand what the network learned during the training process may require too much effort and expertise for an end-user. In such cases, future directions could involve the employment of summarization techniques and circuits~\cite{Olah2020} that go beyond the currently supported \gls{XAI} methods. 

Another interesting aspect is that most systems targeting end-users deal with time series and text data, while only one system deals with image data. This finding is surprising given the long-standing maturity of feature attribution methods for image data. The popularity of time series may be attributed to its relevance in healthcare, a primary domain for \gls{XAI} techniques~\cite{Tjoa2021, Albahri2023, Onari2023}. Conversely, the low presence of image data suggests that a few real-world problems can benefit from the progress in this research yet.

\paragraph{Target Domains.}
\begin{figure}[t]
    \centering
    \includegraphics[scale=0.7]{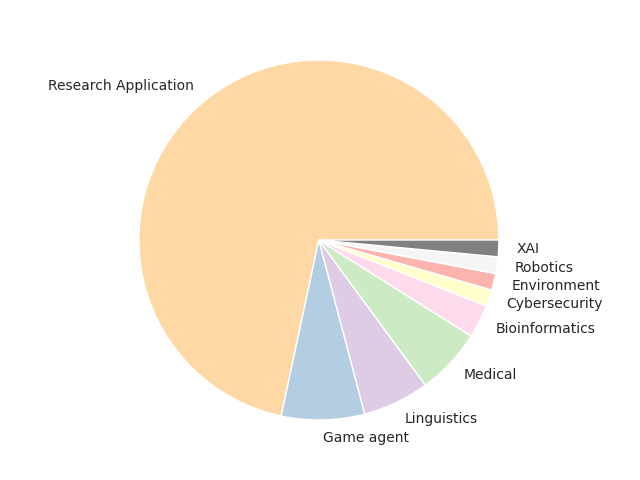}
    \caption{Target domain of VA systems. Research refers to research papers targeting general applications.}
    \label{fig:domain}
\end{figure}
The second analysis discusses the domains investigated in the collected systems. \Cref{fig:domain} corroborates the results regarding the users by showing that the majority of papers target research applications. The other applications target mainly \emph{games}, \emph{linguistics}, and \emph{medical domains}. The inclusion of games and linguistics may be tied to the recent surge in reinforcement learning and large language models, while the medical domain represents a common testing scenario for \gls{XAI} approaches, as noted before. Regarding the supported categories, in alignment with the analysis for the users, most systems targeting non-research rely on feature attribution methods to provide explainability. However, in this case, these methods are often complemented with model behavior (40\%), explanations by examples (10\%), or counterfactuals (10\%) categories. As for end-users, the learned features category is not supported by any system, suggesting that this type of explanation is not mature enough to be digested by domain experts with limited knowledge of \gls{DL}.

\paragraph{Interactions.}
\begin{figure}
    \centering
    \includegraphics[scale=0.7]{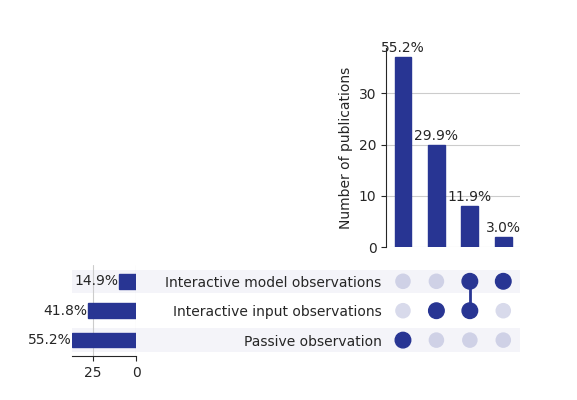}
    \caption{Distribution of interactions supported for the user.}
    \label{fig:interactions}
\end{figure}
\Cref{fig:interactions} presents the distribution of the interaction degree these systems offer to the target users. Notably, over half of the systems provide only \emph{passive observations}, particularly those relying on feature attribution, learned features, and model behavior. These systems exploit interactivity to support exploration, summarization, and filtering capabilities mainly. One-third of the systems additionally support \emph{interactive input observations}, allowing users to modify the input features. Systems providing this type of interaction support feature attribution, counterfactuals, and model behavior. These interactions are essential for the latter categories to support counterfactual reasoning~\cite{Wexler2019} and verify hypotheses on the model behavior. 

Finally, only 15\% of systems support \emph{interactive model observations}. Most of these systems employ feature attribution to provide explainability and target attention or prototype-based models. Among them, only a couple of works support steering ability and the possibility of monitoring the training process. Thus, better support for these tasks represents a promising future research direction.

\paragraph{Evaluation.}
The evaluation of the efficacy of the analyzed systems adheres to the established procedures in the  \gls{VA} field. Specifically, these systems are usually evaluated based on user satisfaction.

\begin{figure}[t]
    \centering
    \includegraphics[scale=0.5]{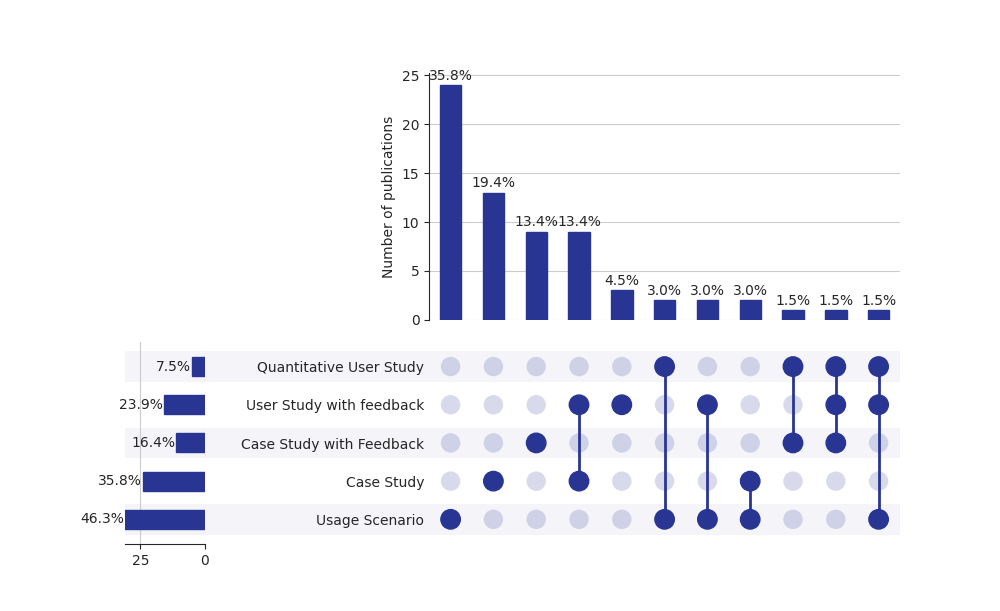}
    \caption{Distribution of evaluation procedures used in the analyzed systems.}
    \label{fig:va_evaluation}
\end{figure}

Nearly half of the papers include a usage scenario, offering potential yet fictitious scenarios where their systems could prove beneficial (\Cref{fig:va_evaluation}). Since no participants or experts are involved in the process, the usage scenario can be considered proof-of-concept or entry points for further evaluation. Slightly more reliable evaluations are provided through case studies. They include a description of workflows and discoveries of real usage sessions carried out by a user in controlled settings. The case study and the usage scenario lack a quantitative and qualitative evaluation from the users. 

The qualitative assessment is provided through case studies with feedback and user studies with feedback. These evaluation procedures are used in 40\% of the analyzed papers. The case study with feedback follows the same workflow as the case study without them, but the user provides feedback in the form of an interview at the end of the test session. The goal of this evaluation is to assess the generalization of the results over similar scenarios. Conversely, user studies with feedback relax the controlled settings and involve multiple testing sessions with different users, collecting their feedback. Finally, quantitative user studies provide quantitative evaluations, asking users to fill out questionnaires and collecting quantitative metrics during interactions with the system. Quantitative user studies represent the most general, reliable, and costly evaluation procedure. Unfortunately, less than 10\% of the analyzed systems incorporate this evaluation process (\Cref{fig:va_evaluation}).

Overall, the evaluation practices align with the \gls*{VA} standards, which heavily rely on user satisfaction. In scenarios where these systems employ explanation methods from the \gls{XAI} literature and apply them to standard settings, these evaluations can further validate the quality of such approaches. However, we observe a lack of a proper quantitative evaluation, in terms of user-independent metrics (\Cref{sec:back_xai}), of explanation methods when these are novel and proposed alongside the system and when \gls{XAI} techniques are applied on settings different from their original ones.

\paragraph{Support for State-of-the-Art models and methods.}
\begin{figure*}[!t]
     \centering
     \begin{subfigure}[b]{0.45\textwidth}
         \centering
         \includegraphics[width=\textwidth]{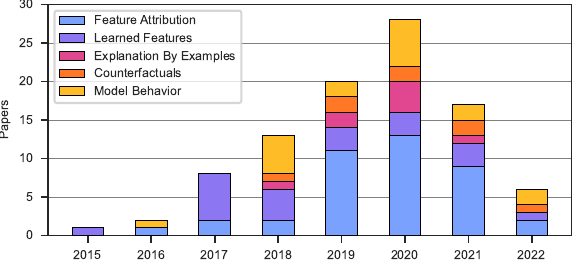}
         \caption{}
         \label{fig:xai_year}
     \end{subfigure}
     \hfill
     \begin{subfigure}[b]{0.45\textwidth}
         \centering
         \includegraphics[width=\textwidth]{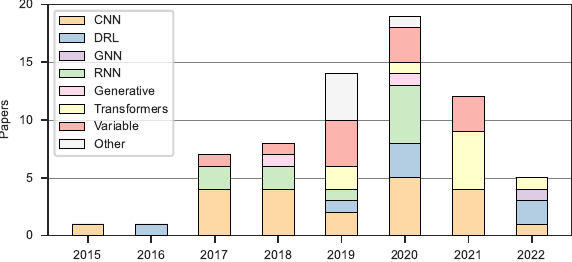}
         \caption{}
         \label{fig:model_year}
     \end{subfigure}
        \caption{ {(a)} Explanation categories and {(b)} models coverage over the years.}
        \label{fig:temporal_trends}
\end{figure*}
Here, we analyze the diversity and coverage in \gls{VA} systems of \glspl{DNN} and explanation methods used to support the different categories. We observe that (\Cref{fig:model_year}) convolutional neural networks were the most supported architecture until 2018.  Subsequently, other architectures, especially Transformers, gained popularity, following the trend in AI research.

Regarding the supported categories (\Cref{fig:xai_year}), feature attribution emerged as the preferred method for enhancing interpretability in 2019, leading to a reduction in the percentage of learned features. Looking more in detail on the specific methods used to support the categories, we noted a large gap between the availability of approaches in \gls{XAI} research and adoption in \gls{VA} systems. Indeed, the most recent supported methods are SHAP~\cite{Lundberg2017}, Grad-CAM~\cite{Selvaraju2017}, and LIME~\cite{Ribeiro2016}, which nowadays are considered baselines in the \gls{XAI} field and have been outperformed by several other methods. Moreover,  only five systems include these methods, while the others prefer even older methods like Deconvolution~\cite{Zeiler2014}  and vanilla Saliency Maps~\cite{Simonyan2014}. These choices may limit the widespread of these systems and their adoption from \gls*{DL} experts beyond the specific cases for which these systems are built.

\section{Towards a Better Integration}
\label{sec:va_future}
The previous section analyzed the current state-of-the-art \gls{VA} systems, discussing their strengths and weaknesses. This section, building upon the previous analysis, outlines research directions and action items necessary to mitigate such issues and advance the field.

We argue that the weaknesses related to models and explanations coverage are connected to the lack of \textbf{involvement of \gls{DL} and \gls{XAI}} experts in the design process. In fact, analyzing the papers' authors, most of \gls{VA} systems are solely designed by \gls{VA} experts, with \gls{DL} experts considered only as users. Few papers list \gls{XAI} experts among the authors, and they are rarely involved as users of the systems. This behavior results in limited support for recent and more effective explanation methods and restricted coverage of \gls{DL} models. Additionally, despite the availability of popular libraries for both \gls{XAI} and \gls{DL}, these systems rarely support their integration and typically favor one specific \gls{XAI} method. Conversely, supporting and displaying explanations returned by different methods of the same class would enhance the reliability of such systems, considering the lack of ground truth and the ongoing debate about evaluation in \gls{XAI} (\Cref{sec:back_xai}). 

Several actions could foster collaboration between these fields and mitigate these issues. Firstly, researchers should develop a \textbf{greater awareness} of each other's work. The knowledge of weaknesses and strengths of each field can help developers build more effective \gls{VA} systems. The awareness can be increased by hosting workshops and events in venues of both fields. While some conferences have hosted workshops discussing \gls{VA} for explaining \gls{DL}, these discussions have often been peripheral rather than central. Given the increasing importance of interactivity in \gls{XAI} with the rise of chatbots and more sophisticated dialogue systems, specialized workshops focused on this topic could significantly enhance the success of the discussed systems. 

A closer collaboration would make the support for novel models and explanation methods faster, aligning the state-of-the-art of both fields. As \gls{VA} systems become more widespread, \gls{XAI} research stands to benefit from their analytics tools, using them to test and refine their methods. Finally, as the field matures, the development of \textbf{common libraries} between \gls{VA}, \gls{DL}, and \gls{XAI} could facilitate research by providing an interface that abstracts access to \gls{DL} frameworks, \gls{XAI} methods, and \gls{VA} components. This would alleviate the difficulty researchers face when adapting or changing components of a system, such as datasets, models, and workflows, speeding up research in the field.

\section{Contributions}
\label{sec:va_contributions}
This chapter reviewed and analyzed visual analytics systems employing \gls{XAI} methods to explain \glspl{DNN}. It presented  \gls{XAI} categories supported by these systems and introduced a novel category specific to these approaches: the model behavior category. The analysis highlighted the current support for different users, domains, models, and \gls{XAI} methods, along with the evaluation procedures employed in papers proposing such systems.  The chapter underscored the potentiality of \gls{VA} solutions and the areas where future research could yield additional benefits. It advocates for increased collaboration between \gls{XAI}, \gls{VA}, and \gls{DL} fields and the promotion of joint events to align research findings. Additionally, it highlighted drawbacks of current solutions, such as the lack of quantitative evaluation of proposed \gls{XAI} methods and the limited presence of \gls{VA} solutions tailored for end-users without \gls{DL} knowledge.
\part{Conclusions}
\label{part:conclusion}
\chapter{Conclusions}
\label{chapter:conclusion}
This final chapter concludes the thesis by summarizing its contributions (\Cref{sec:summary}), discussing its limitations (\Cref{sec:limitations}), and outlining potential future research directions stemming from this work (\Cref{sec:concl_future}).
\section{Summary}
\label{sec:summary}
This thesis contributed to the field of explainable Artificial Intelligence by focusing on intrinsic techniques to enhance the interpretability of deep neural networks. The overarching goal was to leverage elements already embedded within modern neural networks to probe their behavior, thereby enabling users to understand and trust the decisions made by these models. The underlying idea is to treat the layers preceding the classifier as a black-box feature extractor and focus on the last part to enhance interpretability. In particular, the thesis investigated methods that add elements after the feature extractor or make its representations more interpretable. These approaches address a common limitation of self-explainable deep neural networks: constrained generalization across different architectures and lower performance than black-box models. 

The first part of the thesis introduced novel designs for self-explainable neural networks aimed at improving the interpretability of deep neural networks while preserving or enhancing their performance.
\Cref{chapter:pignn} introduced Prototype-based Interpretable Graph Neural Networks to address the opacity of graph neural networks by integrating prototype layers between the feature extractor and the classifier. Through extensive experiments, we demonstrated that this design preserves the performance of the black-box models while enhancing their interpretability, particularly in the context of graph data and chemical domains.

Building upon the insights gained from \Cref{chapter:pignn}, \Cref{chapter:senn} introduced Self-Explainable Memory-Augmented Deep Neural Networks, which augmented black-box DNNs with memory modules. The memory facilitates the storage and utilization of crucial information during the decision process. The chapter proposed tracking mechanisms of writing and reading processes to provide interpretability. The proposed designs are model agnostics and, thus, can be applied to any neural network and preserve the performance of the black-box models. Moreover, these designs have been proven flexible in the types of explanations that can be derived, spanning from single and group feature attributions to explanations by example and counterfactuals.

The second part of this thesis aimed at making the interpretation of latent representations easier.
\Cref{chapter:whitening} explored Graph Concept Whitening, which aims to make the interpretation of the semantics encoded in the latent representations of graph neural networks easier. By modifying the normalization layers, this technique encourages the latent space to represent molecular properties on its axes. In this way, it provides a useful entry point for network inspection and interpretability in the chemical domain while preserving model performance.

\Cref{chapter:compositional} pushed forward the interpretation of latent representations by introducing Clustered Compositional Explanations, a method applicable to black-box DNNs without any modification of its structure for analyzing alignment between latent representations and a predefined set of concepts. By studying the full spectrum of neuron activations and employing clustering algorithms, this method provided novel insights into the behavior of neurons and improved the efficiency and quality of explanations for DNNs.

Lastly, the last part and \Cref{chapter:va} analyzed a promising direction for delivering explanations to users and allowing them to interact with models and explanations. Namely, \Cref{chapter:va} reviewed the emerging field of visual analytics, which is working on the problem of explaining deep learning to the user through interactive interfaces. Specifically, the chapter analyzed visual analytics systems employing XAI methods to explain DNNs, discussing the potentiality and steps needed for full integration and alignment between visual analytics and the XAI field.

In conclusion, this thesis has contributed significantly to the ongoing research on explainable deep learning. By exploring various techniques and methodologies, we have advanced the state-of-the-art of self-explainable neural networks, particularly in graph neural networks, memory-augmented neural networks, and explainable latent space analysis. These contributions enhance the interpretability of deep learning models and pave the way for more transparent and trustworthy AI systems in various domains.

\section{Limitations}
\label{sec:limitations}

Despite the advancements made in this thesis, several open challenges related to the discussed approaches need further research and investigation.

Concerning prototype-based layers, a distinction from the vision domain is the difficulty in projecting the prototypes back to the input space. As the design relies on node embeddings, the interpretability of prototypes depends on the ratio between the number of layers preceding the prototype layers and the size of the relevant subgraphs. In the context of very deep neural networks, which is currently uncommon for graph neural networks, and relatively small graphs, each projection could cover big portions of the graph, thereby undermining the design principles of this architecture type. This aspect, along with the difficulty in choosing the correct number of prototypes and the induced bottleneck of prototype layers, represent the main limitations to be addressed in the future.

Conversely, memory-based architectures avoid the bottleneck problem by employing skip connections. However, these connections also pose a limitation to such methods as they enable the network to ignore the interpretable elements, thereby constraining the interpretability assurances of these designs. Striking a balance between skip connections and the induced bottlenecks remains an open problem to be tackled. Additionally, these architectures require enhancements in terms of memory footprint and computational time needed for training to compete with black-box models in resource-limited scenarios.

Probing processes and concept-based approaches rely on the availability of concept datasets~\cite{Bau2017, Mu2020, LaRosa2023Towards} or external multi-modal models~\cite{Hernandez2022, Oikarinen2023}. These dependencies introduce bias in the process that can compromise the interpretability of these methods. Probing processes may overlook factors captured by the networks that are not aligned with human knowledge. Concept-based approaches are susceptible to concept leakage and noise from factors correlated with the presence of a concept (e.g., background).

While the thesis contributes to the generalization of eXplainable Artificial Intelligence (XAI) methods across architectures and domains, it primarily focuses on classification problems. Most of the proposed techniques do not make assumptions about the problem's structure. However, it is not guaranteed that their extension to other types of problems is straightforward, thus representing an open question.

\section{Future Research Directions}
\label{sec:concl_future}
Moving forward, the insights and methodologies presented in this thesis lay a solid foundation for future research in explainable deep learning. In this section, we discuss potential future research directions and areas for further development of the proposed techniques that can mitigate the limitations of current approaches.

This thesis discussed two designs of self-explainable memory-augmented neural networks and how the memory tracking mechanism is used to compute explanations from the proposed designs. In this direction, we foresee a unified framework to apply the same concept to any memory-augmented neural network, studying the impact of different solutions and laying the foundation for this research area. Other research areas related to memory-based architectures include reduction mechanisms for the large footprint and the adaptation strategy for the training process of the black-box models. In this context, integrating the proposed techniques in a visual analytics system could open the door to a human-centered evaluation. This evaluation could assess the effectiveness and usability of the proposed methods in real-world scenarios, providing valuable insights into their practical utility and informing further improvements based on user feedback.

Regarding neural explanations, most dataset-based techniques are based on the beam search algorithm, which does not guarantee the optimality and completeness of the explanations. A research direction could examine the effectiveness of current approaches and identify the challenges connected to the search for optimality. In particular, by leveraging state-of-the-art clustering algorithms and exploring novel approaches for identifying meaningful clusters in high-dimensional activation spaces, we could improve the granularity and interpretability of explanations provided by this method. Additionally, future research could work on methods that mitigate biases and ensure the reliability and generalization of the probing process, for example, investigating the application of methods rooted in pattern recognition. 

Finally, looking at the long-term vision for the research on self-explainable deep neural networks, the \gls{XAI} field needs a unified framework that can be applied and modified easily to generate different types of explanations for several problems. Such a design could have an impact similar to that of Transformers for the deep learning field, applied nowadays to multiple different problems and domains, and be able to push research beyond its limits.

By following the research directions described above, we can continue to advance the field of explainable deep learning and develop more transparent, interpretable, and trustworthy AI systems.

\cleardoublepage
\addcontentsline{toc}{chapter}{Bibliography}
\bibliography{Include/Backmatter/Bibliography}

\begin{thebibliography}{292}
\providecommand{\natexlab}[1]{#1}
\providecommand{\url}[1]{\texttt{#1}}
\expandafter\ifx\csname urlstyle\endcsname\relax
  \providecommand{\doi}[1]{doi: #1}\else
  \providecommand{\doi}{doi: \begingroup \urlstyle{rm}\Url}\fi

\bibitem[Gra(2023)]{Grau2023}
Uncovering the hidden significance of activities location in predictive process monitoring.
\newblock In \emph{Pre-proceedings of the ML4PM 2023 Fourth International Workshop On Leveraging Machine Learning In Process Mining}, pages 1--12, November 2023.
\newblock 4th International Workshop on Leveraging Machine Learning in Process Mining, ML4PM 2023, ML4PM 2023 ; Conference date: 23-10-2023 Through 27-10-2023.

\bibitem[Abnar and Zuidema(2020)]{Abnar2020}
Samira Abnar and Willem Zuidema.
\newblock Quantifying attention flow in transformers.
\newblock In \emph{Proceedings of the 58th Annual Meeting of the Association for Computational Linguistics}. Association for Computational Linguistics, 2020.
\newblock \doi{10.18653/v1/2020.acl-main.385}.

\bibitem[Adadi and Berrada(2018)]{Adadi2018}
Amina Adadi and Mohammed Berrada.
\newblock Peeking inside the black-box: A survey on explainable artificial intelligence ({XAI}).
\newblock \emph{{IEEE} Access}, 6:\penalty0 52138--52160, 2018.
\newblock \doi{10.1109/access.2018.2870052}.

\bibitem[Adebayo et~al.(2018)Adebayo, Gilmer, Muelly, Goodfellow, Hardt, and Kim]{Adebayo2018}
Julius Adebayo, Justin Gilmer, Michael Muelly, Ian Goodfellow, Moritz Hardt, and Been Kim.
\newblock Sanity checks for saliency maps.
\newblock In S.~Bengio, H.~Wallach, H.~Larochelle, K.~Grauman, N.~Cesa-Bianchi, and R.~Garnett, editors, \emph{Advances in Neural Information Processing Systems}, volume~31. Curran Associates, Inc., 2018.

\bibitem[Adhikari et~al.(2019)Adhikari, Tax, Satta, and Faeth]{Adhikari2019}
Ajaya Adhikari, David M.~J. Tax, Riccardo Satta, and Matthias Faeth.
\newblock Leafage: Example-based and feature importance-based explanations for black-box ml models.
\newblock In \emph{2019 IEEE International Conference on Fuzzy Systems (FUZZ-IEEE)}. IEEE, June 2019.
\newblock \doi{10.1109/fuzz-ieee.2019.8858846}.

\bibitem[Albahri et~al.(2023)Albahri, Duhaim, Fadhel, Alnoor, Baqer, Alzubaidi, Albahri, Alamoodi, Bai, Salhi, Santamaría, Ouyang, Gupta, Gu, and Deveci]{Albahri2023}
A.S. Albahri, Ali~M. Duhaim, Mohammed~A. Fadhel, Alhamzah Alnoor, Noor~S. Baqer, Laith Alzubaidi, O.S. Albahri, A.H. Alamoodi, Jinshuai Bai, Asma Salhi, Jose Santamaría, Chun Ouyang, Ashish Gupta, Yuantong Gu, and Muhammet Deveci.
\newblock A systematic review of trustworthy and explainable artificial intelligence in healthcare: Assessment of quality, bias risk, and data fusion.
\newblock \emph{Information Fusion}, 96:\penalty0 156--191, August 2023.
\newblock ISSN 1566-2535.
\newblock \doi{10.1016/j.inffus.2023.03.008}.

\bibitem[Alicioglu and Sun(2021)]{Alicioglu2021}
Gulsum Alicioglu and Bo~Sun.
\newblock A survey of visual analytics for explainable artificial intelligence methods.
\newblock \emph{Computers \& Graphics}, 2021.
\newblock ISSN 0097-8493.
\newblock \doi{https://doi.org/10.1016/j.cag.2021.09.002}.

\bibitem[Alvarez~Melis and Jaakkola(2018)]{Melis2018}
David Alvarez~Melis and Tommi Jaakkola.
\newblock Towards robust interpretability with self-explaining neural networks.
\newblock In S.~Bengio, H.~Wallach, H.~Larochelle, K.~Grauman, N.~Cesa-Bianchi, and R.~Garnett, editors, \emph{Advances in Neural Information Processing Systems}, volume~31. Curran Associates, Inc., 2018.
\newblock URL \url{https://proceedings.neurips.cc/paper/2018/file/3e9f0fc9b2f89e043bc6233994dfcf76-Paper.pdf}.

\bibitem[Antverg and Belinkov(2022)]{Antverg2022}
Omer Antverg and Yonatan Belinkov.
\newblock On the pitfalls of analyzing individual neurons in language models.
\newblock In \emph{International Conference on Learning Representations}, 2022.
\newblock URL \url{https://openreview.net/forum?id=8uz0EWPQIMu}.

\bibitem[Arrieta et~al.(2020)Arrieta, D{\'{\i}}az-Rodr{\'{\i}}guez, Ser, Bennetot, Tabik, Barbado, Garcia, Gil-Lopez, Molina, Benjamins, Chatila, and Herrera]{Arrieta2020}
Alejandro~Barredo Arrieta, Natalia D{\'{\i}}az-Rodr{\'{\i}}guez, Javier~Del Ser, Adrien Bennetot, Siham Tabik, Alberto Barbado, Salvador Garcia, Sergio Gil-Lopez, Daniel Molina, Richard Benjamins, Raja Chatila, and Francisco Herrera.
\newblock Explainable artificial intelligence ({XAI}): Concepts, taxonomies, opportunities and challenges toward responsible {AI}.
\newblock \emph{Information Fusion}, 58:\penalty0 82--115, jun 2020.
\newblock \doi{10.1016/j.inffus.2019.12.012}.

\bibitem[Bach et~al.(2015)Bach, Binder, Montavon, Klauschen, Müller, and Samek]{Bach2015}
Sebastian Bach, Alexander Binder, Gr{\'{e}}goire Montavon, Frederick Klauschen, Klaus-Robert Müller, and Wojciech Samek.
\newblock On pixel-wise explanations for non-linear classifier decisions by layer-wise relevance propagation.
\newblock \emph{{PLOS} {ONE}}, 10\penalty0 (7):\penalty0 e0130140, jul 2015.
\newblock \doi{10.1371/journal.pone.0130140}.

\bibitem[Bahdanau et~al.(2015{\natexlab{a}})Bahdanau, Cho, and Bengio]{Bahdanau2015}
Dzmitry Bahdanau, {Kyung Hyun} Cho, and Yoshua Bengio.
\newblock Neural machine translation by jointly learning to align and translate.
\newblock January 2015{\natexlab{a}}.
\newblock 3rd International Conference on Learning Representations, ICLR 2015;.

\bibitem[Bahdanau et~al.(2015{\natexlab{b}})Bahdanau, Cho, and Bengio]{Bahdanau2014}
Dzmitry Bahdanau, Kyunghyun Cho, and Yoshua Bengio.
\newblock Neural machine translation by jointly learning to align and translate.
\newblock In Yoshua Bengio and Yann LeCun, editors, \emph{3rd International Conference on Learning Representations, {ICLR} 2015, San Diego, CA, USA, May 7-9, 2015, Conference Track Proceedings}, 2015{\natexlab{b}}.
\newblock URL \url{http://arxiv.org/abs/1409.0473}.

\bibitem[Bahng et~al.(2020)Bahng, Chun, Yun, Choo, and Oh]{pmlr-v119-bahng20a}
Hyojin Bahng, Sanghyuk Chun, Sangdoo Yun, Jaegul Choo, and Seong~Joon Oh.
\newblock Learning de-biased representations with biased representations.
\newblock In Hal~Daumé III and Aarti Singh, editors, \emph{Proceedings of the 37th International Conference on Machine Learning}, volume 119 of \emph{Proceedings of Machine Learning Research}, pages 528--539. PMLR, 13--18 Jul 2020.
\newblock URL \url{https://proceedings.mlr.press/v119/bahng20a.html}.

\bibitem[Bakker(2001)]{Bakker2002}
Bram Bakker.
\newblock Reinforcement learning with long short-term memory.
\newblock In T.~Dietterich, S.~Becker, and Z.~Ghahramani, editors, \emph{Advances in Neural Information Processing Systems}, volume~14. MIT Press, 2001.
\newblock URL \url{https://proceedings.neurips.cc/paper_files/paper/2001/file/a38b16173474ba8b1a95bcbc30d3b8a5-Paper.pdf}.

\bibitem[Barredo-Arrieta and Del~Ser(2020)]{BarredoArrieta2020}
Alejandro Barredo-Arrieta and Javier Del~Ser.
\newblock Plausible counterfactuals: Auditing deep learning classifiers with realistic adversarial examples.
\newblock In \emph{2020 International Joint Conference on Neural Networks (IJCNN)}. IEEE, July 2020.
\newblock \doi{10.1109/ijcnn48605.2020.9206728}.

\bibitem[Bau et~al.(2017)Bau, Zhou, Khosla, Oliva, and Torralba]{Bau2017}
David Bau, Bolei Zhou, Aditya Khosla, Aude Oliva, and Antonio Torralba.
\newblock Network dissection: Quantifying interpretability of deep visual representations.
\newblock In \emph{Proceedings of the IEEE Conference on Computer Vision and Pattern Recognition (CVPR)}, July 2017.

\bibitem[Becker et~al.(2020)Becker, Drichel, Muller, and Ertl]{Becker2020}
Franziska Becker, Arthur Drichel, Christoph Muller, and Thomas Ertl.
\newblock Interpretable visualizations of deep neural networks for domain generation algorithm detection.
\newblock In \emph{2020 {IEEE} Symposium on Visualization for Cyber Security ({VizSec})}. {IEEE}, oct 2020.
\newblock \doi{10.1109/vizsec51108.2020.00010}.

\bibitem[Beltagy et~al.(2020)Beltagy, Peters, and Cohan]{Beltagy2020}
Iz~Beltagy, Matthew~E. Peters, and Arman Cohan.
\newblock Longformer: The long-document transformer.
\newblock April 2020.
\newblock \doi{10.48550/ARXIV.2004.05150}.

\bibitem[Bickerton et~al.(2012)Bickerton, Paolini, Besnard, Muresan, and Hopkins]{Bickerton2012}
Richard Bickerton, Gaia Paolini, Jérémy Besnard, Sorel Muresan, and Andrew Hopkins.
\newblock Quantifying the chemical beauty of drugs.
\newblock \emph{Nature chemistry}, 4:\penalty0 90--8, 02 2012.
\newblock \doi{10.1038/nchem.1243}.

\bibitem[Bilal et~al.(2018)Bilal, Jourabloo, Ye, Liu, and Ren]{Bilal2018}
Alsallakh Bilal, Amin Jourabloo, Mao Ye, Xiaoming Liu, and Liu Ren.
\newblock Do convolutional neural networks learn class hierarchy?
\newblock \emph{{IEEE} Transactions on Visualization and Computer Graphics}, 24\penalty0 (1):\penalty0 152--162, jan 2018.
\newblock \doi{10.1109/tvcg.2017.2744683}.

\bibitem[Borowski et~al.(2021)Borowski, Zimmermann, Schepers, Geirhos, Wallis, Bethge, and Brendel]{Borowski2021}
Judy Borowski, Roland~Simon Zimmermann, Judith Schepers, Robert Geirhos, Thomas S.~A. Wallis, Matthias Bethge, and Wieland Brendel.
\newblock Exemplary natural images explain {\{}cnn{\}} activations better than state-of-the-art feature visualization.
\newblock In \emph{International Conference on Learning Representations}, 2021.
\newblock URL \url{https://openreview.net/forum?id=QO9-y8also-}.

\bibitem[Bricken et~al.(2023)Bricken, Templeton, Batson, Chen, Jermyn, Conerly, Turner, Anil, Denison, Askell, Lasenby, Wu, Kravec, Schiefer, Maxwell, Joseph, Hatfield-Dodds, Tamkin, Nguyen, McLean, Burke, Hume, Carter, Henighan, and Olah]{bricken2023monosemanticity}
Trenton Bricken, Adly Templeton, Joshua Batson, Brian Chen, Adam Jermyn, Tom Conerly, Nick Turner, Cem Anil, Carson Denison, Amanda Askell, Robert Lasenby, Yifan Wu, Shauna Kravec, Nicholas Schiefer, Tim Maxwell, Nicholas Joseph, Zac Hatfield-Dodds, Alex Tamkin, Karina Nguyen, Brayden McLean, Josiah~E Burke, Tristan Hume, Shan Carter, Tom Henighan, and Christopher Olah.
\newblock Towards monosemanticity: Decomposing language models with dictionary learning.
\newblock \emph{Transformer Circuits Thread}, 2023.

\bibitem[Brown et~al.(2020)Brown, Mann, Ryder, Subbiah, Kaplan, Dhariwal, Neelakantan, Shyam, Sastry, Askell, Agarwal, Herbert-Voss, Krueger, Henighan, Child, Ramesh, Ziegler, Wu, Winter, Hesse, Chen, Sigler, Litwin, Gray, Chess, Clark, Berner, McCandlish, Radford, Sutskever, and Amodei]{Brown2020}
Tom Brown, Benjamin Mann, Nick Ryder, Melanie Subbiah, Jared~D Kaplan, Prafulla Dhariwal, Arvind Neelakantan, Pranav Shyam, Girish Sastry, Amanda Askell, Sandhini Agarwal, Ariel Herbert-Voss, Gretchen Krueger, Tom Henighan, Rewon Child, Aditya Ramesh, Daniel Ziegler, Jeffrey Wu, Clemens Winter, Chris Hesse, Mark Chen, Eric Sigler, Mateusz Litwin, Scott Gray, Benjamin Chess, Jack Clark, Christopher Berner, Sam McCandlish, Alec Radford, Ilya Sutskever, and Dario Amodei.
\newblock Language models are few-shot learners.
\newblock In H.~Larochelle, M.~Ranzato, R.~Hadsell, M.F. Balcan, and H.~Lin, editors, \emph{Advances in Neural Information Processing Systems}, volume~33, pages 1877--1901. Curran Associates, Inc., 2020.
\newblock URL \url{https://proceedings.neurips.cc/paper_files/paper/2020/file/1457c0d6bfcb4967418bfb8ac142f64a-Paper.pdf}.

\bibitem[Bykov et~al.(2022)Bykov, Hedström, Nakajima, and Höhne]{Bykov2022}
Kirill Bykov, Anna Hedström, Shinichi Nakajima, and Marina M.-C. Höhne.
\newblock Noisegrad — enhancing explanations by introducing stochasticity to model weights.
\newblock \emph{Proceedings of the AAAI Conference on Artificial Intelligence}, 36\penalty0 (6):\penalty0 6132--6140, June 2022.
\newblock ISSN 2159-5399.
\newblock \doi{10.1609/aaai.v36i6.20561}.

\bibitem[Bykov et~al.(2023)Bykov, Kopf, Nakajima, Kloft, and H{\"o}hne]{bykov2023labeling}
Kirill Bykov, Laura Kopf, Shinichi Nakajima, Marius Kloft, and Marina~MC H{\"o}hne.
\newblock Labeling neural representations with inverse recognition.
\newblock In \emph{Thirty-seventh Conference on Neural Information Processing Systems}, 2023.
\newblock URL \url{https://openreview.net/forum?id=gLfgyIWiWW}.

\bibitem[Casper et~al.(2023)Casper, Rauker, Ho, and Hadfield-Menell]{Casper2023}
Stephen Casper, Tilman Rauker, Anson Ho, and Dylan Hadfield-Menell.
\newblock Sok: Toward transparent {AI}: A survey on interpreting the inner structures of deep neural networks.
\newblock In \emph{First IEEE Conference on Secure and Trustworthy Machine Learning}, 2023.
\newblock URL \url{https://openreview.net/forum?id=8C5zt-0Utdn}.

\bibitem[Chae et~al.(2017)Chae, Gao, Ramanathan, Steed, and Tourassi]{Chae2017}
Junghoon Chae, Shang Gao, Arvind Ramanathan, Chad~A. Steed, and Georgia Tourassi.
\newblock Visualization for classification in deep neural networks.
\newblock Technical report, Oak Ridge National Lab.(ORNL), Oak Ridge, TN (United States), 10 2017.
\newblock URL \url{https://www.osti.gov/biblio/1407764}.

\bibitem[Chan et~al.(2020)Chan, Bertini, Nonato, Barr, and Silva]{Chan2020}
Gromit Yeuk-Yin Chan, Enrico Bertini, Luis~Gustavo Nonato, Brian Barr, and Claudio~T. Silva.
\newblock Melody: Generating and visualizing machine learning model summary to understand data and classifiers together.
\newblock \emph{arXiv preprint arXiv:2007.10609}, July 2020.

\bibitem[Chattopadhay et~al.(2018)Chattopadhay, Sarkar, Howlader, and Balasubramanian]{Chattopadhay2018}
Aditya Chattopadhay, Anirban Sarkar, Prantik Howlader, and Vineeth~N Balasubramanian.
\newblock Grad-cam++: Generalized gradient-based visual explanations for deep convolutional networks.
\newblock In \emph{2018 IEEE Winter Conference on Applications of Computer Vision (WACV)}. IEEE, March 2018.
\newblock \doi{10.1109/wacv.2018.00097}.

\bibitem[Chawla et~al.(2020)Chawla, Hazarika, and Shen]{Chawla2020}
Piyush Chawla, Subhashis Hazarika, and Han-Wei Shen.
\newblock Token-wise sentiment decomposition for {ConvNet}: Visualizing a sentiment classifier.
\newblock \emph{Visual Informatics}, 4\penalty0 (2):\penalty0 132--141, jun 2020.
\newblock \doi{10.1016/j.visinf.2020.04.006}.

\bibitem[Chefer et~al.(2023)Chefer, Alaluf, Vinker, Wolf, and Cohen-Or]{Chefer2023}
Hila Chefer, Yuval Alaluf, Yael Vinker, Lior Wolf, and Daniel Cohen-Or.
\newblock Attend-and-excite: Attention-based semantic guidance for text-to-image diffusion models.
\newblock \emph{ACM Transactions on Graphics}, 42\penalty0 (4):\penalty0 1--10, July 2023.
\newblock ISSN 1557-7368.
\newblock \doi{10.1145/3592116}.

\bibitem[Chen et~al.(2019)Chen, Li, Tao, Barnett, Rudin, and Su]{Chen2019}
Chaofan Chen, Oscar Li, Daniel Tao, Alina Barnett, Cynthia Rudin, and Jonathan~K Su.
\newblock This looks like that: Deep learning for interpretable image recognition.
\newblock In H.~Wallach, H.~Larochelle, A.~Beygelzimer, F.~d\textquotesingle Alch\'{e}-Buc, E.~Fox, and R.~Garnett, editors, \emph{Advances in Neural Information Processing Systems}, volume~32. Curran Associates, Inc., 2019.
\newblock URL \url{https://proceedings.neurips.cc/paper/2019/file/adf7ee2dcf142b0e11888e72b43fcb75-Paper.pdf}.

\bibitem[Chen et~al.(2020{\natexlab{a}})Chen, Dong, Wang, Lu, Kaymak, and Huang]{Chen2020healthcare}
Peipei Chen, Wei Dong, Jinliang Wang, Xudong Lu, Uzay Kaymak, and Zhengxing Huang.
\newblock Interpretable clinical prediction via attention-based neural network.
\newblock \emph{BMC Medical Informatics and Decision Making}, 20\penalty0 (S3), July 2020{\natexlab{a}}.
\newblock ISSN 1472-6947.
\newblock \doi{10.1186/s12911-020-1110-7}.

\bibitem[Chen et~al.(2020{\natexlab{b}})Chen, Bei, and Rudin]{Chen2020}
Zhi Chen, Yijie Bei, and Cynthia Rudin.
\newblock Concept whitening for interpretable image recognition.
\newblock \emph{Nature Machine Intelligence}, 2\penalty0 (12):\penalty0 772--782, December 2020{\natexlab{b}}.
\newblock ISSN 2522-5839.
\newblock \doi{10.1038/s42256-020-00265-z}.

\bibitem[Cheng et~al.(2021)Cheng, Ming, and Qu]{Cheng2021}
Furui Cheng, Yao Ming, and Huamin Qu.
\newblock {DECE}: Decision explorer with counterfactual explanations for machine learning models.
\newblock \emph{{IEEE} Transactions on Visualization and Computer Graphics}, 27\penalty0 (2):\penalty0 1438--1447, feb 2021.
\newblock \doi{10.1109/tvcg.2020.3030342}.

\bibitem[Choo and Liu(2018)]{Choo2018}
Jaegul Choo and Shixia Liu.
\newblock Visual analytics for explainable deep learning.
\newblock \emph{{IEEE} Computer Graphics and Applications}, 38\penalty0 (4):\penalty0 84--92, jul 2018.
\newblock \doi{10.1109/mcg.2018.042731661}.

\bibitem[Clark et~al.(2019)Clark, Khandelwal, Levy, and Manning]{Clark2019}
Kevin Clark, Urvashi Khandelwal, Omer Levy, and Christopher~D. Manning.
\newblock What does bert look at? an analysis of bert’s attention.
\newblock In \emph{Proceedings of the 2019 ACL Workshop BlackboxNLP: Analyzing and Interpreting Neural Networks for NLP}. Association for Computational Linguistics, 2019.
\newblock \doi{10.18653/v1/w19-4828}.

\bibitem[Cover and Hart(1967)]{Cover1967}
T.~Cover and P.~Hart.
\newblock Nearest neighbor pattern classification.
\newblock \emph{{IEEE} Transactions on Information Theory}, 13\penalty0 (1):\penalty0 21--27, jan 1967.
\newblock \doi{10.1109/tit.1967.1053964}.

\bibitem[Dai and Wang(2021)]{Dai2021}
Enyan Dai and Suhang Wang.
\newblock Towards self-explainable graph neural network.
\newblock In \emph{Proceedings of the 30th ACM International Conference on Information \&; Knowledge Management}, CIKM '21, page 302–311, New York, NY, USA, 2021. Association for Computing Machinery.
\newblock ISBN 9781450384469.
\newblock \doi{10.1145/3459637.3482306}.
\newblock URL \url{https://doi.org/10.1145/3459637.3482306}.

\bibitem[Dalvi et~al.(2019)Dalvi, Durrani, Sajjad, Belinkov, Bau, and Glass]{Dalvi2019}
Fahim Dalvi, Nadir Durrani, Hassan Sajjad, Yonatan Belinkov, Anthony Bau, and James Glass.
\newblock What is one grain of sand in the desert? analyzing individual neurons in deep {NLP} models.
\newblock \emph{Proceedings of the {AAAI} Conference on Artificial Intelligence}, 33\penalty0 (01):\penalty0 6309--6317, jul 2019.
\newblock \doi{10.1609/aaai.v33i01.33016309}.

\bibitem[Darlow et~al.(2018)Darlow, Crowley, Antoniou, and Storkey]{Darlow2018}
Luke~N. Darlow, Elliot~J Crowley, Antreas Antoniou, and Amos~J. Storkey.
\newblock Cinic-10 is not imagenet or cifar-10.
\newblock Technical report, University of Edinburgh, 2018.
\newblock [dataset].

\bibitem[Das et~al.(2020)Das, Park, Wang, Hohman, Firstman, Rogers, and Chau]{Das2020}
Nilaksh Das, Haekyu Park, Zijie~J. Wang, Fred Hohman, Robert Firstman, Emily Rogers, and Duen Horng~Polo Chau.
\newblock Bluff: Interactively deciphering adversarial attacks on deep neural networks.
\newblock In \emph{2020 {IEEE} Visualization Conference ({VIS})}. {IEEE}, oct 2020.
\newblock \doi{10.1109/vis47514.2020.00061}.

\bibitem[Debnath et~al.(1991)Debnath, Lopez~de Compadre, Debnath, Shusterman, and Hansch]{Debnath1991}
Asim~Kumar Debnath, Rosa~L. Lopez~de Compadre, Gargi Debnath, Alan~J. Shusterman, and Corwin Hansch.
\newblock Structure-activity relationship of mutagenic aromatic and heteroaromatic nitro compounds. correlation with molecular orbital energies and hydrophobicity.
\newblock \emph{Journal of Medicinal Chemistry}, 34\penalty0 (2):\penalty0 786--797, Feb 1991.
\newblock ISSN 0022-2623.
\newblock \doi{10.1021/jm00106a046}.
\newblock URL \url{https://doi.org/10.1021/jm00106a046}.

\bibitem[DeRose et~al.(2021)DeRose, Wang, and Berger]{DeRose2021}
Joseph~F. DeRose, Jiayao Wang, and Matthew Berger.
\newblock Attention flows: Analyzing and comparing attention mechanisms in language models.
\newblock \emph{{IEEE} Transactions on Visualization and Computer Graphics}, 27\penalty0 (2):\penalty0 1160--1170, feb 2021.
\newblock \doi{10.1109/tvcg.2020.3028976}.

\bibitem[Dhurandhar et~al.(2018)Dhurandhar, Chen, Luss, Tu, Ting, Shanmugam, and Das]{Dhurandhar2018}
Amit Dhurandhar, Pin-Yu Chen, Ronny Luss, Chun-Chen Tu, Paishun Ting, Karthikeyan Shanmugam, and Payel Das.
\newblock Explanations based on the missing: Towards contrastive explanations with pertinent negatives.
\newblock In S.~Bengio, H.~Wallach, H.~Larochelle, K.~Grauman, N.~Cesa-Bianchi, and R.~Garnett, editors, \emph{Advances in Neural Information Processing Systems}, volume~31. Curran Associates, Inc., 2018.
\newblock URL \url{https://proceedings.neurips.cc/paper_files/paper/2018/file/c5ff2543b53f4cc0ad3819a36752467b-Paper.pdf}.

\bibitem[Dong et~al.(2020)Dong, Wu, Song, and Zhang]{Dong2020}
Zhihang Dong, Tongshuang Wu, Sicheng Song, and Mingrui Zhang.
\newblock Interactive attention model explorer for natural language processing tasks with unbalanced data sizes.
\newblock In \emph{2020 {IEEE} Pacific Visualization Symposium ({PacificVis})}. {IEEE}, jun 2020.
\newblock \doi{10.1109/pacificvis48177.2020.1031}.

\bibitem[Doshi-Velez and Kim(2017)]{DoshiVelez2017}
Finale Doshi-Velez and Been Kim.
\newblock Towards a rigorous science of interpretable machine learning.
\newblock February 2017.
\newblock \doi{10.48550/ARXIV.1702.08608}.

\bibitem[Dosovitskiy et~al.(2021)Dosovitskiy, Beyer, Kolesnikov, Weissenborn, Zhai, Unterthiner, Dehghani, Minderer, Heigold, Gelly, Uszkoreit, and Houlsby]{Dosovitskiy2021}
Alexey Dosovitskiy, Lucas Beyer, Alexander Kolesnikov, Dirk Weissenborn, Xiaohua Zhai, Thomas Unterthiner, Mostafa Dehghani, Matthias Minderer, Georg Heigold, Sylvain Gelly, Jakob Uszkoreit, and Neil Houlsby.
\newblock An image is worth 16x16 words: Transformers for image recognition at scale.
\newblock In \emph{International Conference on Learning Representations}, 2021.

\bibitem[Duchi et~al.(2010)Duchi, Hazan, and Singer]{Duchi2010}
John~C. Duchi, Elad Hazan, and Yoram Singer.
\newblock Adaptive subgradient methods for online learning and stochastic optimization.
\newblock In Adam~Tauman Kalai and Mehryar Mohri, editors, \emph{{COLT} 2010 - The 23rd Conference on Learning Theory, Haifa, Israel, June 27-29, 2010}, pages 257--269. Omnipress, 2010.

\bibitem[Durrani et~al.(2020)Durrani, Sajjad, Dalvi, and Belinkov]{Durrani2020}
Nadir Durrani, Hassan Sajjad, Fahim Dalvi, and Yonatan Belinkov.
\newblock Analyzing individual neurons in pre-trained language models.
\newblock In \emph{Proceedings of the 2020 Conference on Empirical Methods in Natural Language Processing ({EMNLP})}. Association for Computational Linguistics, 2020.
\newblock \doi{10.18653/v1/2020.emnlp-main.395}.

\bibitem[Elhage et~al.(2022)Elhage, Hume, Olsson, Schiefer, Henighan, Kravec, Hatfield-Dodds, Lasenby, Drain, Chen, Grosse, McCandlish, Kaplan, Amodei, Wattenberg, and Olah]{elhage2022superposition}
Nelson Elhage, Tristan Hume, Catherine Olsson, Nicholas Schiefer, Tom Henighan, Shauna Kravec, Zac Hatfield-Dodds, Robert Lasenby, Dawn Drain, Carol Chen, Roger Grosse, Sam McCandlish, Jared Kaplan, Dario Amodei, Martin Wattenberg, and Christopher Olah.
\newblock Toy models of superposition.
\newblock \emph{Transformer Circuits Thread}, 2022.

\bibitem[Endert et~al.(2012)Endert, Fiaux, and North]{Endert2012}
Alex Endert, Patrick Fiaux, and Chris North.
\newblock Semantic interaction for sensemaking: Inferring analytical reasoning for model steering.
\newblock \emph{IEEE Transactions on Visualization and Computer Graphics}, 18\penalty0 (12):\penalty0 2879--2888, 2012.
\newblock \doi{10.1109/TVCG.2012.260}.

\bibitem[Erhan et~al.(2009)Erhan, Bengio, Courville, and Vincent]{Erhan2009}
Dumitru Erhan, Yoshua Bengio, Aaron Courville, and Pascal Vincent.
\newblock Visualizing higher-layer features of a deep network.
\newblock \emph{University of Montreal}, 1341\penalty0 (3):\penalty0 1, 2009.

\bibitem[Espinosa~Zarlenga et~al.(2022)Espinosa~Zarlenga, Barbiero, Ciravegna, Marra, Giannini, Diligenti, Shams, Precioso, Melacci, Weller, Li\'{o}, and Jamnik]{Zarlenga2022}
Mateo Espinosa~Zarlenga, Pietro Barbiero, Gabriele Ciravegna, Giuseppe Marra, Francesco Giannini, Michelangelo Diligenti, Zohreh Shams, Frederic Precioso, Stefano Melacci, Adrian Weller, Pietro Li\'{o}, and Mateja Jamnik.
\newblock Concept embedding models: Beyond the accuracy-explainability trade-off.
\newblock In S.~Koyejo, S.~Mohamed, A.~Agarwal, D.~Belgrave, K.~Cho, and A.~Oh, editors, \emph{Advances in Neural Information Processing Systems}, volume~35, pages 21400--21413. Curran Associates, Inc., 2022.
\newblock URL \url{https://proceedings.neurips.cc/paper_files/paper/2022/file/867c06823281e506e8059f5c13a57f75-Paper-Conference.pdf}.

\bibitem[Everingham et~al.(2009)Everingham, Gool, Williams, Winn, and Zisserman]{Everingham2009}
Mark Everingham, Luc~Van Gool, Christopher K.~I. Williams, John Winn, and Andrew Zisserman.
\newblock The pascal visual object classes ({VOC}) challenge.
\newblock \emph{International Journal of Computer Vision}, 88\penalty0 (2):\penalty0 303--338, sep 2009.
\newblock \doi{10.1007/s11263-009-0275-4}.

\bibitem[Fang et~al.(2020)Fang, Kuang, Lin, Wu, and Yao]{Fang2020}
Zhengqing Fang, Kun Kuang, Yuxiao Lin, Fei Wu, and Yu-Feng Yao.
\newblock Concept-based explanation for fine-grained images and its application in infectious keratitis classification.
\newblock In \emph{Proceedings of the 28th {ACM} International Conference on Multimedia}. {ACM}, October 2020.
\newblock \doi{10.1145/3394171.3413557}.

\bibitem[FEL et~al.(2023)FEL, Boissin, Boutin, Picard, Novello, Colin, Linsley, ROUSSEAU, Cadene, Goetschalckx, Gardes, and Serre]{Fel2023}
Thomas FEL, Thibaut Boissin, Victor Boutin, Agustin~Martin Picard, Paul Novello, Julien Colin, Drew Linsley, Tom ROUSSEAU, Remi Cadene, Lore Goetschalckx, Laurent Gardes, and Thomas Serre.
\newblock Unlocking feature visualization for deep network with {MA}gnitude constrained optimization.
\newblock In \emph{Thirty-seventh Conference on Neural Information Processing Systems}, 2023.
\newblock URL \url{https://openreview.net/forum?id=J7VoDuzuKs}.

\bibitem[Fong and Vedaldi(2017)]{Fong2017}
Ruth~C. Fong and Andrea Vedaldi.
\newblock Interpretable explanations of black boxes by meaningful perturbation.
\newblock In \emph{Proceedings of the IEEE International Conference on Computer Vision (ICCV)}, Oct 2017.

\bibitem[Franke et~al.(2018)Franke, Niehues, and Waibel]{Franke2018}
Jörg Franke, Jan Niehues, and Alex Waibel.
\newblock Robust and scalable differentiable neural computer for question answering.
\newblock In \emph{Proceedings of the Workshop on Machine Reading for Question Answering}. Association for Computational Linguistics, 2018.
\newblock \doi{10.18653/v1/w18-2606}.

\bibitem[Fu et~al.(2017)Fu, Zheng, and Mei]{Fu2017}
Jianlong Fu, Heliang Zheng, and Tao Mei.
\newblock Look closer to see better: Recurrent attention convolutional neural network for fine-grained image recognition.
\newblock In \emph{2017 IEEE Conference on Computer Vision and Pattern Recognition (CVPR)}. IEEE, July 2017.
\newblock \doi{10.1109/cvpr.2017.476}.

\bibitem[Gao and Ji(2019)]{Gao2019}
Hongyang Gao and Shuiwang Ji.
\newblock Graph u-nets.
\newblock In Kamalika Chaudhuri and Ruslan Salakhutdinov, editors, \emph{Proceedings of the 36th International Conference on Machine Learning}, volume~97 of \emph{Proceedings of Machine Learning Research}, pages 2083--2092. PMLR, 09--15 Jun 2019.
\newblock URL \url{https://proceedings.mlr.press/v97/gao19a.html}.

\bibitem[Ghorbani et~al.(2019)Ghorbani, Wexler, Zou, and Kim]{ghorbani2019towards}
Amirata Ghorbani, James Wexler, James~Y Zou, and Been Kim.
\newblock Towards automatic concept-based explanations.
\newblock In \emph{Advances in Neural Information Processing Systems}, pages 9273--9282, 2019.

\bibitem[Giles et~al.(1998)Giles, Bollacker, and Lawrence]{Giles1998}
C.~Lee Giles, Kurt~D. Bollacker, and Steve Lawrence.
\newblock Citeseer: an automatic citation indexing system.
\newblock In \emph{Proceedings of the third ACM conference on Digital libraries - DL ’98}, DL ’98. ACM Press, 1998.
\newblock \doi{10.1145/276675.276685}.

\bibitem[Gilmer et~al.(2017)Gilmer, Schoenholz, Riley, Vinyals, and Dahl]{Gilmer17}
Justin Gilmer, Samuel~S. Schoenholz, Patrick~F. Riley, Oriol Vinyals, and George~E. Dahl.
\newblock Neural message passing for quantum chemistry.
\newblock In Doina Precup and Yee~Whye Teh, editors, \emph{Proceedings of the 34th International Conference on Machine Learning}, volume~70 of \emph{Proceedings of Machine Learning Research}, pages 1263--1272. PMLR, 06--11 Aug 2017.
\newblock URL \url{https://proceedings.mlr.press/v70/gilmer17a.html}.

\bibitem[Gilpin et~al.(2018)Gilpin, Bau, Yuan, Bajwa, Specter, and Kagal]{Gilpin2018}
Leilani~H. Gilpin, David Bau, Ben~Z. Yuan, Ayesha Bajwa, Michael Specter, and Lalana Kagal.
\newblock Explaining explanations: An overview of interpretability of machine learning.
\newblock In \emph{2018 IEEE 5th International Conference on Data Science and Advanced Analytics (DSAA)}. IEEE, October 2018.
\newblock \doi{10.1109/dsaa.2018.00018}.

\bibitem[Goyal et~al.(2019)Goyal, Wu, Ernst, Batra, Parikh, and Lee]{goyal19}
Yash Goyal, Ziyan Wu, Jan Ernst, Dhruv Batra, Devi Parikh, and Stefan Lee.
\newblock Counterfactual visual explanations.
\newblock In Kamalika Chaudhuri and Ruslan Salakhutdinov, editors, \emph{Proceedings of the 36th International Conference on Machine Learning}, volume~97 of \emph{Proceedings of Machine Learning Research}, pages 2376--2384. PMLR, 09--15 Jun 2019.
\newblock URL \url{https://proceedings.mlr.press/v97/goyal19a.html}.

\bibitem[Grau et~al.(2024)Grau, Nápoles, Bello, Salgueiro, and Jastrzebska]{Grau2024}
Isel Grau, Gonzalo Nápoles, Marilyn Bello, Yamisleydi Salgueiro, and Agnieszka Jastrzebska.
\newblock Forward composition propagation for explainable neural reasoning.
\newblock \emph{IEEE Computational Intelligence Magazine}, 19\penalty0 (1):\penalty0 26--35, February 2024.
\newblock ISSN 1556-6048.
\newblock \doi{10.1109/mci.2023.3327834}.

\bibitem[Graves et~al.(2013)Graves, Mohamed, and Hinton]{Graves2013}
Alex Graves, Abdel-rahman Mohamed, and Geoffrey Hinton.
\newblock Speech recognition with deep recurrent neural networks.
\newblock In \emph{2013 IEEE International Conference on Acoustics, Speech and Signal Processing}. IEEE, May 2013.
\newblock \doi{10.1109/icassp.2013.6638947}.

\bibitem[Graves et~al.(2016)Graves, Wayne, Reynolds, Harley, Danihelka, Grabska-Barwi{\'{n}}ska, Colmenarejo, Grefenstette, Ramalho, Agapiou, Badia, Hermann, Zwols, Ostrovski, Cain, King, Summerfield, Blunsom, Kavukcuoglu, and Hassabis]{Graves2016}
Alex Graves, Greg Wayne, Malcolm Reynolds, Tim Harley, Ivo Danihelka, Agnieszka Grabska-Barwi{\'{n}}ska, Sergio~G{\'{o}}mez Colmenarejo, Edward Grefenstette, Tiago Ramalho, John Agapiou, Adri{\`{a}}~Puigdom{\`{e}}nech Badia, Karl~Moritz Hermann, Yori Zwols, Georg Ostrovski, Adam Cain, Helen King, Christopher Summerfield, Phil Blunsom, Koray Kavukcuoglu, and Demis Hassabis.
\newblock Hybrid computing using a neural network with dynamic external memory.
\newblock \emph{Nature}, 538\penalty0 (7626):\penalty0 471--476, oct 2016.
\newblock \doi{10.1038/nature20101}.

\bibitem[Gui et~al.(2023)Gui, Yuan, Wang, Lao, Li, and Ji]{Gui2023}
Shurui Gui, Hao Yuan, Jie Wang, Qicheng Lao, Kang Li, and Shuiwang Ji.
\newblock Flowx: Towards explainable graph neural networks via message flows.
\newblock \emph{IEEE Transactions on Pattern Analysis and Machine Intelligence}, pages 1--12, 2023.
\newblock ISSN 1939-3539.
\newblock \doi{10.1109/tpami.2023.3347470}.

\bibitem[Guidotti(2022)]{Guidotti2022}
Riccardo Guidotti.
\newblock Counterfactual explanations and how to find them: literature review and benchmarking.
\newblock \emph{Data Mining and Knowledge Discovery}, April 2022.
\newblock ISSN 1573-756X.
\newblock \doi{10.1007/s10618-022-00831-6}.

\bibitem[Gunning and Aha(2019)]{Gunning2019}
David Gunning and David~W. Aha.
\newblock Darpa’s explainable artificial intelligence program.
\newblock \emph{AI Magazine}, 40\penalty0 (2):\penalty0 44--58, June 2019.
\newblock ISSN 2371-9621.
\newblock \doi{10.1609/aimag.v40i2.2850}.

\bibitem[Gurumoorthy et~al.(2019)Gurumoorthy, Dhurandhar, Cecchi, and Aggarwal]{Gurumoorthy2019}
Karthik~S. Gurumoorthy, Amit Dhurandhar, Guillermo Cecchi, and Charu Aggarwal.
\newblock Efficient data representation by selecting prototypes with importance weights.
\newblock In \emph{2019 {IEEE} International Conference on Data Mining ({ICDM})}. {IEEE}, nov 2019.
\newblock \doi{10.1109/icdm.2019.00036}.

\bibitem[Han et~al.(2021)Han, Shen, Li, Lu, Shan, and Wang]{Han2021}
Xiaoyang Han, Han-Wei Shen, Guan Li, Xuyi Lu, Guihua Shan, and Yangang Wang.
\newblock {IVDAS}: an interactive visual design and analysis system for image data symmetry detection of {CNN} models.
\newblock \emph{Journal of Visualization}, 24\penalty0 (3):\penalty0 615--629, jan 2021.
\newblock \doi{10.1007/s12650-020-00721-3}.

\bibitem[Harth(2022)]{Harth2022}
Rafael Harth.
\newblock Understanding individual neurons of~{ResNet} through improved compositional formulas.
\newblock In \emph{Pattern Recognition and Artificial Intelligence}, pages 283--294. Springer International Publishing, 2022.
\newblock \doi{10.1007/978-3-031-09282-4_24}.

\bibitem[Hase et~al.(2019)Hase, Chen, Li, and Rudin]{Hase2019}
Peter Hase, Chaofan Chen, Oscar Li, and Cynthia Rudin.
\newblock Interpretable image recognition with hierarchical prototypes.
\newblock \emph{Proceedings of the AAAI Conference on Human Computation and Crowdsourcing}, 7:\penalty0 32--40, October 2019.
\newblock ISSN 2769-1330.
\newblock \doi{10.1609/hcomp.v7i1.5265}.

\bibitem[Havasi et~al.(2022)Havasi, Parbhoo, and Doshi-Velez]{havasi2022addressing}
Marton Havasi, Sonali Parbhoo, and Finale Doshi-Velez.
\newblock Addressing leakage in concept bottleneck models.
\newblock In Alice~H. Oh, Alekh Agarwal, Danielle Belgrave, and Kyunghyun Cho, editors, \emph{Advances in Neural Information Processing Systems}, 2022.
\newblock URL \url{https://openreview.net/forum?id=tglniD_fn9}.

\bibitem[Hazarika et~al.(2019)Hazarika, Li, Wang, Shen, and Chou]{Hazarika2019}
Subhashis Hazarika, Haoyu Li, Ko-Chih Wang, Han-Wei Shen, and Ching-Shan Chou.
\newblock {NNVA}: Neural network assisted visual analysis of yeast cell polarization simulation.
\newblock \emph{{IEEE} Transactions on Visualization and Computer Graphics}, pages 1--1, 2019.
\newblock \doi{10.1109/tvcg.2019.2934591}.

\bibitem[He et~al.(2016)He, Zhang, Ren, and Sun]{He2016}
Kaiming He, Xiangyu Zhang, Shaoqing Ren, and Jian Sun.
\newblock Deep residual learning for image recognition.
\newblock In \emph{2016 {IEEE} Conference on Computer Vision and Pattern Recognition ({CVPR})}, pages 770--778. {IEEE}, 2016.

\bibitem[He et~al.(2020)He, Lee, van Baar, Wittenburg, and Shen]{He2020}
Wenbin He, Teng-Yok Lee, Jeroen van Baar, Kent Wittenburg, and Han-Wei Shen.
\newblock {DynamicsExplorer}: Visual analytics for robot control tasks involving dynamics and {LSTM}-based control policies.
\newblock In \emph{2020 {IEEE} Pacific Visualization Symposium ({PacificVis})}. {IEEE}, jun 2020.
\newblock \doi{10.1109/pacificvis48177.2020.7127}.

\bibitem[Hearst et~al.(1998)Hearst, Dumais, Osuna, Platt, and Scholkopf]{Hearst1998}
M.A. Hearst, S.T. Dumais, E.~Osuna, J.~Platt, and B.~Scholkopf.
\newblock Support vector machines.
\newblock \emph{IEEE Intelligent Systems and their Applications}, 13\penalty0 (4):\penalty0 18--28, July 1998.
\newblock ISSN 1094-7167.
\newblock \doi{10.1109/5254.708428}.

\bibitem[Hennigen et~al.(2020)Hennigen, Williams, and Cotterell]{Hennigen2020}
Lucas~Torroba Hennigen, Adina Williams, and Ryan Cotterell.
\newblock Intrinsic probing through dimension selection.
\newblock In \emph{Proceedings of the 2020 Conference on Empirical Methods in Natural Language Processing ({EMNLP})}. Association for Computational Linguistics, 2020.
\newblock \doi{10.18653/v1/2020.emnlp-main.15}.

\bibitem[Hernandez et~al.(2022)Hernandez, Schwettmann, Bau, Bagashvili, Torralba, and Andreas]{Hernandez2022}
Evan Hernandez, Sarah Schwettmann, David Bau, Teona Bagashvili, Antonio Torralba, and Jacob Andreas.
\newblock Natural language descriptions of deep features.
\newblock In \emph{International Conference on Learning Representations}, 2022.
\newblock URL \url{https://openreview.net/forum?id=NudBMY-tzDr}.

\bibitem[Hilton et~al.(2020)Hilton, Cammarata, Carter, Goh, and Olah]{Hilton2020}
Jacob Hilton, Nick Cammarata, Shan Carter, Gabriel Goh, and Chris Olah.
\newblock Understanding {RL} vision.
\newblock \emph{Distill}, 5\penalty0 (11), nov 2020.
\newblock \doi{10.23915/distill.00029}.

\bibitem[Hochreiter and Schmidhuber(1997)]{Hochreiter1997}
Sepp Hochreiter and Jürgen Schmidhuber.
\newblock Long short-term memory.
\newblock \emph{Neural Computation}, 9\penalty0 (8):\penalty0 1735--1780, November 1997.
\newblock ISSN 1530-888X.
\newblock \doi{10.1162/neco.1997.9.8.1735}.

\bibitem[Hogr\"{a}fer et~al.(2022)Hogr\"{a}fer, Angelini, Santucci, and Schulz]{Hografer2022}
Marius Hogr\"{a}fer, Marco Angelini, Giuseppe Santucci, and Hans-J\"{o}rg Schulz.
\newblock Steering-by-example for progressive visual analytics.
\newblock \emph{ACM Trans. Intell. Syst. Technol.}, 13\penalty0 (6), 9 2022.
\newblock ISSN 2157-6904.
\newblock \doi{10.1145/3531229}.
\newblock URL \url{https://doi.org/10.1145/3531229}.

\bibitem[Hohman et~al.(2017)Hohman, Hodas, and Chau]{Hohman2017}
Fred Hohman, Nathan Hodas, and Duen~Horng Chau.
\newblock {ShapeShop}.
\newblock In \emph{Proceedings of the 2017 {CHI} Conference Extended Abstracts on Human Factors in Computing Systems}. {ACM}, may 2017.
\newblock \doi{10.1145/3027063.3053103}.

\bibitem[Hohman et~al.(2019)Hohman, Kahng, Pienta, and Chau]{Hohman2019}
Fred Hohman, Minsuk Kahng, Robert Pienta, and Duen~Horng Chau.
\newblock Visual analytics in deep learning: An interrogative survey for the next frontiers.
\newblock \emph{{IEEE} Transactions on Visualization and Computer Graphics}, 25\penalty0 (8):\penalty0 2674--2693, aug 2019.
\newblock \doi{10.1109/tvcg.2018.2843369}.

\bibitem[Hohman et~al.(2020)Hohman, Park, Robinson, and Chau]{Hohman2020}
Fred Hohman, Haekyu Park, Caleb Robinson, and Duen Horng~Polo Chau.
\newblock Summit: Scaling deep learning interpretability by visualizing activation and attribution summarizations.
\newblock \emph{{IEEE} Transactions on Visualization and Computer Graphics}, 26\penalty0 (1):\penalty0 1096--1106, jan 2020.
\newblock \doi{10.1109/tvcg.2019.2934659}.

\bibitem[Hoover et~al.(2020)Hoover, Strobelt, and Gehrmann]{Hoover2020}
Benjamin Hoover, Hendrik Strobelt, and Sebastian Gehrmann.
\newblock {exBERT}: A visual analysis tool to explore learned representations in transformer models.
\newblock In \emph{Proceedings of the 58th Annual Meeting of the Association for Computational Linguistics: System Demonstrations}. Association for Computational Linguistics, 2020.
\newblock \doi{10.18653/v1/2020.acl-demos.22}.

\bibitem[Huang et~al.(2017)Huang, Liu, Maaten, and Weinberger]{Huang2017}
Gao Huang, Zhuang Liu, Laurens Van~Der Maaten, and Kilian~Q. Weinberger.
\newblock Densely connected convolutional networks.
\newblock In \emph{2017 {IEEE} Conference on Computer Vision and Pattern Recognition ({CVPR})}, pages 4700--4708. {IEEE}, 2017.

\bibitem[Huang et~al.(2018)Huang, Liu, Lang, Yu, Wang, and Li]{Huang2018}
Lei Huang, Xianglong Liu, Bo~Lang, Adams~Wei Yu, Yongliang Wang, and Bo~Li.
\newblock Orthogonal weight normalization: Solution to optimization over multiple dependent stiefel manifolds in deep neural networks.
\newblock In \emph{AAAI}, pages 3271--3278, 2018.

\bibitem[Huang et~al.(2021)Huang, Jamonnak, Zhao, Wu, and Xu]{Huang2021}
Xinyi Huang, Suphanut Jamonnak, Ye~Zhao, Tsung~Heng Wu, and Wei Xu.
\newblock A visual designer of layer-wise relevance propagation models.
\newblock \emph{Computer Graphics Forum}, 40\penalty0 (3):\penalty0 227--238, jun 2021.
\newblock \doi{10.1111/cgf.14302}.

\bibitem[Jaunet et~al.(2020)Jaunet, Vuillemot, and Wolf]{Jaunet2020}
T.~Jaunet, R.~Vuillemot, and C.~Wolf.
\newblock Drlviz: Understanding decisions and memory in deep reinforcement learning.
\newblock \emph{Computer Graphics Forum}, 39\penalty0 (3):\penalty0 49--61, 2020.
\newblock \doi{https://doi.org/10.1111/cgf.13962}.

\bibitem[Jaunet et~al.(2022)Jaunet, Kervadec, Vuillemot, Antipov, Baccouche, and Wolf]{Jaunet2021}
Théo Jaunet, Corentin Kervadec, Romain Vuillemot, Grigory Antipov, Moez Baccouche, and Christian Wolf.
\newblock Visqa: X-raying vision and language reasoning in transformers.
\newblock \emph{IEEE Transactions on Visualization and Computer Graphics}, 28\penalty0 (1):\penalty0 976--986, 2022.
\newblock \doi{10.1109/TVCG.2021.3114683}.

\bibitem[Ji et~al.(2021)Ji, Tu, He, Wang, Shen, and Yen]{Ji2021}
Xiaonan Ji, Yamei Tu, Wenbin He, Junpeng Wang, Han-Wei Shen, and Po-Yin Yen.
\newblock {USEVis}: Visual analytics of attention-based neural embedding in information retrieval.
\newblock \emph{Visual Informatics}, 5\penalty0 (2):\penalty0 1--12, jun 2021.
\newblock \doi{10.1016/j.visinf.2021.03.003}.

\bibitem[Jia et~al.(2019)Jia, Lin, Li, Zhang, and Liu]{Jia2019}
Shichao Jia, Peiwen Lin, Zeyu Li, Jiawan Zhang, and Shixia Liu.
\newblock Visualizing surrogate decision trees of convolutional neural networks.
\newblock \emph{Journal of Visualization}, 23\penalty0 (1):\penalty0 141--156, nov 2019.
\newblock \doi{10.1007/s12650-019-00607-z}.

\bibitem[Jin et~al.(2023)Jin, Li, and Hamarneh]{Jin2023}
Weina Jin, Xiaoxiao Li, and Ghassan Hamarneh.
\newblock The xai alignment problem: Rethinking how should we evaluate human-centered ai explainability techniques.
\newblock March 2023.
\newblock \doi{10.48550/ARXIV.2303.17707}.

\bibitem[Jin et~al.(2022)Jin, Wang, Wang, Ming, Ma, and Qu]{Jin2020va}
Zhihua Jin, Yong Wang, Qianwen Wang, Yao Ming, Tengfei Ma, and Huamin Qu.
\newblock Gnnlens: A visual analytics approach for prediction error diagnosis of graph neural networks.
\newblock \emph{IEEE Transactions on Visualization and Computer Graphics}, pages 1--1, 2022.
\newblock \doi{10.1109/TVCG.2022.3148107}.

\bibitem[Jung et~al.(2022)Jung, Kang, Kim, Won, and Lee]{Jung2022}
Hong-Gyu Jung, Sin-Han Kang, Hee-Dong Kim, Dong-Ok Won, and Seong-Whan Lee.
\newblock Counterfactual explanation based on gradual construction for deep networks.
\newblock \emph{Pattern Recognition}, 132:\penalty0 108958, December 2022.
\newblock ISSN 0031-3203.
\newblock \doi{10.1016/j.patcog.2022.108958}.

\bibitem[Kahng et~al.(2018)Kahng, Andrews, Kalro, and Chau]{Kahng2018}
Minsuk Kahng, Pierre~Y. Andrews, Aditya Kalro, and Duen~Horng Chau.
\newblock {ActiVis}: Visual exploration of industry-scale deep neural network models.
\newblock \emph{IEEE transactions on visualization and computer graphics}, 24\penalty0 (1):\penalty0 88--97, jan 2018.
\newblock \doi{10.1109/tvcg.2017.2744718}.

\bibitem[Kapishnikov et~al.(2021)Kapishnikov, Venugopalan, Avci, Wedin, Terry, and Bolukbasi]{Kapishnikov2021}
Andrei Kapishnikov, Subhashini Venugopalan, Besim Avci, Ben Wedin, Michael Terry, and Tolga Bolukbasi.
\newblock Guided integrated gradients: an adaptive path method for removing noise.
\newblock In \emph{2021 IEEE/CVF Conference on Computer Vision and Pattern Recognition (CVPR)}. IEEE, June 2021.
\newblock \doi{10.1109/cvpr46437.2021.00501}.

\bibitem[Keim et~al.()Keim, Andrienko, Fekete, Görg, Kohlhammer, and Melançon]{Keim2008}
Daniel Keim, Gennady Andrienko, Jean-Daniel Fekete, Carsten Görg, Jörn Kohlhammer, and Guy Melançon.
\newblock \emph{Visual Analytics: Definition, Process, and Challenges}, pages 154--175.
\newblock Springer Berlin Heidelberg.
\newblock ISBN 9783540709565.
\newblock \doi{10.1007/978-3-540-70956-5_7}.

\bibitem[Kenny and Keane(2019)]{Kenny2019}
Eoin~M. Kenny and Mark~T. Keane.
\newblock Twin-systems to explain artificial neural networks using case-based reasoning: Comparative tests of feature-weighting methods in {ANN}-{CBR} twins for {XAI}.
\newblock In \emph{Proceedings of the Twenty-Eighth International Joint Conference on Artificial Intelligence}, pages 2708--2715. International Joint Conferences on Artificial Intelligence Organization, 2019.

\bibitem[Kenny and Keane(2021{\natexlab{a}})]{Kenny2021}
Eoin~M. Kenny and Mark~T. Keane.
\newblock Explaining deep learning using examples: Optimal feature weighting methods for twin systems using post-hoc, explanation-by-example in {XAI}.
\newblock \emph{Knowledge-Based Systems}, 233:\penalty0 107530, dec 2021{\natexlab{a}}.
\newblock \doi{10.1016/j.knosys.2021.107530}.

\bibitem[Kenny and Keane(2021{\natexlab{b}})]{Kenny2021counter}
Eoin~M. Kenny and Mark~T Keane.
\newblock On generating plausible counterfactual and semi-factual explanations for deep learning.
\newblock \emph{Proceedings of the AAAI Conference on Artificial Intelligence}, 35\penalty0 (13):\penalty0 11575--11585, May 2021{\natexlab{b}}.
\newblock ISSN 2159-5399.
\newblock \doi{10.1609/aaai.v35i13.17377}.

\bibitem[Kenny et~al.(2023{\natexlab{a}})Kenny, Delaney, and Keane]{Kenny2023features}
Eoin~M. Kenny, Eoin Delaney, and Mark~T. Keane.
\newblock Advancing post-hoc case-based explanation with feature highlighting.
\newblock In \emph{Proceedings of the Thirty-Second International Joint Conference on Artificial Intelligence}, IJCAI-2023. International Joint Conferences on Artificial Intelligence Organization, August 2023{\natexlab{a}}.
\newblock \doi{10.24963/ijcai.2023/48}.

\bibitem[Kenny et~al.(2023{\natexlab{b}})Kenny, Tucker, and Shah]{Kenny2023}
Eoin~M. Kenny, Mycal Tucker, and Julie Shah.
\newblock Towards interpretable deep reinforcement learning with human-friendly prototypes.
\newblock In \emph{The Eleventh International Conference on Learning Representations}, 2023{\natexlab{b}}.
\newblock URL \url{https://openreview.net/forum?id=hWwY_Jq0xsN}.

\bibitem[Kim et~al.(2016)Kim, Khanna, and Koyejo]{Kim2016}
Been Kim, Rajiv Khanna, and Oluwasanmi~O Koyejo.
\newblock Examples are not enough, learn to criticize! criticism for interpretability.
\newblock In D.~Lee, M.~Sugiyama, U.~Luxburg, I.~Guyon, and R.~Garnett, editors, \emph{Advances in Neural Information Processing Systems}, volume~29. Curran Associates, Inc., 2016.

\bibitem[Kim et~al.(2018)Kim, Wattenberg, Gilmer, Cai, Wexler, Viegas, and sayres]{Kim2018}
Been Kim, Martin Wattenberg, Justin Gilmer, Carrie Cai, James Wexler, Fernanda Viegas, and Rory sayres.
\newblock Interpretability beyond feature attribution: Quantitative testing with concept activation vectors ({TCAV}).
\newblock In Jennifer Dy and Andreas Krause, editors, \emph{Proceedings of the 35th International Conference on Machine Learning}, volume~80 of \emph{Proceedings of Machine Learning Research}, pages 2668--2677. PMLR, 10--15 Jul 2018.
\newblock URL \url{https://proceedings.mlr.press/v80/kim18d.html}.

\bibitem[Kingma and Ba(2015)]{Kingma2014}
Diederik~P. Kingma and Jimmy Ba.
\newblock Adam: {A} method for stochastic optimization.
\newblock In Yoshua Bengio and Yann LeCun, editors, \emph{3rd International Conference on Learning Representations, {ICLR} 2015, San Diego, CA, USA, May 7-9, 2015, Conference Track Proceedings}, 2015.

\bibitem[Kipf and Welling(2017)]{Kipf2017}
Thomas~N. Kipf and Max Welling.
\newblock Semi-supervised classification with graph convolutional networks.
\newblock In \emph{International Conference on Learning Representations (ICLR)}, 2017.

\bibitem[Kiralj and Ferreira(2003)]{Kiralj2003}
Rudolf Kiralj and Márcia~M.C. Ferreira.
\newblock A priori molecular descriptors in qsar: a case of hiv-1 protease inhibitors: I. the chemometric approach.
\newblock \emph{Journal of Molecular Graphics and Modelling}, 21\penalty0 (5):\penalty0 435--448, 2003.
\newblock ISSN 1093-3263.
\newblock \doi{https://doi.org/10.1016/S1093-3263(02)00201-2}.
\newblock URL \url{https://www.sciencedirect.com/science/article/pii/S1093326302002012}.

\bibitem[Koh et~al.(2020)Koh, Nguyen, Tang, Mussmann, Pierson, Kim, and Liang]{Koh20}
Pang~Wei Koh, Thao Nguyen, Yew~Siang Tang, Stephen Mussmann, Emma Pierson, Been Kim, and Percy Liang.
\newblock Concept bottleneck models.
\newblock In Hal~Daumé III and Aarti Singh, editors, \emph{Proceedings of the 37th International Conference on Machine Learning}, volume 119 of \emph{Proceedings of Machine Learning Research}, pages 5338--5348. PMLR, 13--18 Jul 2020.
\newblock URL \url{https://proceedings.mlr.press/v119/koh20a.html}.

\bibitem[Kohlhammer et~al.(2011)Kohlhammer, Keim, Pohl, Santucci, and Andrienko]{Kohlhammer2011}
Jörn Kohlhammer, Daniel Keim, Margit Pohl, Giuseppe Santucci, and Gennady Andrienko.
\newblock Solving problems with visual analytics.
\newblock \emph{Procedia Computer Science}, 7:\penalty0 117--120, 2011.
\newblock ISSN 1877-0509.
\newblock \doi{https://doi.org/10.1016/j.procs.2011.12.035}.
\newblock Proceedings of the 2nd European Future Technologies Conference and Exhibition 2011 (FET 11).

\bibitem[Krizhevsky(2009)]{Krizhevsky2009}
A.~Krizhevsky.
\newblock Learning multiple layers of features from tiny images.
\newblock Technical report, University of Toronto, 2009.
\newblock [dataset].

\bibitem[Krizhevsky et~al.(2012)Krizhevsky, Sutskever, and Hinton]{Krizhevsky2012}
Alex Krizhevsky, Ilya Sutskever, and Geoffrey~E Hinton.
\newblock Imagenet classification with deep convolutional neural networks.
\newblock In F.~Pereira, C.J. Burges, L.~Bottou, and K.Q. Weinberger, editors, \emph{Advances in Neural Information Processing Systems}, volume~25. Curran Associates, Inc., 2012.
\newblock URL \url{https://proceedings.neurips.cc/paper_files/paper/2012/file/c399862d3b9d6b76c8436e924a68c45b-Paper.pdf}.

\bibitem[Kujawski et~al.(2012)Kujawski, Popielarska, Myka, Drabińska, and Bernard]{Kujawski2012}
Jacek Kujawski, Hanna Popielarska, Anna Myka, Beata Drabińska, and Marek Bernard.
\newblock The log p parameter as a molecular descriptor in the computer-aided drug design – an overview.
\newblock \emph{Computational Methods in Science and Technology}, 18:\penalty0 81--88, 08 2012.
\newblock \doi{10.12921/cmst.2012.18.02.81-88}.

\bibitem[{Kwon} et~al.(2019){Kwon}, {Choi}, {Kim}, {Choi}, {Kim}, {Kwon}, {Sun}, and {Choo}]{Kwon2019}
B.~C. {Kwon}, M.~{Choi}, J.~T. {Kim}, E.~{Choi}, Y.~B. {Kim}, S.~{Kwon}, J.~{Sun}, and J.~{Choo}.
\newblock Retainvis: Visual analytics with interpretable and interactive recurrent neural networks on electronic medical records.
\newblock \emph{IEEE Transactions on Visualization and Computer Graphics}, 25\penalty0 (1):\penalty0 299--309, Jan 2019.
\newblock ISSN 1941-0506.
\newblock \doi{10.1109/TVCG.2018.2865027}.

\bibitem[{La Rosa} et~al.(2023{\natexlab{a}}){La Rosa}, Blasilli, Bourqui, Auber, Santucci, Capobianco, Bertini, Giot, and Angelini]{LaRosa2023survey}
B.~{La Rosa}, G.~Blasilli, R.~Bourqui, D.~Auber, G.~Santucci, R.~Capobianco, E.~Bertini, R.~Giot, and M.~Angelini.
\newblock State of the art of visual analytics for explainable deep learning.
\newblock \emph{Computer Graphics Forum}, 42\penalty0 (1):\penalty0 319--355, February 2023{\natexlab{a}}.
\newblock ISSN 1467-8659.
\newblock \doi{10.1111/cgf.14733}.

\bibitem[La~Rosa et~al.(2020)La~Rosa, Capobianco, and Nardi]{LaRosa2020}
Biagio La~Rosa, Roberto Capobianco, and Daniele Nardi.
\newblock Explainable inference on sequential data via memory-tracking.
\newblock In \emph{Proceedings of the Twenty-Ninth International Joint Conference on Artificial Intelligence}, pages 2006--2013. International Joint Conferences on Artificial Intelligence Organization, 2020.

\bibitem[{La Rosa} et~al.(2021){La Rosa}, Capobianco, and Nardi]{LaRosa2021}
Biagio {La Rosa}, Roberto Capobianco, and Daniele Nardi.
\newblock A discussion about explainable inference on sequential data via memory-tracking.
\newblock In \emph{AIxIA 2021 Discussion Papers}, volume 3078, pages 33--44. CEUR Workshop Proceedings, 2021.
\newblock URL \url{http://ceur-ws.org/Vol-3078/paper-24.pdf}.

\bibitem[{La Rosa} et~al.(2022){La Rosa}, Capobianco, and Nardi]{LaRosa2022}
Biagio {La Rosa}, Roberto Capobianco, and Daniele Nardi.
\newblock A self-interpretable module for deep image classification on small data.
\newblock \emph{Applied Intelligence}, aug 2022.
\newblock \doi{10.1007/s10489-022-03886-6}.

\bibitem[{La Rosa} et~al.(2023{\natexlab{b}}){La Rosa}, Gilpin, and Capobianco]{LaRosa2023Towards}
Biagio {La Rosa}, Leilani~H. Gilpin, and Roberto Capobianco.
\newblock Towards a fuller understanding of neurons with clustered compositional explanations.
\newblock In \emph{Thirty-seventh Conference on Neural Information Processing Systems}, 2023{\natexlab{b}}.
\newblock URL \url{https://openreview.net/forum?id=51PLYhMFWz}.

\bibitem[Lapuschkin et~al.(2019)Lapuschkin, Wäldchen, Binder, Montavon, Samek, and Müller]{Lapuschkin2019}
Sebastian Lapuschkin, Stephan Wäldchen, Alexander Binder, Grégoire Montavon, Wojciech Samek, and Klaus-Robert Müller.
\newblock Unmasking clever hans predictors and assessing what machines really learn.
\newblock \emph{Nature Communications}, 10\penalty0 (1), March 2019.
\newblock ISSN 2041-1723.
\newblock \doi{10.1038/s41467-019-08987-4}.

\bibitem[Laugel et~al.(2018)Laugel, Lesot, Marsala, Renard, and Detyniecki]{Laugel2018}
Thibault Laugel, Marie-Jeanne Lesot, Christophe Marsala, Xavier Renard, and Marcin Detyniecki.
\newblock Comparison-based inverse classification for interpretability in machine learning.
\newblock In \emph{Communications in Computer and Information Science}, pages 100--111. Springer International Publishing, 2018.
\newblock \doi{10.1007/978-3-319-91473-2_9}.

\bibitem[Levine et~al.(2016)Levine, Finn, Darrell, and Abbeel]{Levine2016}
Sergey Levine, Chelsea Finn, Trevor Darrell, and Pieter Abbeel.
\newblock End-to-end training of deep visuomotor policies.
\newblock \emph{Journal of Machine Learning Research}, 17\penalty0 (39):\penalty0 1--40, 2016.
\newblock URL \url{http://jmlr.org/papers/v17/15-522.html}.

\bibitem[Li et~al.(2021)Li, Wang, Verma, Nakashima, Kawasaki, and Nagahara]{Li2021}
Liangzhi Li, Bowen Wang, Manisha Verma, Yuta Nakashima, Ryo Kawasaki, and Hajime Nagahara.
\newblock Scouter: Slot attention-based classifier for explainable image recognition.
\newblock In \emph{Proceedings of the IEEE/CVF International Conference on Computer Vision (ICCV)}, pages 1046--1055, October 2021.

\bibitem[Lillicrap et~al.(2016)Lillicrap, Hunt, Pritzel, Heess, Erez, Tassa, Silver, and Wierstra]{Lillicrap2016}
Timothy~P. Lillicrap, Jonathan~J. Hunt, Alexander Pritzel, Nicolas Heess, Tom Erez, Yuval Tassa, David Silver, and Daan Wierstra.
\newblock Continuous control with deep reinforcement learning.
\newblock In Yoshua Bengio and Yann LeCun, editors, \emph{4th International Conference on Learning Representations, {ICLR} 2016, San Juan, Puerto Rico, May 2-4, 2016, Conference Track Proceedings}, 2016.

\bibitem[Lipinski et~al.(2001)Lipinski, Lombardo, Dominy, and Feeney]{Lipinski2001}
Christopher~A. Lipinski, Franco Lombardo, Beryl~William Dominy, and P.~J. Feeney.
\newblock Experimental and computational approaches to estimate solubility and permeability in drug discovery and development settings.
\newblock \emph{Advanced drug delivery reviews}, 46 1-3:\penalty0 3--26, 2001.

\bibitem[Liu et~al.(2019{\natexlab{a}})Liu, Yin, and Wang]{liu2019towards}
Hui Liu, Qingyu Yin, and William~Yang Wang.
\newblock Towards explainable nlp: A generative explanation framework for text classification.
\newblock In \emph{Proceedings of the 57th Annual Meeting of the Association for Computational Linguistics}, pages 5570--5581, 2019{\natexlab{a}}.

\bibitem[{Liu} et~al.(2017){Liu}, {Shi}, {Li}, {Li}, {Zhu}, and {Liu}]{Liu2017}
M.~{Liu}, J.~{Shi}, Z.~{Li}, C.~{Li}, J.~{Zhu}, and S.~{Liu}.
\newblock Towards better analysis of deep convolutional neural networks.
\newblock \emph{IEEE Transactions on Visualization and Computer Graphics}, 23\penalty0 (1):\penalty0 91--100, Jan 2017.
\newblock ISSN 1941-0506.
\newblock \doi{10.1109/TVCG.2016.2598831}.

\bibitem[Liu et~al.(2018)Liu, Liu, Su, Cao, and Zhu]{Liu2018}
Mengchen Liu, Shixia Liu, Hang Su, Kelei Cao, and Jun Zhu.
\newblock Analyzing the noise robustness of deep neural networks.
\newblock In \emph{2018 {IEEE} Conference on Visual Analytics Science and Technology ({VAST})}. {IEEE}, oct 2018.
\newblock \doi{10.1109/vast.2018.8802509}.

\bibitem[Liu et~al.(2019{\natexlab{b}})Liu, Kailkhura, Loveland, and Han]{Liu2019counterfactuals}
Shusen Liu, Bhavya Kailkhura, Donald Loveland, and Yong Han.
\newblock Generative counterfactual introspection for explainable deep learning.
\newblock In \emph{2019 IEEE Global Conference on Signal and Information Processing (GlobalSIP)}, pages 1--5, 2019{\natexlab{b}}.

\bibitem[Liu et~al.(2019{\natexlab{c}})Liu, Li, Li, Srikumar, Pascucci, and Bremer]{Liu2019nlize}
Shusen Liu, Zhimin Li, Tao Li, Vivek Srikumar, Valerio Pascucci, and Peer-Timo Bremer.
\newblock {NLIZE}: A perturbation-driven visual interrogation tool for analyzing and interpreting natural language inference models.
\newblock \emph{{IEEE} Transactions on Visualization and Computer Graphics}, 25\penalty0 (1):\penalty0 651--660, jan 2019{\natexlab{c}}.
\newblock \doi{10.1109/tvcg.2018.2865230}.

\bibitem[Liu et~al.(2021)Liu, Khandagale, White, and Neiswanger]{liu2021synthetic}
Yang Liu, Sujay Khandagale, Colin White, and Willie Neiswanger.
\newblock Synthetic benchmarks for scientific research in explainable machine learning.
\newblock In \emph{Thirty-fifth Conference on Neural Information Processing Systems Datasets and Benchmarks Track (Round 2)}, 2021.
\newblock URL \url{https://openreview.net/forum?id=R7vr14ffhF9}.

\bibitem[Looveren and Klaise(2021)]{Looveren2021}
Arnaud~Van Looveren and Janis Klaise.
\newblock Interpretable counterfactual explanations guided by prototypes.
\newblock pages 650--665. Springer International Publishing, 2021.
\newblock \doi{10.1007/978-3-030-86520-7_40}.

\bibitem[Losch et~al.(2021)Losch, Fritz, and Schiele]{Losch2021}
Max~Maria Losch, Mario Fritz, and Bernt Schiele.
\newblock \emph{Semantic Bottlenecks: Quantifying and Improving Inspectability of Deep Representations}, pages 15--29.
\newblock Springer International Publishing, 2021.
\newblock ISBN 9783030712785.
\newblock \doi{10.1007/978-3-030-71278-5_2}.

\bibitem[Lundberg and Lee(2017)]{Lundberg2017}
Scott~M Lundberg and Su-In Lee.
\newblock A unified approach to interpreting model predictions.
\newblock In I.~Guyon, U.~V. Luxburg, S.~Bengio, H.~Wallach, R.~Fergus, S.~Vishwanathan, and R.~Garnett, editors, \emph{Advances in Neural Information Processing Systems}, volume~30. Curran Associates, Inc., 2017.

\bibitem[Luo et~al.(2020)Luo, Cheng, Xu, Yu, Zong, Chen, and Zhang]{Luo2020}
Dongsheng Luo, Wei Cheng, Dongkuan Xu, Wenchao Yu, Bo~Zong, Haifeng Chen, and Xiang Zhang.
\newblock Parameterized explainer for graph neural network.
\newblock \emph{Advances in Neural Information Processing Systems}, 33, 2020.

\bibitem[Luong et~al.(2015)Luong, Pham, and Manning]{Luong2015}
Thang Luong, Hieu Pham, and Christopher~D. Manning.
\newblock Effective approaches to attention-based neural machine translation.
\newblock In \emph{Proceedings of the 2015 Conference on Empirical Methods in Natural Language Processing}. Association for Computational Linguistics, 2015.
\newblock \doi{10.18653/v1/d15-1166}.

\bibitem[Ma et~al.(2021)Ma, Fan, He, Nelakurthi, and Maciejewski]{Ma2021}
Yuxin Ma, Arlen Fan, Jingrui He, Arun~Reddy Nelakurthi, and Ross Maciejewski.
\newblock A visual analytics framework for explaining and diagnosing transfer learning processes.
\newblock \emph{{IEEE} Transactions on Visualization and Computer Graphics}, 27\penalty0 (2):\penalty0 1385--1395, feb 2021.
\newblock \doi{10.1109/tvcg.2020.3028888}.

\bibitem[Mahendran and Vedaldi(2015)]{Mahendran2015}
Aravindh Mahendran and Andrea Vedaldi.
\newblock Understanding deep image representations by inverting them.
\newblock In \emph{Proceedings of the IEEE Conference on Computer Vision and Pattern Recognition (CVPR)}, June 2015.

\bibitem[Makinwa et~al.(2022)Makinwa, {La Rosa}, and Capobianco]{Makinwa2022}
Sayo~M. Makinwa, Biagio {La Rosa}, and Roberto Capobianco.
\newblock Detection accuracy for~evaluating compositional explanations of~units.
\newblock In \emph{AIxIA 2021 - Advances in Artificial Intelligence}, pages 550--563. Springer International Publishing, 2022.
\newblock \doi{10.1007/978-3-031-08421-8_38}.

\bibitem[Malaviya et~al.(2018)Malaviya, Ferreira, and Martins]{Malaviya2018}
Chaitanya Malaviya, Pedro Ferreira, and Andr{\'{e}} F.~T. Martins.
\newblock Sparse and constrained attention for neural machine translation.
\newblock In \emph{Proceedings of the 56th Annual Meeting of the Association for Computational Linguistics (Volume 2: Short Papers)}. Association for Computational Linguistics, 2018.
\newblock \doi{10.18653/v1/p18-2059}.

\bibitem[Martins and Astudillo(2016{\natexlab{a}})]{Martins16}
Andre Martins and Ramon Astudillo.
\newblock From softmax to sparsemax: A sparse model of attention and multi-label classification.
\newblock In Maria~Florina Balcan and Kilian~Q. Weinberger, editors, \emph{Proceedings of The 33rd International Conference on Machine Learning}, volume~48 of \emph{Proceedings of Machine Learning Research}, pages 1614--1623, New York, New York, USA, 20--22 Jun 2016{\natexlab{a}}. PMLR.
\newblock URL \url{https://proceedings.mlr.press/v48/martins16.html}.

\bibitem[Martins and Astudillo(2016{\natexlab{b}})]{Martins2016}
Andr\'{e} F.~T. Martins and Ram\'{o}n~F. Astudillo.
\newblock From softmax to sparsemax: A sparse model of attention and multi-label classification.
\newblock In \emph{Proceedings of the 33rd International Conference on International Conference on Machine Learning - Volume 48}, ICML'16, page 1614–1623. JMLR.org, 2016{\natexlab{b}}.

\bibitem[Martins et~al.(2012)Martins, Teixeira, Pinheiro, and Falcao]{Martins2012}
Ines~Filipa Martins, Ana~L. Teixeira, Luis Pinheiro, and Andre~O. Falcao.
\newblock A bayesian approach to in silico blood-brain barrier penetration modeling.
\newblock \emph{Journal of Chemical Information and Modeling}, 52\penalty0 (6):\penalty0 1686--1697, jun 2012.
\newblock \doi{10.1021/ci300124c}.
\newblock URL \url{https://doi.org/10.1021%2Fci300124c}.

\bibitem[Massidda and Bacciu(2023)]{Massidda2023}
Riccardo Massidda and Davide Bacciu.
\newblock Knowledge-driven interpretation of~convolutional neural networks.
\newblock In \emph{Machine Learning and Knowledge Discovery in Databases}, pages 356--371. Springer International Publishing, 2023.
\newblock \doi{10.1007/978-3-031-26387-3_22}.

\bibitem[McCallum et~al.(2000)McCallum, Nigam, Rennie, and Seymore]{McCallum2000}
Andrew~Kachites McCallum, Kamal Nigam, Jason Rennie, and Kristie Seymore.
\newblock Automating the construction of internet portals with machine learning.
\newblock \emph{Information Retrieval}, 3\penalty0 (2):\penalty0 127--163, 2000.
\newblock ISSN 1386-4564.
\newblock \doi{10.1023/a:1009953814988}.

\bibitem[Meng et~al.(2023)Meng, Sharma, Andonian, Belinkov, and Bau]{meng2023massediting}
Kevin Meng, Arnab~Sen Sharma, Alex~J Andonian, Yonatan Belinkov, and David Bau.
\newblock Mass-editing memory in a transformer.
\newblock In \emph{The Eleventh International Conference on Learning Representations}, 2023.
\newblock URL \url{https://openreview.net/forum?id=MkbcAHIYgyS}.

\bibitem[Mihaylov and Frank(2017)]{Mihaylov2017}
Todor Mihaylov and Anette Frank.
\newblock Story cloze ending selection baselines and data examination.
\newblock In Michael Roth, Nasrin Mostafazadeh, Nathanael Chambers, and Annie Louis, editors, \emph{Proceedings of the 2nd Workshop on Linking Models of Lexical, Sentential and Discourse-level Semantics}, pages 87--92, Valencia, Spain, April 2017. Association for Computational Linguistics.
\newblock \doi{10.18653/v1/W17-0913}.
\newblock URL \url{https://aclanthology.org/W17-0913}.

\bibitem[Mikolov et~al.(2013)Mikolov, Yih, and Zweig]{Mikolov2013}
Tom{\'a}{\v{s}} Mikolov, Wen-tau Yih, and Geoffrey Zweig.
\newblock Linguistic regularities in continuous space word representations.
\newblock In \emph{Proceedings of the 2013 conference of the north american chapter of the association for computational linguistics: Human language technologies}, pages 746--751, 2013.

\bibitem[Miller(2019)]{Miller2019}
Tim Miller.
\newblock Explanation in artificial intelligence: Insights from the social sciences.
\newblock \emph{Artificial Intelligence}, 267:\penalty0 1--38, February 2019.
\newblock ISSN 0004-3702.
\newblock \doi{10.1016/j.artint.2018.07.007}.

\bibitem[Ming et~al.(2019)Ming, Xu, Qu, and Ren]{Ming2019}
Yao Ming, Panpan Xu, Huamin Qu, and Liu Ren.
\newblock Interpretable and steerable sequence learning via prototypes.
\newblock In \emph{Proceedings of the 25th ACM SIGKDD International Conference on Knowledge Discovery \& Data Mining}, KDD ’19. ACM, July 2019.
\newblock \doi{10.1145/3292500.3330908}.

\bibitem[Ming et~al.(2020)Ming, Xu, Cheng, Qu, and Ren]{Ming2020}
Yao Ming, Panpan Xu, Furui Cheng, Huamin Qu, and Liu Ren.
\newblock {ProtoSteer}: Steering deep sequence model with prototypes.
\newblock \emph{{IEEE} Transactions on Visualization and Computer Graphics}, 26\penalty0 (1):\penalty0 238--248, jan 2020.
\newblock \doi{10.1109/tvcg.2019.2934267}.

\bibitem[Mishra et~al.(2022)Mishra, Soni, Huang, and Bryan]{Mishra2021}
A.~Mishra, U.~Soni, J.~Huang, and C.~Bryan.
\newblock Why? why not? when? visual explanations of agent behaviour in reinforcement learning.
\newblock In \emph{2022 IEEE 15th Pacific Visualization Symposium (PacificVis)}, pages 111--120, Los Alamitos, CA, USA, apr 2022. IEEE Computer Society.
\newblock \doi{10.1109/PacificVis53943.2022.00020}.
\newblock URL \url{https://doi.ieeecomputersociety.org/10.1109/PacificVis53943.2022.00020}.

\bibitem[Mnih et~al.(2015)Mnih, Kavukcuoglu, Silver, Rusu, Veness, Bellemare, Graves, Riedmiller, Fidjeland, Ostrovski, Petersen, Beattie, Sadik, Antonoglou, King, Kumaran, Wierstra, Legg, and Hassabis]{Mnih2015}
Volodymyr Mnih, Koray Kavukcuoglu, David Silver, Andrei~A. Rusu, Joel Veness, Marc~G. Bellemare, Alex Graves, Martin Riedmiller, Andreas~K. Fidjeland, Georg Ostrovski, Stig Petersen, Charles Beattie, Amir Sadik, Ioannis Antonoglou, Helen King, Dharshan Kumaran, Daan Wierstra, Shane Legg, and Demis Hassabis.
\newblock Human-level control through deep reinforcement learning.
\newblock \emph{Nature}, 518\penalty0 (7540):\penalty0 529--533, February 2015.
\newblock ISSN 1476-4687.
\newblock \doi{10.1038/nature14236}.

\bibitem[Mohammadjafari et~al.(2021)Mohammadjafari, Çevik, Thanabalasingam, Basar, and Initiative]{Mohammadjafari2021}
Sanaz Mohammadjafari, Mucahit Çevik, Mathusan Thanabalasingam, Ayse Basar, and Alzheimer’s Initiative.
\newblock Using protopnet for interpretable alzheimer’s disease classification.
\newblock \emph{Proceedings of the Canadian Conference on Artificial Intelligence}, 06 2021.
\newblock \doi{10.21428/594757db.fb59ce6c}.

\bibitem[Mordvintsev et~al.(2018)Mordvintsev, Pezzotti, Schubert, and Olah]{Mordvintsev2018}
Alexander Mordvintsev, Nicola Pezzotti, Ludwig Schubert, and Chris Olah.
\newblock Differentiable image parameterizations.
\newblock \emph{Distill}, 3\penalty0 (7), jul 2018.
\newblock \doi{10.23915/distill.00012}.

\bibitem[Mostafazadeh et~al.(2016)Mostafazadeh, Chambers, He, Parikh, Batra, Vanderwende, Kohli, and Allen]{mostafazadeh2016}
Nasrin Mostafazadeh, Nathanael Chambers, Xiaodong He, Devi Parikh, Dhruv Batra, Lucy Vanderwende, Pushmeet Kohli, and James Allen.
\newblock A corpus and cloze evaluation for deeper understanding of commonsense stories.
\newblock In Kevin Knight, Ani Nenkova, and Owen Rambow, editors, \emph{Proceedings of the 2016 Conference of the North {A}merican Chapter of the Association for Computational Linguistics: Human Language Technologies}, pages 839--849, San Diego, California, June 2016. Association for Computational Linguistics.
\newblock \doi{10.18653/v1/N16-1098}.
\newblock URL \url{https://aclanthology.org/N16-1098}.

\bibitem[Mu and Andreas(2020)]{Mu2020}
Jesse Mu and Jacob Andreas.
\newblock Compositional explanations of neurons.
\newblock In H.~Larochelle, M.~Ranzato, R.~Hadsell, M.F. Balcan, and H.~Lin, editors, \emph{Advances in Neural Information Processing Systems}, volume~33, pages 17153--17163. Curran Associates, Inc., 2020.
\newblock URL \url{https://proceedings.neurips.cc/paper/2020/file/c74956ffb38ba48ed6ce977af6727275-Paper.pdf}.

\bibitem[Na et~al.(2019)Na, Choe, Lee, and Kim]{Na2019}
Seil Na, Yo~Joong Choe, Dong-Hyun Lee, and Gunhee Kim.
\newblock {Discovery of Natural Language Concepts in Individual Units of CNNs}.
\newblock In \emph{International Conference on Learning Representations}, 2019.
\newblock URL \url{https://openreview.net/forum?id=S1EERs09YQ}.

\bibitem[Natesan~Ramamurthy et~al.(2020)Natesan~Ramamurthy, Vinzamuri, Zhang, and Dhurandhar]{Ramamurthy2020}
Karthikeyan Natesan~Ramamurthy, Bhanukiran Vinzamuri, Yunfeng Zhang, and Amit Dhurandhar.
\newblock Model agnostic multilevel explanations.
\newblock In H.~Larochelle, M.~Ranzato, R.~Hadsell, M.F. Balcan, and H.~Lin, editors, \emph{Advances in Neural Information Processing Systems}, volume~33, pages 5968--5979. Curran Associates, Inc., 2020.
\newblock URL \url{https://proceedings.neurips.cc/paper_files/paper/2020/file/426f990b332ef8193a61cc90516c1245-Paper.pdf}.

\bibitem[Nauta et~al.(2021)Nauta, van Bree, and Seifert]{Nauta2021}
Meike Nauta, Ron van Bree, and Christin Seifert.
\newblock Neural prototype trees for interpretable fine-grained image recognition.
\newblock In \emph{2021 IEEE/CVF Conference on Computer Vision and Pattern Recognition (CVPR)}. IEEE, June 2021.
\newblock \doi{10.1109/cvpr46437.2021.01469}.

\bibitem[Nauta et~al.(2023)Nauta, Schlötterer, van Keulen, and Seifert]{Nauta2023}
Meike Nauta, Jörg Schlötterer, Maurice van Keulen, and Christin Seifert.
\newblock Pip-net: Patch-based intuitive prototypes for interpretable image classification.
\newblock In \emph{2023 IEEE/CVF Conference on Computer Vision and Pattern Recognition (CVPR)}. IEEE, June 2023.
\newblock \doi{10.1109/cvpr52729.2023.00269}.

\bibitem[Netzer et~al.(2011)Netzer, Wang, Coates, Bissacco, Wu, and Ng]{Netzer2011}
Yuval Netzer, Tao Wang, Adam Coates, Alessandro Bissacco, Bo~Wu, and Andrew~Y. Ng.
\newblock Reading digits in natural images with unsupervised feature learning.
\newblock In \emph{NIPS Workshop on Deep Learning and Unsupervised Feature Learning}, 2011.

\bibitem[Nguyen et~al.(2016{\natexlab{a}})Nguyen, Dosovitskiy, Yosinski, Brox, and Clune]{Nguyen2016nips}
Anh Nguyen, Alexey Dosovitskiy, Jason Yosinski, Thomas Brox, and Jeff Clune.
\newblock Synthesizing the preferred inputs for neurons in neural networks via deep generator networks.
\newblock In D.~Lee, M.~Sugiyama, U.~Luxburg, I.~Guyon, and R.~Garnett, editors, \emph{Advances in Neural Information Processing Systems}, volume~29. Curran Associates, Inc., 2016{\natexlab{a}}.
\newblock URL \url{https://proceedings.neurips.cc/paper_files/paper/2016/file/5d79099fcdf499f12b79770834c0164a-Paper.pdf}.

\bibitem[Nguyen et~al.(2016{\natexlab{b}})Nguyen, Yosinski, and Clune]{Nguyen2016feat}
Anh Nguyen, Jason Yosinski, and Jeff Clune.
\newblock Multifaceted feature visualization: Uncovering the different types of features learned by each neuron in deep neural networks.
\newblock February 2016{\natexlab{b}}.
\newblock \doi{10.48550/ARXIV.1602.03616}.

\bibitem[Nguyen et~al.(2017)Nguyen, Clune, Bengio, Dosovitskiy, and Yosinski]{Nguyen2017}
Anh Nguyen, Jeff Clune, Yoshua Bengio, Alexey Dosovitskiy, and Jason Yosinski.
\newblock Plug \& play generative networks: Conditional iterative generation of images in latent space.
\newblock In \emph{Proceedings of the IEEE Conference on Computer Vision and Pattern Recognition (CVPR)}, July 2017.

\bibitem[Nugent and Cunningham(2005)]{Nugent2005}
Conor Nugent and P{\'{a}}draig Cunningham.
\newblock A case-based explanation system for black-box systems.
\newblock \emph{Artificial Intelligence Review}, 24\penalty0 (2):\penalty0 163--178, oct 2005.
\newblock \doi{10.1007/s10462-005-4609-5}.

\bibitem[Oikarinen and Weng(2023)]{Oikarinen2023}
Tuomas Oikarinen and Tsui-Wei Weng.
\newblock {CLIP}-dissect: Automatic description of neuron representations in deep vision networks.
\newblock In \emph{International Conference on Learning Representations}, 2023.
\newblock URL \url{https://openreview.net/forum?id=iPWiwWHc1V}.

\bibitem[Oikarinen et~al.(2023)Oikarinen, Das, Nguyen, and Weng]{oikarinen2023labelfree}
Tuomas Oikarinen, Subhro Das, Lam~M. Nguyen, and Tsui-Wei Weng.
\newblock Label-free concept bottleneck models.
\newblock In \emph{The Eleventh International Conference on Learning Representations}, 2023.
\newblock URL \url{https://openreview.net/forum?id=FlCg47MNvBA}.

\bibitem[Olah et~al.(2017)Olah, Mordvintsev, and Schubert]{Olah2017}
Chris Olah, Alexander Mordvintsev, and Ludwig Schubert.
\newblock Feature visualization.
\newblock \emph{Distill}, 2\penalty0 (11), nov 2017.

\bibitem[Olah et~al.(2020)Olah, Cammarata, Schubert, Goh, Petrov, and Carter]{Olah2020}
Chris Olah, Nick Cammarata, Ludwig Schubert, Gabriel Goh, Michael Petrov, and Shan Carter.
\newblock Zoom in: An introduction to circuits.
\newblock \emph{Distill}, 5\penalty0 (3), March 2020.
\newblock ISSN 2476-0757.
\newblock \doi{10.23915/distill.00024.001}.

\bibitem[Onari et~al.(2023)Onari, Grau, Nobile, and Zhang]{Onari2023}
Mohsen~Abbaspour Onari, Isel Grau, Marco~S. Nobile, and Yingqian Zhang.
\newblock Trustworthy artificial intelligence in medical applications: A mini survey.
\newblock In \emph{2023 IEEE Conference on Computational Intelligence in Bioinformatics and Computational Biology (CIBCB)}. IEEE, August 2023.
\newblock \doi{10.1109/cibcb56990.2023.10264883}.

\bibitem[Palm et~al.(1997)Palm, Stenberg, Luthman, and Artursson1]{Palm1997}
Katrin Palm, Patric Stenberg, Kristina Luthman, and Per Artursson1.
\newblock \emph{Pharmaceutical Research}, 14\penalty0 (5):\penalty0 568--571, 1997.
\newblock \doi{10.1023/a:1012188625088}.

\bibitem[Papernot and McDaniel(2018)]{Papernot2018}
Nicolas Papernot and Patrick McDaniel.
\newblock Deep k-nearest neighbors: Towards confident, interpretable and robust deep learning.
\newblock March 2018.

\bibitem[Park et~al.(2019)Park, Choo, Na, Jo, Shin, Yoo, Kwon, Zhao, Noh, and Lee]{Park2019}
Cheonbok Park, Jaegul Choo, Inyoup Na, Yongjang Jo, Sungbok Shin, Jaehyo Yoo, Bum~Chul Kwon, Jian Zhao, Hyungjong Noh, and Yeonsoo Lee.
\newblock {SANVis}: Visual analytics for understanding self-attention networks.
\newblock In \emph{2019 IEEE Visualization Conference (VIS)}. {IEEE}, oct 2019.
\newblock \doi{10.1109/visual.2019.8933677}.

\bibitem[Park et~al.(2021)Park, Yang, Na, Chung, Shin, Kwon, Park, and Choo]{Park2021}
Cheonbok Park, Soyoung Yang, Inyoup Na, Sunghyo Chung, Sungbok Shin, Bum~Chul Kwon, Deokgun Park, and Jaegul Choo.
\newblock Vatun: Visual analytics for testing and understanding convolutional neural networks.
\newblock \emph{EuroVis 2021 - Short Papers}, 2021.
\newblock \doi{10.2312/EVS.20211047}.

\bibitem[Park et~al.(2022)Park, Das, Duggal, Wright, Shaikh, Hohman, and Chau]{Park2022}
Haekyu Park, Nilaksh Das, Rahul Duggal, Austin~P. Wright, Omar Shaikh, Fred Hohman, and Duen Horng~Polo Chau.
\newblock {NeuroCartography}: Scalable automatic visual summarization of concepts in deep neural networks.
\newblock \emph{{IEEE} Transactions on Visualization and Computer Graphics}, 28\penalty0 (1):\penalty0 813--823, jan 2022.
\newblock \doi{10.1109/tvcg.2021.3114858}.

\bibitem[Park et~al.(2004)Park, Im, Shin, and Park]{Park2004}
Jae~Heon Park, Kwang~Hyuk Im, Chung-Kwan Shin, and Sang~Chan Park.
\newblock Mbnr: Case-based reasoning with local feature weighting by neural network.
\newblock \emph{Applied Intelligence}, 21\penalty0 (3):\penalty0 265--276, November 2004.
\newblock ISSN 0924-669X.
\newblock \doi{10.1023/b:apin.0000043559.83167.3d}.

\bibitem[Peters et~al.(2019)Peters, Niculae, and Martins]{Peters2019}
Ben Peters, Vlad Niculae, and Andr{\'e} F.~T. Martins.
\newblock Sparse sequence-to-sequence models.
\newblock In Anna Korhonen, David Traum, and Llu{\'\i}s M{\`a}rquez, editors, \emph{Proceedings of the 57th Annual Meeting of the Association for Computational Linguistics}, pages 1504--1519, Florence, Italy, July 2019. Association for Computational Linguistics.
\newblock \doi{10.18653/v1/P19-1146}.
\newblock URL \url{https://aclanthology.org/P19-1146}.

\bibitem[Petsiuk et~al.(2018)Petsiuk, Das, and Saenko]{Petsiuk2018rise}
Vitali Petsiuk, Abir Das, and Kate Saenko.
\newblock Rise: Randomized input sampling for explanation of black-box models.
\newblock In \emph{Proceedings of the British Machine Vision Conference (BMVC)}, 2018.

\bibitem[Pezzotti et~al.(2016)Pezzotti, Höllt, Lelieveldt, Eisemann, and Vilanova]{Pezzotti2016}
N.~Pezzotti, T.~Höllt, B.~Lelieveldt, E.~Eisemann, and A.~Vilanova.
\newblock Hierarchical stochastic neighbor embedding.
\newblock \emph{Computer Graphics Forum}, 35\penalty0 (3):\penalty0 21--30, jun 2016.
\newblock \doi{10.1111/cgf.12878}.

\bibitem[Pezzotti et~al.(2018)Pezzotti, Hollt, Gemert, Lelieveldt, Eisemann, and Vilanova]{Pezzotti2018}
Nicola Pezzotti, Thomas Hollt, Jan~Van Gemert, Boudewijn~P.F. Lelieveldt, Elmar Eisemann, and Anna Vilanova.
\newblock {DeepEyes}: Progressive visual analytics for designing deep neural networks.
\newblock \emph{{IEEE} Transactions on Visualization and Computer Graphics}, 24\penalty0 (1):\penalty0 98--108, jan 2018.
\newblock \doi{10.1109/tvcg.2017.2744358}.

\bibitem[phi Nguyen and Martínez(2020)]{Nguyen2020}
An\ phi Nguyen and María~Rodríguez Martínez.
\newblock On quantitative aspects of model interpretability.
\newblock July 2020.

\bibitem[Poyiadzi et~al.(2020)Poyiadzi, Sokol, Santos-Rodriguez, De~Bie, and Flach]{Poyiadzi2020}
Rafael Poyiadzi, Kacper Sokol, Raul Santos-Rodriguez, Tijl De~Bie, and Peter Flach.
\newblock Face: Feasible and actionable counterfactual explanations.
\newblock In \emph{Proceedings of the AAAI/ACM Conference on AI, Ethics, and Society}, AIES ’20. ACM, February 2020.
\newblock \doi{10.1145/3375627.3375850}.

\bibitem[Proietti et~al.(2023)Proietti, Ragno, {La Rosa}, Ragno, and Capobianco]{Proietti2023}
Michela Proietti, Alessio Ragno, Biagio {La Rosa}, Rino Ragno, and Roberto Capobianco.
\newblock Explainable {AI} in drug discovery: self-interpretable graph neural network for molecular property prediction using concept whitening.
\newblock \emph{Machine Learning}, oct 2023.
\newblock \doi{10.1007/s10994-023-06369-y}.

\bibitem[Ragodos et~al.(2022)Ragodos, Wang, Lin, and Zhou]{Ragodos2022}
Ronilo Ragodos, Tong Wang, Qihang Lin, and Xun Zhou.
\newblock Explaining a reinforcement learning agent via prototyping.
\newblock In Alice~H. Oh, Alekh Agarwal, Danielle Belgrave, and Kyunghyun Cho, editors, \emph{Advances in Neural Information Processing Systems}, 2022.
\newblock URL \url{https://openreview.net/forum?id=nyBJcnhjAoy}.

\bibitem[Ramaswamy et~al.(2022)Ramaswamy, Kim, Fong, and Russakovsky]{Ramaswamy2022}
Vikram~V. Ramaswamy, Sunnie S.~Y. Kim, Ruth Fong, and Olga Russakovsky.
\newblock Overlooked factors in concept-based explanations: Dataset choice, concept salience, and human capability.
\newblock July 2022.

\bibitem[Ramon et~al.(2020)Ramon, Martens, Provost, and Evgeniou]{Ramon2020}
Yanou Ramon, David Martens, Foster Provost, and Theodoros Evgeniou.
\newblock A comparison of instance-level counterfactual explanation algorithms for behavioral and textual data: Sedc, lime-c and shap-c.
\newblock \emph{Advances in Data Analysis and Classification}, 14\penalty0 (4):\penalty0 801--819, September 2020.
\newblock ISSN 1862-5355.
\newblock \doi{10.1007/s11634-020-00418-3}.

\bibitem[Rathore et~al.(2021)Rathore, Chalapathi, Palande, and Wang]{rathore_topoact_2021}
Archit Rathore, Nithin Chalapathi, Sourabh Palande, and Bei Wang.
\newblock Topoact: Visually exploring the shape of activations in deep learning.
\newblock \emph{Computer Graphics Forum}, 40\penalty0 (1):\penalty0 382--397, 2021.
\newblock \doi{https://doi.org/10.1111/cgf.14195}.
\newblock URL \url{https://onlinelibrary.wiley.com/doi/abs/10.1111/cgf.14195}.

\bibitem[Rayhan and Hashem(2023)]{Rayhan2023}
Yeasir Rayhan and Tanzima Hashem.
\newblock Aist: An interpretable attention-based deep learning model for crime prediction.
\newblock \emph{ACM Transactions on Spatial Algorithms and Systems}, 9\penalty0 (2):\penalty0 1--31, April 2023.
\newblock ISSN 2374-0361.
\newblock \doi{10.1145/3582274}.

\bibitem[Redmon et~al.(2016)Redmon, Divvala, Girshick, and Farhadi]{Redmon2016}
Joseph Redmon, Santosh Divvala, Ross Girshick, and Ali Farhadi.
\newblock You only look once: Unified, real-time object detection.
\newblock In \emph{2016 IEEE Conference on Computer Vision and Pattern Recognition (CVPR)}. IEEE, June 2016.
\newblock \doi{10.1109/cvpr.2016.91}.

\bibitem[Ribeiro et~al.(2016)Ribeiro, Singh, and Guestrin]{Ribeiro2016}
Marco~Tulio Ribeiro, Sameer Singh, and Carlos Guestrin.
\newblock “why should i trust you?”: Explaining the predictions of any classifier.
\newblock In \emph{Proceedings of the 22nd ACM SIGKDD International Conference on Knowledge Discovery and Data Mining}, KDD ’16. ACM, August 2016.
\newblock \doi{10.1145/2939672.2939778}.

\bibitem[Rigotti et~al.(2022)Rigotti, Miksovic, Giurgiu, Gschwind, and Scotton]{rigotti2022attentionbased}
Mattia Rigotti, Christoph Miksovic, Ioana Giurgiu, Thomas Gschwind, and Paolo Scotton.
\newblock Attention-based interpretability with concept transformers.
\newblock In \emph{International Conference on Learning Representations}, 2022.
\newblock URL \url{https://openreview.net/forum?id=kAa9eDS0RdO}.

\bibitem[Rosenblatt(1958)]{Rosenblatt1958}
F.~Rosenblatt.
\newblock The perceptron: A probabilistic model for information storage and organization in the brain.
\newblock \emph{Psychological Review}, 65\penalty0 (6):\penalty0 386--408, 1958.
\newblock ISSN 0033-295X.
\newblock \doi{10.1037/h0042519}.

\bibitem[Rudin(2019)]{Rudin2019}
Cynthia Rudin.
\newblock Stop explaining black box machine learning models for high stakes decisions and use interpretable models instead.
\newblock \emph{Nature Machine Intelligence}, 1\penalty0 (5):\penalty0 206--215, May 2019.
\newblock ISSN 2522-5839.
\newblock \doi{10.1038/s42256-019-0048-x}.

\bibitem[Rymarczyk et~al.(2021)Rymarczyk, Struski, Tabor, and Zieliński]{Rymarczyk2021}
Dawid Rymarczyk, Łukasz Struski, Jacek Tabor, and Bartosz Zieliński.
\newblock Protopshare: Prototypical parts sharing for similarity discovery in interpretable image classification.
\newblock In \emph{Proceedings of the 27th ACM SIGKDD Conference on Knowledge Discovery \&; Data Mining}, KDD ’21. ACM, August 2021.
\newblock \doi{10.1145/3447548.3467245}.

\bibitem[Rymarczyk et~al.(2022)Rymarczyk, Struski, Górszczak, Lewandowska, Tabor, and Zieliński]{Rymarczyk2022}
Dawid Rymarczyk, Łukasz Struski, Michał Górszczak, Koryna Lewandowska, Jacek Tabor, and Bartosz Zieliński.
\newblock \emph{Interpretable Image Classification with Differentiable Prototypes Assignment}, pages 351--368.
\newblock Springer Nature Switzerland, 2022.
\newblock ISBN 9783031197758.
\newblock \doi{10.1007/978-3-031-19775-8_21}.

\bibitem[Sacha et~al.(2014)Sacha, Stoffel, Stoffel, Kwon, Ellis, and Keim]{Sacha2014}
Dominik Sacha, Andreas Stoffel, Florian Stoffel, Bum~Chul Kwon, Geoffrey Ellis, and Daniel~A. Keim.
\newblock Knowledge generation model for visual analytics.
\newblock \emph{IEEE Transactions on Visualization and Computer Graphics}, 20\penalty0 (12):\penalty0 1604--1613, 2014.
\newblock \doi{10.1109/TVCG.2014.2346481}.

\bibitem[Sajjad et~al.(2022)Sajjad, Durrani, and Dalvi]{Sajjad2022}
Hassan Sajjad, Nadir Durrani, and Fahim Dalvi.
\newblock Neuron-level interpretation of deep {NLP} models: A survey.
\newblock \emph{Transactions of the Association for Computational Linguistics}, 10:\penalty0 1285--1303, 2022.
\newblock \doi{10.1162/tacl_a_00519}.

\bibitem[Sakiyama et~al.(2021)Sakiyama, Fukuda, and Okuno]{Sakiyama2021}
Hiroshi Sakiyama, Motohisa Fukuda, and Takashi Okuno.
\newblock Prediction of blood-brain barrier penetration (bbbp) based on molecular descriptors of the free-form and in-blood-form datasets.
\newblock \emph{Molecules}, 2021.

\bibitem[Salman et~al.(2020)Salman, Payrovnaziri, Liu, Rengifo-Moreno, and He]{Salman2020}
Shaeke Salman, Seyedeh~Neelufar Payrovnaziri, Xiuwen Liu, Pablo Rengifo-Moreno, and Zhe He.
\newblock Deepconsensus: Consensus-based interpretable deep neural networks with application to mortality prediction.
\newblock In \emph{2020 International Joint Conference on Neural Networks (IJCNN)}. IEEE, July 2020.
\newblock \doi{10.1109/ijcnn48605.2020.9206678}.

\bibitem[Sandler et~al.(2018)Sandler, Howard, Zhu, Zhmoginov, and Chen]{Sandler2018}
Mark Sandler, Andrew Howard, Menglong Zhu, Andrey Zhmoginov, and Liang-Chieh Chen.
\newblock Mobilenetv2: Inverted residuals and linear bottlenecks.
\newblock In \emph{Proceedings of the IEEE Conference on Computer Vision and Pattern Recognition (CVPR)}, June 2018.

\bibitem[Selvaraju et~al.(2017)Selvaraju, Cogswell, Das, Vedantam, Parikh, and Batra]{Selvaraju2017}
Ramprasaath~R. Selvaraju, Michael Cogswell, Abhishek Das, Ramakrishna Vedantam, Devi Parikh, and Dhruv Batra.
\newblock Grad-{CAM}: Visual explanations from deep networks via gradient-based localization.
\newblock In \emph{2017 {IEEE} International Conference on Computer Vision ({ICCV})}. {IEEE}, oct 2017.

\bibitem[Selvaraju et~al.(2020)Selvaraju, Cogswell, Das, Vedantam, Parikh, and Batra]{Selvaraju2020}
Ramprasaath~R. Selvaraju, Michael Cogswell, Abhishek Das, Ramakrishna Vedantam, Devi Parikh, and Dhruv Batra.
\newblock Grad-cam: Visual explanations from deep networks via gradient-based localization.
\newblock \emph{Int. J. Comput. Vision}, 128\penalty0 (2):\penalty0 336–359, feb 2020.
\newblock ISSN 0920-5691.
\newblock \doi{10.1007/s11263-019-01228-7}.
\newblock URL \url{https://doi.org/10.1007/s11263-019-01228-7}.

\bibitem[Sen et~al.(2008)Sen, Namata, Bilgic, Getoor, Gallagher, and Eliassi‐Rad]{Sen2008}
Prithviraj Sen, Galileo Namata, Mustafa Bilgic, Lise Getoor, Brian Gallagher, and Tina Eliassi‐Rad.
\newblock Collective classification in network data.
\newblock \emph{AI Magazine}, 29\penalty0 (3):\penalty0 93--106, September 2008.
\newblock ISSN 2371-9621.
\newblock \doi{10.1609/aimag.v29i3.2157}.

\bibitem[Serrano and Smith(2019)]{Serrano2019}
Sofia Serrano and Noah~A. Smith.
\newblock Is attention interpretable?
\newblock In \emph{Proceedings of the 57th Annual Meeting of the Association for Computational Linguistics}. Association for Computational Linguistics, 2019.
\newblock \doi{10.18653/v1/p19-1282}.

\bibitem[Sharma et~al.(2020)Sharma, Henderson, and Ghosh]{Sharma2020}
Shubham Sharma, Jette Henderson, and Joydeep Ghosh.
\newblock Certifai: A common framework to provide explanations and analyse the fairness and robustness of black-box models.
\newblock In \emph{Proceedings of the AAAI/ACM Conference on AI, Ethics, and Society}, AIES ’20. ACM, February 2020.
\newblock \doi{10.1145/3375627.3375812}.

\bibitem[Shen et~al.(2020)Shen, Wu, Jiang, Zeng, LAU, Vianova, and Qu]{Shen2020}
Qiaomu Shen, Yanhong Wu, Yuzhe Jiang, Wei Zeng, Alexis K~H LAU, Anna Vianova, and Huamin Qu.
\newblock Visual interpretation of recurrent neural network on multi-dimensional time-series forecast.
\newblock In \emph{2020 {IEEE} Pacific Visualization Symposium ({PacificVis})}. {IEEE}, jun 2020.
\newblock \doi{10.1109/pacificvis48177.2020.2785}.

\bibitem[Shrikumar et~al.(2017)Shrikumar, Greenside, and Kundaje]{Shrikumar2017}
Avanti Shrikumar, Peyton Greenside, and Anshul Kundaje.
\newblock Learning important features through propagating activation differences.
\newblock In \emph{Proceedings of the 34th International Conference on Machine Learning - Volume 70}, ICML'17, page 3145–3153. JMLR.org, 2017.

\bibitem[Silver et~al.(2017)Silver, Schrittwieser, Simonyan, Antonoglou, Huang, Guez, Hubert, Baker, Lai, Bolton, Chen, Lillicrap, Hui, Sifre, van~den Driessche, Graepel, and Hassabis]{Silver2017}
David Silver, Julian Schrittwieser, Karen Simonyan, Ioannis Antonoglou, Aja Huang, Arthur Guez, Thomas Hubert, Lucas Baker, Matthew Lai, Adrian Bolton, Yutian Chen, Timothy Lillicrap, Fan Hui, Laurent Sifre, George van~den Driessche, Thore Graepel, and Demis Hassabis.
\newblock Mastering the game of go without human knowledge.
\newblock \emph{Nature}, 550\penalty0 (7676):\penalty0 354--359, October 2017.
\newblock ISSN 1476-4687.
\newblock \doi{10.1038/nature24270}.

\bibitem[Simonyan and Zisserman(2014)]{Simonyan2014}
Karen Simonyan and Andrew Zisserman.
\newblock Very deep convolutional networks for large-scale image recognition.
\newblock September 2014.
\newblock \doi{10.48550/ARXIV.1409.1556}.

\bibitem[Simonyan et~al.(2014)Simonyan, Vedaldi, and Zisserman]{Simonyan2013}
Karen Simonyan, Andrea Vedaldi, and Andrew Zisserman.
\newblock Deep inside convolutional networks: Visualising image classification models and saliency maps.
\newblock In Yoshua Bengio and Yann LeCun, editors, \emph{2nd International Conference on Learning Representations, {ICLR} 2014, Banff, AB, Canada, April 14-16, 2014, Workshop Track Proceedings}, 2014.

\bibitem[Singh and Yow(2021)]{Singh2021}
Gurmail Singh and Kin-Choong Yow.
\newblock These do not look like those: An interpretable deep learning model for image recognition.
\newblock \emph{{IEEE} Access}, 9:\penalty0 41482--41493, 2021.
\newblock \doi{10.1109/access.2021.3064838}.

\bibitem[Sirois et~al.(2005)Sirois, Tsoukas, Chou, Wei, Boucher, and Hatzakis]{Sirois2005}
Suzanne Sirois, Christos Tsoukas, Kuo-Chen Chou, Dong-Qing Wei, Christina Boucher, and G~Hatzakis.
\newblock Selection of molecular descriptors with artificial intelligence for the understanding of hiv-1 protease peptidomimetic inhibitors-activity.
\newblock \emph{Medicinal chemistry (Shāriqah (United Arab Emirates))}, 1:\penalty0 173--84, 04 2005.
\newblock \doi{10.2174/1573406053175238}.

\bibitem[{\v{S}}krlj et~al.(2021){\v{S}}krlj, Sheehan, Er{\v{z}}en, Robnik-{\v{S}}ikonja, Luz, and Pollak]{Skrlj2021}
Bla{\v{z}} {\v{S}}krlj, Shane Sheehan, Nika Er{\v{z}}en, Marko Robnik-{\v{S}}ikonja, Saturnino Luz, and Senja Pollak.
\newblock Exploring neural language models via analysis of local and global self-attention spaces.
\newblock In \emph{Proceedings of the EACL Hackashop on News Media Content Analysis and Automated Report Generation}, pages 76--83, Online, April 2021. Association for Computational Linguistics.
\newblock URL \url{https://aclanthology.org/2021.hackashop-1.11}.

\bibitem[Slack et~al.(2023)Slack, Krishna, Lakkaraju, and Singh]{Slack2023}
Dylan Slack, Satyapriya Krishna, Himabindu Lakkaraju, and Sameer Singh.
\newblock Explaining machine learning models with interactive natural language conversations using talktomodel.
\newblock \emph{Nature Machine Intelligence}, 5\penalty0 (8):\penalty0 873--883, July 2023.
\newblock ISSN 2522-5839.
\newblock \doi{10.1038/s42256-023-00692-8}.

\bibitem[Smilkov et~al.(2016)Smilkov, Thorat, Nicholson, Reif, Viégas, and Wattenberg]{Smilkov2016}
Daniel Smilkov, Nikhil Thorat, Charles Nicholson, Emily Reif, Fernanda~B. Viégas, and Martin Wattenberg.
\newblock Embedding projector: Interactive visualization and interpretation of embeddings.
\newblock \emph{arXiv preprint arXiv:1611.05469}, November 2016.

\bibitem[Snell et~al.(2017)Snell, Swersky, and Zemel]{Snell2017}
Jake Snell, Kevin Swersky, and Richard Zemel.
\newblock Prototypical networks for few-shot learning.
\newblock In \emph{Proceedings of the 31st International Conference on Neural Information Processing Systems}, NIPS'17, page 4080–4090. Curran Associates Inc., 2017.
\newblock ISBN 9781510860964.

\bibitem[Springenberg et~al.(2015)Springenberg, Dosovitskiy, Brox, and Riedmiller]{Springenberg2015}
J~Springenberg, Alexey Dosovitskiy, Thomas Brox, and M~Riedmiller.
\newblock Striving for simplicity: The all convolutional net.
\newblock In \emph{ICLR (workshop track)}, 2015.

\bibitem[Stigler(1989)]{Stigler1989}
Stephen~M. Stigler.
\newblock Francis galton{\textquotesingle}s account of the invention of correlation.
\newblock \emph{Statistical Science}, 4\penalty0 (2), may 1989.
\newblock \doi{10.1214/ss/1177012580}.

\bibitem[Strezoski and Worring(2017)]{Strezoski2017}
Gjorgji Strezoski and Marcel Worring.
\newblock Plug-and-play interactive deep network visualization.
\newblock \emph{VADL: Visual Analytics for Deep Learning}, pages 0100--0106, 2017.

\bibitem[Strobelt et~al.(2018)Strobelt, Gehrmann, Pfister, and Rush]{Strobelt2018}
Hendrik Strobelt, Sebastian Gehrmann, Hanspeter Pfister, and Alexander~M. Rush.
\newblock {LSTMVis}: A tool for visual analysis of hidden state dynamics in recurrent neural networks.
\newblock \emph{{IEEE} Transactions on Visualization and Computer Graphics}, 24\penalty0 (1):\penalty0 667--676, jan 2018.
\newblock \doi{10.1109/tvcg.2017.2744158}.

\bibitem[Strobelt et~al.(2019)Strobelt, Gehrmann, Behrisch, Perer, Pfister, and Rush]{Strobelt2019}
Hendrik Strobelt, Sebastian Gehrmann, Michael Behrisch, Adam Perer, Hanspeter Pfister, and Alexander~M. Rush.
\newblock Seq2seq-vis: A visual debugging tool for sequence-to-sequence models.
\newblock \emph{{IEEE} Transactions on Visualization and Computer Graphics}, 25\penalty0 (1):\penalty0 353--363, jan 2019.
\newblock \doi{10.1109/tvcg.2018.2865044}.

\bibitem[Subramanian et~al.(2018)Subramanian, Pruthi, Jhamtani, Berg-Kirkpatrick, and Hovy]{Subramanian2018}
Anant Subramanian, Danish Pruthi, Harsh Jhamtani, Taylor Berg-Kirkpatrick, and Eduard Hovy.
\newblock Spine: Sparse interpretable neural embeddings.
\newblock \emph{Proceedings of the AAAI Conference on Artificial Intelligence}, 32\penalty0 (1), April 2018.
\newblock ISSN 2159-5399.
\newblock \doi{10.1609/aaai.v32i1.11935}.

\bibitem[Subramanian et~al.(2016)Subramanian, Ramsundar, Pande, and Denny]{Subramanian2016}
Govindan Subramanian, Bharath Ramsundar, Vijay Pande, and Rajiah~Aldrin Denny.
\newblock Computational modeling of $\beta$-secretase 1 (bace-1) inhibitors using ligand based approaches.
\newblock \emph{Journal of Chemical Information and Modeling}, 56\penalty0 (10):\penalty0 1936--1949, October 2016.
\newblock ISSN 1549-960X.
\newblock \doi{10.1021/acs.jcim.6b00290}.

\bibitem[Sundararajan et~al.(2017)Sundararajan, Taly, and Yan]{Sundararajan2017}
Mukund Sundararajan, Ankur Taly, and Qiqi Yan.
\newblock Axiomatic attribution for deep networks.
\newblock In \emph{Proceedings of the 34th International Conference on Machine Learning - Volume 70}, ICML'17, page 3319–3328. JMLR.org, 2017.

\bibitem[Szegedy et~al.(2015)Szegedy, Liu, Jia, Sermanet, Reed, Anguelov, Erhan, Vanhoucke, and Rabinovich]{Szegedy2015}
Christian Szegedy, Wei Liu, Yangqing Jia, Pierre Sermanet, Scott Reed, Dragomir Anguelov, Dumitru Erhan, Vincent Vanhoucke, and Andrew Rabinovich.
\newblock Going deeper with convolutions.
\newblock In \emph{Proceedings of the IEEE Conference on Computer Vision and Pattern Recognition (CVPR)}, pages 1--9, June 2015.

\bibitem[Taigman et~al.(2014)Taigman, Yang, Ranzato, and Wolf]{Taigman2014}
Yaniv Taigman, Ming Yang, Marc’Aurelio Ranzato, and Lior Wolf.
\newblock Deepface: Closing the gap to human-level performance in face verification.
\newblock In \emph{2014 IEEE Conference on Computer Vision and Pattern Recognition}. IEEE, June 2014.
\newblock \doi{10.1109/cvpr.2014.220}.

\bibitem[Tan and Le(2019)]{Tan2019}
Mingxing Tan and Quoc Le.
\newblock {E}fficient{N}et: Rethinking model scaling for convolutional neural networks.
\newblock In Kamalika Chaudhuri and Ruslan Salakhutdinov, editors, \emph{Proceedings of the 36th International Conference on Machine Learning}, volume~97 of \emph{Proceedings of Machine Learning Research}, pages 6105--6114. PMLR, 2019.

\bibitem[Tenney et~al.(2019)Tenney, Das, and Pavlick]{Tenney2019}
Ian Tenney, Dipanjan Das, and Ellie Pavlick.
\newblock Bert rediscovers the classical nlp pipeline.
\newblock In \emph{Proceedings of the 57th Annual Meeting of the Association for Computational Linguistics}. Association for Computational Linguistics, 2019.
\newblock \doi{10.18653/v1/p19-1452}.

\bibitem[Tenney et~al.(2020)Tenney, Wexler, Bastings, Bolukbasi, Coenen, Gehrmann, Jiang, Pushkarna, Radebaugh, Reif, and Yuan]{Tenney2020}
Ian Tenney, James Wexler, Jasmijn Bastings, Tolga Bolukbasi, Andy Coenen, Sebastian Gehrmann, Ellen Jiang, Mahima Pushkarna, Carey Radebaugh, Emily Reif, and Ann Yuan.
\newblock The language interpretability tool: Extensible, interactive visualizations and analysis for nlp models.
\newblock In \emph{Proceedings of the 2020 Conference on Empirical Methods in Natural Language Processing: System Demonstrations}. Association for Computational Linguistics, 2020.
\newblock \doi{10.18653/v1/2020.emnlp-demos.15}.

\bibitem[Teoh and Ma(2003)]{Teoh2003}
Soon~Tee Teoh and Kwan-Liu Ma.
\newblock Paintingclass: interactive construction, visualization and exploration of decision trees.
\newblock In \emph{Proceedings of the ninth ACM SIGKDD international conference on Knowledge discovery and data mining}, KDD03. ACM, August 2003.
\newblock \doi{10.1145/956750.956837}.

\bibitem[Thomas and Cook(2006)]{Thomas2006}
J.J. Thomas and K.A. Cook.
\newblock A visual analytics agenda.
\newblock \emph{IEEE Computer Graphics and Applications}, 26\penalty0 (1):\penalty0 10--13, 2006.
\newblock \doi{10.1109/MCG.2006.5}.

\bibitem[Tjoa and Guan(2021)]{Tjoa2021}
Erico Tjoa and Cuntai Guan.
\newblock A survey on explainable artificial intelligence (xai): Toward medical xai.
\newblock \emph{IEEE Transactions on Neural Networks and Learning Systems}, 32\penalty0 (11):\penalty0 4793--4813, November 2021.
\newblock ISSN 2162-2388.
\newblock \doi{10.1109/tnnls.2020.3027314}.

\bibitem[Tominski and Schumann(2020)]{Tominski20IVDA}
Christian Tominski and Heidrun Schumann.
\newblock \emph{{Interactive Visual Data Analysis}}.
\newblock AK Peters Visualization Series. CRC Press, 2020.
\newblock ISBN 9781498753982.
\newblock \doi{10.1201/9781315152707}.
\newblock URL \url{https://ivda-book.de}.

\bibitem[Tsallis(1988)]{Tsallis1988}
Constantino Tsallis.
\newblock Possible generalization of boltzmann-gibbs statistics.
\newblock \emph{Journal of Statistical Physics}, 52\penalty0 (1-2):\penalty0 479--487, jul 1988.
\newblock \doi{10.1007/bf01016429}.

\bibitem[van~den Brandt et~al.(2020)van~den Brandt, Christopher, Rezapour, Welsbie, Camp, Baxter, Do, Moghimi, Belghith, Bowd, et~al.]{VanDenBrandt2020}
Astrid van~den Brandt, Mark Christopher, Jasmin Rezapour, Derek~S Welsbie, Andrew~S Camp, Sally~L Baxter, Jiun Do, Sasan Moghimi, Akram Belghith, Christopher Bowd, et~al.
\newblock Glance: A visual analytics approach for opening the black box to explain deep learning predictions of glaucomatous visual field damage from optical coherence tomography scans.
\newblock \emph{Investigative Ophthalmology \& Visual Science}, 61\penalty0 (7):\penalty0 4527--4527, 2020.

\bibitem[van~der Maaten and Hinton(2008)]{Vandermaaten08}
Laurens van~der Maaten and Geoffrey Hinton.
\newblock Visualizing data using t-sne.
\newblock \emph{Journal of Machine Learning Research}, 9\penalty0 (86):\penalty0 2579--2605, 2008.
\newblock URL \url{http://jmlr.org/papers/v9/vandermaaten08a.html}.

\bibitem[Varshneya et~al.(2021)Varshneya, Ledent, Vandermeulen, Lei, Enders, Borth, and Kloft]{Varshneya2021}
Saurabh Varshneya, Antoine Ledent, Robert~A. Vandermeulen, Yunwen Lei, Matthias Enders, Damian Borth, and Marius Kloft.
\newblock Learning interpretable concept groups in {CNNs}.
\newblock In \emph{Proceedings of the Thirtieth International Joint Conference on Artificial Intelligence}. International Joint Conferences on Artificial Intelligence Organization, aug 2021.
\newblock \doi{10.24963/ijcai.2021/147}.

\bibitem[Vaswani et~al.(2017)Vaswani, Shazeer, Parmar, Uszkoreit, Jones, Gomez, Kaiser, and Polosukhin]{Vaswani2017}
Ashish Vaswani, Noam Shazeer, Niki Parmar, Jakob Uszkoreit, Llion Jones, Aidan~N Gomez, \L~ukasz Kaiser, and Illia Polosukhin.
\newblock Attention is all you need.
\newblock In I.~Guyon, U.~Von Luxburg, S.~Bengio, H.~Wallach, R.~Fergus, S.~Vishwanathan, and R.~Garnett, editors, \emph{Advances in Neural Information Processing Systems}, volume~30. Curran Associates, Inc., 2017.
\newblock URL \url{https://proceedings.neurips.cc/paper_files/paper/2017/file/3f5ee243547dee91fbd053c1c4a845aa-Paper.pdf}.

\bibitem[Veli{\v{c}}kovi{\'{c}} et~al.(2018)Veli{\v{c}}kovi{\'{c}}, Cucurull, Casanova, Romero, Li{\`{o}}, and Bengio]{Velickovic2018}
Petar Veli{\v{c}}kovi{\'{c}}, Guillem Cucurull, Arantxa Casanova, Adriana Romero, Pietro Li{\`{o}}, and Yoshua Bengio.
\newblock {Graph Attention Networks}.
\newblock \emph{International Conference on Learning Representations}, 2018.
\newblock URL \url{https://openreview.net/forum?id=rJXMpikCZ}.

\bibitem[Vermeire et~al.(2022)Vermeire, Brughmans, Goethals, de~Oliveira, and Martens]{Vermeire2022}
Tom Vermeire, Dieter Brughmans, Sofie Goethals, Raphael Mazzine~Barbossa de~Oliveira, and David Martens.
\newblock Explainable image classification with evidence counterfactual.
\newblock \emph{Pattern Analysis and Applications}, 25\penalty0 (2):\penalty0 315--335, January 2022.
\newblock ISSN 1433-755X.
\newblock \doi{10.1007/s10044-021-01055-y}.

\bibitem[Vig(2019)]{Vig2019}
Jesse Vig.
\newblock A multiscale visualization of attention in the transformer model.
\newblock In \emph{Proceedings of the 57th Annual Meeting of the Association for Computational Linguistics: System Demonstrations}, pages 37--42, 2019.

\bibitem[Vig and Belinkov(2019)]{Vig2019nlp}
Jesse Vig and Yonatan Belinkov.
\newblock Analyzing the structure of attention in a transformer language model.
\newblock In \emph{Proceedings of the 2019 ACL Workshop BlackboxNLP: Analyzing and Interpreting Neural Networks for NLP}. Association for Computational Linguistics, 2019.
\newblock \doi{10.18653/v1/w19-4808}.

\bibitem[Vinyals et~al.(2016)Vinyals, Blundell, Lillicrap, Kavukcuoglu, and Wierstra]{Vinyals2016}
Oriol Vinyals, Charles Blundell, Timothy Lillicrap, Koray Kavukcuoglu, and Daan Wierstra.
\newblock Matching networks for one shot learning.
\newblock In \emph{Proceedings of the 30th International Conference on Neural Information Processing Systems}, NIPS'16, page 3637–3645. Curran Associates Inc., 2016.
\newblock ISBN 9781510838819.

\bibitem[Wachter et~al.(2018)Wachter, Mittelstadt, and Russell]{Wachter2017}
Sandra Wachter, Brent Mittelstadt, and Chris Russell.
\newblock Counterfactual explanations without opening the black box: Automated decisions and the gdpr.
\newblock \emph{Harvard Journal of Law and Technology}, 31\penalty0 (2):\penalty0 841--887, 2018.

\bibitem[Wang and Ma(2020)]{wang_hyppersteer_2020}
Chuan Wang and Kwan-Liu Ma.
\newblock Hyppersteer: Hypothetical steering and data perturbation in sequence prediction with deep learning.
\newblock \emph{arXiv preprint arXiv:2011.02149}, November 2020.

\bibitem[Wang et~al.(2018{\natexlab{a}})Wang, Onishi, Nemoto, and Ma]{Wang2018rnn}
Chuan Wang, Takeshi Onishi, Keiichi Nemoto, and Kwan-Liu Ma.
\newblock Visual reasoning of feature attribution with deep recurrent neural networks.
\newblock In \emph{2018 IEEE International Conference on Big Data (Big Data)}, pages 1661--1668, 2018{\natexlab{a}}.
\newblock \doi{10.1109/BigData.2018.8622502}.

\bibitem[Wang et~al.(2018{\natexlab{b}})Wang, Liu, and Cheng]{Wang2018features}
Feng Wang, Haijun Liu, and Jian Cheng.
\newblock Visualizing deep neural network by alternately image blurring and deblurring.
\newblock \emph{Neural Networks}, 97:\penalty0 162--172, January 2018{\natexlab{b}}.
\newblock ISSN 0893-6080.
\newblock \doi{10.1016/j.neunet.2017.09.007}.

\bibitem[{Wang} et~al.(2019){Wang}, {Gou}, {Shen}, and {Yang}]{Wang2019}
J.~{Wang}, L.~{Gou}, H.~{Shen}, and H.~{Yang}.
\newblock Dqnviz: A visual analytics approach to understand deep q-networks.
\newblock \emph{IEEE Transactions on Visualization and Computer Graphics}, 25\penalty0 (1):\penalty0 288--298, Jan 2019.
\newblock ISSN 1941-0506.
\newblock \doi{10.1109/TVCG.2018.2864504}.

\bibitem[Wang et~al.(2021{\natexlab{a}})Wang, Liu, Wang, and Jing]{Wang2021tesnet}
Jiaqi Wang, Huafeng Liu, Xinyue Wang, and Liping Jing.
\newblock Interpretable image recognition by constructing transparent embedding space.
\newblock In \emph{Proceedings of the IEEE/CVF International Conference on Computer Vision (ICCV)}, pages 895--904, October 2021{\natexlab{a}}.

\bibitem[Wang et~al.(2019)Wang, Gou, Zhang, Yang, and Shen]{Wang2019DeepVid}
Junpeng Wang, Liang Gou, Wei Zhang, Hao Yang, and Han-Wei Shen.
\newblock {DeepVID}: Deep visual interpretation and diagnosis for image classifiers via knowledge distillation.
\newblock \emph{{IEEE} Transactions on Visualization and Computer Graphics}, 25\penalty0 (6):\penalty0 2168--2180, jun 2019.
\newblock \doi{10.1109/tvcg.2019.2903943}.

\bibitem[Wang et~al.(2020)Wang, Zhang, and Yang]{Wang2020scan}
Junpeng Wang, Wei Zhang, and Hao Yang.
\newblock {SCANViz}: Interpreting the symbol-concept association captured by deep neural networks through visual analytics.
\newblock In \emph{2020 IEEE Pacific Visualization Symposium (PacificVis)}. {IEEE}, jun 2020.
\newblock \doi{10.1109/pacificvis48177.2020.3542}.

\bibitem[Wang et~al.(2021{\natexlab{b}})Wang, Zhang, Yang, Yeh, and Wang]{Wang2021drl}
Junpeng Wang, Wei Zhang, Hao Yang, Chin-Chia~Michael Yeh, and Liang Wang.
\newblock Visual analytics for {RNN}-based deep reinforcement learning.
\newblock \emph{{IEEE} Transactions on Visualization and Computer Graphics}, pages 1--1, 2021{\natexlab{b}}.
\newblock \doi{10.1109/tvcg.2021.3076749}.

\bibitem[Wang and Vasconcelos(2020)]{Wang2020counter}
Pei Wang and Nuno Vasconcelos.
\newblock Scout: Self-aware discriminant counterfactual explanations.
\newblock In \emph{2020 IEEE/CVF Conference on Computer Vision and Pattern Recognition (CVPR)}. IEEE, June 2020.
\newblock \doi{10.1109/cvpr42600.2020.00900}.

\bibitem[Wang et~al.(2022)Wang, He, Jin, Yang, Wang, and Qu]{Wang2021a}
Xingbo Wang, Jianben He, Zhihua Jin, Muqiao Yang, Yong Wang, and Huamin Qu.
\newblock M2lens: Visualizing and explaining multimodal models for sentiment analysis.
\newblock \emph{IEEE Transactions on Visualization and Computer Graphics}, 28\penalty0 (1):\penalty0 802--812, 2022.
\newblock \doi{10.1109/TVCG.2021.3114794}.

\bibitem[Wang et~al.(2021{\natexlab{c}})Wang, Turko, and Chau]{Wang2021dodrio}
Zijie~J. Wang, Robert Turko, and Duen~Horng Chau.
\newblock Dodrio: Exploring transformer models with interactive visualization.
\newblock In \emph{Proceedings of the 59th Annual Meeting of the Association for Computational Linguistics and the 11th International Joint Conference on Natural Language Processing: System Demonstrations}, pages 132--141, Online, August 2021{\natexlab{c}}. Association for Computational Linguistics.
\newblock \doi{10.18653/v1/2021.acl-demo.16}.

\bibitem[Wexler et~al.(2019)Wexler, Pushkarna, Bolukbasi, Wattenberg, Viegas, and Wilson]{Wexler2019}
James Wexler, Mahima Pushkarna, Tolga Bolukbasi, Martin Wattenberg, Fernanda Viegas, and Jimbo Wilson.
\newblock The what-if tool: Interactive probing of machine learning models.
\newblock \emph{{IEEE} Transactions on Visualization and Computer Graphics}, pages 1--1, 2019.
\newblock \doi{10.1109/tvcg.2019.2934619}.

\bibitem[Wiegreffe and Pinter(2019{\natexlab{a}})]{Wiegreffe2019}
Sarah Wiegreffe and Yuval Pinter.
\newblock Attention is not not explanation.
\newblock In Kentaro Inui, Jing Jiang, Vincent Ng, and Xiaojun Wan, editors, \emph{Proceedings of the 2019 Conference on Empirical Methods in Natural Language Processing and the 9th International Joint Conference on Natural Language Processing (EMNLP-IJCNLP)}, pages 11--20, Hong Kong, China, November 2019{\natexlab{a}}. Association for Computational Linguistics.
\newblock \doi{10.18653/v1/D19-1002}.
\newblock URL \url{https://aclanthology.org/D19-1002}.

\bibitem[Wiegreffe and Pinter(2019{\natexlab{b}})]{Wiegreffe2019notnot}
Sarah Wiegreffe and Yuval Pinter.
\newblock Attention is not not explanation.
\newblock In \emph{Proceedings of the 2019 Conference on Empirical Methods in Natural Language Processing and the 9th International Joint Conference on Natural Language Processing (EMNLP-IJCNLP)}. Association for Computational Linguistics, 2019{\natexlab{b}}.
\newblock \doi{10.18653/v1/d19-1002}.

\bibitem[Wright et~al.(2021)Wright, Shaikh, Park, Epperson, Ahmed, Pinel, Chau, and Yang]{Wright2021}
Austin~P. Wright, Omar Shaikh, Haekyu Park, Will Epperson, Muhammed Ahmed, Stephane Pinel, Duen Horng~(Polo) Chau, and Diyi Yang.
\newblock {RECAST}: Enabling user recourse and interpretability of toxicity detection models with interactive visualization.
\newblock \emph{Proceedings of the {ACM} on Human-Computer Interaction}, 5\penalty0 ({CSCW}1):\penalty0 1--26, apr 2021.
\newblock \doi{10.1145/3449280}.

\bibitem[Wu et~al.(2021)Wu, Pan, Chen, Long, Zhang, and Yu]{Wu2021}
Zonghan Wu, Shirui Pan, Fengwen Chen, Guodong Long, Chengqi Zhang, and Philip~S. Yu.
\newblock A comprehensive survey on graph neural networks.
\newblock \emph{{IEEE} Transactions on Neural Networks and Learning Systems}, 32\penalty0 (1):\penalty0 4--24, January 2021.
\newblock \doi{10.1109/tnnls.2020.2978386}.
\newblock URL \url{https://doi.org/10.1109/tnnls.2020.2978386}.

\bibitem[Xu et~al.(2019)Xu, Hu, Leskovec, and Jegelka]{Xu2019}
Keyulu Xu, Weihua Hu, Jure Leskovec, and Stefanie Jegelka.
\newblock How powerful are graph neural networks?
\newblock In \emph{International Conference on Learning Representations}, 2019.
\newblock URL \url{https://openreview.net/forum?id=ryGs6iA5Km}.

\bibitem[Xu et~al.(2020)Xu, Venugopalan, and Sundararajan]{Xu2020}
Shawn Xu, Subhashini Venugopalan, and Mukund Sundararajan.
\newblock Attribution in scale and space.
\newblock In \emph{2020 IEEE/CVF Conference on Computer Vision and Pattern Recognition (CVPR)}. IEEE, June 2020.
\newblock \doi{10.1109/cvpr42600.2020.00970}.

\bibitem[Xu et~al.(2017)Xu, Cheng, Gu, Yang, Chang, and Zhou]{Xu2017}
Shuangjie Xu, Yu~Cheng, Kang Gu, Yang Yang, Shiyu Chang, and Pan Zhou.
\newblock Jointly attentive spatial-temporal pooling networks for video-based person re-identification.
\newblock In \emph{Proceedings of the IEEE International Conference on Computer Vision (ICCV)}, Oct 2017.

\bibitem[Yang et~al.(2020)Yang, Kenny, Ng, Yang, Smyth, and Dong]{Yang2020}
Linyi Yang, Eoin Kenny, Tin Lok~James Ng, Yi~Yang, Barry Smyth, and Ruihai Dong.
\newblock Generating plausible counterfactual explanations for deep transformers in financial text classification.
\newblock In \emph{Proceedings of the 28th International Conference on Computational Linguistics}. International Committee on Computational Linguistics, 2020.
\newblock \doi{10.18653/v1/2020.coling-main.541}.

\bibitem[Yang et~al.(2023)Yang, Panagopoulou, Zhou, Jin, Callison-Burch, and Yatskar]{Yang2023}
Yue Yang, Artemis Panagopoulou, Shenghao Zhou, Daniel Jin, Chris Callison-Burch, and Mark Yatskar.
\newblock Language in a bottle: Language model guided concept bottlenecks for interpretable image classification.
\newblock In \emph{Proceedings of the IEEE/CVF Conference on Computer Vision and Pattern Recognition (CVPR)}, pages 19187--19197, June 2023.

\bibitem[Ying et~al.(2019)Ying, Bourgeois, You, Zitnik, and Leskovec]{Ying2019}
Zhitao Ying, Dylan Bourgeois, Jiaxuan You, Marinka Zitnik, and Jure Leskovec.
\newblock Gnnexplainer: Generating explanations for graph neural networks.
\newblock In H.~Wallach, H.~Larochelle, A.~Beygelzimer, F.~d\textquotesingle Alch\'{e}-Buc, E.~Fox, and R.~Garnett, editors, \emph{Advances in Neural Information Processing Systems}, volume~32. Curran Associates, Inc., 2019.
\newblock URL \url{https://proceedings.neurips.cc/paper/2019/file/d80b7040b773199015de6d3b4293c8ff-Paper.pdf}.

\bibitem[Yosinski et~al.(2015)Yosinski, Clune, Nguyen, Fuchs, and Lipson]{Yosinski2015}
Jason Yosinski, Jeff Clune, Anh Nguyen, Thomas Fuchs, and Hod Lipson.
\newblock Understanding neural networks through deep visualization.
\newblock \emph{arXiv preprint arXiv:1506.06579}, 2015.

\bibitem[Yuan et~al.(2022)Yuan, Yu, Gui, and Ji]{Yuan2022}
Hao Yuan, Haiyang Yu, Shurui Gui, and Shuiwang Ji.
\newblock Explainability in graph neural networks: A taxonomic survey.
\newblock \emph{IEEE Transactions on Pattern Analysis and Machine Intelligence}, pages 1--19, 2022.
\newblock ISSN 1939-3539.
\newblock \doi{10.1109/tpami.2022.3204236}.

\bibitem[Yuksekgonul et~al.(2023)Yuksekgonul, Wang, and Zou]{yuksekgonul2023posthoc}
Mert Yuksekgonul, Maggie Wang, and James Zou.
\newblock Post-hoc concept bottleneck models.
\newblock In \emph{The Eleventh International Conference on Learning Representations}, 2023.
\newblock URL \url{https://openreview.net/forum?id=nA5AZ8CEyow}.

\bibitem[Zagoruyko and Komodakis(2016)]{Zagoruyko2016}
Sergey Zagoruyko and Nikos Komodakis.
\newblock Wide residual networks.
\newblock In \emph{Procedings of the British Machine Vision Conference 2016}. British Machine Vision Association, 2016.
\newblock \doi{10.5244/c.30.87}.

\bibitem[Zahavy et~al.(2016)Zahavy, Ben-Zrihem, and Mannor]{Zahavy2016}
Tom Zahavy, Nir Ben-Zrihem, and Shie Mannor.
\newblock Graying the black box: Understanding dqns.
\newblock In \emph{International Conference on Machine Learning}, pages 1899--1908. PMLR, 2016.

\bibitem[Zeiler and Fergus(2014)]{Zeiler2014}
Matthew~D. Zeiler and Rob Fergus.
\newblock Visualizing and understanding convolutional networks.
\newblock In \emph{Computer Vision {\textendash} {ECCV} 2014}, pages 818--833. Springer International Publishing, 2014.
\newblock \doi{10.1007/978-3-319-10590-1_53}.

\bibitem[Zeng et~al.(2017)Zeng, Haleem, Plantaz, Cao, and Qu]{Zeng2017}
Haipeng Zeng, Hammad Haleem, Xavier Plantaz, Nan Cao, and Huamin Qu.
\newblock Cnncomparator: Comparative analytics of convolutional neural networks.
\newblock \emph{arXiv preprint arXiv:1710.05285}, October 2017.

\bibitem[Zhang et~al.(2018)Zhang, Zhou, Lin, and Sun]{Zhang2018}
Xiangyu Zhang, Xinyu Zhou, Mengxiao Lin, and Jian Sun.
\newblock {ShuffleNet}: An extremely efficient convolutional neural network for mobile devices.
\newblock In \emph{2018 {IEEE}/{CVF} Conference on Computer Vision and Pattern Recognition}, pages 6848--6856. {IEEE}, jun 2018.
\newblock \doi{10.1109/cvpr.2018.00716}.

\bibitem[Zhang et~al.(2021)Zhang, Tino, Leonardis, and Tang]{Zhang2021}
Yu~Zhang, Peter Tino, Ales Leonardis, and Ke~Tang.
\newblock A survey on neural network interpretability.
\newblock \emph{{IEEE} Transactions on Emerging Topics in Computational Intelligence}, 5\penalty0 (5):\penalty0 726--742, oct 2021.
\newblock \doi{10.1109/tetci.2021.3100641}.

\bibitem[Zhang et~al.(2022)Zhang, Liu, Wang, Lu, and Lee]{Zhang2022}
Zaixi Zhang, Qi~Liu, Hao Wang, Chengqiang Lu, and Cheekong Lee.
\newblock {ProtGNN}: Towards self-explaining graph neural networks.
\newblock \emph{Proceedings of the {AAAI} Conference on Artificial Intelligence}, 36\penalty0 (8):\penalty0 9127--9135, June 2022.
\newblock \doi{10.1609/aaai.v36i8.20898}.
\newblock URL \url{https://doi.org/10.1609/aaai.v36i8.20898}.

\bibitem[Zhao et~al.(2020)Zhao, Dai, Xu, and Ren]{Zhao2020}
Junhan Zhao, Zeng Dai, Panpan Xu, and Liu Ren.
\newblock {ProtoViewer}: Visual interpretation and diagnostics of deep neural networks with factorized prototypes.
\newblock In \emph{2020 {IEEE} Visualization Conference ({VIS})}. {IEEE}, oct 2020.
\newblock \doi{10.1109/vis47514.2020.00064}.

\bibitem[Zheng et~al.(2017)Zheng, Fu, Mei, and Luo]{Zheng2017}
Heliang Zheng, Jianlong Fu, Tao Mei, and Jiebo Luo.
\newblock Learning multi-attention convolutional neural network for fine-grained image recognition.
\newblock In \emph{2017 {IEEE} International Conference on Computer Vision ({ICCV})}. {IEEE}, oct 2017.
\newblock \doi{10.1109/iccv.2017.557}.

\bibitem[Zhong et~al.(2017)Zhong, Xie, Zhong, Wang, Xu, Cheng, and Mueller]{Zhong2017}
Wen Zhong, Cong Xie, Yuan Zhong, Yang Wang, Wei Xu, Shenghui Cheng, and Klaus Mueller.
\newblock Evolutionary visual analysis of deep neural networks.
\newblock In \emph{ICML Workshop on Visualization for Deep Learning}, page~9, 2017.

\bibitem[Zhou et~al.(2017)Zhou, Zhao, Puig, Fidler, Barriuso, and Torralba]{zhou2017scene}
Bolei Zhou, Hang Zhao, Xavier Puig, Sanja Fidler, Adela Barriuso, and Antonio Torralba.
\newblock Scene parsing through ade20k dataset.
\newblock In \emph{Proceedings of the IEEE Conference on Computer Vision and Pattern Recognition}, 2017.

\bibitem[Zhou et~al.(2018)Zhou, Lapedriza, Khosla, Oliva, and Torralba]{Zhou2018}
Bolei Zhou, Agata Lapedriza, Aditya Khosla, Aude Oliva, and Antonio Torralba.
\newblock Places: A 10 million image database for scene recognition.
\newblock \emph{{IEEE} Transactions on Pattern Analysis and Machine Intelligence}, 40\penalty0 (6):\penalty0 1452--1464, jun 2018.
\newblock \doi{10.1109/tpami.2017.2723009}.

\bibitem[Zhou et~al.(2019)Zhou, Zhao, Puig, Xiao, Fidler, Barriuso, and Torralba]{zhou2019semantic}
Bolei Zhou, Hang Zhao, Xavier Puig, Tete Xiao, Sanja Fidler, Adela Barriuso, and Antonio Torralba.
\newblock Semantic understanding of scenes through the ade20k dataset.
\newblock \emph{International Journal of Computer Vision}, 127\penalty0 (3):\penalty0 302--321, 2019.

\bibitem[Zimmermann et~al.(2021)Zimmermann, Borowski, Geirhos, Bethge, Wallis, and Brendel]{Zimmermann2021}
Roland~Simon Zimmermann, Judy Borowski, Robert Geirhos, Matthias Bethge, Thomas S.~A. Wallis, and Wieland Brendel.
\newblock How well do feature visualizations support causal understanding of {CNN} activations?
\newblock In A.~Beygelzimer, Y.~Dauphin, P.~Liang, and J.~Wortman Vaughan, editors, \emph{Advances in Neural Information Processing Systems}, 2021.
\newblock URL \url{https://openreview.net/forum?id=vLPqnPf9k0}.

\bibitem[Zintgraf et~al.(2017)Zintgraf, Cohen, Adel, and Welling]{zintgraf2017}
Luisa~M Zintgraf, Taco~S Cohen, Tameem Adel, and Max Welling.
\newblock Visualizing deep neural network decisions: Prediction difference analysis.
\newblock In \emph{International Conference on Learning Representations}, 2017.
\newblock URL \url{https://openreview.net/forum?id=BJ5UeU9xx}.

\bibitem[Štrumbelj and Kononenko(2013)]{Strumbelj2013}
Erik Štrumbelj and Igor Kononenko.
\newblock Explaining prediction models and individual predictions with feature contributions.
\newblock \emph{Knowledge and Information Systems}, 41\penalty0 (3):\penalty0 647--665, August 2013.
\newblock ISSN 0219-3116.
\newblock \doi{10.1007/s10115-013-0679-x}.

\end{thebibliography}

\cleardoublepage

\pagenumbering{gobble}

\end{document}